%% file: main.tex
\documentclass[12pt]{report}
\usepackage{lingmacros}
\usepackage[T1]{fontenc}    
\usepackage{tree-dvips}
\usepackage{amsmath}
\usepackage{tabu}
\usepackage[euler]{textgreek}
\usepackage[utf8]{inputenc}
\usepackage{graphicx}
\usepackage{cite}
\usepackage{tabularx}
\usepackage[english]{babel}
\usepackage{makecell}
\addto{\captionsenglish}{}
\usepackage[utf8]{inputenc}
\usepackage{subfigure}
\usepackage{caption}
\usepackage{booktabs}
\usepackage{multirow}
\usepackage{siunitx}
\usepackage{rotating}
\usepackage{float}
\usepackage{csquotes}
\usepackage[colorlinks, citecolor=black, urlcolor=black]{hyperref}
\usepackage{url}            
\usepackage{amsfonts}       
\usepackage{nicefrac}       
\usepackage{microtype}      
\usepackage{xcolor}
\usepackage{amssymb}
\usepackage{wrapfig}
\usepackage{color}
\usepackage{hyperref}
\usepackage{float} 
\usepackage{lipsum}
\usepackage{stackengine}
\usepackage{authblk}
\usepackage{amsthm}
\usepackage{enumitem}
\usepackage{times}

\usepackage{algorithmic}
\usepackage[ruled,vlined]{algorithm2e}

\usepackage{tikz}
\usepackage{comment}
\usepackage{orcidlink}
\usepackage{nicematrix}
\usepackage{mathtools} 
\usepackage[accsupp]{axessibility}  

\DeclareMathOperator*{\argminA}{arg\,min} 

\DeclareMathOperator*{\argmaxA}{arg\,max} 

\input{math_commands.tex}

\newtheorem{theorem}{Theorem}
\newtheorem{lemma}{Lemma}

\DeclareMathOperator{\birth}{birth}
\DeclareMathOperator{\death}{death}
\DeclareMathOperator{\dgm}{Dgm}

\newcommand{\etal}{\textit{et al.}}
\newcommand{\myparagraph}[1]{\noindent\textbf{#1}}

\newcommand{\highlight}[1]{#1}

\begin{document}
\pagenumbering{gobble}

\vspace*{3\baselineskip}
\centerline{\bf{Learning Topological Representations for Deep Image Understanding}}
\vspace*{1\baselineskip}
\centerline{A Dissertation presented}
\vspace*{1\baselineskip}
\centerline{by} 
\vspace*{1\baselineskip}
\centerline{\bf{Xiaoling Hu}}
\vspace*{1\baselineskip}
\centerline{to} 
\vspace*{1\baselineskip}
\centerline{The Graduate School}
\vspace*{1\baselineskip}
\centerline{in Partial Fulfillment of the}
\vspace*{1\baselineskip}
\centerline{Requirements}
\vspace*{1\baselineskip}
\centerline{for the Degree of}
\vspace*{1\baselineskip}
\centerline{\bf{Doctor of Philosophy}}
\vspace*{1\baselineskip}
\centerline{in}
\vspace*{1\baselineskip}
\centerline{\bf{Computer Science}}
\vspace*{2\baselineskip}
\centerline{Stony Brook University}
\vspace*{2\baselineskip}
\centerline{\bf{August 2023}}




\newpage

\pagenumbering{roman}
\setcounter{page}{2}

\centerline{\bf{Stony Brook University}}
\vspace*{1\baselineskip}
\centerline{The Graduate School}
\vspace*{2\baselineskip}
\centerline{\bf{Xiaoling Hu}}
\vspace*{2\baselineskip}
\centerline{We, the dissertation committee for the above candidate for the}
\vspace*{1\baselineskip}
\centerline{Doctor of Philosophy degree, hereby recommend}
\vspace*{1\baselineskip}
\centerline{acceptance of this dissertation}
\vspace*{3\baselineskip}

\centerline{\bf{Chao Chen, Advisor}}
\centerline{\bf{Assistant Professor, Department of Biomedical Informatics}}
\vspace*{2\baselineskip}
\centerline{\bf{Dimitris Samaras, Committee Member}}
\centerline{\bf{SUNY Empire Innovation Professor, Department of Computer Science}}
\vspace*{2\baselineskip}
\centerline{\bf{Haibin Ling, Committee Member}}
\centerline{\bf{SUNY Empire Innovation Professor, Department of Computer Science}}
\vspace*{2\baselineskip}
\centerline{\bf{Fuxin Li, Outside Committee Member}}
\centerline{\bf{Associate Professor, Department of EECS, Oregon State University}}

\vspace*{2\baselineskip}
\centerline{This dissertation is accepted by the Graduate School}
\vspace*{3\baselineskip}
\centerline{Celia Marshik}
\centerline{Dean of the Graduate School}

\centerline{Abstract of the Dissertation}
\vspace*{1\baselineskip}
\centerline{\bf{Learning Topological Representations for Deep Image Understanding}}
\vspace*{1\baselineskip}
\centerline{by}
\vspace*{1\baselineskip}
\centerline{\bf{Xiaoling Hu}}
\vspace*{1\baselineskip}
\centerline{\bf{Doctor of Philosophy}}
\vspace*{1\baselineskip}
\centerline{in}
\vspace*{1\baselineskip}
\centerline{\bf{Computer Science}}
\vspace*{1\baselineskip}
\centerline{Stony Brook University}
\vspace*{1\baselineskip}
\centerline{\bf{2023}}
\vspace*{2\baselineskip}

In many scenarios, especially biomedical applications, the correct delineation of complex fine-scaled structures such as neurons, tissues, and vessels is critical for downstream analysis. Despite the strong predictive power of deep learning methods, they do not provide a satisfactory representation of these structures, thus creating significant barriers in scalable annotation and downstream analysis. In this dissertation, we tackle such challenges by proposing novel representations of these topological structures in a deep learning framework. We leverage the mathematical tools from topological data analysis, i.e., persistent homology and discrete Morse theory, to develop principled methods for better segmentation and uncertainty estimation, which will become powerful tools for scalable annotation.

We focus on a few specific problems. First, we propose novel topological losses for fully-supervised segmentation. Although deep-learning-based segmentation methods have achieved satisfactory segmentation performance in terms of per-pixel accuracy, most of them are still prone to structural errors, e.g., broken connections and missing connected components. We propose topological losses to teach neural networks to segment with correct topology. The continuous-valued loss functions enforce a segmentation to have a better topology by penalizing topologically critical pixels/locations. Second, we focus on the interactive setting, where uncertainty measurement of a neural network segmenter is crucial for scalable annotation. Existing methods only learn pixel-wise feature representations. We move from pixel space to structure space using the classic discrete Morse theory. We decompose an input image into structural elements such as branches and patches and then learn a probabilistic model over such structural space. Our method effectively identifies hypothetical structures that a model is uncertain about and asks a domain expert to confirm. This will significantly improve the annotation speed. 



\newpage

\vspace*{\fill}
\centerline{Dedicated to my family.}
\vspace*{\fill}

\cleardoublepage

{\hypersetup{linkcolor=black}
\newpage
\tableofcontents

\listoffigures
 \addcontentsline{toc}{chapter}{\numberline{}List of Figures}
\listoftables
 \addcontentsline{toc}{chapter}{\numberline{}List of Tables}
}

\cleardoublepage
\pagenumbering{arabic}

\input{intro}
\input{related}

\input{topoloss}
\input{trojan}
\input{warping}
\input{dmt_loss}
\input{uncertainty}

\input{conclusion}

\bibliographystyle{unsrt}
\bibliography{ref}
\addcontentsline{toc}{chapter}{\numberline{}References}

\end{document}

%% file: math_commands.tex
\usepackage{amsmath,amsfonts,bm}

\def\eqref#1{equation~\ref{#1}}

\def\1{\bm{1}}

\DeclareMathAlphabet{\mathsfit}{\encodingdefault}{\sfdefault}{m}{sl}
\SetMathAlphabet{\mathsfit}{bold}{\encodingdefault}{\sfdefault}{bx}{n}

\newcommand{\R}{\mathbb{R}}

\DeclareMathOperator*{\argmax}{arg\,max}
\DeclareMathOperator*{\argmin}{arg\,min}

\newcommand{\reals}  {{\mathbb{R}}}
\newcommand{\Stm}   {{\mathsf{S}}}
\newcommand{\myVF}   {{\mathsf{M}}}
\newcommand{\myP}   {{\mathcal{P}}}
\newcommand{\myS}   {{\mathcal{S}}}
\newcommand{\pers}  {{\mathrm{pers}}}
\newcommand{\pp}    {{\mathsf{p}}}
\newcommand{\hatS}  {{\widehat{\myS}}}

%% file: intro.tex
\chapter{Introduction and Overview}



In many scenarios, especially biomedical applications, the correct delineation of complex fine-scaled structures such as neurons, tissues, and vessels is critical for downstream analysis. Despite the strong predictive power of deep learning methods, they are only learning pixel-wise representations, thus creating significant barriers in scalable annotation and downstream analysis. 

The goal of this dissertation is to explore beyond pixel-wise representations: \textit{How can we be able to learn topological representations to understand the topology/structures for image tasks}? 

In this dissertation, we tackle such challenges by proposing novel representations of these topological structures in a deep learning framework. We leverage the mathematical tools from topological data analysis, i.e., persistent homology and discrete Morse theory, to develop principled methods for better segmentation and uncertainty estimation, which will become powerful tools for scalable annotation.

\section{Topological Loss for Image Segmentation}
Image segmentation, i.e., assigning labels to all pixels of an input image, is crucial in many computer vision tasks. State-of-the-art deep segmentation methods~\cite{long2015fully,he2017mask,chen2014semantic,chen2018deeplab,chen2017rethinking} learn high quality feature representations through an end-to-end trained deep network and achieve satisfactory per-pixel accuracy. 

However, these segmentation algorithms are still prone to make errors on fine-scaled structures, such as small object instances, instances with multiple connected components, and thin connections. These fine-scaled structures may be crucial in analyzing the $\emph{functionality}$ of the objects. For example, accurate extraction of thin parts such as ropes and handles is crucial in planning robot actions, e.g., dragging or grasping. In biomedical images, correct delineation of thin objects such as neuron membranes and vessels is crucial in providing accurate morphological and structural quantification of the underlying system. A broken connection or a missing component may only induce marginal per-pixel error but can cause catastrophic functional mistakes. 

\begin{figure*}[ht]
\centering 
\subfigure[Original]{
\includegraphics[width=0.18\textwidth]{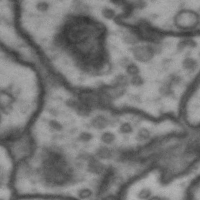}}
\subfigure[GT]{
\includegraphics[width=0.18\textwidth]{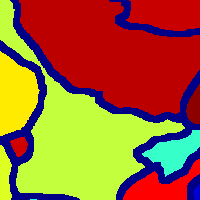}}
\subfigure[DIVE~\cite{fakhry2016deep}]{
\includegraphics[width=0.18\textwidth]{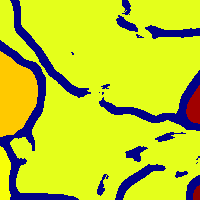}}
\subfigure[Ours]{
\includegraphics[width=0.18\textwidth]{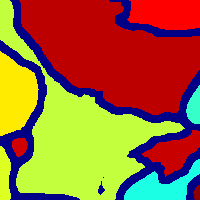}}
\caption{Illustration of the importance of topological correctness in a neuron image segmentation task. The goal of this task is to segment membranes that partition the image into regions corresponding to neurons. \textbf{(a)} an input neuron image. \textbf{(b)} ground truth segmentation of the membranes (dark blue) and the result neuron regions. \textbf{(c)} result of a baseline method without topological guarantee~\cite{fakhry2016deep}. Small pixel-wise errors lead to broken membranes, resulting in the merging of many neurons into one. \textbf{(d)} Our method produces the correct topology and the correct partitioning of neurons~\cite{hu2019topology}.}
\label{fig:teaser1}

\end{figure*}

One example is neuron image segmentation. When we do segmentation for neuron images, the goal of this task is to segment membranes that partition the image into regions corresponding to neurons. As illustrated in Fig.~\ref{fig:teaser1}(c), though the method achieves good enough performance in terms of pixel accuracy, it makes errors at some connection parts, resulting in the wrong partition of some regions. These wrong partitions will definitely affect the downstream functionality analysis of the neurons. 

At the initial step, with the help of topological data analysis, a novel method based on persistent homology is proposed to tackle structural accuracy directly. The proposed topology-preserving loss function is differentiable and is easy to be incorporated into deep neural networks. Also, we extend the proposed method to 3D cases. The proposed method achieves much better performance in terms of several topology-relevant metrics, e.g., the Adjusted Rand Index and the Variation of Information. We conduct experiments on several natural and medical image datasets.  We discuss the approach and present our results for this work in Chapter~\ref{chapter:topoloss}. 

Persistent homology is a powerful tool to analyze the geometric properties of low dimensional data, including images, which can be regarded as 2-dimensional data. Encouraged by the promising performances on the segmentation tasks, we further explore the power of the topological loss in a quite different context, i.e., trojan detection. More specifically, we propose a novel target-label-agnostic reverse engineering method. First, to improve the quality of the recovered triggers, we need a prior that can localize the triggers, but in a flexible manner. We, therefore, propose to enforce a \emph{topological prior} to the optimization process of reverse engineering, i.e., the recovered trigger should have fewer connected components. This prior is implemented through a topological loss introduced in Chapter~\ref{chapter:topoloss}. It allows the recovered trigger to have arbitrary shape and size. Meanwhile, it ensures the trigger is not scattered and is reasonably localized.  We discuss the approach and present our results for this work in Chapter~\ref{chapter:trojan}. 

\begin{figure}[ht]
  \centering
\subfigure[Original image]{
   \includegraphics[width=.25\textwidth]{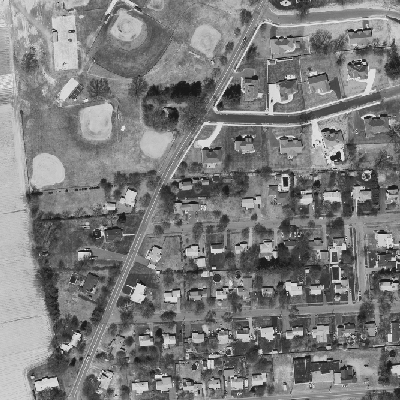}}
\subfigure[GT mask]{
     \includegraphics[width=.25\textwidth]{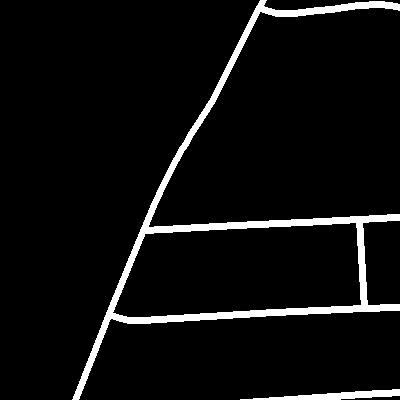}}
\subfigure[Likelihood map]{
     \includegraphics[width=.25\textwidth]{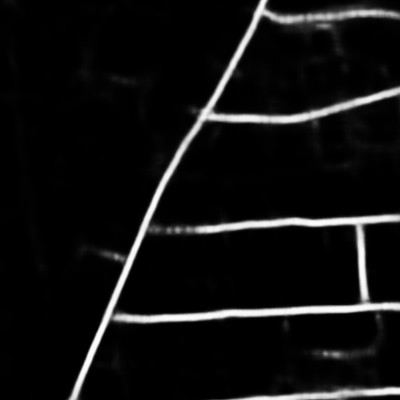}}
\subfigure[Thresholded mask]{
    \includegraphics[width=.25\textwidth]{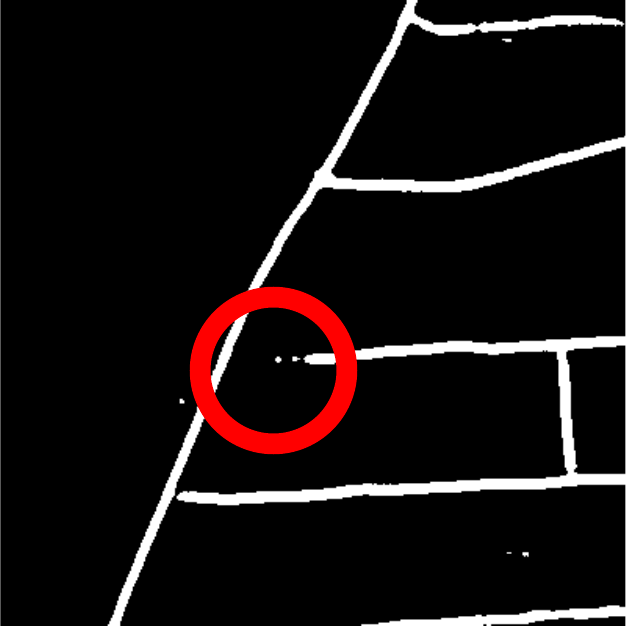}}
\subfigure[TopoNet]{
   \includegraphics[width=.25\textwidth]{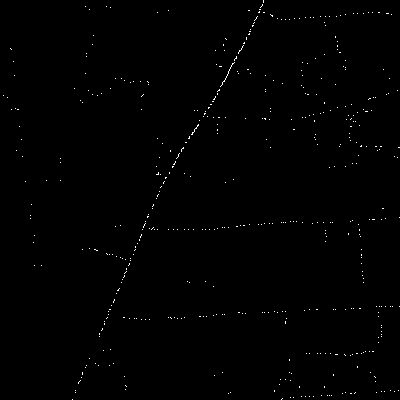}}
\subfigure[Warping]{
     \includegraphics[width=.25\textwidth]{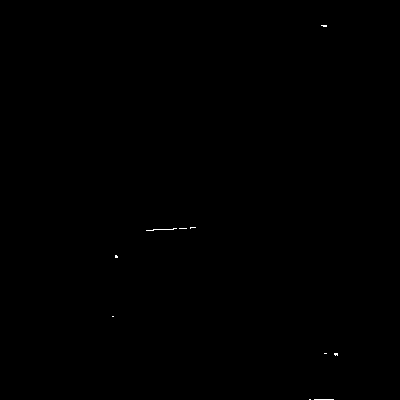}}
  \caption{Illustration of the critical points identified by different methods. \textbf{(a)}: Original image. \textbf{(b)}: GT mask. \textbf{(c)}: Predicted likelihood map. \textbf{(d)}: Segmentation (Thresholded mask from likelihood map). \textbf{(e)}: Critical points identified by~\cite{hu2019topology}. \textbf{(f)}: Critical points identified by our homotopy warping. Please zoom-in for better viewing. }
  \label{fig:cpt}
\end{figure}

Though TopoLoss achieves quite good performances, the identified critical points can be very noisy and often are not relevant to topological errors. See Fig.~\ref{fig:cpt} as an illustration. 
Moreover, the computation of persistent homology is expensive, making it difficult to evaluate the loss and gradient at every training iteration. Compared to the proposed method (Fig.~\ref{fig:cpt}(f)), the critical points identified from~\cite{hu2019topology} are very noisy and often are not relevant to the topological errors. 

Instead, we propose a novel \textit{homotopy warping loss}, which penalizes errors on topologically critical pixels. These locations are defined by homotopic warping of predicted and ground truth masks. The loss can be incorporated into the training of topology-preserving deep segmentation networks. We discuss the approach and present our results for this work in Chapter~\ref{chapter:warping}.

\section{Beyond Pixel-Wise Representation}
In both Chapter~\ref{chapter:topoloss} and Chapter~\ref{chapter:warping}, we introduce methods identifying a set of critical points of the likelihood function, e.g., saddles and extrema, as topologically critical locations for the neural network to memorize. However, only identifying a \emph{sparse set} of critical points at every epoch is inefficient in terms of training. 
Instead, our method identifies a much bigger set of critical locations at each epoch, i.e., 1D or 2D Morse skeletons (curves and patches). This is beneficial in both training efficiency and model performance. Extending the critical location sets from points to 1D curves and 2D patches makes it much more efficient in training. Furthermore, by focusing on more critical locations early, our method is more likely to escape poor local minima of the loss landscape. Thus it achieves better topological accuracy than TopoLoss. The shorter training time may also contribute to better stability of the SGD algorithm, and thus better test accuracy~\cite{hardt2016train}.

However, these loss-based methods are still not ideal. They are based on a standard segmentation network, and thus \emph{only learn pixel-wise feature representations}. 
This causes several issues. First, a standard segmentation network makes pixel-wise predictions. Thus, at the inference stage, topological errors, e.g.~broken connections, can still happen, even though they may be mitigated by the topology-inspired losses.
Another issue is in uncertainty estimation, i.e., estimating how certain a segmentation network is at different locations. Uncertainty maps can direct the focus of human annotators for efficient proofreading. However, for fine-scale structures, the existing pixel-wise uncertainty map is not effective.

To fundamentally address these issues, we propose to directly model and reason about the structures. In this paper, we propose \emph{the first deep learning based method that directly learns the topological/structural representation of images}. 
To move from pixel space to structure space, we apply the classic discrete Morse theory~\cite{milnor1963morse,forman2002user, barannikov1994framed} to decompose an image into a Morse complex, consisting of structural elements like branches, patches, etc. These structural elements are hypothetical structures one can infer from the input image.
Their combinations constitute a space of structures arising from the input image. We discuss the approach and present our results for this work in Chapter~\ref{chapter:uncertainty}.

\section{Outline}
The rest of this dissertation is organized as follows: In Chapter~\ref{chapter:related}, we review different works related to the research of this dissertation. In Chapter~\ref{chapter:topoloss}, Chapter~\ref{chapter:warping} and Chapter~\ref{chapter:dmt}, we introduce different methods to learn to segment with correct topology. In Chapter~\ref{chapter:trojan}, we explore the application of topological prior in the trojan detection context. In Chapter~\ref{chapter:uncertainty}, we introduce the first deep learning based method that directly learns the topological/structural representation of images. 
Chapter~\ref{chapter:conclusion} is the conclusion and future work.

%% file: related.tex
\chapter{Related Work}
\label{chapter:related}
In this chapter, we review the existing works relevant to our research which will be presented in Chapters~\ref{chapter:topoloss} through~\ref{chapter:uncertainty}. 

\section{Topological Data Analysis and Persistent Homology}
Topological data analysis (TDA) is a field in which one analyzes datasets using topological tools such as persistent homology~\cite{edelsbrunner2010computational, edelsbrunner2000topological, chazal2012structure, fasy2014confidence,carlsson2005persistence, carriere2019general, rosenfeld1979digital}. The theory has been applied to different applications~\cite{wu2017optimal,kwitt2015statistical,wong2016kernel,chazal2013persistence,ni2017composing,adams2017persistence,bubenik2015statistical,varshney2015persistent, hu2019topology, hu2021topology,zhao2020persistence,yan2021link,wu2020topological, kulp2015ventricular, chazal2016structure}. 

With the advent of deep learning, some works have tried to incorporate topological information into deep neural networks, and the differentiable property of persistent homology makes it possible. The main idea is that the persistence diagram/barcodes can capture all the topological changes, and it is differentiable to the original data. \cite{hu2019topology} first propose a topological loss to learn to segment images with correct topology, by matching persistence diagrams in a supervised manner. 
Similarly, \cite{clough2020topological} use the persistence barcodes to enforce a given topological prior to the target object. These methods achieve better results, especially in structural accuracy. Persistent-homology-based losses have been applied to other imaging \cite{abousamra2021localization,wang2020topogan} and learning problems~\cite{hofer2019connectivity,hofer2020graph,carriere2020perslay,chen2019topological}. 

The aforementioned methods use topological priors in supervised learning tasks (namely, segmentation). Instead, in this work, we propose to leverage the topological prior in an unsupervised setting; we use a topological prior for the reverse engineering pipeline to reduce the search space of triggers and enforce the recovered triggers to have fewer connected components. 

\section{Attention Mechanism}
Attention modules model relationships between pixels/channels/feature maps and have been widely applied in both vision and natural language processing tasks~\cite{lin2016efficient,lin2017structured,shen2018disan}. Specifically,  the self-attention mechanism~\cite{vaswani2017attention} is proposed to draw global dependencies of inputs and has been used in machine translation tasks. \cite{zhang2019self} tries to learn a better image generator via a self-attention mechanism. \cite{wang2018non} mainly explores the effectiveness of non-local operation, which is similar to the self-attention mechanism. \cite{zhao2018psanet} learns an attention map to aggregate contextual information for each individual point for scene parsing.


\section{Deep Image Segmentation} 

Deep learning methods (CNNs) have achieved satisfying performances for image segmentation~\cite{long2015fully,chen2014semantic,chen2018deeplab,chen2017rethinking,noh2015learning,ronneberger2015u,ciresan2012deep,funke2018large,kirillov2020pointrend,shen2017multi,shuai2017scene,dollar2006supervised, maire2008using, cour2005spectral, farabet2012learning, pinheiro2014recurrent, milletari2016v, sudre2017generalised, zhao2019region, farabet2013learning, nnUNet, delong2009globally, kappes2016higher, chen2018masklab, dai2016instance, li2017fully, huang2019mask, liu2018path, kirillov2017instancecut, bai2017deep, liang2017proposal}. By replacing fully connected layer with fully convolutional layers, FCN~\cite{long2015fully} transforms a classification CNNs (e.g., AlexNet~\cite{krizhevsky2012imagenet,chen2016combining}, VGG~\cite{simonyan2014very}, or ResNet~\cite{he2016deep}) to fully-convolutional neural networks. In this way, FCN successfully transfers the success of image classification~\cite{krizhevsky2012imagenet,simonyan2014very,szegedy2015going} to dense prediction/image segmentation. Instead of using Conditional Random Field (CRF) as post-processing, Deeplab (v1-v2)~\cite{chen2014semantic,chen2018deeplab} methods add another fully connected CRF after the last CNN layer to make use of global information. Moreover, Deeplab v3~\cite{chen2017rethinking} introduces dilated/atrous convolution to increase the receptive field and make better use of context information to achieve better performance.

Besides the methods mentioned above, UNet~\cite{ronneberger2015u} has also been one of the most popular methods for image segmentation, especially for images with fine structures. UNet architecture is based on FCN with two major modifications: 1) Similar to the encoder-decoder, UNet is symmetric. The output is the same size as input images, thus suitable for dense prediction/image segmentation, and 2) Skip connections between downsampling and upsampling paths. 
The skip connections of UNet are able to combine low level/local information with high level/global information, resulting in better segmentation performance. 

We also note that many deep learning techniques have been proposed to ensure the segmentation output preserves details, and thus preserves topology implicitly~\cite{badrinarayanan2017segnet, ding2019boundary, kervadec2019boundary, karimi2019reducing}.

Though obtaining satisfying pixel performances, these methods are still prone to structural/topological errors, as they are usually optimized via pixel-wise loss functions, such as mean-square-error loss (MSE) and cross-entropy loss.

\section{Topology-Aware Deep Image Segmentation}
Beyond pixel-wise accuracy, researchers have tried to segment with the correct shape/topology/geometry~\cite{chan2018topology, xue2019shape, chen2019learning, bentaieb2016topology, reddy2019brain, haralick1987image}.

One of the closest methods to ours is by Mosinska \etal~\cite{mosinska2018beyond}, which also proposes a topology-aware loss. Instead of actually computing and comparing the topology, their approach uses the response of selected filters from a pretrained VGG19 network to construct the loss. These filters prefer elongated shapes and thus alleviate the broken connection issue. But this method is hard to generalize to more complex settings with connections of arbitrary shapes. Furthermore, even if this method achieves zero loss, its segmentation is not guaranteed to be topologically correct.

Another closely related to our method is recent works on persistent-homology-based losses~\cite{clough2020topological}. These methods identify a set of critical points of the likelihood function, e.g., saddles and extrema, as topologically critical locations for the neural network to memorize. However, only identifying a \emph{sparse set} of critical points at every epoch is inefficient in terms of training. 
Instead, our method identifies a much bigger set of critical locations at each epoch, i.e., 1D or 2D Morse skeletons (curves and patches).
This is beneficial in both training efficiency and model performance.
Extending the critical location sets from points to 1D curves and 2D patches makes it much more efficient in training. 

Different ideas have been proposed to capture fine details of objects, mostly revolving around deconvolution and upsampling~\cite{long2015fully,chen2014semantic,chen2018deeplab,chen2017rethinking,noh2015learning,ronneberger2015u}.
However, these methods focus on the prediction accuracy of individual pixels and are intrinsically topology-agnostic.
Topological constraints, e.g., connectivity and loop-freeness, have been incorporated into variational~\cite{han2003topology,le2008self,sundaramoorthi2007global,segonne2008active,wu2017optimal,gao2013segmenting} and MRF/CRF-based segmentation methods~\cite{vicente2008graph,nowozin2009global,zeng2008topology,chen2011enforcing,andres2011probabilistic,stuhmer2013tree,oswald2014generalized,estrada2014tree}. 
However, these methods focus on enforcing topological constraints in the inference stage, while the trained model is agnostic of the topological prior. 
In neuron image segmentation, some methods~\cite{funke2018large,briggman2009maximin} directly find an optimal partition of the image into neurons and thus avoid segmenting membranes. These methods cannot be generalized to other structures, e.g., vessels, cracks, and roads.  

Other methods indirectly preserve topology by enhancing the curvilinear structures.
Several methods extract the skeletons of the masks and penalize heavily on pixels of the skeletons. This ensures the prediction is correct along the skeletons and thus is likely correct in topology.
clDice~\cite{shit2021cldice} extracts the skeleton through min/max-pooling operations over the likelihood map.  

One may also use topological constraints as postprocessing steps once we have the predicted likelihood maps~\cite{han2003topology,le2008self,sundaramoorthi2007global,segonne2008active,wu2017optimal,gao2013segmenting,vicente2008graph,nowozin2009global,zeng2008topology,chen2011enforcing,andres2011probabilistic,stuhmer2013tree,oswald2014generalized,estrada2014tree}. Compared to end-to-end methods, postprocessing methods usually contain self-defined parameters or hand-crafted features, making it difficult to generalize to different situations.

For completeness, we also refer to other existing works on topological features and their applications~\cite{adams2017persistence,reininghaus2015stable,kusano2016persistence,carriere2017sliced,ni2017composing,chen2016clustering,wu2017optimal, chen2010topology, chazal2009proximity}. 
In graphics, the topological similarity was used to simplify and align shapes~\cite{poulenard2018topological}.
Chen et al.~\cite{chen2019topological} proposed a topological regularizer to simplify the decision boundary of a classifier.
Deep neural networks have also been proposed to learn from topological features directly extracted from data~\cite{hofer2017deep,carriere2019general}. As for deep neural networks, Hofer \etal~\cite{hofer2017deep} proposed a CNN-based topological classifier. This method directly extracts topological information from an input image/shape/graph as input for CNN, hence cannot generate segmentations that preserve topological priors learned from the training set.

Persistent-homology-inspired objective functions have been proposed for graphics~\cite{poulenard2018topological} machine learning~\cite{chen2019topological,hofer2019connectivity}. 
Discrete Morse theory has been used to identify skeleton structures from images; e.g.,~\cite{DRS15,RWS11,WWL15}. The resulting 1D Morse structure has been used to enhance neural network architecture: e.g., in~\cite{DWW19} it is used for both pre- and post-process images, while in~\cite{banerjee2020semantic}, the 1D Morse skeleton is used as a topological prior (part of the input) for an encoder-decoder deep network for semantic segmentation of microscopic neuroanatomical data. Our work, in contrast, uses Morse structures (beyond 1D) of the output to strengthen the global structural signal more explicitly in an end-to-end training of a network.

Specific to neuron image segmentation, some methods~\cite{funke2018large,briggman2009maximin,januszewski2018high,uzunbas2016efficient,ye2019diverse} directly find neuron regions instead of their boundary/membranes. These methods cannot be generalized to other types of data such as satellite images, retinal images, vessel images, etc.

Additionally, the discrete Morse complex has been used for image analysis, but only as a preprocessing step~\cite{DRS15,RWS11,WWL15,DWW19} 
or as a conditional input of a neural network~\cite{banerjee2020semantic}.

\section{Trojan Detection}
Many Trojan detection methods~\cite{jang2019need, gu2017badnets, zhang2020cassandra, kurakin2016adversarial, szegedy2013intriguing, chen2017targeted, xiang2020revealing, liu2018fine, bajcsy2021baseline} have been proposed recently.
Some focus on detecting poisoned inputs via anomaly detection~\cite{chou2020sentinet, gao2019strip, liu2017neural, ma2019nic}. For example, SentiNet~\cite{chou2020sentinet} tries to identify adversarial inputs and uses the behaviors of these adversarial inputs to detect Trojaned models. Others focus on analyzing the behaviors of the trained models~\cite{chen2019detecting, guo2019tabor, shen2021backdoor, sun2020poisoned}. Specifically, \cite{chen2019detecting} propose the Activation Clustering (AC) methodology to analyze the activations of neural networks to determine if a model has been poisoned or not.

While early works require all training data to detect Trojans~\cite{chen2019detecting, gao2019strip, tran2018spectral}, recent approaches have been focusing on a more realistic setting -- when one has limited access to the training data.
A particularly promising direction is reverse engineering approaches, which recover Trojan triggers with only a few clean samples. Neural cleanse (NC)~\cite{wang2019neural} develops a Trojan detection method by identifying if there is a trigger that would produce misclassified results when added to an input. However, as pointed out by~\cite{guo2019tabor}, NC becomes futile when triggers vary in terms of size, shape, and location. 

Since NC, different approaches have been proposed, extending the reverse engineering idea.
Using a conditional generative model, DeepInspect~\cite{chen2019deepinspect} learns the probability distribution of potential triggers from a model of interest. \cite{kolouri2020universal} propose to learn universal patterns that change predictions of the model (called Universal Litmus Patterns (ULPs)). The method is efficient as it only involves forward passes through a CNN and avoids backpropagation. ABS~\cite{liu2019abs} analyzes inner neuron behaviors by measuring how extra stimulation can change the network's prediction. 
\cite{wang2020practical} propose a data-limited TrojanNet detector (TND) by comparing the impact of the per-sample attack and universal attack.
\cite{guo2019tabor} cast Trojan detection as a non-convex optimization problem and it is solved by optimizing an objective function. \cite{huster2021top} solve the problem by observing that, compared with clean models, adversarial perturbations transfer from image to image more readily in poisoned models. \cite{zheng2021topological} inspect neural network structure using persistent homology and identify structural cues differentiating Trojaned and clean models.

Existing methods are generally demanding training data access, neural network architectures, types of triggers, target class, etc. This limits their deployment to real-world applications. As for reverse engineering approaches, it remains challenging, if not entirely infeasible, to recover the true triggers. We propose a novel reverse engineering approach that can recover the triggers with high quality using the novel diversity and topological prior. Our method shares the common benefit of reverse engineering methods; it only needs a few clean input images per model. Meanwhile, our approach is agnostic of model architectures, trigger types, and target labels.

\section{Segmentation Uncertainty} 
Uncertainty estimation has been the focus of research in recent years~\cite{graves2011practical, gal2016dropout, lakshminarayanan2017simple, moon2020confidence, guzman2012multiple, lee2015m, lee2016stochastic, nguyen2015deep}. However, most existing work focuses on the classification problem. In terms of image segmentation, the research is still relatively limited. Some existing methods directly apply classification uncertainty to individual pixels, e.g., dropout~\cite{kendall2015bayesian,kendall2017uncertainties}. This, however, is not taking into consideration the image structures. Several methods estimate the uncertainty by generating an ensemble of segmentation~\cite{lakshminarayanan2017simple} or using multi-heads~\cite{rupprecht2017learning,ilg2018uncertainty}. Notably, Probabilistic-UNet~\cite{kohl2018probabilistic} learns a distribution over the latent space and then samples over the latent space to produce segmentation samples. When it comes down to uncertainty, however, these methods can still only generate a pixel-wise uncertainty map, using the frequency of appearance of each pixel in the sample segmentations. These methods are fundamentally different from ours, which makes predictions on the structures.

%% file: topoloss.tex
\chapter{Topology-Preserving Deep Image Segmentation via TopoLoss}
\label{chapter:topoloss}

In this chapter,
we introduce our first try to learn to segment with correct topology by leveraging the theory of persistent homology. We propose a differentiable loss that can be incorporated into any segmentation backbones to improve the topology-aware segmentation accuracy.

\section{Introduction}
\begin{figure*}[ht]
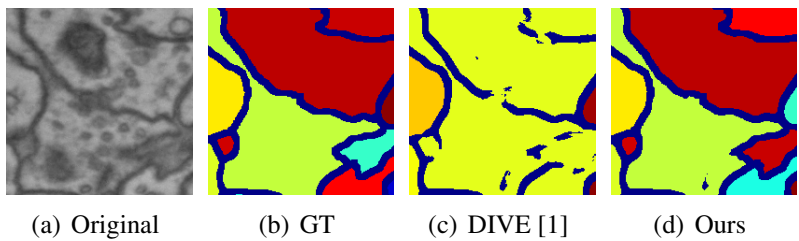

\centering 
\subfigure[Original]{
\includegraphics[width=0.18\textwidth]{topoloss/patch2_ori.png}}
\subfigure[GT]{
\includegraphics[width=0.18\textwidth]{topoloss/gt_color_revised2.png}}
\subfigure[DIVE~\cite{fakhry2016deep}]{
\includegraphics[width=0.18\textwidth]{topoloss/a1_color_revised2.png}}
\subfigure[Ours]{
\includegraphics[width=0.18\textwidth]{topoloss/our_color_revised2.png}}
\caption{ Illustration of the importance of topological correctness in a neuron image segmentation task. The goal of this task is to segment membranes that partition the image into regions corresponding to neurons. \textbf{(a)} an input neuron image. \textbf{(b)} ground truth segmentation of the membranes (dark blue) and the result neuron regions. \textbf{(c)} result of a baseline method without topological guarantee~\cite{fakhry2016deep}. Small pixel-wise errors lead to broken membranes, resulting in the merging of many neurons into one. \textbf{(d)} Our method produces the correct topology and the correct partitioning of neurons.}
\label{fig:teaser}
\end{figure*}

Image segmentation, i.e., assigning  labels to all pixels of an input image, is crucial in many computer vision tasks. 
State-of-the-art deep segmentation methods~\cite{long2015fully,he2017mask,chen2014semantic,chen2018deeplab,chen2017rethinking} learn high quality feature representations through an end-to-end trained deep network and achieve satisfactory per-pixel accuracy. However, these segmentation algorithms are still prone to make errors on fine-scale structures, such as small object instances, instances with multiple connected components, and thin connections. These fine-scale structures may be crucial in analyzing the \emph{functionality} of the objects. For example, accurate extraction of thin parts such as ropes and handles is crucial in planning robot actions, e.g., dragging or grasping. In biomedical images, correct delineation of thin objects such as neuron membranes and vessels is crucial in providing accurate morphological and structural quantification of the underlying system. A broken connection or a missing component may only induce marginal per-pixel error but can cause catastrophic functional mistakes. 
See Fig.~\ref{fig:teaser} for an example.

We propose a novel deep segmentation method that \emph{learns to segment with correct topology}. In particular, we propose a \emph{topological loss} that enforces the segmentation results to have the same topology as the ground truth, i.e., having the same \emph{Betti number} (number of connected components and handles). A neural network trained with such loss will achieve high topological fidelity without sacrificing per-pixel accuracy. The main challenge in designing such loss is that topological information, namely, Betti numbers, are discrete values. We need a continuous-valued measurement of the topological similarity between a prediction and the ground truth, and such measurement needs to be differentiable in order to backpropagate through the network. 

To this end, we propose to use theory from computational topology~\cite{edelsbrunner2010computational}, which summarizes the topological information from a continuous-valued function (in our case, the likelihood function $f$ is predicted by a neural network). Instead of acquiring the segmentation by thresholding $f$ at 0.5 and inspecting its topology, \emph{persistent homology}~\cite{edelsbrunner2010computational,edelsbrunner2000topological,zomorodian2005computing} captures topological information carried by $f$ over all possible thresholds. This provides a unified, differentiable approach of measuring the topological similarity between $f$ and the ground truth, called the \emph{topological loss}.
We derive the gradient of the loss so that the network predicting $f$ can be optimized accordingly. We focus on 1-dimensional topology (connections) in 2-dimensional images. And we will focus on 2-dimensional topology (number of voids) when dealing with 3-dimensional images.

\emph{Our method is the first end-to-end deep segmentation network with guaranteed topological correctness.} We show that when the topological loss is decreased to zero, the segmentation is guaranteed to be topologically correct, i.e., have identical topology as the ground truth. Our method is empirically validated by comparing with state-of-the-art on natural and biomedical datasets with fine-scaled structures. It achieves superior performance on metrics that encourage structural accuracy. In particular, our method significantly outperforms others on the Betti number error which exactly measures the topological accuracy. Fig.~\ref{fig:teaser} shows a qualitative result. 

Our method shows how topological computation and deep learning can be mutually beneficial. While our method empowers deep nets with advanced topological constraints, it is also a powerful approach to topological analysis; the observed function is now learned with a highly nonlinear deep network. This enables topology to be estimated based on a semantically informed and denoised observation. 

\section{Method}
Our method achieves both per-pixel accuracy and topological correctness by training a deep neural network with a new topological loss, $L_{topo}(f,g)$. Here $f$ is the likelihood map predicted by the network and $g$ is the ground truth. The loss function on each training image is a weighted sum of the per-pixel cross-entropy loss, $L_{bce}$, and the topological loss:
\begin{equation}
    L(f,g)= L_{bce}(f,g)+ \lambda L_{topo}(f,g),
\label{total_loss}
\end{equation}
in which $\lambda$ controls the weight of the topological loss. We assume a binary segmentation task. Thus, there is one single likelihood function $f$, whose value ranges between 0 and 1. 

\begin{figure*}[ht]
\vskip -5pt
  \centering
  \noindent\makebox[\textwidth][c] {
    \includegraphics[width=0.6\paperwidth]{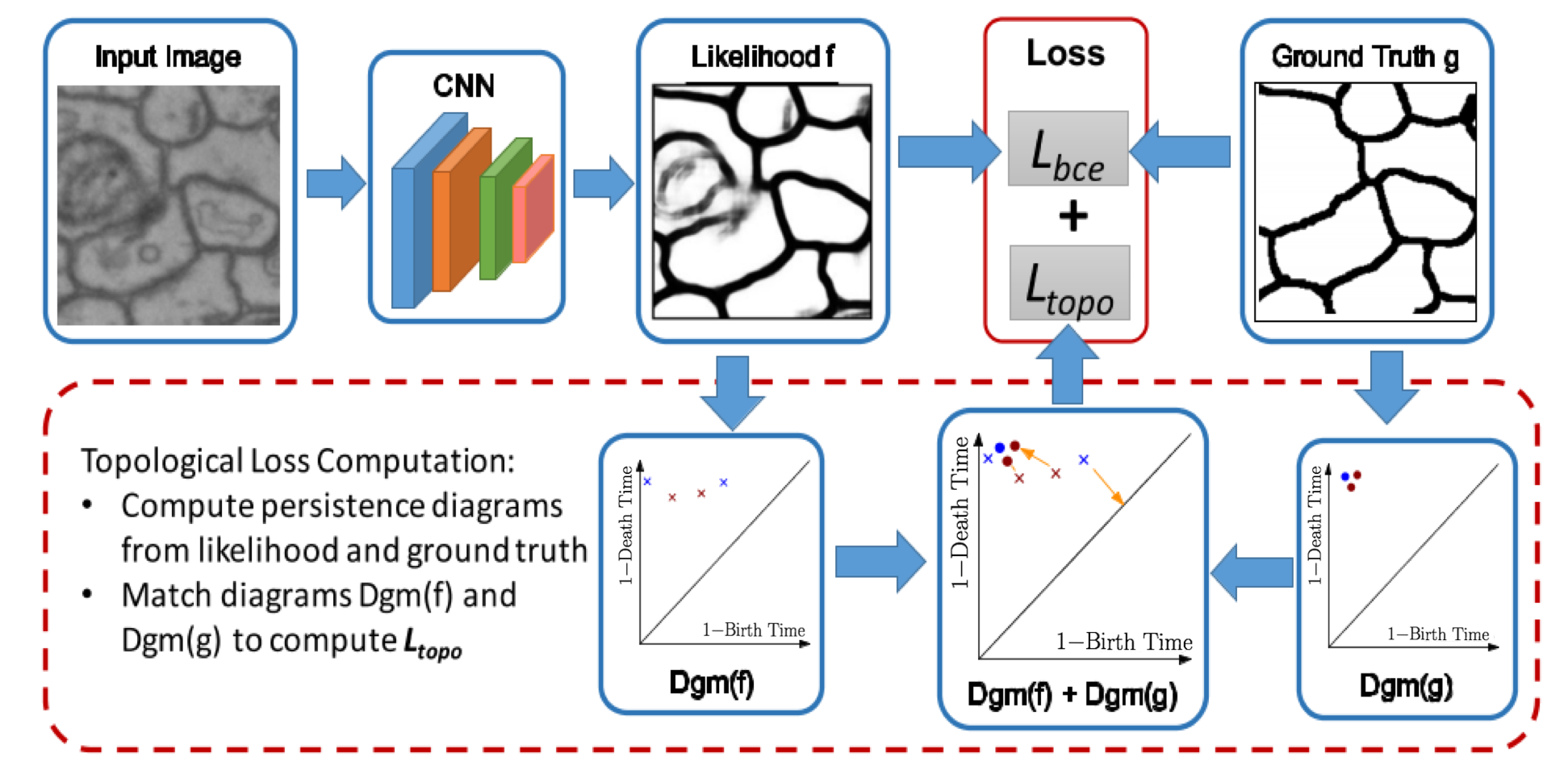}}
    \caption{An overview of our method.}
      \label{fig:architecture}
     \vskip -10pt
\end{figure*}

 In Sec.~\ref{topo_ph}, we introduce the mathematical foundation of topology and how to measure the topology of a likelihood map robustly using persistent homology. In Sec.~\ref{loss_gradient}, we formalize the topological loss as the difference between the persistent homology of $f$ and $g$. We derive the gradient of the loss and prove its correctness.  In Sec.~\ref{details} we explain how to incorporate the loss into the training of a neural network. Although we fix one architecture in experiments, our method is general and can use any neural network that provides pixel-wise prediction. Fig.~\ref{fig:architecture} illustrates the overview of our method.
 
\subsection{Topology and Persistent Homology}
\label{topo_ph}
\begin{figure*}[ht]
\centering 
\subfigure[Segmentation]{\label{fig:2a} 
\includegraphics[width=0.18\textwidth]{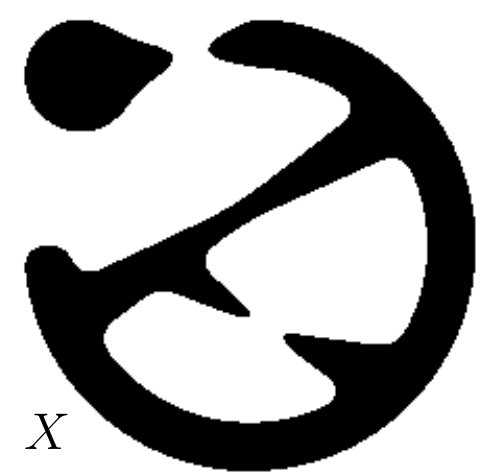}}
\subfigure[GT]{\label{fig:2b} 
\includegraphics[width=0.18\textwidth]{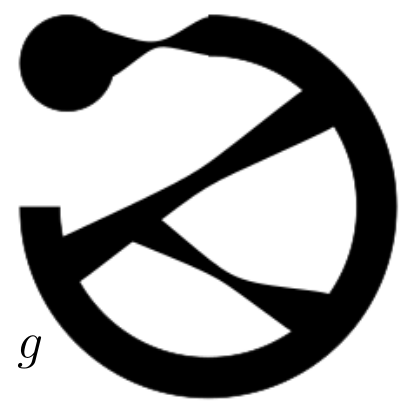}}
\subfigure[Likelihood $f$]{\label{fig:2c} 
\includegraphics[width=0.27\textwidth]{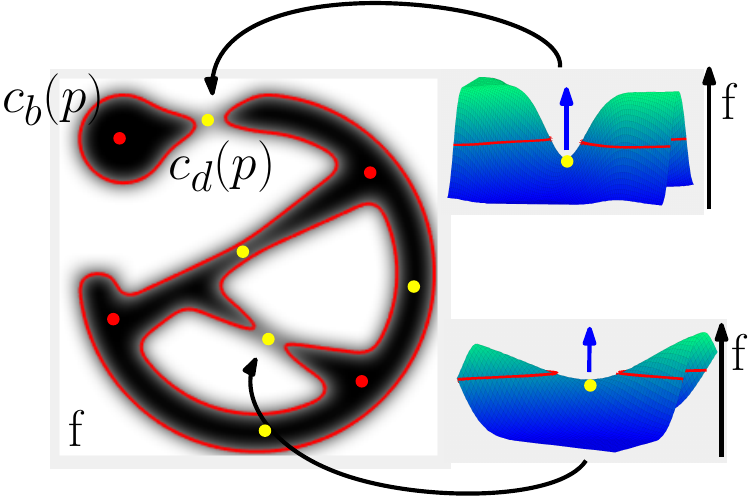}}
\subfigure[Likelihood $f'$]{\label{fig:2d} 
\includegraphics[width=0.27\textwidth]{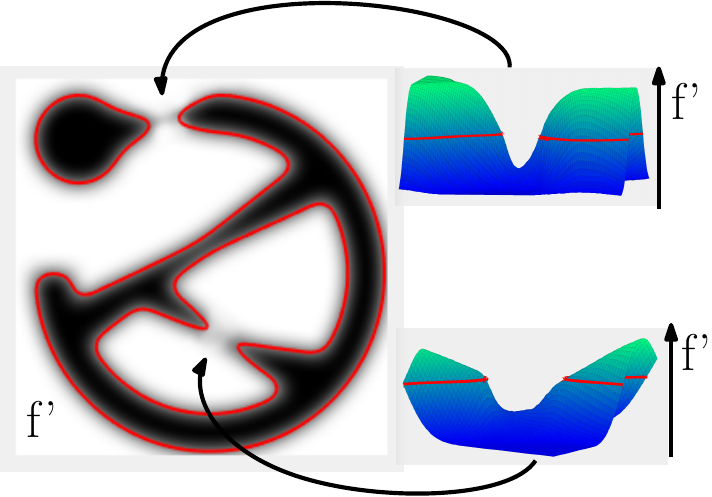}}
\caption{Illustration of topology and topology of a likelihood. For visualization purposes, the higher the function values are, the darker the area is. \textbf{(a)} an example segmentation $X$ with two connected components and one handle. \textbf{(b)} The ground truth with one connected component and two handles. It can also be viewed as a binary valued function $g$. \textbf{(c)} a likelihood map $f$ whose segmentation (bounded by the red curve) is $X$. The landscape views near the broken bridge and handle are drawn. Critical points are highlighted in the segmentation.
\textbf{(d)} another likelihood map $f'$ with the same segmentation as $f$. But the landscape views reveal that $f'$ is worse than $f$ due to deeper gaps. }
\label{fig:similarity}
\end{figure*}

Given a continuous image domain, $\Omega \subseteq \mathbb{R}^2$ (e.g., a 2D rectangle), we study a likelihood map $f(x): \Omega \rightarrow \mathbb{R}$, which is predicted by a deep neural network (Fig.~\ref{fig:2c}).~\footnote{$f$ depends on the network parameter $\omega$, which will be optimized during training. For convenience, we only use $x$ as the argument of $f$.} Note that in practice, we only have samples of $f$ at all pixels. In such a case, we extend $f$ to the whole image domain $\Omega$ by linear interpolation. Therefore, $f$ is piecewise-linear and is controlled by values at all pixels. A segmentation, $X \subseteq \Omega$ (Fig.~\ref{fig:2a}), is calculated by thresholding $f$ at a given value $\alpha$ (often set to $0.5$). 

Given $X$, its $d$-dimension topological structure called a \textit{homology class}~\cite{edelsbrunner2010computational,munkres2018elements}, is an equivalence class of $d$-manifolds which can be deformed into each other within $X$.~\footnote{To be exact, a homology class is an equivalent class of cycles whose difference is the boundary of a $(d+1)$-dimensional patch.} In particular, 0-dim and 1-dim structures are connected components and handles, respectively. For example, in Fig.~\ref{fig:2a}, the segmentation $X$ has two connected components and one handle. Meanwhile, the ground truth (Fig.~\ref{fig:2b}) has one connected component and two handles. Given  $X$, we can compute the number of topological structures, called the \textit{Betti number}, and compare it with the topology of the ground truth.

However, simply comparing Betti numbers of $X$ and $g$ will result in a discrete-valued topological error function. To incorporate topological prior into deep neural networks, we need a continuous-valued function that can reveal subtle differences between similar structures. Fig.~\ref{fig:2c} and \ref{fig:2d} show two likelihood maps $f$ and $f'$ with identical segmentations, both with incorrect topology comparing with the ground truth $g$ (Fig.~\ref{fig:2b}). However, $f$ is preferable as we need much less effort to change it so that the thresholded segmentation $X$ has a correct topology. In particular, look closely to Fig.~\ref{fig:2c} and \ref{fig:2d} near the broken handles and view the landscape of the function. To restore the broken handle in Fig.~\ref{fig:2d}, we need to spend more effort to fill a much deeper gap than in Fig.~\ref{fig:2c}. The same situation happens near the missing bridge between the two connected components.

To capture such subtle structural differences between different likelihood maps, we need a holistic view. In particular, we use the theory of \emph{persistent homology}~\cite{edelsbrunner2000topological,edelsbrunner2010computational}.
Instead of choosing a fixed threshold, persistent homology theory captures all possible topological structures from all thresholds, and summarizes all these information in a concise format, called a \textit{persistence diagram}.

Fig.~\ref{fig:similarity} shows that only considering one threshold $\alpha = 0.5$ is insufficient. We consider thresholding the likelihood function with all possible thresholds. The thresholded results, 
$ f^{\alpha}:= \{x \in \Omega | f(x) \geq \alpha \}$ at different $\alpha$'s, constitute a filtration, i.e., a monotonically growing sequence induced by decreasing the threshold $\alpha: \varnothing \subseteq f^{\alpha_1} \subseteq f^{\alpha_2} \subseteq ... \subseteq f^{\alpha_n} = \Omega$, where $\alpha_1 \geq \alpha_2 \geq ... \geq \alpha_n$. As $\alpha$ decreases, the topology of $f^{\alpha}$ changes. Some new topological structures are born while existing ones are killed. When $\alpha<\alpha_n$, only one connected component survives and never gets killed. See Fig.~\ref{fig:a} and \ref{fig:d} for filtrations induced by the ground truth $g$ (as a binary-valued function) and the likelihood  $f$.

For a continuous-valued function $f$, its \textit{persistence diagram}, $\dgm(f)$, contains a finite number of dots in 2-dimensional plane, called \textit{persistent dots}. Each persistent dot $p\in \dgm(f)$ corresponds to a topological structure born and dies in the filtration. Denote by $\birth(p)$ and $\death(p)$ the birth and death time/threshold of the structure. For the connected component born at a global minimum and never dies, we say it dies at $\max_x f(x) = 1$. The coordinates of the dot $p$ in the diagram are $(1-\birth(p),1-\death(p))$.~\footnote{Unlike traditional setting, we use $1-\birth$ and $1-\death$ as the x and y axes, because we are using an upperstar filtration, i.e., using the superlevel set, and decreasing $\alpha$ value.}
 Fig.~\ref{fig:b} and \ref{fig:e} show the diagrams of $g$ and $f$, respectively. Instead of comparing discrete Betti numbers, we can use the information from persistence diagrams to compare a likelihood $f$ with the ground truth $g$ in terms of topology. 

To compute $\dgm(f)$, we use the classic algorithm~\cite{edelsbrunner2010computational,edelsbrunner2000topological} with an efficient implementation~\cite{chen2011persistent,wagner2012efficient}:
we first discretize an image patch into vertices (pixels), edges, and squares. Note we adopt a cubical complex discretization, which is more suitable for images. The adjacency relationship between these discretized elements and their likelihood function values are encoded in a boundary matrix, whose rows and columns correspond to vertices/edges/squares. 
The matrix is reduced using a modified Gaussian elimination algorithm. The pivoting entries of the reduced matrix correspond to all the dots in $\dgm(f)$. This algorithm is cubic to the matrix dimension, which is linear to the image size. 

\subsection{Topological Loss and its Gradient}
\label{loss_gradient}
\begin{figure*}[ht]
\centering 
\subfigure[Filtration induced by the ground truth function, $g$.]{\label{fig:a} 
\includegraphics[width=0.5\textwidth]{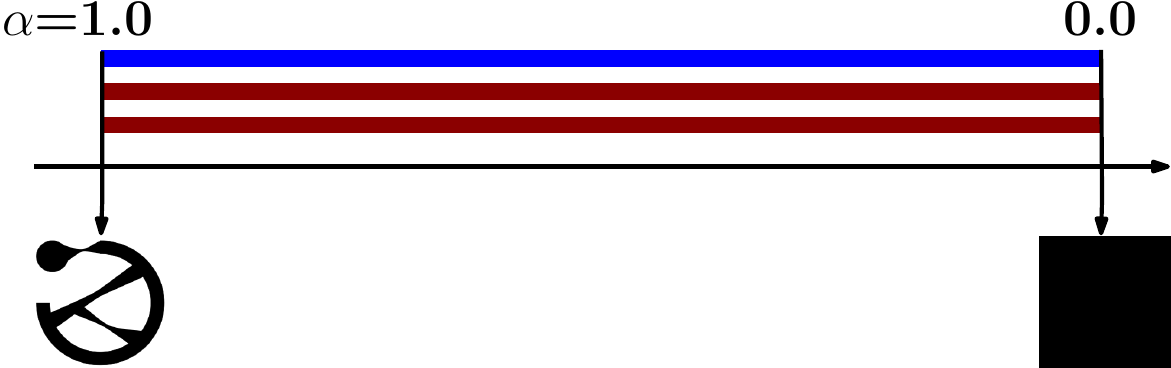}}
\subfigure[$\dgm(g)$]{\label{fig:b} 
\includegraphics[width=0.15\textwidth]{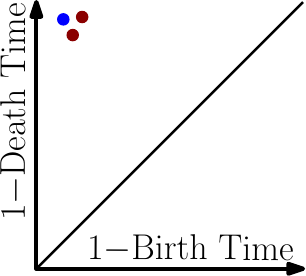}}
\subfigure[$\dgm(g)$+$\dgm(f)$]{\label{fig:c} 
\includegraphics[width=0.15\textwidth]{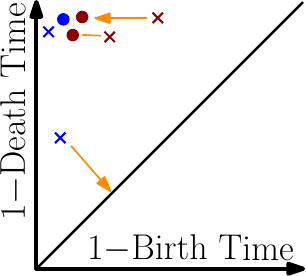}}
\subfigure[Filtration induced by the likelihood function, $f$.]{\label{fig:d} 
\includegraphics[width=0.5\textwidth]{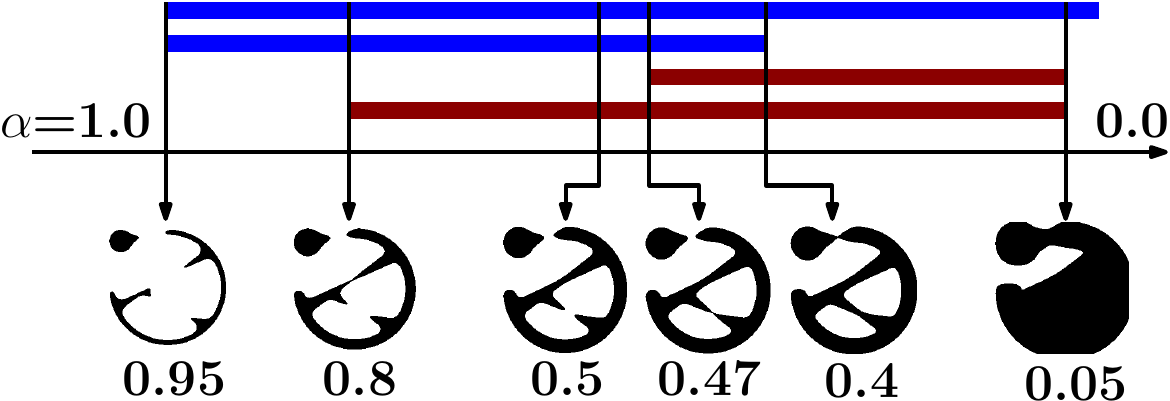}}
\subfigure[$\dgm(f)$]{\label{fig:e}
\includegraphics[width=0.15\textwidth]{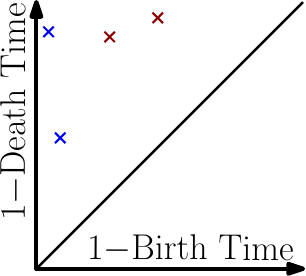}}
\subfigure[$\dgm(g)$+$\dgm({f'})$]{\label{fig:f} 
\includegraphics[width=0.15\textwidth]{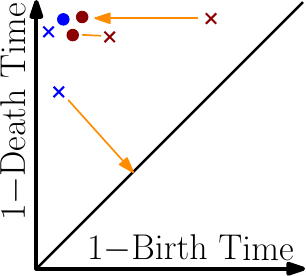}}
\vspace{-8pt}
\caption{An illustration of persistent homology. \textbf{Left} the filtrations on the ground truth function $g$ and the likelihood function $f$. The bars of blue and burgundy colors are connected components and handles respectively. \textbf{(a)} For $g$, all structures are born at $\alpha = 1.0$ and die at $\alpha = 0$. \textbf{(d)} For $f$, from left to right, the birth of two components, birth of the longer handle, segmentation at $\alpha = 0.5$, the birth of the shorter handle, death of the extra component, death of both handles. \textbf{(b)} and \textbf{(e)} the persistence diagrams of $g$ and $f$. \textbf{(c)} the overlay of the two diagrams. Orange arrows denote the matching between the persistent dots. The extra component (a blue cross) from the likelihood is matched to the diagonal line and will be removed if we move $\dgm(f)$ to $\dgm(g)$. \textbf{(f)} the overlay of the diagrams of $g$ and the worse likelihood $\dgm({f'})$. The matching is obviously more expensive.}
\label{fig:persistent}
\end{figure*}

We are now ready to formalize the topological loss, which measures the topological similarity between the likelihood $f$ and the ground truth $g$. We abuse the notation and also view $g$ as a binary valued function.
We use the dots in the persistence diagram of $f$ as they capture all possible topological structures $f$ potentially has. We slightly modify the \textit{Wasserstein distance} for persistence diagrams~\cite{cohen2010lipschitz, memoli2011gromov, carriere2017sliced}.
For persistence diagrams $\dgm(f)$ and $\dgm(g)$, we find the best one-to-one correspondence between the two sets of dots, and measure the total squared distance between them.~\footnote{To be exact, the matching needs to be done on separate dimensions. Dots of 0-dim structures (blue markers in Fig.~\ref{fig:b} and \ref{fig:e}) should be matched to the diagram of 0-dim structures. Dots of 1-dim structures (red markers in Fig.~\ref{fig:b} and \ref{fig:e}) should be matched to the diagram of 1-dim  structures.} 
An unmatched dot will be matched to the diagonal line.
Fig.~\ref{fig:c} shows the optimal matching of the diagrams of $g$ and $f$. Fig.~\ref{fig:f} shows the optimal matching of $\dgm(g)$ and $\dgm(f')$. The latter is clearly more expensive.

 The matching algorithm is as follows.
A total of $k$ (=Betti number) dots from ground truth ($\dgm(g)$) are at the upper-left corner $p_{ul}=(0,1)$, with $\birth(p_{ul})=1$ and $\death(p_{ul})=0$ (Fig.~\ref{fig:b}). In $\dgm(f)$, we find the $k$ dots closest to the corner $p_{ul}$ and match them to the ground truth dots. The remaining dots in $\dgm(f)$ are matched to the diagonal line. The algorithm computes and sorts the squared distances from all dots in $\dgm(f)$ to $p_{ul}$. The complexity is $O(n \log n)$, $n$ = the number of dots in $\dgm(f)$. 
In general, the state-of-the-art matches two arbitrary diagrams in $O(n^{3/2})$ time~\cite{kerber2017geometry}.

Let $\Gamma$ be the set of all possible bijections between $\dgm(f)$ and $\dgm(g)$.
The loss $L_{topo}(f,g)$ is: 
\begin{equation}
\label{topo_loss}
\begin{aligned}
     \min \limits_{\gamma \in \Gamma} \sum_{p \in \dgm(f)}||p-\gamma(p)||^2= 
    \sum_{p \in \dgm(f)}[\birth(p)-\birth(\gamma^*(p))]^2 \\
    +[\death(p)-\death(\gamma^*(p))]^2  
\end{aligned}
\end{equation}
where $\gamma^{*}$ is the optimal matching between two different point sets. 

Intuitively, this loss measures the minimal amount of necessary effort to modify the diagram of $\dgm(f)$ to $\dgm(g)$ by moving all dots toward their matches. Note there are more dots in $\dgm(f)$  (Fig.~\ref{fig:c}) than in $\dgm(g)$ (Fig.~\ref{fig:b});  there will usually be some noise in the predicted likelihood map. If a dot $p$ cannot be matched, we match it to its projection on the diagonal line, $\{(1-b,1-d)|b=d\}$. This means we consider it as noise that should be removed. 
The dots matched to the diagonal line correspond to small noisy components or noisy loops. These dots will be pushed to the diagonal. And their corresponding components/loops will be removed or merged with others. 

In this example, the extra connected component (a blue cross) in $\dgm(f)$ will be removed. For comparison, we also show in Fig.~\ref{fig:f} the matching between diagrams of the worse likelihood $f'$ and $g$. The cost of the matching is obviously higher, i.e., $L_{topo}(f',g) > L_{topo}(f,g)$. As a theoretical reassurance, it has been proven that this metric for diagrams is stable, and the loss function $L_{topo}(f,g)$ is Lipschitz with regard to the likelihood function $f$~\cite{cohen2007stability}. 

The following theorem guarantees that the topological loss, when minimized to zero, enforces the constraint that the segmentation has the same topology and the ground truth. 
\begin{theorem}[Topological Correctness]
\label{thm:correctness}
When the loss function $L_{topo}(f,g)$ is zero, the segmentation by thresholding $f$ at 0.5 has the same Betti number as $g$.
\end{theorem}
\myparagraph{Proof.} Assume $L_{topo}(f,g)$ is zero.
By Eq.~\ref{topo_loss},  $\dgm(f)$ and $\dgm(g)$ are matched perfectly, i.e.,  $p=\gamma^\ast(p), \forall p\in \dgm(f)$. The two diagrams are identical and have the same number of dots. 

Since $g$ is a binary-valued function, as we decrease the threshold $\alpha$ continuously, all topological
structures are created at $\alpha=1$. 
The number of topological structures (Betti number) of $g^\alpha$ for any $0<\alpha<1$ is the same as the number of dots in $\dgm(g)$. Note that for any $\alpha\in (0,1)$, $g^\alpha$ is the ground truth segmentation. Therefore, the Betti number of the ground truth is the number of dots in $\dgm(g)$. 
Similarly, for any $\alpha\in (0,1)$, the Betti number of $f^{\alpha}$ equals to the number of dots in $\dgm(f)$. Since the two diagrams $\dgm(f)$ and $\dgm(g)$ are identical, the Betti number of the segmentation $f^{0.5}$ is the same as the ground truth segmentation.~\footnote{Note that a more careful proof should be done for diagrams of 0- and 1-dimension separately.}

\myparagraph{Topological Gradient.} The loss function (Eq.~\ref{topo_loss}) depends on crucial thresholds at which topological changes happen, e.g., birth and death times of different dots in the diagram. These crucial thresholds are uniquely determined by the locations at which the topological changes happen. When the underlying function $f$ is differentiable, these crucial locations are exactly \emph{critical points}, i.e., points with zero gradients. In the training context, our likelihood function $f$ is a piecewise-linear function controlled by the neural network predictions at pixels. For such $f$, a critical point is always a pixel, since topological changes always happen at pixels. Denote by $\omega$ the neural network parameters. For each dot $p \in \dgm(f)$ , we denote by $c_b(p)$ and $c_d(p)$ the birth and death critical points of the corresponding topological structure (See Fig.~\ref{fig:2c} for examples).

Formally, we can show that the gradient of the topological loss $\nabla_\omega L_{topo}(f,g)$  is:
\begin{equation}
\label{gradient}
\begin{aligned}
 \sum_{p \in \dgm(f)}2[f(c_b(p))-\birth(\gamma^*(p))]\frac{\partial f(c_b(p))}{\partial \omega}   \\
    +2[f(c_d(p))-\death(\gamma^*(p))]\frac{\partial f(c_d(p))}{\partial \omega}  
\end{aligned}
\end{equation}
To see this, within a sufficiently small neighborhood of $f$, any other piecewise linear function will have the same super level set filtration as $f$. The critical points of each persistent dot in $\dgm(f)$ remain constant within such a small neighborhood. So does the optimal mapping $\gamma^*$. Therefore, the gradient can be straightforwardly computed based on the chain rule, as Eq.~\ref{gradient}. When function values at different vertices are the same, or when the matching is ambiguous, the gradient does not exist. However, these cases constitute a measure of zero subspace in the space of likelihood functions. In summary, $L_{topo}(f,g)$ is a piecewise differentiable loss function over the space of all possible likelihood functions $f$. 

\myparagraph{Intuition.} During training, we take the negative gradient, i.e., 
$-\nabla_ \omega L_{topo}(f,g)$. For each topological structure, the gradient descent step is pushing the corresponding dot $p \in \dgm(f)$ toward its match $\gamma^*(p) \in \dgm(g)$. These coordinates are the function values of the critical points $c_b(p)$ and $c_d(p)$. They are both moved closer to the matched persistent dot in $\dgm(g)$. We also show the negative gradient force in the landscape view of the function $f$ (blue arrow in Fig.~\ref{fig:2c}). Intuitively, forcing from the topological gradient will push the saddle points up so that the broken bridge gets connected.

\subsection{Training a Neural Network}
\label{details}
We present some crucial details of our training algorithm.
Although our method is architecture-agnostic, we select one architecture inspired by DIVE~\cite{fakhry2016deep}, which was designed for neuron image segmentation tasks. Our network contains six trainable weight layers, four convolutional layers, and two fully connected layers. The first, second, and fourth convolutional layers are followed by a single max pooling layer of size $2 \times 2$ and stride 2 by the end of the layer. Particularly, because of the computational complexity, we use a patch size of $65 \times 65$ during the training process.

We use small patches ($65 \times 65$) instead of big patches/whole images. The reason is twofold. First, the computation of topological information is relatively expensive. Second, the matching process between the persistence diagrams of predicted likelihood maps and ground truth can be quite difficult. For example, if the patch size is too big, there will be many persistent dots in $\dgm(g)$ and even more dots in $\dgm(g)$. The matching process is too complex and prone to errors. \emph{By focusing on smaller patches, we localize topological structures and fix them one by one.}

\myparagraph{Topology of Small Patches and Relative Homology.} 
The small patches ($65\times 65$) often only contain partial branching structures rather than closed loops. To have a meaningful topological measure on these small patches, we apply \textit{relative persistent homology} as a more localized approach for the computation of topological structures. Particularly, for each patch, we consider the topological structures relative to the boundary. It is equivalent to adding a black frame to the boundary and computing the topology to avoid trivial topological structures. As shown in the figure on the right, with the additional frame, a $Y$-shaped branching structure cropped within the patch will create two handles and be captured by persistent homology. 
\begin{wrapfigure}{r}{0.2\textwidth}
\centering 
    \includegraphics[width=0.2\textwidth]{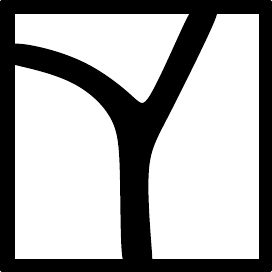}
\label{fig:padding}
\end{wrapfigure}

Training using this localized topological loss can be very efficient via random patch sampling. Specifically, we do not partition the image into patches. Instead, we randomly and densely sample patches that can overlap. As Theorem~\ref{thm:correctness} guarantees, Our loss enforces correct topology within each sampled patch. These overlaps between patches propagate correct topology everywhere. On the other hand, correct topology within a patch means the segmentation can be a deformation of the ground truth. But the deformation is constrained within the patch. The patch size controls the tolerable geometric deformation.
During training, even for the same patch, the diagram $\dgm(f)$, the critical pixels, and the gradients change. At each epoch, we resample patches and reevaluate their persistence diagrams, and the loss gradients. After computing topological gradients of all sampled patches from a mini-batch, we aggregate them for backpropagation.

\section{Experiments}

\subsection{Datasets}
\label{sec:dataset_topoloss}
We evaluate our method on six natural and biomedical datasets:
 \textbf{CREMI}~\footnote{https://cremi.org/}, \textbf{ISBI12}~\cite{arganda2015crowdsourcing}, \textbf{ISBI13}~\cite{arganda20133d},
\textbf{CrackTree}~\cite{zou2012cracktree}, \textbf{Mass.}~\cite{mnih2013machine} and  \textbf{DRIVE}~\cite{staal2004ridge}. 
\begin{itemize}
    \item \textbf{CREMI} contains 125 images of size 1250x1250. 
    \item \textbf{ISBI12}~\cite{arganda2015crowdsourcing} contains 30 images of size 512x512.
        \item \textbf{ISBI13}~\cite{arganda20133d} contains 100 images of size 1024x1024.
    \item \textbf{CrackTree}~\cite{zou2012cracktree} contains 206 images of cracks in road (resolution 600x800). 
    \item \textbf{Mass.}~\cite{mnih2013machine} has 1108 images from the Massachusetts Roads Dataset. The resolution is 1500x1500. 
    \item \textbf{DRIVE}~\cite{staal2004ridge} is a retinal vessel segmentation dataset with 20 images. The resolution is 584x565.
\end{itemize}
The first three are neuron image segmentation datasets. The task is to segment membranes and eventually partition the image into neuron regions. For all datasets, we use a three-fold cross-validation and report the mean performance over the validation set.

\subsection{Evaluation Metrics} 
\label{sec:metric_topoloss}
We use four different evaluation metrics.
\begin{itemize}
    \item \textbf{Pixel-wise accuracy} is the percentage of correctly classified pixels. 
    \item \textbf{Betti number error} directly compares the topology (number of handles) between the segmentation and the ground truth.~\footnote{Note we focus on 1-dimensional topology in evaluation and training as they are more crucial in practice.}
We randomly sample patches over the segmentation and report the average absolute difference between their Betti numbers and the corresponding ground truth patches.
\item \textbf{Adapted Rand Index (ARI)} is the maximal F-score of the foreground-restricted Rand index, a measure of similarity between two clusters. On this version of the Rand index, we exclude the zero component of the original labels (background pixels of the ground truth).
\item \textbf{Variation of Information (VOI)} is a measure of the distance between two clusterings. It is closely related to mutual information; indeed, it is a simple linear expression involving mutual information. 
\end{itemize}

The first one is pixel-wise evaluation metric, and the remaining three metrics are topology-relevant.

\subsection{Baselines} 
\label{sec:baseline_topoloss}
We compare the proposed method with three popular baselines.
\begin{itemize}
    \item \textbf{{DIVE}}~\cite{fakhry2016deep} is a state-of-the-art neural network that predicts the probability of every individual pixel in a given image being a membrane (border) pixel or not.
    \item  \textbf{{UNet}}~\cite{ronneberger2015u} is a popular image segmentation method trained with cross-entropy loss.
    \item  \textbf{{UNet-VGG}}~\cite{mosinska2018beyond} uses the response of selected filters from a pretrained CNN to construct the topology aware loss.
For all methods, we generate segmentations by thresholding the predicted likelihood maps at 0.5. 
\end{itemize}

\begin{table*}[ht]
\begin{center}
\small
\caption{Quantitative results for different models on several medical datasets.}
\label{table:medical}
\begin{tabular}{ccccccc}
Dataset & Method & Accuracy & ARI & VOI  & Betti Error\\
\hline
\multirow{5}{*}{\textbf{ISBI12}} &
\textbf{DIVE} & 0.9640  & 0.9434 & 1.235 & 3.187\\
~ & \textbf{UNet} & \textbf{0.9678} & 0.9338 & 1.367 & 2.785\\
~ & \textbf{UNet-VGG} & 0.9532 & 0.9312 & 0.983 &1.238\\
~ & \textbf{TopoNet} & 0.9626 & \textbf{0.9444} & \textbf{0.782}  &\textbf{0.429} \\
\hline
\multirow{5}{*}{\textbf{ISBI13}} &
\textbf{DIVE} & \textbf{0.9642} & 0.6923 & 2.790 & 3.875\\
~ & \textbf{UNet} & 0.9631 & 0.7031 &  2.583 &3.463\\
~ & \textbf{UNet-VGG} & 0.9578 & 0.7483 & 1.534 & 2.952\\
~ & \textbf{TopoNet} & 0.9569 & \textbf{0.8064} & \textbf{1.436}& \textbf{1.253}\\
\hline
\multirow{5}{*}{\textbf{CREMI}} &
\textbf{DIVE} & \textbf{0.9498} & 0.6532 & 2.513  & 4.378\\
~ & \textbf{UNet} & 0.9468 & 0.6723 & 2.346 & 3.016\\
~ & \textbf{UNet-VGG} & 0.9467 & 0.7853 & 1.623 & 1.973\\
~ & \textbf{TopoNet} & 0.9456 & \textbf{0.8083} & \textbf{1.462}  & \textbf{1.113}\\
\hline
\end{tabular}
\end{center}
\end{table*}

\begin{table*}[ht]
\centering
\begin{center}
\small
\caption{Quantitative results for different models on some other datasets.}
\label{table:ensemble}
\begin{tabular}{ccccccc}

Dataset & Method & Accuracy & ARI & VOI & Betti Error\\
\hline
\multirow{5}{*}{\textbf{DRIVE}} &
\textbf{DIVE} & \textbf{0.9549} & 0.8407 & 1.936 & 3.276 \\
~ & \textbf{UNet} & 0.9452 & 0.8343 &  1.975 & 3.643\\
~ & \textbf{UNet-VGG} & 0.9543 & 0.8870   & 1.167 & 2.784\\
 ~ & \textbf{TopoNet} & 0.9521 & \textbf{0.9024} & \textbf{1.083} & \textbf{1.076}\\
\hline
\multirow{5}{*}{\textbf{CrackTree}} &
\textbf{DIVE} & \textbf{0.9854}  & 0.8634 & 1.570   & 1.576\\
~ & \textbf{UNet} & 0.9821 & 0.8749 & 1.625  & 1.785\\
~ & \textbf{UNet-VGG} & 0.9833 & 0.8897 & 1.113  & 1.045\\
~ & \textbf{TopoNet} & 0.9826 & \textbf{0.9291} & \textbf{0.997} & \textbf{0.672} \\
\hline
\multirow{5}{*}{\textbf{Mass.}} &
\textbf{DIVE} & 0.9734 & 0.8201 & 2.368  & 3.598\\
~ & \textbf{UNet} & \textbf{0.9786} & 0.8189 & 2.249  & 3.439\\
~ & \textbf{UNet-VGG} & 0.9754 & 0.8456 & 1.457  & 2.781\\
~ & \textbf{TopoNet} & 0.9728  & \textbf{0.8671} & \textbf{1.234}  & \textbf{1.275}\\
\hline
\end{tabular}
\end{center}
\end{table*}

\subsection{Results}
\begin{figure*}[ht]
\centering 
\subfigure{
\includegraphics[width=0.14\textwidth]{topoloss/patch2_ori.png}}
\subfigure{
\includegraphics[width=0.14\textwidth]{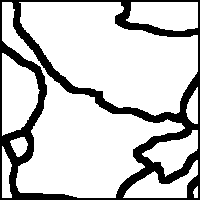}}
\subfigure{
\includegraphics[width=0.14\textwidth]{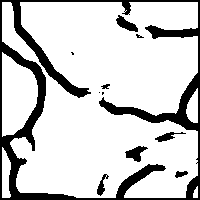}}
\subfigure{
\includegraphics[width=0.14\textwidth]{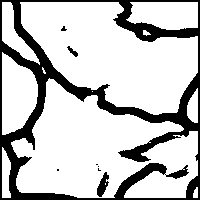}}
\subfigure{ 
\includegraphics[width=0.14\textwidth]{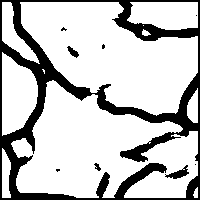}}
\subfigure{
\includegraphics[width=0.14\textwidth]{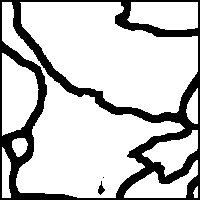}}

\subfigure{
\includegraphics[width=0.14\textwidth]{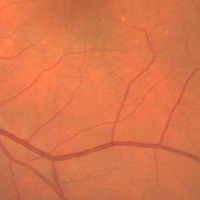}}
\subfigure{
\includegraphics[width=0.14\textwidth]{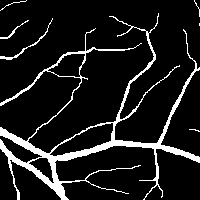}}
\subfigure{
\includegraphics[width=0.14\textwidth]{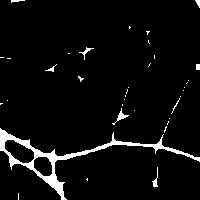}}
\subfigure{
\includegraphics[width=0.14\textwidth]{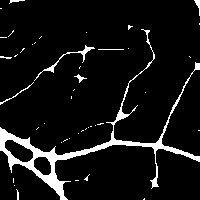}}
\subfigure{ 
\includegraphics[width=0.14\textwidth]{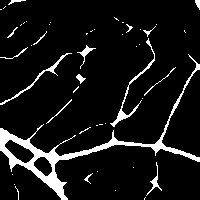}}
\subfigure{
\includegraphics[width=0.14\textwidth]{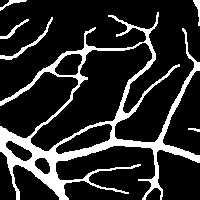}}

\subfigure{
\includegraphics[width=0.14\textwidth]{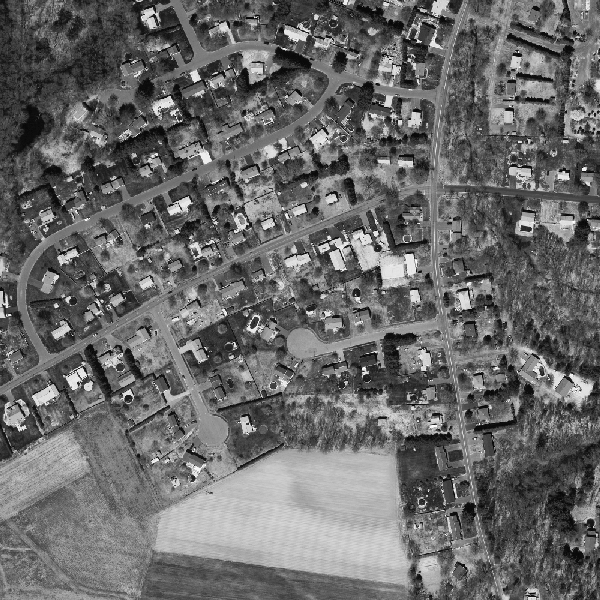}}
\subfigure{
\includegraphics[width=0.14\textwidth]{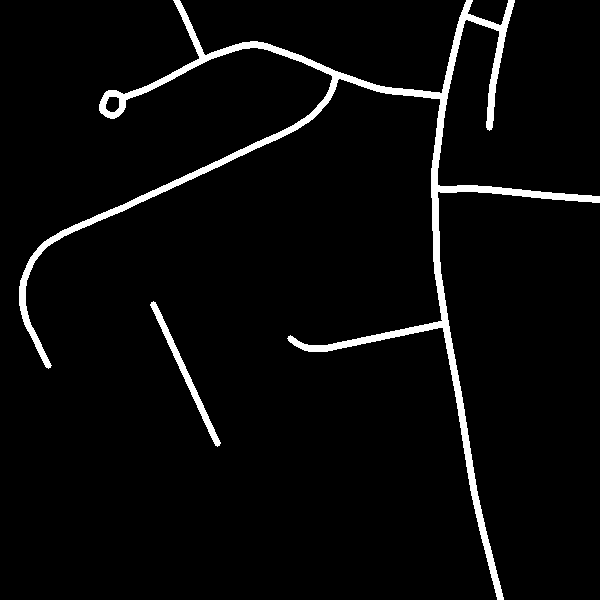}}
\subfigure{
\includegraphics[width=0.14\textwidth]{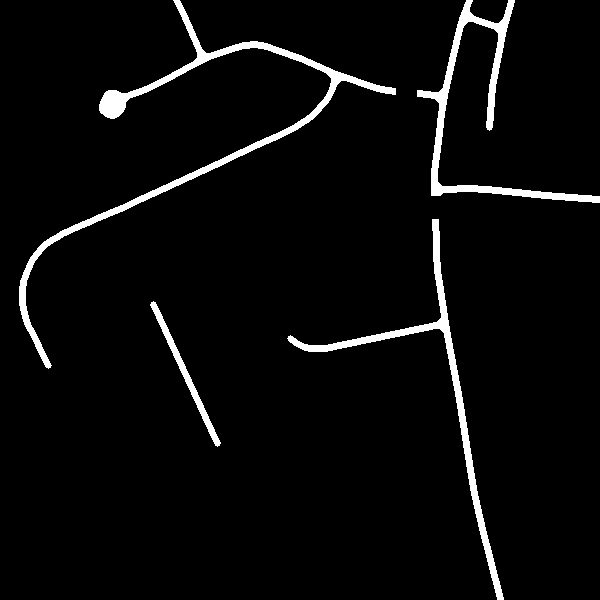}}
\subfigure{
\includegraphics[width=0.14\textwidth]{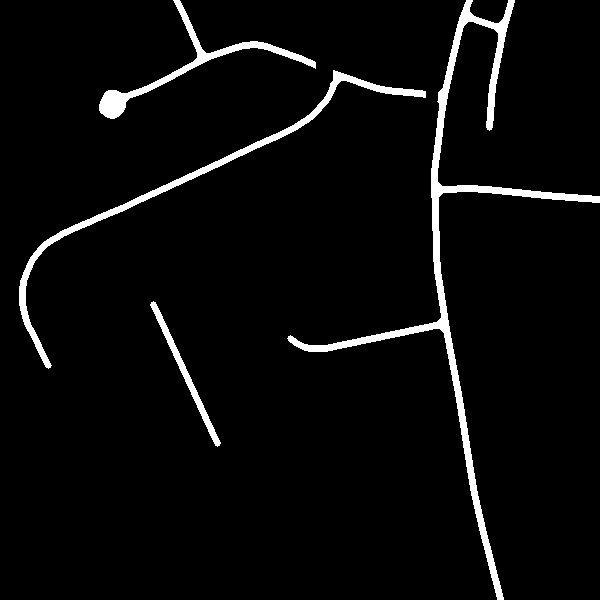}}
\subfigure{ 
\includegraphics[width=0.14\textwidth]{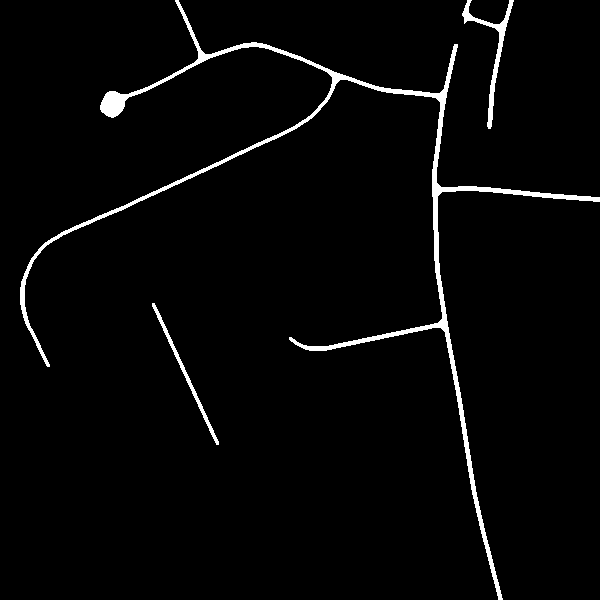}}
\subfigure{
\includegraphics[width=0.14\textwidth]{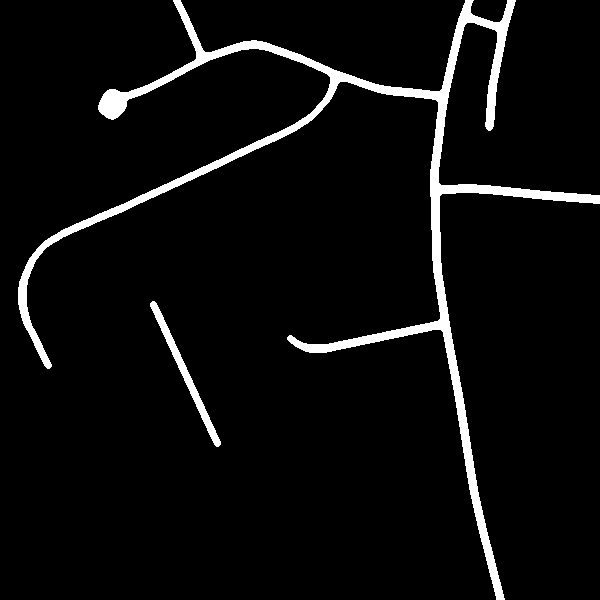}}

\subfigure{
\stackunder{\includegraphics[width=0.14\textwidth]{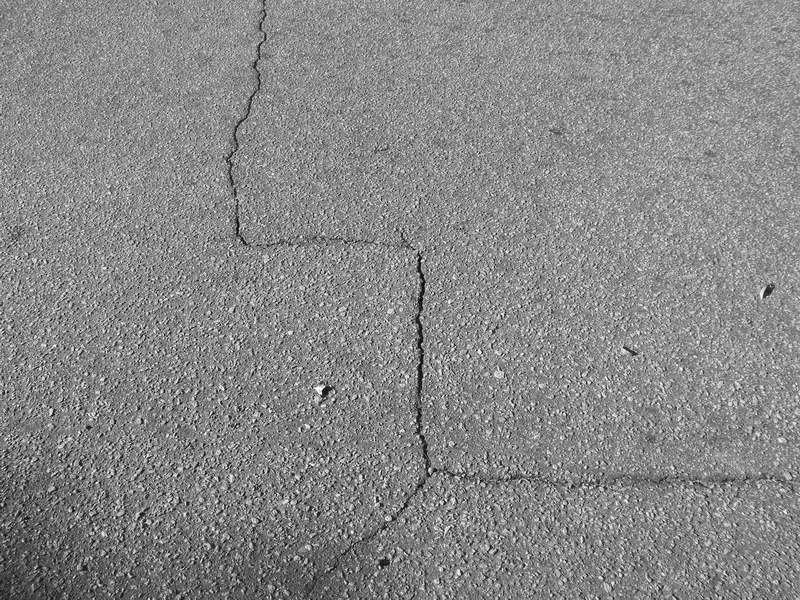}}{(a)}}
\subfigure{
\stackunder{\includegraphics[width=0.14\textwidth]{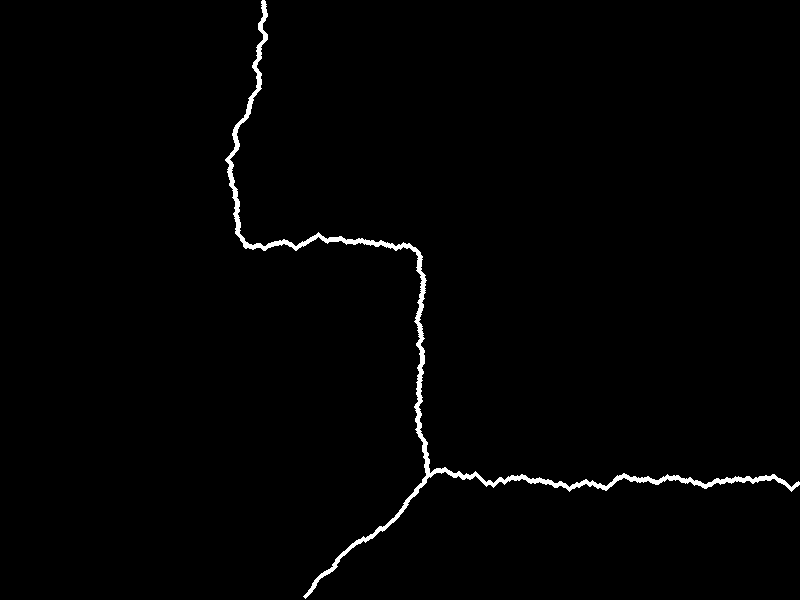}}{(b)}}
\subfigure{
\stackunder{\includegraphics[width=0.14\textwidth]{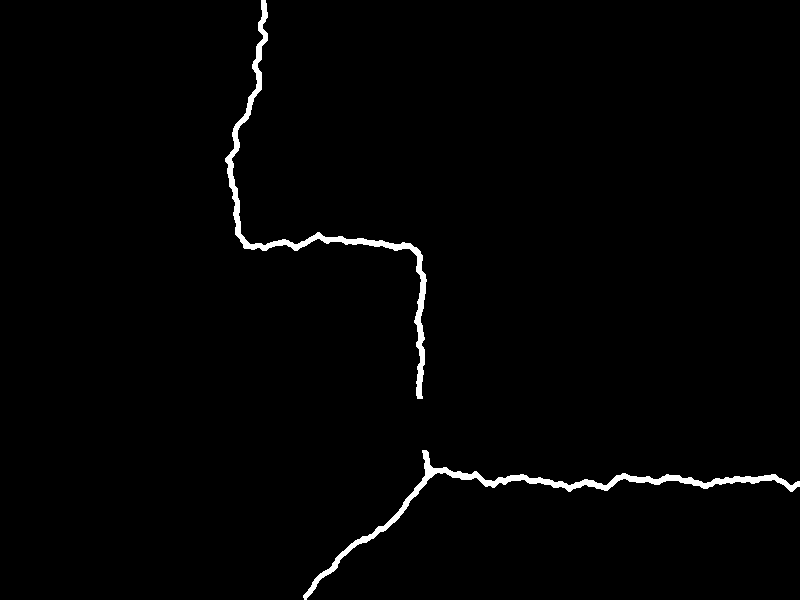}}{(c)}}
\subfigure{
\stackunder{\includegraphics[width=0.14\textwidth]{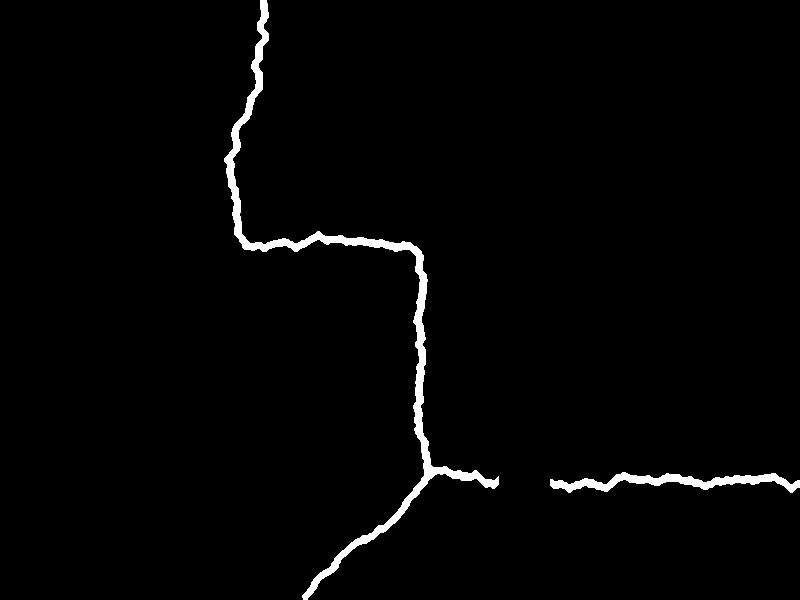}}{(d)}}
\subfigure{ 
\stackunder{\includegraphics[width=0.14\textwidth]{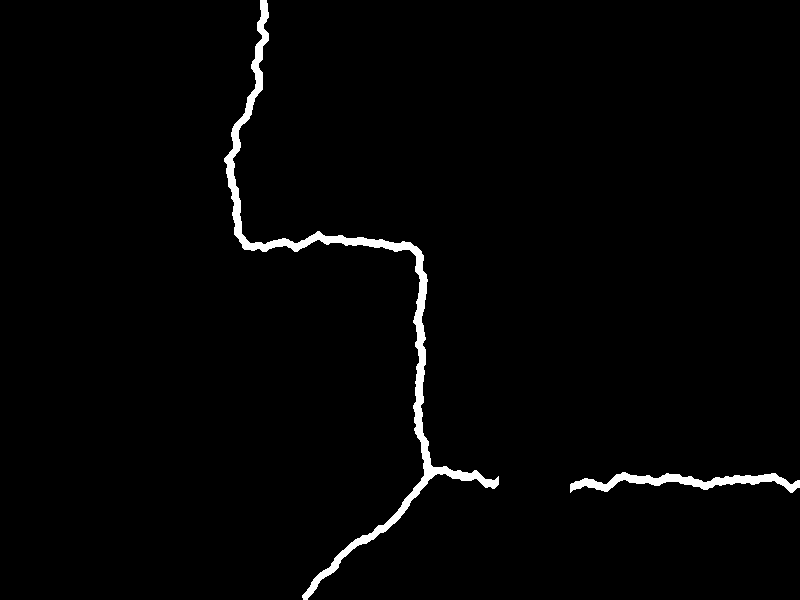}}{(e)}}
\subfigure{
\stackunder{\includegraphics[width=0.14\textwidth]{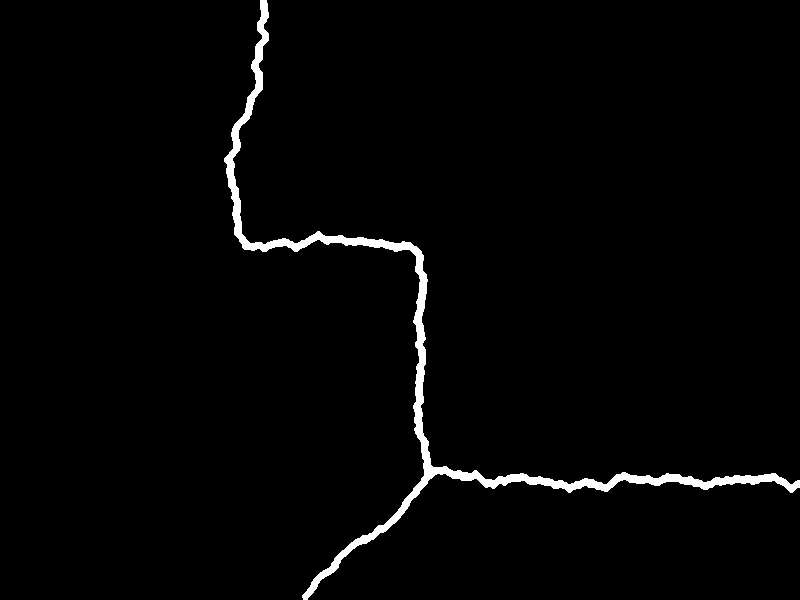}}{(f)}}
\caption{Qualitative results of the proposed method compared to other models. From left to right, sample images, ground truth, results for \textbf{DIVE}, \textbf{UNet}, \textbf{UNet-VGG} and our proposed \textbf{TopoLoss}.}
\label{fig:result}
\end{figure*}

Tab.~\ref{table:medical} shows the quantitative results for three different neuron image datasets, ISBI12, ISBI13, and CREMI. Tab.~\ref{table:ensemble} shows the quantitative results for DRIVE, CrackTree, and Mass.. Our method significantly outperforms existing methods in topological accuracy (in all three topology-aware metrics), without sacrificing pixel accuracy. 
Fig.~\ref{fig:result} shows qualitative results. Our method demonstrates more consistency in terms of structures and topology. It correctly segments fine structures such as membranes, roads, and vessels, while all other methods fail to do so.  
Note that the topological error cannot be solved by training with dilated ground truth masks.
We run additional experiments on the CREMI dataset by training a topology-agnostic model with dilated ground truth masks. For 1 and 2 pixel dilation, We have Betti Error 4.126 and 4.431, respectively. They are still significantly worse than TopoLoss (Betti Error = 1.113).

\begin{figure*}[ht]
\centering 
\subfigure[Loss Curve]{\label{fig:loss}
\includegraphics[width=0.31\textwidth]{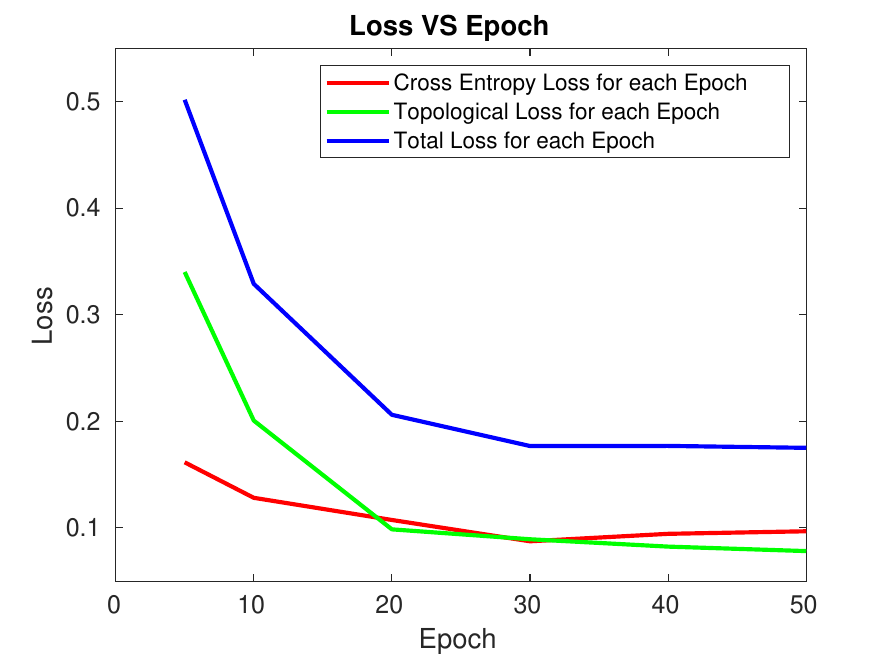}}
\subfigure[Performance VS $\lambda$]{\label{fig:per}
\includegraphics[width=0.31\textwidth]{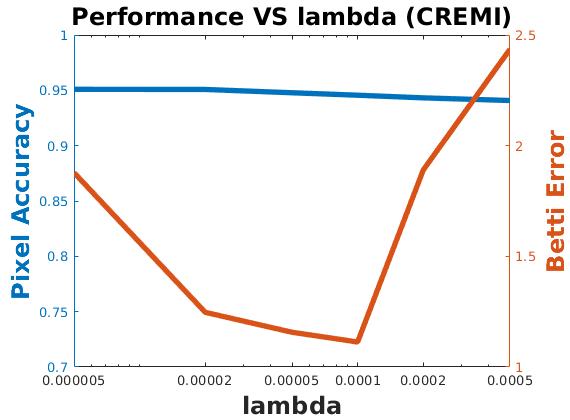}}
\subfigure[Time VS $\lambda$]{\label{fig:time}
\includegraphics[width=0.31\textwidth]{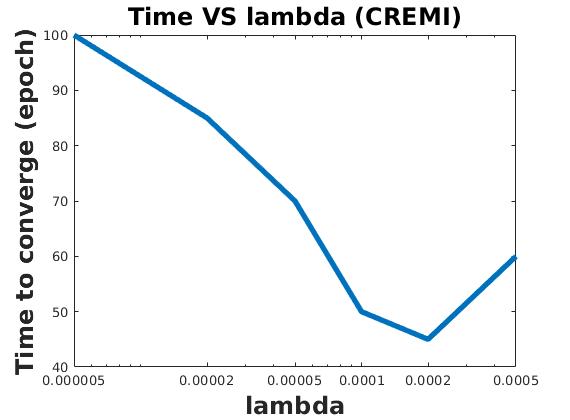}}
\caption{\textbf{(a)} Cross Entropy loss, Topological loss, and total loss in terms of training epochs. \textbf{(b)} Ablation studies of lambda on CREMI w.r.t. accuracy, Betti error. \textbf{(c)} Ablation study of lambda on CREMI w.r.t. convergence rate.}
\label{fig:ablation}
\end{figure*}

\subsection{Ablation Study: Loss Weights}
Our loss (Eq.~\ref{total_loss}) is a weighted combination of cross entropy loss and topological loss. For convenience, we drop the weight of cross entropy loss and weight the topological loss with $\lambda$. Fig.~\ref{fig:per} and~\ref{fig:time} show ablation studies of $\lambda$ on CREMI w.r.t. accuracy, Betti error and convergence rate. As we increase lambda, per-pixel accuracy is slightly compromised. The Betti error decreases at first but increases later. One important observation is that a certain amount of topological loss improves the convergence rate significantly. 
Empirically, we choose $\lambda$ via cross-validation. Different datasets have different $\lambda$'s. In general, $\lambda$ is at the magnitude of 1/10000. This is understandable; while cross entropy loss gradient is applied to all pixels, the topological gradient is only applied to a sparse set of critical pixels. Therefore, the weight needs to be much smaller to avoid overfitting with these critical pixels.

Fig.~\ref{fig:loss} shows the weighted topological loss ($\lambda L_{topo}$), cross entropy loss ($L_{bce}$) and total loss ($L$) at different training epochs. After 30 epochs, the total loss becomes stable. Meanwhile, while $L_{bce}$ increases slightly, $L_{topo}$ decreases. This is reasonable; incorporating of topological loss may force the network to overtrain on certain locations (near critical pixels), and thus may hurt the overall pixel accuracy slightly. 
This is confirmed by the pixel accuracy of TopoLoss in Tab.~\ref{table:medical} and Tab.~\ref{table:ensemble}.

\subsection{Rationale of the Proposed Algorithm} 
\begin{figure*}[ht]
\centering 

\subfigure{\label{fig:patch_ori}
\includegraphics[width=0.16\textwidth]{topoloss/patch2_ori.png}}
\subfigure{\label{fig:epoch1_lh}
\includegraphics[width=0.16\textwidth]{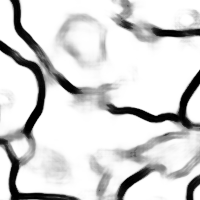}}
\subfigure{\label{fig:epoch2_lh}
\includegraphics[width=0.16\textwidth]{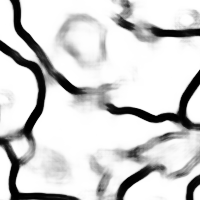}}
\subfigure{\label{fig:epoch3_lh}
\includegraphics[width=0.16\textwidth]{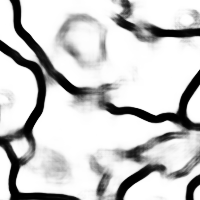}}
\subfigure{\label{fig:epoch4_lh}
\includegraphics[width=0.16\textwidth]{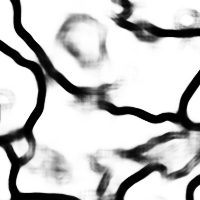}}

\subfigure{\label{fig:gt}
\includegraphics[width=0.16\textwidth]{topoloss/patch2_gt_revised.png}}
\subfigure{\label{fig:epoch1}
\includegraphics[width=0.16\textwidth]{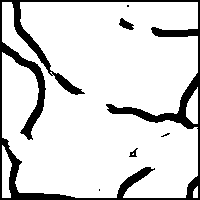}}
\subfigure{\label{fig:epoch2}
\includegraphics[width=0.16\textwidth]{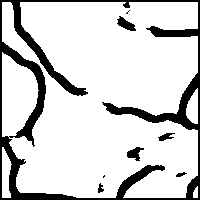}}
\subfigure{\label{fig:epoch3}
\includegraphics[width=0.16\textwidth]{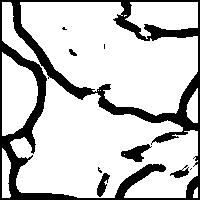}}
\subfigure{\label{fig:epoch4}
\includegraphics[width=0.16\textwidth]{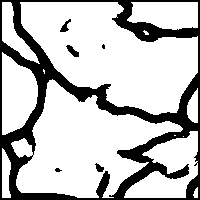}}
\caption{For a sample patch from CREMI, we show the likelihood map and segmentation at different training epochs. The first row corresponds to likelihood maps and the second row are thresholded results. From left to right, the original patch/ground truth, results after 10, 20, 30, and 40 epochs.}
\label{fig:gradual}
\end{figure*}

To further explain the rationale of topological loss, we first study an example training patch.
In Fig.~\ref{fig:gradual}, we plot the likelihood map and the segmentation at different epochs. 
Within a short period, the likelihood map and the segmentation are stabilized globally, mostly thanks to the cross-entropy loss. 
After epoch 20, topological errors are gradually fixed by the topological loss. Notice the change of the likelihood map is only at specific topology-relevant locations. 

\emph{Our topological loss compliments cross-entropy loss by combating sampling bias.}
In Fig.~\ref{fig:gradual}, for most membrane pixels, the network learns to make correct predictions quickly. However, for a small number of difficult locations (blurred regions), it is much harder to learn to predict correctly. The issue is these locations only take a small portion of training pixel samples. Such disproportion cannot be changed even with more annotated training images. The topological loss essentially identifies these difficult locations during training (as critical pixels). It then forces the network to learn patterns near these locations, at the expense of overfitting and consequently slightly compromised per-pixel accuracy.
On the other hand, we stress that topological loss cannot succeed alone. Without cross-entropy loss, inferring topology from a completely random likelihood map is meaningless. Cross-entropy loss finds a reasonable likelihood map so that the topological loss can improve its topology. 

\section{Conclusion}

In this chapter, we introduce a new topological loss driven by persistent homology and incorporate it into the end-to-end training of deep neural networks. Our method is particularly suitable for fine structure image segmentation. Quantitative and qualitative results show that the proposed topological loss term helps to achieve better performance in terms of topological-relevant metrics. Our proposed topological loss term is generic and can be incorporated into different deep learning architectures.

%% file: trojan.tex
\chapter{Trigger Hunting with a Topological Prior for Trojan Detection}
\label{chapter:trojan}
In the previous chapter, we introduced the proposed differentiable persistent-homology based topological loss and demonstrated its effectiveness in image segmentation tasks. In this chapter,
we further explore the power of the topological loss in a quite different context, i.e., trojan detection.

\section{Introduction}

Deep learning has achieved superior performance in various computer vision tasks, such as image classification~\cite{krizhevsky2012imagenet}, image segmentation~\cite{long2015fully}, object detection~\cite{girshick2014rich}, etc. However, the vulnerability of DNNs against backdoor attacks raises serious concerns. In this chapter, we address the problem of \emph{Trojan attacks}, where during training, an attacker injects \emph{polluted samples}. While resembling normal samples, these polluted samples contain a specific type of perturbation (called triggers). These polluted samples are assigned with \emph{target labels}, which are usually different from the expected class labels. Training with this polluted dataset results in a \emph{Trojaned model}. At the inference stage, a Trojaned model behaves normally given clean samples. But when the trigger is present, it makes unexpected, yet consistently incorrect predictions.

One major constraint for Trojan detection is the limited access to polluted training data. In practice, the end-users, who need to detect the Trojaned models, often only have access to the weights and architectures of the trained DNNs. State-of-the-art (SOTA) Trojan detection methods generally adopt a \emph{reverse engineering} approach~\cite{guo2019tabor, wang2019neural, huster2021top, wang2020practical, chen2019deepinspect, liu2019abs}. They start with a few clean samples, using either gradient descent or careful stimuli crafting, to find a potential trigger that alters model prediction. Characteristics of the recovered triggers along with the associated network activations are used as features to determine whether a model is Trojaned or not.

\begin{figure*}[ht]
\centering 
\subfigure[]{
\includegraphics[width=0.145\textwidth]{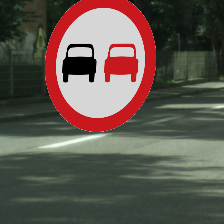}}
\subfigure[]{
\includegraphics[width=0.145\textwidth]{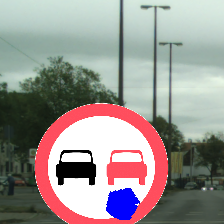}}
\subfigure[]{
\includegraphics[width=0.145\textwidth]{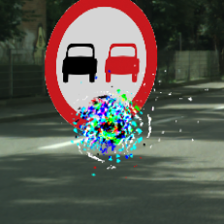}}
\subfigure[]{
\includegraphics[width=0.145\textwidth]{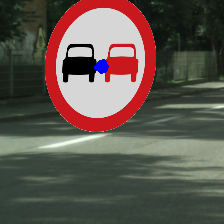}}
\subfigure[]{
\includegraphics[width=0.145\textwidth]{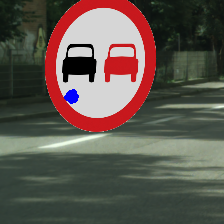}}
\subfigure[]{
\includegraphics[width=0.145\textwidth]{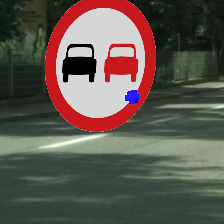}}
\caption{Illustration of recovered triggers: \textbf{(a)} clean image, \textbf{(b)} poisoned image, \textbf{(c)} image with a trigger recovered without topological prior, \textbf{(d)-(f)} images with candidate triggers recovered with the proposed method. Topological prior contributes to improved compactness. We run the trigger reconstruction for multiple rounds with a diversity prior to ensuring a diverse set of trigger candidates.}
\label{fig:teaser_trojan}
\end{figure*}

Trojan triggers can be of arbitrary patterns (e.g., shape, color, texture) at arbitrary locations of an input image (e.g., Fig.~\ref{fig:teaser_trojan}). As a result, one major challenge of the reverse engineering-based approach is the enormous search space for potential triggers.
Meanwhile, just like the trigger is unknown, the target label (i.e., the class label to which a triggered model predicts) is also unknown in practice. 
Gradient descent may flip a model's prediction to the closest alternative label, which may not necessarily be the target label. This makes it even more challenging to recover the true trigger. 
Note that many existing methods \cite{guo2019tabor, wang2019neural, wang2020practical} require a target label. These methods achieve target label independence by enumerating all possible labels, which can be computationally prohibitive especially when the label space is huge. 

We propose a novel target-label-agnostic reverse engineering method. First, to improve the quality of the recovered triggers, we need a prior that can localize the triggers, but in a flexible manner. We, therefore, propose to enforce a \emph{topological prior} to the optimization process of reverse engineering, i.e., the recovered trigger should have fewer connected components. This prior is implemented through a topological loss based on the theory of persistent homology \cite{edelsbrunner2010computational}. It allows the recovered trigger to have arbitrary shape and size. Meanwhile, it ensures the trigger is not scattered and is reasonably localized. 
See Fig.~\ref{fig:teaser_trojan} for an example -- comparing (d)-(f) vs. (c).

\begin{figure}[ht]
\centering 
    \includegraphics[width=0.6\textwidth]{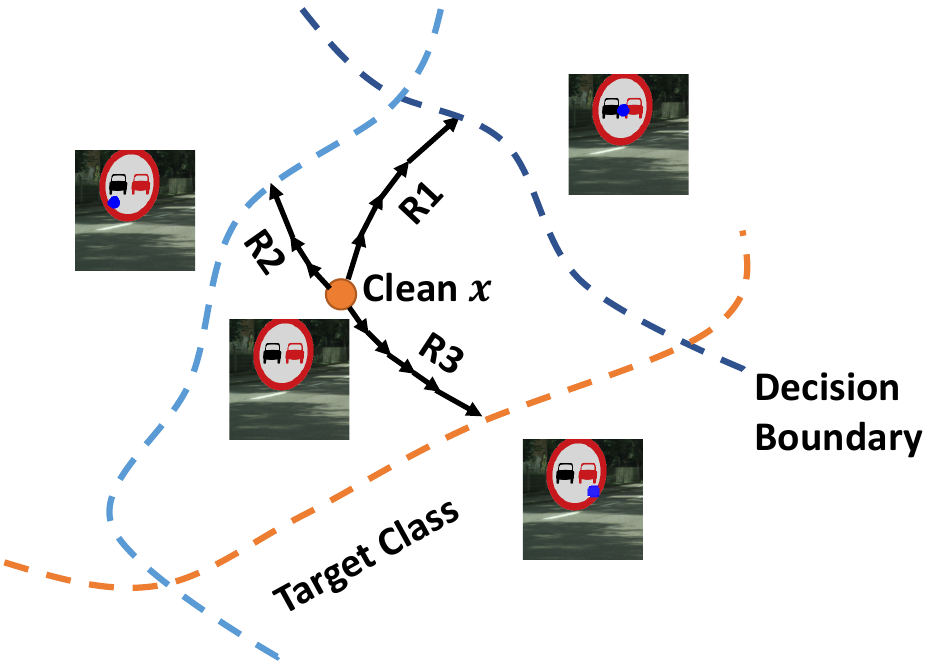}
  \caption{Illustration of generating a diverse set of trigger candidates to increase the chance of finding the true trigger, especially for scenarios with unknown target labels.}
\label{fig:trigger}
\end{figure}

As a second contribution, we propose to reverse engineer \emph{multiple diverse trigger candidates}. Instead of running gradient descent once, we run it for multiple rounds, each time producing one trigger candidate, e.g., Fig.~\ref{fig:teaser_trojan} (d)-(f). Furthermore, we propose a \emph{trigger diversity loss} to ensure the trigger candidates are sufficiently different from each other (see Fig.~\ref{fig:trigger}). Generating multiple diverse trigger candidates can increase the chance of finding the true trigger. It also mitigates the risk of unknown target labels.
In the example of Fig.~\ref{fig:trigger}, the first trigger candidate flips the model prediction to a label different from the target, while only the third candidate hits the true target label.

Generating multiple trigger candidates, however, adds difficulties in filtering out already-subtle cues for Trojan detection. We also note reverse engineering approaches often suffer from false positive triggers such as adversarial perturbations or direct modification of the crucial objects of the image.~\footnote{Modification of crucial objects is usually not a valid trigger strategy; it is too obvious to end-users and is against the principle of adversaries.}
In practice, we systematically extract a rich set of features to describe the characteristics of the reconstructed trigger candidates based on geometry, color, and topology, as well as network activations. A Trojan-detection network is then trained to detect Trojaned models based on these features. Our main contributions are summarized as follows:

\begin{itemize}
  \item We propose a topological prior to regularize the optimization process of reverse engineering. The prior ensures the locality of the recovered triggers, while being sufficiently flexible regarding the appearance. It significantly improves the quality of the reconstructed triggers.
  \item We propose a diversity loss to generate multiple diverse trigger candidates. This increases the chance of recovering the true trigger, especially for cases with unknown target labels. 

\item Combining the topological prior and diversity loss, we propose a novel Trojan detection framework. On both synthetic and public TrojAI benchmarks, our method demonstrates substantial improvement in both trigger recovery and Trojan detection.
\end{itemize}

\section{Method}

\begin{figure*}[ht]
  \centering
  \noindent\makebox[\textwidth][c] {
    \includegraphics[width=0.6\paperwidth]{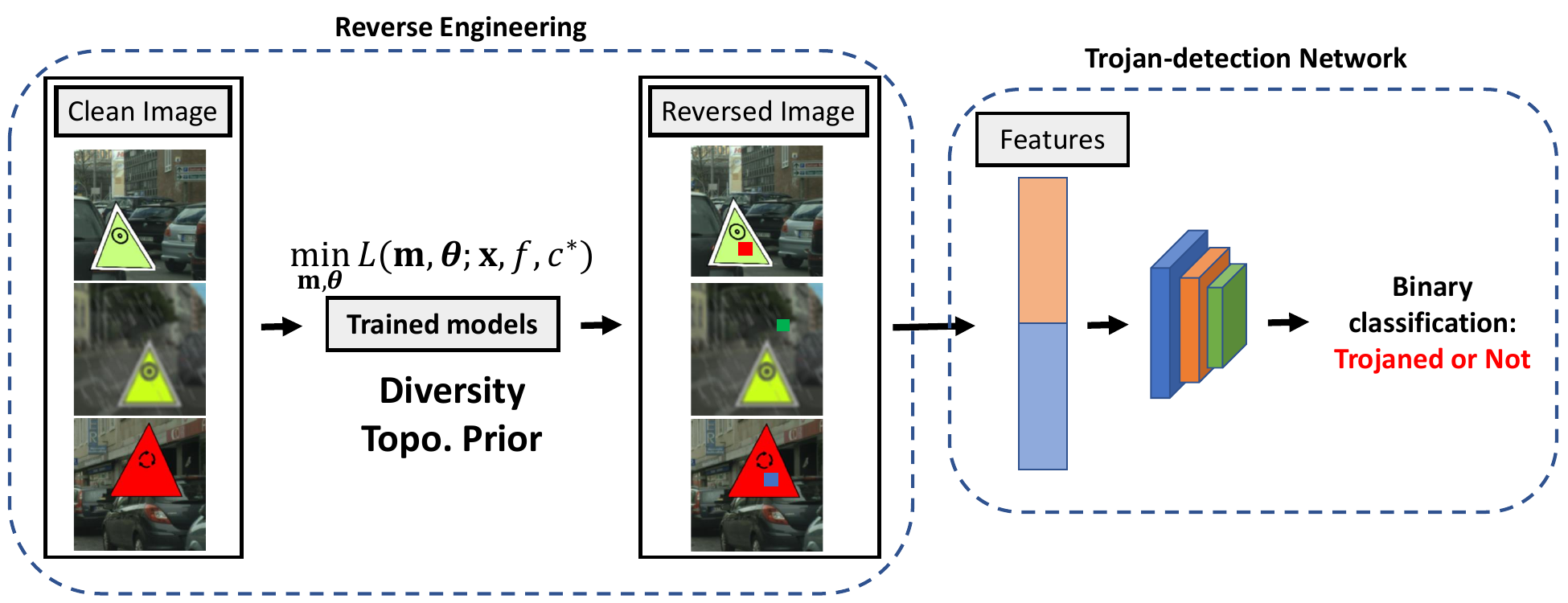}}
    \caption{Our Trojan detection framework. }
      \label{fig:framework}
\end{figure*}

Our reverse engineering framework is illustrated in Fig.~\ref{fig:framework}.
Given a trained DNN model, either clean or Trojaned, and a few clean images, we use gradient descent to reconstruct triggers that can flip the model's prediction. To increase the quality of reconstructed triggers, we introduce novel diversity loss and topological prior. They help recover multiple diverse triggers of high quality. 

The common hypothesis of reverse engineering approaches is that the reconstructed triggers will appear different for Trojaned and clean models. 
To fully exploit the discriminative power of the reconstructed triggers for Trojan detection, we extract features based on trigger characteristics and associated network activations. These features are used to train a classifier, called the Trojan-detection network, to classify a given model as Trojaned or clean. 

We note the discriminative power of the extracted trigger features, and thus the Trojan-detection network is highly dependent on the quality of the reconstructed triggers. Empirical results will show the proposed diversity loss and topological prior are crucial in reconstructing high quality triggers, and ensuring a high quality Trojan-detection network. We will show that our method can learn to detect Trojaned models even when trained with a small amount of labeled DNN models.

For the rest of this section, we mainly focus on the reverse engineering module. We also add details of the Trigger feature extraction to Sec.~\ref{sec:features} and details of the Trojan-detection network to Sec.~\ref{classifier}.

\subsection{Reverse Engineering of Multiple Diverse Trigger Candidates}
\label{Reverse}

\begin{figure}[ht]
\centering 
    \includegraphics[width=0.5\textwidth]{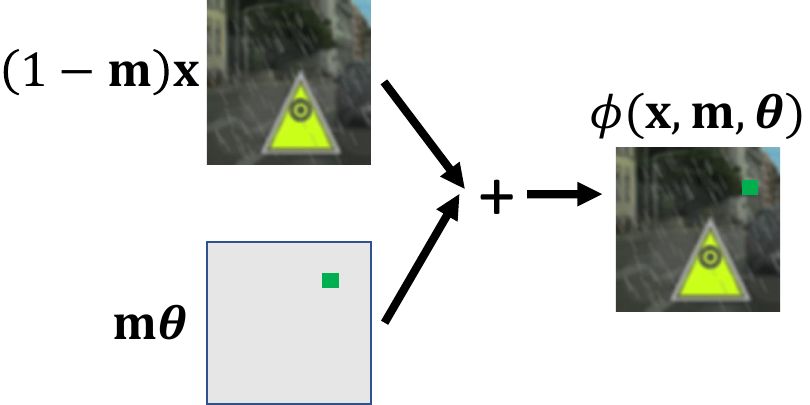}
  \caption{$\mathbf{m}$ and $\boldsymbol{\theta}$ convert an input image $\mathbf{x}$ into an altered one $\phi(\mathbf{x},\mathbf{m},\boldsymbol{\theta})$. The $\odot$ is omitted here for simplification.}
\label{fig:polygon}
\end{figure}

Our method is based on the existing reverse engineering pipeline first proposed by Neural Cleanse~\cite{wang2019neural}. 
Given a trained DNN model, let $f(\cdot)$ be the mapping from an input clean image $\mathbf{x}\in \mathbb{R}^{3 \times M \times N}$ to the output $\mathbf{y} \in \mathbb{R}^K$ with $K$ classes, where $M$ and $N$ denote the height and width of the image, respectively. Denote by $f_k(\cdot)$ the $k$-th output of $f$. The predicted label $c^\ast$ is given by $c^\ast=\argmax_{k} f_k(\mathbf{x})$, $1 \le k \le K$. We introduce parameters $\bm{\theta}$ and $\mathbf{m}$ to convert $\mathbf{x}$ into an altered sample
\begin{equation}
\label{reverse}
    \phi(\mathbf{x}, \mathbf{m}, \bm{\theta}) = (\mathbf{1}-\mathbf{m})\odot \mathbf{x} + \mathbf{m} \odot \bm{\theta},
\end{equation}

where the binary mask $\mathbf{m}\in \{0,1\}^{M \times N}$ and the pattern $\bm{\theta} \in \mathbb{R}^{M \times N}$ determine the trigger. $\mathbf{1}$ denotes an all-one matrix. The symbol ``$\odot$'' denotes Hadamard product.  
See Fig.~\ref{fig:polygon} for an illustration. 
We intend to find a triggered image 
$\hat{\mathbf{x}} = \phi(\mathbf{x},\hat{\mathbf{m}},\hat{\bm{\theta}})$ 
so that the model prediction $\hat{c} = \argmax_k f_k(\hat{\mathbf{x}})$ is different from the prediction on the original image $c^\ast$.

We find the triggered image, $\hat{\mathbf{x}}$, by minimizing a loss over the space of $\mathbf{m}$ and $\bm{\theta}$: 
\begin{equation}
\label{optimization}
L( \mathbf{m}, \bm{\theta}; \mathbf{x}, f, c^\ast) =  
    L_{flip}(\ldots) + \lambda_1 L_{div}(\ldots)  + \lambda_2 L_{topo}(\ldots) + R(\mathbf{m}),
\end{equation}
where $L_{flip}$, $L_{div}$, and $L_{topo}$ denote the label-flipping loss, diversity loss, and topological loss, respectively. We temporarily dropped their arguments for convenience. $\lambda_1$, $\lambda_2$ are the weights to balance the loss terms. 
$R(\mathbf{m})$ is a regularization term penalizing the size and range of the mask (more details will be provided in Sec.~\ref{Regularizer}). 

To facilitate optimization, we relax the constraint on the mask $\mathbf{m}$ and allow it to be a continuous-valued function, ranging between 0 and 1 and defined over the image domain, $\mathbf{m}\in [0,1]^{M\times N}$. 
Next, we introduce the three loss terms one-by-one.

\myparagraph{Label-Flipping Loss $L_{flip}$}: 
The label-flipping loss $L_{flip}$ penalizes the prediction of the model regarding the ground truth label, formally:
\begin{equation}
    L_{flip}(\mathbf{m},\boldsymbol{\theta}; \mathbf{x}, f, c^\ast) = f_{c^\ast}(\phi(\mathbf{x},\mathbf{m},\boldsymbol{\theta})).
\end{equation}
Minimizing $L_{flip}$ means minimizing the probability that the altered image $\phi(\mathbf{x},\mathbf{m},\boldsymbol{\theta})$ is predicted as $c^\ast$. In other words, we are pushing the input image out of its initial decision region.

Note that we do not specify which label we would like to flip the prediction to. This makes the optimization easier. Existing approaches often run optimization to flip the label to a target label and enumerate through all possible target labels~\cite{wang2019neural,wang2020practical}. This can be rather expensive in computation, especially with large label space. 

The downside of not specifying a target label during optimization is we will potentially miss the correct target label, i.e., the label that the Trojaned model predicts on a triggered image. To this end, we propose to reconstruct multiple candidate triggers with diversity constraints. This will increase the chance of hitting the correct target label. See Fig.~\ref{fig:trigger} for an illustration.

\myparagraph{Diversity Loss $L_{div}$}: With the label-flipping loss $L_{flip}$, we flip the label to a different one from the original clean label and recover the corresponding triggers. The new label, however, may not be the same as the true target label. Also considering the huge trigger search space, it is difficult to recover the triggers with only one attempt. Instead, we propose to search for multiple trigger candidates to increase the chance of capturing the true trigger. 

We run our algorithm for $N_T$ rounds, each time reconstructing a different trigger candidate.
To avoid finding similar trigger candidates, we introduce the diversity loss $L_{div}$ to encourage different trigger patterns and locations. Let $\mathbf{m}_j$ and $\bm{\theta}_{j}$ denote the trigger mask and pattern found in the $j$-th round. 
At the $i$-th round, we compare the current candidates with triggers from all previous founds in terms of $L_2$ norm. Formally:
\begin{equation}
    L_{div}(\mathbf{m},\boldsymbol{\theta}) = - \sum\nolimits_{j=1}^{i-1} ||\mathbf{m}\odot \bm{\theta}-\mathbf{m}_j\odot \bm{\theta}_{j}||_2.
\end{equation}
Minimizing $L_{div}$ ensures the eventual trigger $\mathbf{m}_i\odot \bm{\theta}_{i}$ to be different from triggers from previous rounds. Fig.~\ref{fig:teaser_trojan}(d)-(f) demonstrates the multiple candidates recovered with sufficient diversity.

\subsection{Topological Prior}
\label{sec:topology}

Quality control of the trigger reconstruction remains a major challenge in reverse engineering methods, due to the huge search space of triggers. Even with the regularizer $R(\mathbf{m})$, the recovered triggers can still be scattered and unrealistic. See Fig.~\ref{fig:teaser_trojan}(c) for an illustration.
We propose a topological prior to improve the locality of the reconstructed trigger. We introduce a topological loss enforcing that the recovered trigger mask $\mathbf{m}$ has as few connected components as possible. The loss is based on the theory of persistent homology \cite{edelsbrunner2000topological,edelsbrunner2010computational}, which models the topological structures of a continuous signal in a robust manner. \highlight{Note that the assumption holds if the trigger is not spanned over the whole image, which fits the scenarios we are discussing in our work.}

\myparagraph{Persistent Homology.}
We introduce persistent homology in the context of 2D images. A more comprehensive treatment of the topic can be found in \cite{edelsbrunner2010computational,dey2021computational}. Recall we relaxed the mask function $\mathbf{m}$ to a continuous-valued function defined over the image domain (denoted by $\Omega$).
Given any threshold $\alpha$, we can threshold the image domain with regard to $\mathbf{m}$ and obtain the \emph{superlevel set}, $ \Omega^{\alpha}:= \{p \in \Omega | \mathbf{m}(p) \geq \alpha \}$. A superlevel set can have different topological structures, e.g., connected components and holes. 
If we continuously decrease the value $\alpha$, we have a continuously growing superlevel set $\Omega^{\alpha}$. This sequence of superlevel set is called a \emph{filtration}. The topology of $\Omega^{\alpha}$ continuously changes through the filtration. New connected components are born and later die (merged with others). New holes are born and later die (sealed up). For each topological structure, the threshold at which it is born is called its \emph{birth time}. The threshold at which it dies is called its \emph{death time}. The difference between birth and death time is called the \emph{persistence} of the topological structure.

\begin{figure*}[ht]
  \centering
  \noindent\makebox[\textwidth][c] {
    \includegraphics[width=0.6\paperwidth]{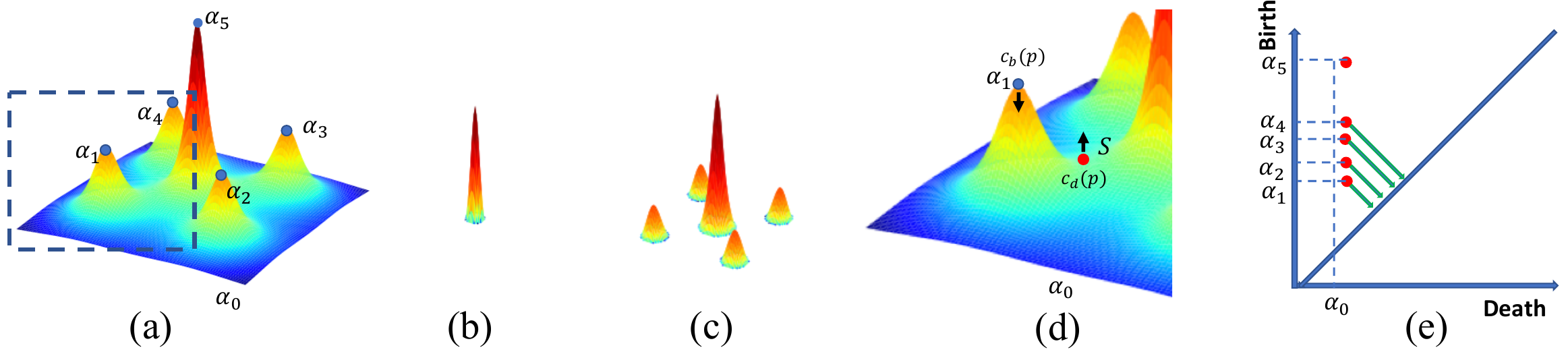}}
    \caption{From the left to right: \textbf{(a)} a sample landscape for a continuous function. The values at the peaks $\alpha_0 < \alpha_1 < \alpha_2< \alpha_3 < \alpha_4< \alpha_5$. As we decrease the threshold, the topological structures of the superlevel set change, (\textbf{b}) and (\textbf{c}) correspond to topological structures captured by different thresholds, (\textbf{d}) highlighted region in (\textbf{a}), (\textbf{e}) the changes are captured by the persistence diagram (right figure). We focus on the 0-dimensional topological structures (connected components). Each persistent dot in the persistence diagram denotes a specific connected component. The topological loss is introduced to reduce the connected components, which means pushing most of the persistent dots to the diagonal (along the green lines).}
      \label{fig:topo_illu}
\end{figure*}

We record the lifespan of all topological structures over the filtration and encode them via a 2D point set called \emph{persistence diagram}, denoted by $\dgm(\mathbf{m})$. Each topological structure is represented by a 2D point within the diagram, $p\in\dgm(\mathbf{m})$, called a \textit{persistent dot}. We use the birth and death times of the topological structure to define the coordinates of the corresponding persistent dot. For each dot $p\in \dgm(\mathbf{m})$, we abuse the notation and call the birth/death time of its corresponding topological structure as $\birth(p)$ and $\death(p)$. Then we have $p=(\death(p), \birth(p))$. 
See Fig.~\ref{fig:topo_illu} for an example function $\mathbf{m}$ (viewed as a terrain function) and its corresponding diagram. There are five dots in the diagram, corresponding to five peaks in the landscape view. 

To compute the persistence diagram, we use the classic algorithm \cite{edelsbrunner2010computational,edelsbrunner2000topological} with an efficient implementation \cite{chen2011persistent,wagner2012efficient}.  The image is first discretized into a cubical complex consisting of vertices (pixels), edges, and squares. A boundary matrix is then created to encode the adjacency relationship between these elements. The algorithm essentially carries out a matrix reduction algorithm over the boundary matrix, and the reduced matrix reads out the persistence diagram.

\myparagraph{Topological Loss $L_{topo}$}: 
We formulate our topological loss based on the persistent homology described above. Minimizing our loss reduces the number of connected components of triggers. We will focus on zero-dimensional topological structures, i.e., connected components. Intuitively speaking, each dot in the diagram corresponds to a connected component. The ones far away from the diagonal line are considered salient as their birth and death times are far apart. And the ones close to the diagonal line are considered trivial. In Fig.~\ref{fig:topo_illu}, there is one salient dot far away from the diagonal line. It corresponds to the highest peak. The other four dots are closer to the diagonal line and correspond to the smaller peaks.
The topological loss will reduce the number of connected components by penalizing the distance of all dots from the diagonal line, except for the most salient one.
Formally, the loss $L_{topo}$ is defined as: 
\begin{equation}
\label{topo_loss_trojan}
    L_{topo}(\mathbf{m}) = \sum\nolimits_{p \in \dgm(m) \setminus \{p^\ast\}}[\birth(p)-\death(p)]^2,
\end{equation}
where $p^\ast$ denotes the persistent dot that is farthest away from the diagonal (with the highest persistence). Minimizing this loss will keep $p^\ast$ intact, while pushing all other dots to the diagonal line, thus making their corresponding components either disappear or merged with the main component. 

\myparagraph{Differentiability and the Gradient}: The loss function (Eq.~\ref{topo_loss_trojan}) is differentiable almost everywhere in the space of functions. To see this, we revisit the filtration, i.e., the growing superlevel set as we continuously decrease the threshold $\alpha$. The topological structures change at specific locations of the image domain. A component is born at the corresponding local maximum. It dies merging with another component at the saddle point between the two peaks. In fact, these locations correspond to critical points of the function. And the function values at these critical points correspond to the birth and death times of these topological structures. For a persistent dot, $p$, we call the critical point corresponding to its birth, $c_b(p)$, and the critical point corresponding to its death, $c_d(p)$. Then we have $\birth(p) = \mathbf{m}(c_b(p))$ and $\death(p) = \mathbf{m}(c_d(p))$. The loss function (Eq.~\ref{topo_loss_trojan}) can be rewritten as a polynomial function of the function $\mathbf{m}$ at different critical points. 

\begin{equation}
\label{gradient_trojan}
    L_{topo}(\mathbf{m}) = \sum\nolimits_{p \in \dgm(m) \setminus \{p^\ast\}}[\mathbf{m}(c_b(p))-\mathbf{m}(c_d(p))]^2.
\end{equation}

The gradient can be computed naturally. $L_{topo}$ is a piecewise differentiable loss function over the space of all possible functions $\mathbf{m}$. 
In a gradient decent step, for all dots except for $p^\ast$, we push up the function at the death critical point $c_d(p)$ (the saddle), and push down the function value at the birth critical point $c_b(p)$ (the local maximum). This is illustrated by the arrows in Fig.~\ref{fig:topo_illu} (Middle-Right). This will kill the non-salient components and push them toward the diagonal.

\subsection{Trigger Feature Extraction and Trojan Detection Network}
\label{sec:features}
Next, we summarize the features we extract from recovered triggers.
The recovered Trojan triggers can be characterized via their capability in flipping model predictions (i.e., the label-flipping loss). Moreover, they are different from adversarial noise as they tend to be more regularly shaped and are also distinct from actual objects which can be recognized by a trained model. 
We introduce appearance-based features to differentiate triggers from adversarial noise and actual objects.

Specifically, for label flipping capability, we directly use the label-flipping loss $L_{flip}$ and diversity loss $L_{div}$ as features. For appearance features, we use trigger size and topological statistics as their features: 1) The number of foreground pixels divided by the total number of pixels in mask $\mathbf{m}$; 2) To capture the size of the triggers in the horizontal and vertical directions, we fit a Gaussian distribution to the mask $\mathbf{m}$ and record \textit{mean} and \textit{std} in both directions; 3) The trigger we find may have multiple connected components. The final formulated topological descriptor includes the topological loss $L_{topo}$, the number of connected components, \textit{mean}, and \textit{std} in terms of the size of each component. 

After the features are extracted, we build a neural network for Trojan detection, which takes the bag of features of the generated triggers as inputs, and outputs a scalar score of whether the model is Trojaned or not. More details are provided in Sec.~\ref{classifier}.

\subsection{Learning-based Trojan-detection Network}
\label{classifier}

\begin{figure}[ht]
\centering 
    \includegraphics[width=0.8\textwidth]{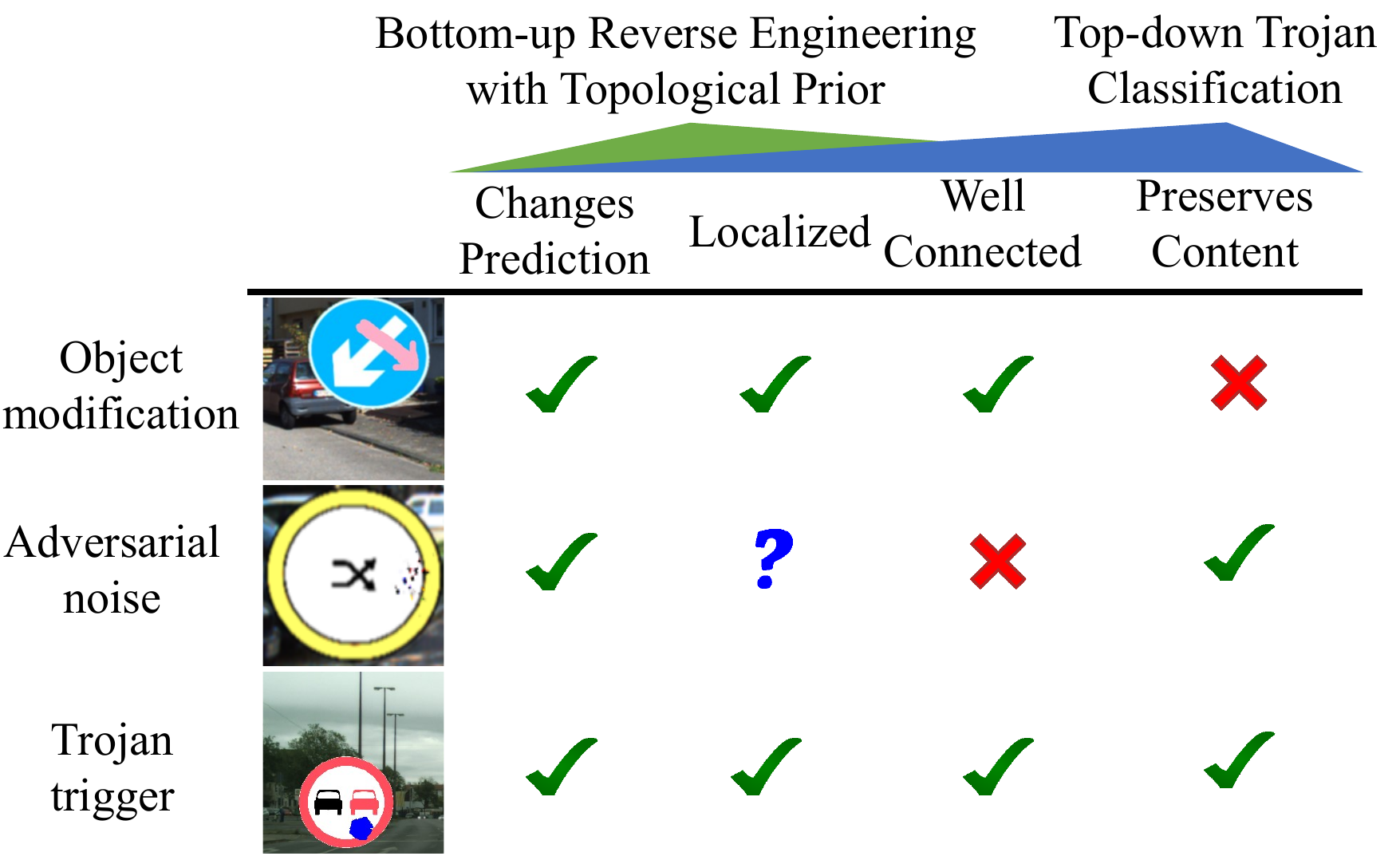}
    \caption{Our Trojan detection method combines bottom-up Trigger reverse engineering under topological constraints, with top-down classification. Such a combination allows us to accurately isolate Trojan triggers from non-Trojan patterns such as adversarial noise and object modifications.}
      \label{fig:hunt}
\end{figure}

While bottom-up trigger generation uses appearance heuristics to search for possible triggers, we cannot guarantee the recovered triggers are true triggers, even for Trojaned models. Many other perturbations can create label flipping effects, such as adversarial samples and modifications specific to the most semantically critical region. See Fig.~\ref{fig:hunt} for illustrations. These examples can easily become false positives for a Trojan detector; they can be good trigger candidates generated by the reverse engineer pipeline.

To fully address these issues, we propose a top-down Trojan detector learned using clean and Trojaned models. It helps separate true Trojan triggers from other false positives. 
To this end, we propose to extract features from the reverse engineered Trojan triggers and train a separate shallow neural network for Trojan detection.

For each model, we generate a diverse set of $N_T$ possible triggers for each of its $K$ output classes to scan for possible triggers. As described above, we extract one feature for each generated trigger. As a result, for each model, we have $N_T\times K$ sets of features. 

\label{Regularizer}
\myparagraph{Regularizer}: The regularizer $R(\mathbf{m})$ consists of a mass term and a size term. For mass, we use $\bar{\mathbf{m}}$ as the average value of $\mathbf{m}$. For size, we normalize the mask $\mathbf{m}$ into a distribution $p(x,y)$ over $x$ and $y$. To capture the spatial extent of mask $m$, we compute the standard deviation of $X \sim p(x)$ as $\delta_X$ and $Y \sim p(y)$ as $\delta_Y$. As a result, $R(\mathbf{m})=\bar{\mathbf{m}} + \delta_X + \delta_Y$ is the regularizer.

\myparagraph{Classification Network}: After the features are extracted, we build a neural network for Trojan detection, which takes as input for a given image classifier, the bag of features of its generated triggers and outputs a scalar score of whether the model is Trojaned or not.

Since different models may have different numbers of output classes $K$, the number of features varies for each model. Therefore, a sequence of modeling architectures -- such as bag-of-words, Recurrent Neural Networks, and Transformers -- could be employed to aggregate the $N_T\times K$ features into a $0/1$ Trojan classification output. 

Given a set of annotated models, clean or infected, supervised learning can be applied to train the classification network. 
Empirically, we found that a simple bag-of-words technique achieved the best Trojan detection performance while also being fast to run and data efficient. Specifically, let the bag of features be $\{v_i\}$, $i=1,\ldots,N_TK$. The features are first transformed individually using an MLP, followed by average pooling across the features and another MLP to output the Trojan classification:
\begin{align}
    \vec{h}_i =MLP_\alpha(\vec{v}_i), \quad
    \vec{h}=\frac{1}{N_TK}\sum_{i}\vec{h}_i, \quad
    s=MLP_\beta(\vec{h}), \quad i=1,\ldots N_TK.
\end{align}

\subsection{Supporting Additional Trigger Classes}
\label{additional}
Different classes of Trojan triggers are being actively explored in Trojan detection benchmarks. For image classification, the TrojAI datasets 
include localized triggers which can be directly applied to objects along with 
global filter-based Trojans, where the idea is that a color filter could be attached to the lens of the camera and results in a global image transformation. 

Our top-down bottom-up Trojan detection framework is designed to support multiple classes of Trojan triggers, where each trigger class gets its dedicated reverse engineering approach and pathway in the Trojan detection network. Adding support to a new class of triggers, e.g. color filters, amounts to adding a reverse engineering approach for color filters and adding an appearance feature descriptor for Trojan classification. 

\begin{figure*}[ht]
  \centering
  \noindent\makebox[\textwidth][c] {
 \includegraphics[width=0.6\paperwidth]{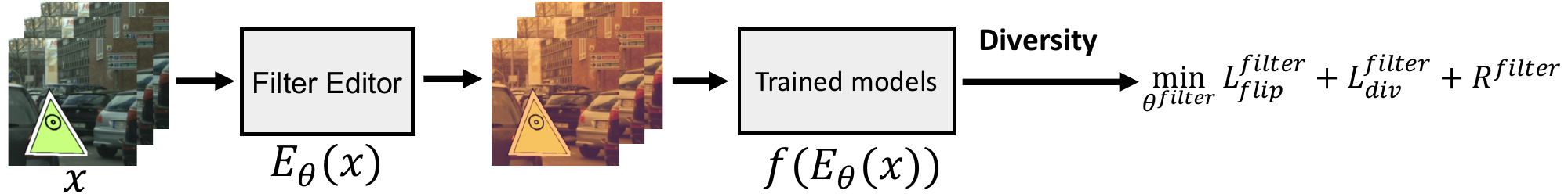}}
 \caption{Reverse engineering of global color filter triggers.}
   \label{fig:instagram}
\end{figure*}

\myparagraph{Reverse Engineering Color Filter Triggers.} The pipeline of reverse engineering color filter triggers is illustrated in Fig.~\ref{fig:instagram}. Compared to reverse engineering local triggers in Fig.~\ref{fig:polygon}, the filter editor and the loss functions are adjusted to find color filter triggers.

We model a color filter trigger using a per-pixel position-dependent color transformation. Let $[r_{ij},g_{ij},b_{ij}]$ be the color of a pixel of input image $\mathbf{x}$ at location $(i,j)$, and we model a color filter trigger using an MLP $E(\cdot ;\theta^{\text{filter}})$ with parameters $\theta^{\text{filter}}$ which performs position-dependent color mapping:

\begin{equation}
[ \hat{r_{ij}},\hat{g_{ij}},\hat{b_{ij}} ]=E([r_{ij},g_{ij},b_{ij},i,j];\theta^{\text{filter}}).
\end{equation}

Here $[ \hat{r_{ij}},\hat{g_{ij}},\hat{b_{ij}} ]$ is the color of pixel $(i,j)$ of the triggered image $\hat{\mathbf{x}}$. For TrojAI datasets we use a 2-layer 16 hidden neurons MLP to model the color filters, as it learns sufficiently complex color transforms while being fast to run:

\begin{equation}
    L^{\text{filter}} = L_{flip}^{\text{filter}} + \lambda_1^{\text{filter}} L_{div}^{\text{filter}}  + \lambda_2^{\text{filter}} R^{\text{filter}}(\theta^{\text{filter}}).
\end{equation}

The label flipping loss $L_{flip}^{\text{filter}}$ remains identical:
\begin{equation}
    L_{flip}^{\text{filter}} = \hat{y}_c.
\end{equation}

The diversity loss $L_{div}^{\text{filter}}$ is designed to induce diverse color transforms. We use how a color filter transforms 16 random $(r,g,b,i,j)$ tuples to characterize a color filter $E(\cdot ;\theta^{\text{filter}})$. We record the 16 output $(\hat{r},\hat{g},\hat{b})$ tuples as a 48-dim descriptor of the color filter, denoted as $u_{\theta^{\text{filter}}}$. The diversity loss is the $L_2$ norm of the current candidate to previously found triggers:

\begin{equation}
    L_{div}^{\text{filter}} = - \sum_{j=1}^{N_T} \sum_{i=1}^{j-1} ||u_{\theta^{\text{filter}}_i}-u_{\theta^{\text{filter}}_j}||_2.
\end{equation}

The regularizer term $R^{\text{filter}}(\theta^{\text{filter}})$ is simply an L2 regularizer on the MLP parameters to reduce the complexity on the color transforms:

\begin{equation}
    R^{\text{filter}}(\theta^{\text{filter}}) = ||\theta^{\text{filter}}||_2^2.
\end{equation}

\myparagraph{Feature Extractor for Color Filter Triggers.} For each model we use reverse engineering to generate $K$ classes by $N_T^{filter}$ diverse color filter triggers. For each color filter trigger, we combine an appearance descriptor with label flipping loss $L_{flip}^{\text{filter}}$ and diversity loss $L_{div}^{\text{filter}}$ as its combined feature descriptor. For the appearance descriptor, we use the same descriptor discussed in diversity: how a color filter $E(\cdot ;\theta^{\text{filter}})$ transforms 16 random $(r,g,b,i,j)$ tuples. We record the 16 output $(\hat{r},\hat{g},\hat{b})$ tuples as a 48-dim appearance descriptor.

As a result for each model, we have $N_T^{filter}K$ features for color filter triggers. 

\myparagraph{Trojan Classifier with Color Filter Triggers.} A bag-of-words model is used to aggregate those features. The aggregated features across multiple trigger classes, e.g. color filters and local triggers, are concatenated and fed through an MLP for final Trojan classification as below:

\begin{align}
    \vec{h}_i^{filter} &=MLP_\alpha^{filter}(\vec{v}_i^{filter}), \quad
    \vec{h}^{filter}=\frac{1}{N_T^{filter}K}\sum_{i}\vec{h}_i^{filter}, \quad i=1,\dot N_T^{filter}K \\
    \vec{h}_i^{local} &=MLP_\alpha^{local}(\vec{v}_i^{local}), \quad
    \vec{h}^{local}=\frac{1}{N_TK}\sum_{i}\vec{h}_i^{local}, \quad i=1,\ldots N_TK \\
    s &=MLP_\beta([\vec{h}^{filter};\vec{h}^{local}]), \quad
\end{align}

For the TrojAI datasets, color filter reverse engineering is conducted using Adam optimizer with a learning rate $3\times 10^{-2}$ for 10 iterations. Hyperparameters are set to $\lambda_1^{\text{filter}}=0.05$ and $\lambda_2^{\text{filter}}=10^{-4}$. We also set $N_T=2$ and $N_T^{filter}=8$.

\section{Experiments}
\label{experiment}
We evaluate our method on both synthetic datasets and publicly available TrojAI benchmarks. 
We provide quantitative and qualitative results, followed by ablation studies, to demonstrate the efficacy of the proposed method.
All clean/Trojaned models are DNNs trained for image classification. 

\subsection{Datasets}
\label{dataset}
\myparagraph{Synthetic Datasets (Trojaned-MNIST and Trojaned-CIFAR10)}: We adopt the codes provided by NIST~\footnote{https://github.com/trojai/trojai} to generate 200 DNNs (50\% of them are Trojaned) trained to classify MNIST and CIFAR10 data, respectively. The Trojaned models are trained with images poisoned by square triggers. The poison rate is set as 0.2.

\myparagraph{TrojAI Benchmarks (TrojAI-Round1, Round2, Round3 and Round4)}: These datasets are
provided by US IARPA/NIST~\footnote{https://pages.nist.gov/trojai/docs/data.html}, who recently organized a Trojan AI competition. Polygon triggers are generated randomly with variations in shape, size, and color. Filter-based triggers are generated by randomly choosing from five distinct filters. Trojan detection is more challenging on these TrojAI datasets as compared to Triggered-MNIST due to the use of deeper DNNs and larger variations in the appearances of foreground/background objects, trigger patterns, etc. Round1, Round2, Round3, and Round4 have 1000, 1104, 1008, and 1008 models, respectively. 

These datasets contain trained models for traffic sign classification (for each round, 50\% of total models are Trojaned). All the models are trained on synthetically created image data of non-real traffic signs superimposed on road background scenes. Trojan detection is more difficult on Round2/Round3/ Round4 compared to Round1 due to following reasons:
\begin{itemize}
\item Round2/Round3 have more number of classes: Round1 has 5 classes while Round2/Round3 have 5-25 classes. 
\item Round2/Round3 have more trigger types: Round1 only has polygon triggers, while Round2/Round3 have both polygon and Instagram filter based triggers. 
\item The number of source classes are different: all classes are poisoned in Round1, while 1, 2, or all classes are poisoned in Round2/Round3. 
\item Round2/Round3 have more type of model architectures: Round1 has 3 architectures, while Round2/Round3 have 23 architectures.
\end{itemize}

Round3 experimental design is identical to Round2 with the addition of Adversarial Training. Two different Adversarial Training approaches: Projected Gradient Descent (PGD), and Fast is Better than Free (FBF)~\cite{wong2019fast} are used.

Unlike the previous rounds, Round4 can have multiple concurrent triggers. Additionally, triggers can have conditions attached to their firing. The differences are listed as follows:
\begin{itemize}
    \item All triggers in Round4 are one to one mappings, which means a trigger flips a single source class to a single target class.
    \item Three possible conditionals, spatial, spectral, and class are attached to triggers within this dataset.
    \item Round4 has removed the very large model architectures to reduce the training time.
\end{itemize}

Round1, Round2, Round3, and Round4 have 1000, 1104, 1008, and 1008 models respectively. 

\subsection{Evaluation Metrics} 
We follow the settings in~\cite{sikka2020detecting}. We report the mean and standard deviation of two metrics: area under the ROC curve (AUC) and accuracy (ACC). Specifically, we evaluate our approach on the whole set by doing an 8-fold cross validation. For each fold, we use 80\% of the models for training, 10\% for validation, and the rest 10\% for testing.

\subsection{Baselines}

We carefully choose the methods with available codes from the authors as our baselines, including NC (Neural Cleanse)~\cite{wang2019neural}, ABS~\cite{liu2019abs}, TABOR~\cite{guo2019tabor}, ULP~\cite{kolouri2020universal}, and DLTND~\cite{wang2020practical}. 
We follow the instructions to obtain the reported results and list all the available repositories here:

\begin{itemize}
    \item \textbf{Neural Cleanse}: https://github.com/bolunwang/backdoor
    \item \textbf{ABS}: https://github.com/naiyeleo/ABS
    \item \textbf{TABOR}: https://github.com/UsmannK/TABOR
\item \textbf{ULP}: https://github.com/UMBCvision/Universal-Litmus-Patterns
\item \textbf{DLTND}: https://github.com/wangren09/TrojanNetDetector
\end{itemize}

\subsection{Implementation Details}
\label{implementaiton}
We set $\lambda_1 =1 $, $\lambda_2=10$, and $N_T=3$ for all our experiments (i.e., we generate 3 trigger candidates for each input image and each model). 
The parameters of the Trojan detection network are learned using a set of clean and Trojaned models with ground truth labeling. We train the detection network by optimizing cross entropy loss using the Adam optimizer~\cite{kingma2014adam}. The hidden state size, the number of layers of $MLP_\alpha$, $MLP_\beta$, as well as optimizer learning rate, weight decay, and the number of epochs are optimized using Bayesian hyperparameter search~\footnote{https://github.com/hyperopt/hyperopt} for 500 rounds on 8-fold cross-validation.

\subsection{Results}
\label{sec:quantitative_results}
Tab.~\ref{table:synthetic} and Tab.~\ref{table:real} show the quantitative results on the Trojaned-MNIST/ CIFAR10 and TrojAI datasets, respectively. The reported performances of baselines are reproduced using source codes provided by the authors or quoted from related papers. The best performing numbers are highlighted in bold. From Tab.~\ref{table:synthetic} and Tab.~\ref{table:real}, we observe that our method performs substantially better than the baselines. It is also worth noting that, compared with these baselines, our proposed method extracts fix-sized features for each model, independent of the number of classes, architectures, trigger types, etc. By using the extracted features, we are able to train a separate Trojan detection network, which is salable and model-agnostic. 

\setlength{\tabcolsep}{5pt}
\begin{table}[ht]
\begin{center}
\small
\caption{Comparison on Trojaned-MNIST/CIFAR10.}
\label{table:synthetic}
\begin{tabular}{cccc}

Method & Metric & Trojaned-MNIST & Trojaned-CIFAR10 \\
\hline
NC & AUC & 0.57 $\pm$ 0.07 & 0.75 $\pm$ 0.07 \\
ABS & AUC & 0.63 $\pm$ 0.04  & 0.67 $\pm$ 0.06 \\
TABOR & AUC & 0.65 $\pm$ 0.07 & 0.71 $\pm$ 0.05 \\
ULP & AUC & 0.59 $\pm$ 0.03 & 0.55 $\pm$ 0.03 \\
DLTND & AUC & 0.62 $\pm$ 0.05 & 0.52 $\pm$ 0.08 \\
Ours  & AUC & \textbf{0.88 $\pm$ 0.04} & \textbf{0.91 $\pm$ 0.05} \\
\hline
NC & ACC & 0.60 $\pm$ 0.04 & 0.73 $\pm$ 0.06 \\
ABS & ACC & 0.65 $\pm$ 0.02 & 0.69 $\pm$ 0.04\\
TABOR& ACC & 0.62 $\pm$ 0.04 & 0.69 $\pm$ 0.08 \\
ULP& ACC & 0.57 $\pm$ 0.02 & 0.59 $\pm$ 0.06 \\
DLTND& ACC & 0.64 $\pm$ 0.07 & 0.55 $\pm$ 0.07 \\
Ours& ACC & \textbf{0.89 $\pm$ 0.02}  & \textbf{0.92 $\pm$ 0.04} \\
\hline
\end{tabular}
\end{center}
\end{table}

\setlength{\tabcolsep}{4pt}
\begin{table*}[ht]
\begin{center}
\footnotesize
\caption{Performance comparison on the TrojAI dataset.}
\label{table:real}
\begin{tabular}{cccccc}

Method & Metric & TrojAI-Round1 & TrojAI-Round2 & TrojAI-Round3 & TrojAI-Round4\\
\hline
NC & AUC & 0.50 $\pm$ 0.03 & 0.63 $\pm$ 0.04 & 0.61 $\pm$ 0.06 & 0.58 $\pm$ 0.05\\
ABS & AUC & 0.68 $\pm$ 0.05 & 0.61 $\pm$ 0.06 & 0.57 $\pm$ 0.04 & 0.53 $\pm$ 0.06 \\
TABOR & AUC & 0.71 $\pm$ 0.04 & 0.66 $\pm$ 0.07 & 0.50 $\pm$ 0.07 & 0.52 $\pm$ 0.04 \\
ULP & AUC & 0.55 $\pm$ 0.06 & 0.48 $\pm$ 0.02 & 0.53 $\pm$ 0.06 & 0.54 $\pm$ 0.02 \\
DLTND & AUC & 0.61 $\pm$ 0.07 & 0.58 $\pm$ 0.04 & 0.62 $\pm$ 0.07 & 0.56 $\pm$ 0.05 \\
Ours & AUC  & \textbf{0.90 $\pm$ 0.02}  & \textbf{0.87 $\pm$ 0.05}  & \textbf{0.89 $\pm$ 0.04}  &  \textbf{0.92 $\pm$ 0.06} \\
\hline
NC & ACC & 0.53 $\pm$ 0.04 & 0.49 $\pm$ 0.02 & 0.59 $\pm$ 0.07 & 0.60 $\pm$ 0.04 \\

ABS & ACC & 0.70 $\pm$ 0.04 & 0.59 $\pm$ 0.05 & 0.56 $\pm$ 0.03 & 0.51 $\pm$ 0.05 \\

TABOR & ACC & 0.70 $\pm$ 0.03 & 0.68 $\pm$ 0.08 & 0.51 $\pm$ 0.05 & 0.55 $\pm$ 0.06 \\

ULP & ACC & 0.58 $\pm$ 0.07 & 0.51 $\pm$ 0.03 & 0.56 $\pm$ 0.04 & 0.57 $\pm$ 0.04 \\

DLTND & ACC & 0.59 $\pm$ 0.04 & 0.61 $\pm$ 0.05 & 0.65$\pm$ 0.04 & 0.59 $\pm$  0.06\\
Ours & ACC & \textbf{0.91 $\pm$ 0.03} & \textbf{0.89 $\pm$ 0.04} & \textbf{0.90 $\pm$ 0.03} &  \textbf{0.91 $\pm$ 0.04} \\
\hline
\end{tabular}
\end{center}
\end{table*}

\begin{figure*}[ht]
\centering 
\subfigure{
\includegraphics[width=0.145\textwidth]{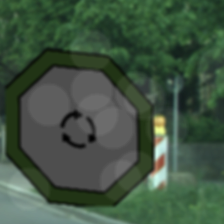}}
\subfigure{
\includegraphics[width=0.145\textwidth]{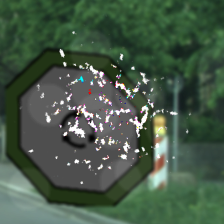}}
\subfigure{
\includegraphics[width=0.145\textwidth]{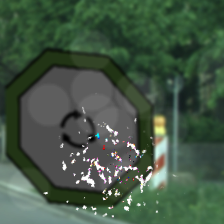}}
\subfigure{
\includegraphics[width=0.145\textwidth]{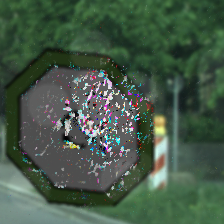}}
\subfigure{
\includegraphics[width=0.145\textwidth]{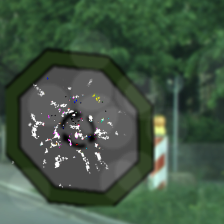}}
\subfigure{
\includegraphics[width=0.145\textwidth]{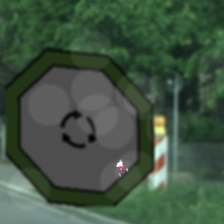}}

\subfigure{
\includegraphics[width=0.145\textwidth]{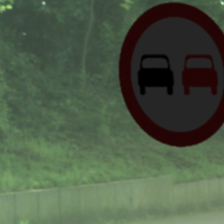}}
\subfigure{
\includegraphics[width=0.145\textwidth]{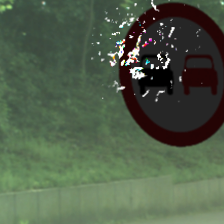}}
\subfigure{
\includegraphics[width=0.145\textwidth]{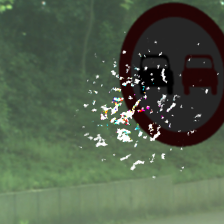}}
\subfigure{
\includegraphics[width=0.145\textwidth]{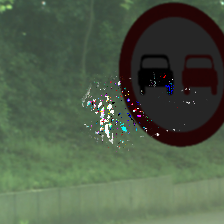}}
\subfigure{
\includegraphics[width=0.145\textwidth]{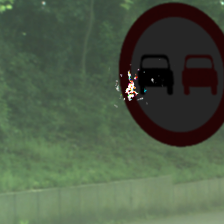}}
\subfigure{
\includegraphics[width=0.145\textwidth]{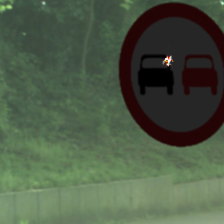}}

\subfigure{
\stackunder{\includegraphics[width=0.145\textwidth]{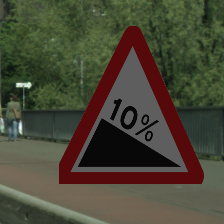}}{(a)}}
\subfigure{
\stackunder{\includegraphics[width=0.145\textwidth]{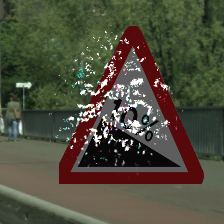}}{(b)}}
\subfigure{
\stackunder{\includegraphics[width=0.145\textwidth]{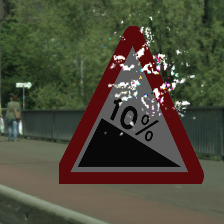}}{(c)}}
\subfigure{
\stackunder{\includegraphics[width=0.145\textwidth]{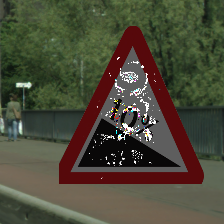}}{(d)}}
\subfigure{
\stackunder{\includegraphics[width=0.145\textwidth]{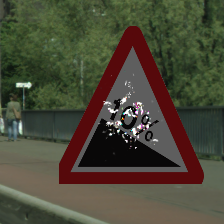}}{(e)}}
\subfigure{
\stackunder{\includegraphics[width=0.14\textwidth]{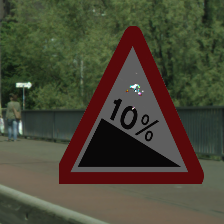}}{(f)}}
\caption{Examples of recovered triggers overlaid on clean images. From left to right: \textbf{(a)} clean image, \textbf{(b)} triggers recovered by~\cite{wang2019neural}, \textbf{(c)} triggers recovered by ~\cite{liu2019abs}, \textbf{(d)} triggers recovered by ~\cite{guo2019tabor}, \textbf{(e)} triggers recovered by our method without topological prior, and \textbf{(f)} triggers recovered by our method with topological prior.}
\label{fig:Qualitative}
\end{figure*}

Fig.~\ref{fig:Qualitative} shows a few examples of recovered triggers. We observe that, compared with the baselines, the triggers found by our method are more compact and of better quality. This is mainly due to the introduction of topological constraints. The improved quality of recovered triggers directly results in improved performance of Trojan detection. 

\subsection{Ablation Studies}
\myparagraph{Ablation Study of Loss Weights}: For the loss weights $\lambda_1$ and $\lambda_2$, we empirically choose the weights which make reverse engineering converge the fastest. This is a reasonable choice as in practice, time is one major concern for reverse engineering pipelines.

\begin{figure}[ht]
\centering 
    \includegraphics[width=0.55\textwidth]{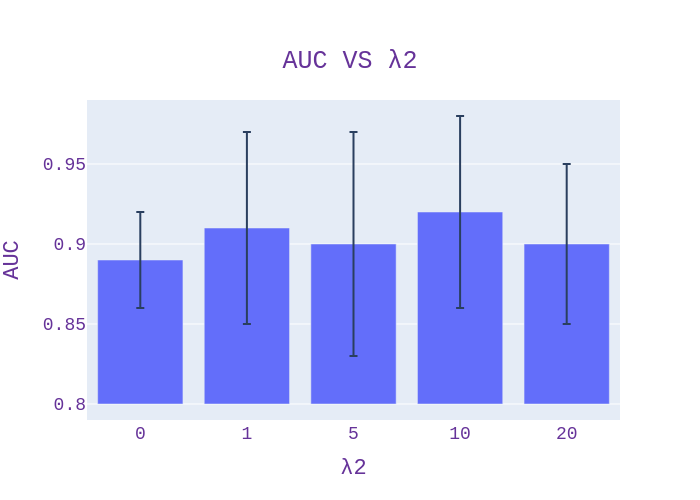}
  \caption{Ablation study results for $\lambda_2$.}
\label{fig:approximation_trojan}
\end{figure}

Despite the seemingly ad hoc choice, we have observed that our performances are quite robust to all these loss weights. As topological loss is a major contribution of this chapter, we conduct an ablation study in terms of its weight ($\lambda_2$) on TrojAI-Round4 dataset. The results are reported in Fig.~\ref{fig:approximation_trojan}. We observe that the proposed method is quite robust to $\lambda_2$, and when $\lambda=10$, it achieves slightly better performance (AUC: 0.92 $\pm$ 0.06) than other choices. 

\myparagraph{Ablation Study of Number of Training Model Samples}: The trigger features and Trojan detection network are important in achieving SOTA performance.
To further demonstrate the efficacy of the proposed diversity and topological loss terms, we conduct another ablation study to investigate the case with fewer training model samples, and thus a weaker Trojan detection network.

The ablation study in terms of the number of training samples on TrojAI-Round4 data is illustrated in Tab.~\ref{table:number_of_sample}. We observe that the proposed topological loss and diversity loss will boost the performance with/without a fully trained Trojan-detection network. These two losses improve the quality of the recovered trigger, in spite of how the trigger information is used. Thus even with a limited number of training samples (e.g., 25), the proposed method could still achieve significantly better performance than the baselines.

\setlength{\tabcolsep}{5pt}
\begin{table}[ht]
\begin{center}
\small
\caption{Ablation study for \# of training samples.}
\label{table:number_of_sample}
\begin{tabular}{cccc}

\# of samples & Ours & w/o topo	 & w/o diversity\\
\hline
25 & \textbf{0.77 $\pm$ 0.04} & 0.73 $\pm$ 0.03 & 0.68 $\pm$ 0.04 \\
50 & \textbf{0.81 $\pm$ 0.03} & 0.76 $\pm$ 0.05 & 0.73 $\pm$ 0.02 \\
100 & \textbf{0.84 $\pm$ 0.05} & 0.78 $\pm$ 0.06 & 0.76 $\pm$ 0.03 \\
200 & \textbf{0.86 $\pm$ 0.04} & 0.82 $\pm$ 0.04 & 0.79 $\pm$ 0.05 \\
400 & \textbf{0.90 $\pm$ 0.05} & 0.85 $\pm$ 0.03 & 0.82 $\pm$ 0.04 \\
800 & \textbf{0.92 $\pm$ 0.06} & 0.89 $\pm$ 0.04 & 0.85 $\pm$ 0.02 \\
\hline
\end{tabular}
\end{center}
\end{table}

\myparagraph{Ablation Study for Loss Terms}: 
We investigate the individual contribution of different loss terms used to search for latent triggers.
Tab.~\ref{table:ablation} lists the corresponding performance on the TrojAI-Round4 dataset. We observe a decrease in AUC (from 0.92 to 0.89) if the topological loss is removed. This drop is expected as the topological loss helps to find more compact triggers. Also, the performance drops significantly (from 0.92 to 0.85 in AUC) if the diversity loss is removed. We also report the performance by setting $N_T=2$; when $N_T=2$, the performance increases from 0.85 to 0.89 in AUC. The reason is that with diversity loss, we are able to generate multiple diverse trigger candidates, which increases the probability of recovering the true trigger when the target class is unknown. Our ablation study justifies the use of both diversity and topological losses. 

\setlength{\tabcolsep}{5pt}
\begin{table}[ht]
\begin{center}
\small
\caption{Ablation results of loss terms.}
\label{table:ablation}

\begin{tabular}{cc}

Method & TrojAI-Round4 \\
\hline
w/o topological loss & 0.89 $\pm$ 0.04 \\
w/o diversity loss ($N_T=1$) &  0.85 $\pm$ 0.02 \\
$N_T$ = 2 &  0.89 $\pm$ 0.05 \\
with all loss terms ($N_T=3$)  & \textbf{0.92 $\pm$ 0.06} \\
\hline
\end{tabular}
\end{center}
\end{table}

In practice, we found that topological loss can improve the convergence of trigger searches. 
Without topological loss, it takes $\approx$50 iterations to find a reasonable trigger (Fig.~\ref{fig:Qualitative}(e)). In contrast, with the topological loss, it takes only $\approx$30 iterations to converge to a better recovered trigger (Fig.~\ref{fig:Qualitative}(f)). The rationale is that, as the topological loss imposes strong constraints on the number of connected components, it largely reduces the search space of triggers, consequently, making the convergence of trigger search much faster. This is worth further investigation.

\subsection{Unsupervised Setting for Trojan Detection}
\label{sec:unsupervised}
\setlength{\tabcolsep}{6pt}
\begin{table}[ht]
\small
\begin{center}
\caption{Unsupervised performances on Trojan.}
\label{table:unsupervised}
\begin{tabular}{ccc}

Method & \textbf{AUC} & \textbf{ACC} \\
\hline
NC & 0.58 & 0.60 \\
ABS &  0.53 & 0.51 \\
TABOR &  0.52 & 0.55 \\
ULP &  0.54 & 0.57 \\
DLTND &  0.56 & 0.59 \\

Ours &  \textbf{0.63} & \textbf{0.65} \\
\hline
\end{tabular}
\end{center}
\end{table}

Indeed, our method outperforms existing methods in different settings: fully supervised settings with annotated models, and unsupervised settings. In (Tab.~\ref{table:number_of_sample}), we have already demonstrated that with a limited number of annotated models, our method outperformed others. To make a fair comparison, following Neural Cleanse and DLTND, we also use the simple technique based on Median Absolute Deviation (MAD) for trojan detection. We report the performance of Round 4 data of TrojAI in Tab.~\ref{table:unsupervised}.

Because of the better trigger quality, due to the proposed losses, our method outperforms baselines such as Neural Cleanse. We also note that, in Tab.~\ref{table:unsupervised}, all methods (including ours) perform unsatisfactorily in the unsupervised setting. This brings us back to the discussion as to whether a supervised setting is justified in Trojan detection (although this is not directly relevant to our method).

From the research point of view, we believe that data-driven methods for Trojan detection are unavoidable as the attack techniques continue to develop. Like in many other security research problems, the Trojan attack and defense are two sides of the same problem that are supposed to advance together. When the problem was first studied, classic unsupervised methods such as Neural Cleanse are sufficient. In recent years, the attack techniques have continued to develop, exploiting the entire dataset and leveraging techniques like adversarial training. Meanwhile, detection methods are confined to only the given model and a few sample data. For defense methods to move forward and to catch up with the attack methods, it seems only natural and necessary to exploit supervised approaches, e.g., learning patterns from public datasets such as the TrojAI benchmarks.

\section{Conclusion}
\label{conclusion}
In this chapter, we propose a diversity loss and a topological prior to improve the quality of the trigger reverse engineering for Trojan detection. These loss terms help to find high quality triggers efficiently. They also avoid the dependence of the method on the target label. On both synthetic datasets and publicly available TrojAI benchmarks, our approach recovers high quality triggers and achieves SOTA Trojan detection performance.

\clearpage

%% file: warping.tex
\chapter{Structure-Aware Image Segmentation with Homotopy Warping}
\label{chapter:warping}
In Chapter~\ref{chapter:topoloss}, we introduced the persistent-homology based topological loss. The results are promising, while the identified critical points are very noisy and are often not relevant to the topological errors.
In this chapter,
we introduce another warping strategy to efficiently identify critical points.

\section{Introduction}
\label{sec:intro}

\begin{figure}[ht]
\centering 
\subfigure[Image]{
\includegraphics[width=0.25\textwidth]{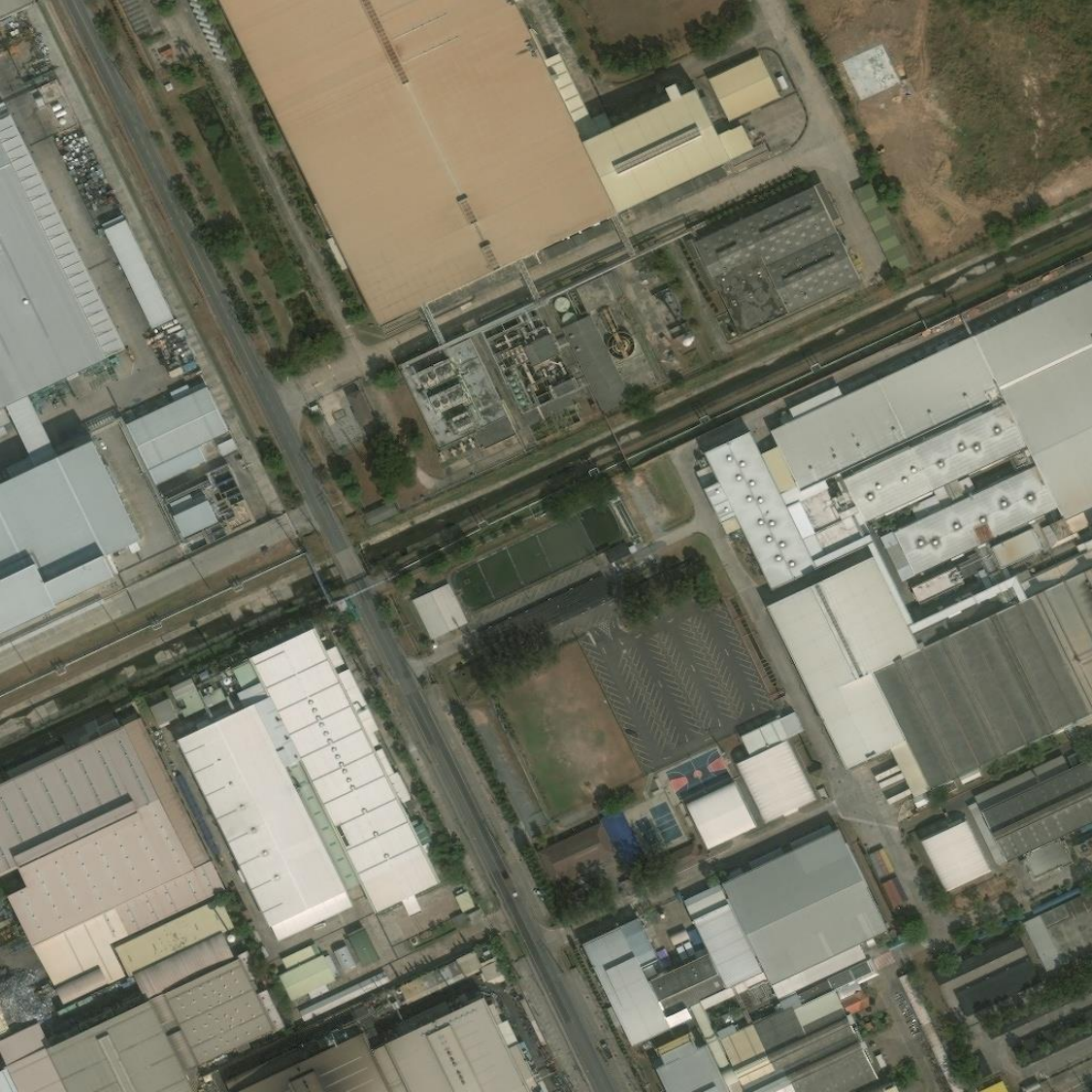}}
\subfigure[GT mask]{
\includegraphics[width=0.25\textwidth]{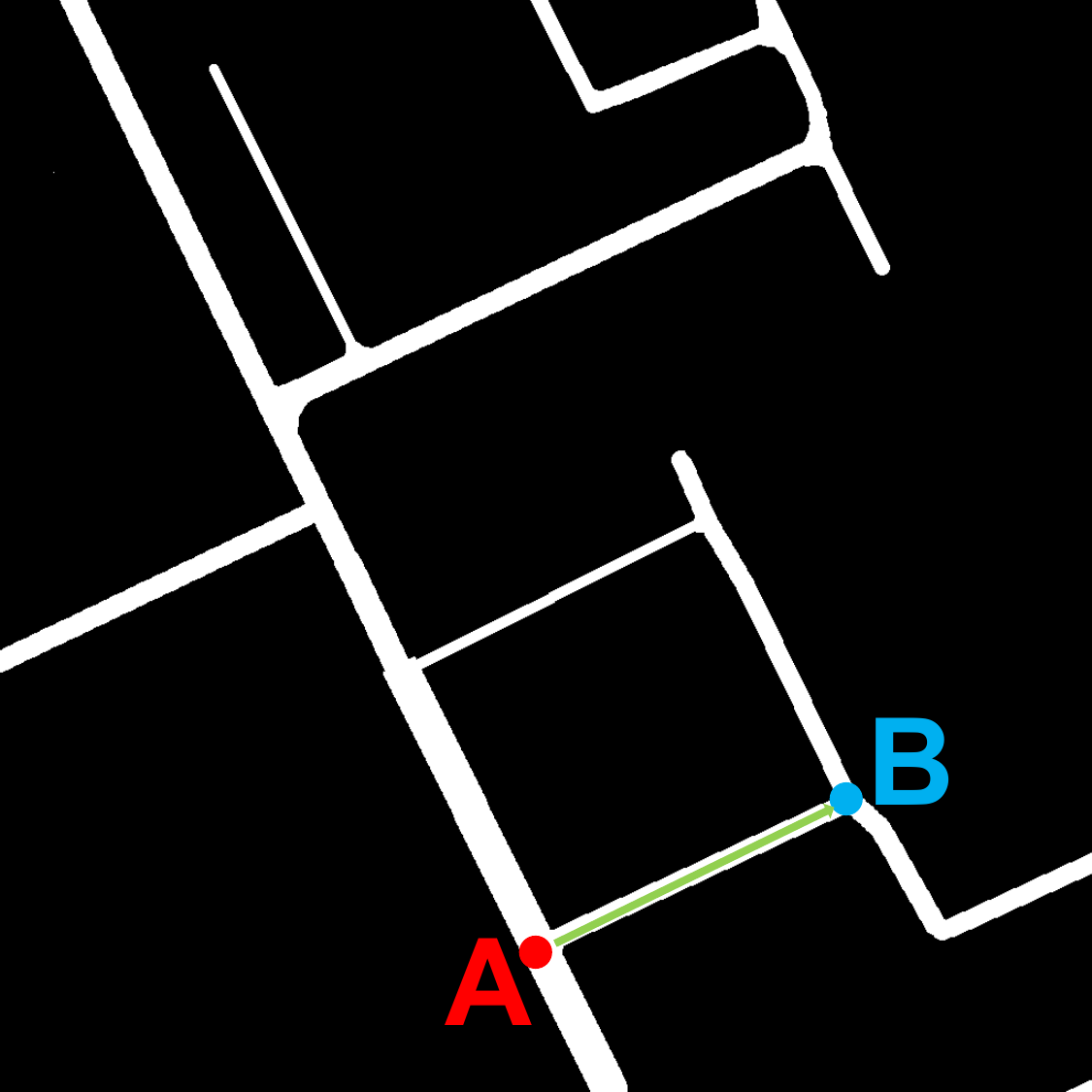}}
\subfigure[UNet pred.]{
\includegraphics[width=0.25\textwidth]{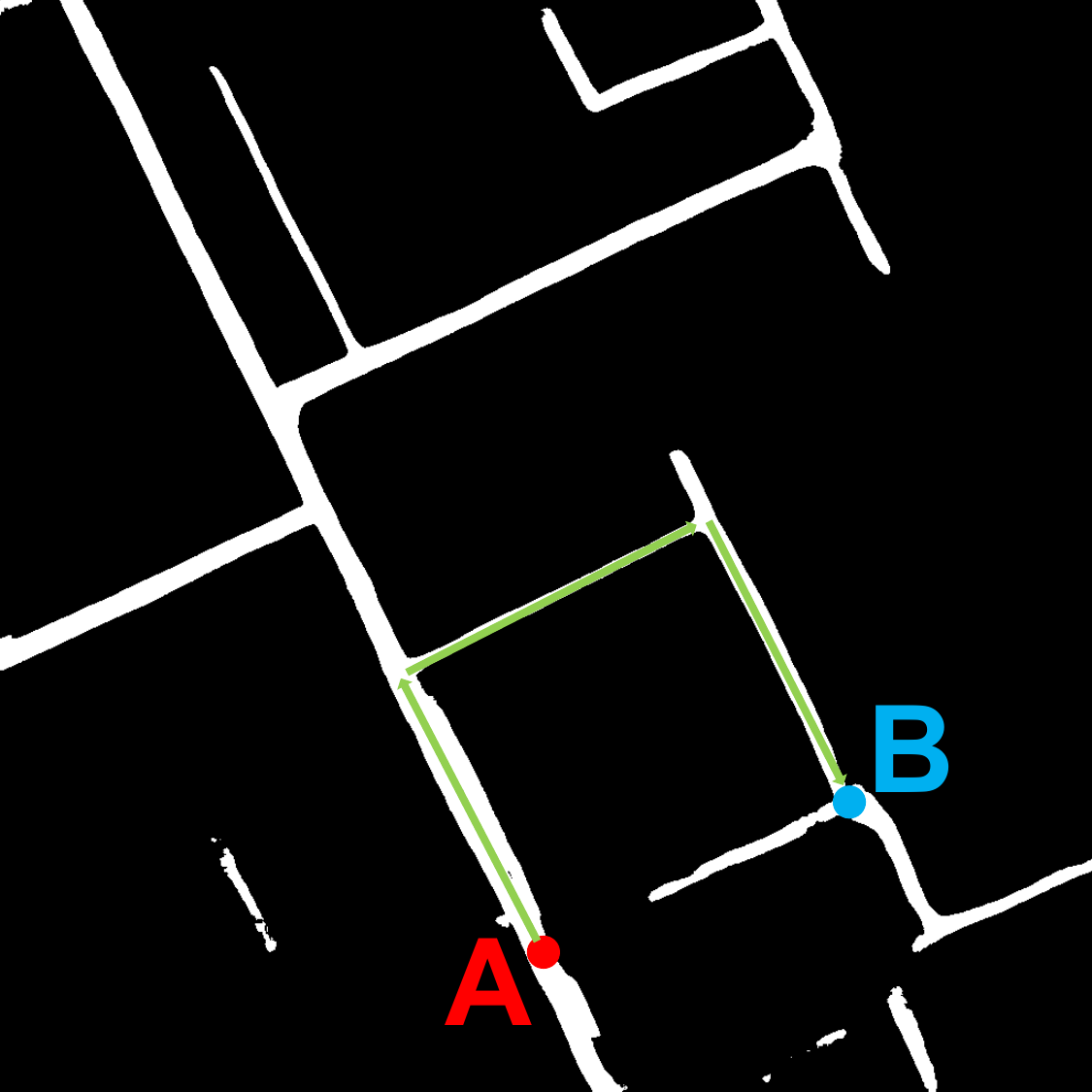}}
\caption{An illustration for the importance of topological correctness. If one wants to go to point $B$ from $A$, the shortest path in the GT is illustrated in \textbf{(b)} (the green path). However, in the result predicted by UNet, though only a few pixels are misclassified, the shortest path from $A$ to $B$ is totally different, which is illustrated by the green path in \textbf{(c)}. Zoom-in for better viewing.}
\label{fig:teaser_warping}
\end{figure}

Image segmentation with topological correctness is a challenging problem,  especially for images with fine-scale structures, e.g., satellite images, neuron images and vessel images. 
Deep learning methods have delivered strong performances in image segmentation tasks~\cite{long2015fully,he2017mask,chen2014semantic,chen2018deeplab,chen2017rethinking}. However, even with satisfying per-pixel accuracy, most existing methods are still prone to topological errors, i.e., broken connections, holes in 2D membranes, missing connected components, etc.  These errors may significantly impact downstream tasks. For example, the reconstructed road maps from satellite images can be used for navigation~\cite{barzohar1996automatic,batra2019improved, van2018spacenet, wegner2013higher, biagioni2012inferring, wiedemann1998empirical, mattyus2017deeproadmapper, vasu2020topoal, mosinska2019joint, yang2019road}. A small amount of pixel errors will result in broken connections, causing incorrect navigation routes. See Fig.~\ref{fig:teaser_warping} for an illustration. In neuron reconstruction~\cite{funke2018large,januszewski2018high,uzunbas2016efficient,ye2019diverse,yang2021topological}, the incorrect topology of the neuron membrane will result in erroneous merge or split of neurons, and thus errors in morphology and connectivity analysis of neuronal circuits.

 Topological errors usually happen at challenging locations, e.g., weak connections or blurred locations. 
But not all challenging locations are topologically relevant; for example, pixels at the boundary of the object of interest can generally be challenging, but not relevant to topology.
 To truly suppress topological errors, we need to focus on \emph{topologically critical pixels}, i.e., challenging locations that are topologically relevant.
 Without identifying and targeting these locations, neural networks that are optimized for standard pixel-wise losses (e.g., cross-entropy loss or mean-square-error loss) cannot avoid topological errors, even if we increase the training set size. 

Existing works have targeted these topologically critical pixels. The closest method to our work is TopoNet~\cite{hu2019topology}, which is based on persistent homology \cite{edelsbrunner2000topological,edelsbrunner2010computational}. The main idea is to identify topologically critical pixels corresponding to critical points of the likelihood map predicted by the neural network. The selected critical points are reweighed in the training loss to force the neural network to focus on them, and thus to avoid topological errors. But there are two main issues with this approach: {1}) The method is based on the likelihood map, which can be noisy with a large amount of irrelevant critical points. This leads to inefficient optimization during training. {2}) The computation for persistent homology is cubic to the image size. It is too expensive to recompute at every iteration. 

In this chapter, we propose a novel approach to identify topologically critical pixels in a more efficient and accurate manner. These locations are penalized in the proposed \emph{homotopy warping loss} to achieve better topological accuracy. 
Our method is partially inspired by the warping error previously proposed to evaluate the topological accuracy~\cite{jain2010boundary}. 
Given a binary predicted mask $f_B$ and a ground truth mask $g$, we ``warp'' one towards another without changing its topology. From the view of topology, to warp mask $f_B$ towards $g$, we find a mask $f_B^{*}$ that is homotopy equivalent to $f_B$ and is as close to $g$ as possible \cite{hatcher2002algebraic}. The difference between the warped mask $f_B^{*}$ and $g$ constitutes the topologically critical pixels. We can also warp $g$ towards $f_B$ and find another set of topologically critical pixels. See Fig.~\ref{fig:warping-synthetic} and Fig.~\ref{fig:error} for illustrations. These locations directly correspond to the topological difference between the prediction and the ground truth, tolerating geometric deformations. Our homotopy warping loss targets them to fix the topological errors of the model.

The warping of a mask is achieved by iteratively flipping labels at pixels without changing the topology of the mask.
These flippable pixels are called \emph{simple points/pixels} in the classic theory of digital topology \cite{kong1989digital}. Note that in this chapter, we focus on the topology of binary masks, simple points, and simple pixels can be used interchangeably.
Finding the optimal warping of a mask towards another mask is challenging due to the huge search space. To this end, we propose a new heuristic method that is computationally efficient. 
We filter the image domain with the distance transform and flip simple pixels based on their distance from the mask. This algorithm is proven efficient and delivers high quality locally optimal warping results.

Overall, our contributions can be summarized as follows:
\begin{itemize}
    \item We propose a novel \textit{homotopy warping loss}, which penalizes errors on topologically critical pixels. These locations are defined by homotopic warping of predicted and ground truth masks. The loss can be incorporated into the training of topology-preserving deep segmentation networks. 
    \item By exploiting distance transforms of binary masks, we propose a novel homotopic warping algorithm to identify topologically critical pixels in an efficient manner. This is essential in incorporating the homotopy warping loss into the training of deep nets.
\end{itemize}
Our loss is a plug-and-play loss function. It can be used to train any segmentation network to achieve better performance in terms of topology. We conduct experiments on both 2D and 3D benchmarks to demonstrate the efficacy of the proposed method. Our method performs strongly in multiple topology-relevant metrics (e.g., ARI, Warping Error, and Betti Error).
 We also conduct several ablation studies to further demonstrate the efficiency and effectiveness of the technical contributions.

\section{Method}
\label{sec:method_warping}

\begin{figure}[ht]
\centering
\includegraphics[width=0.8\textwidth]{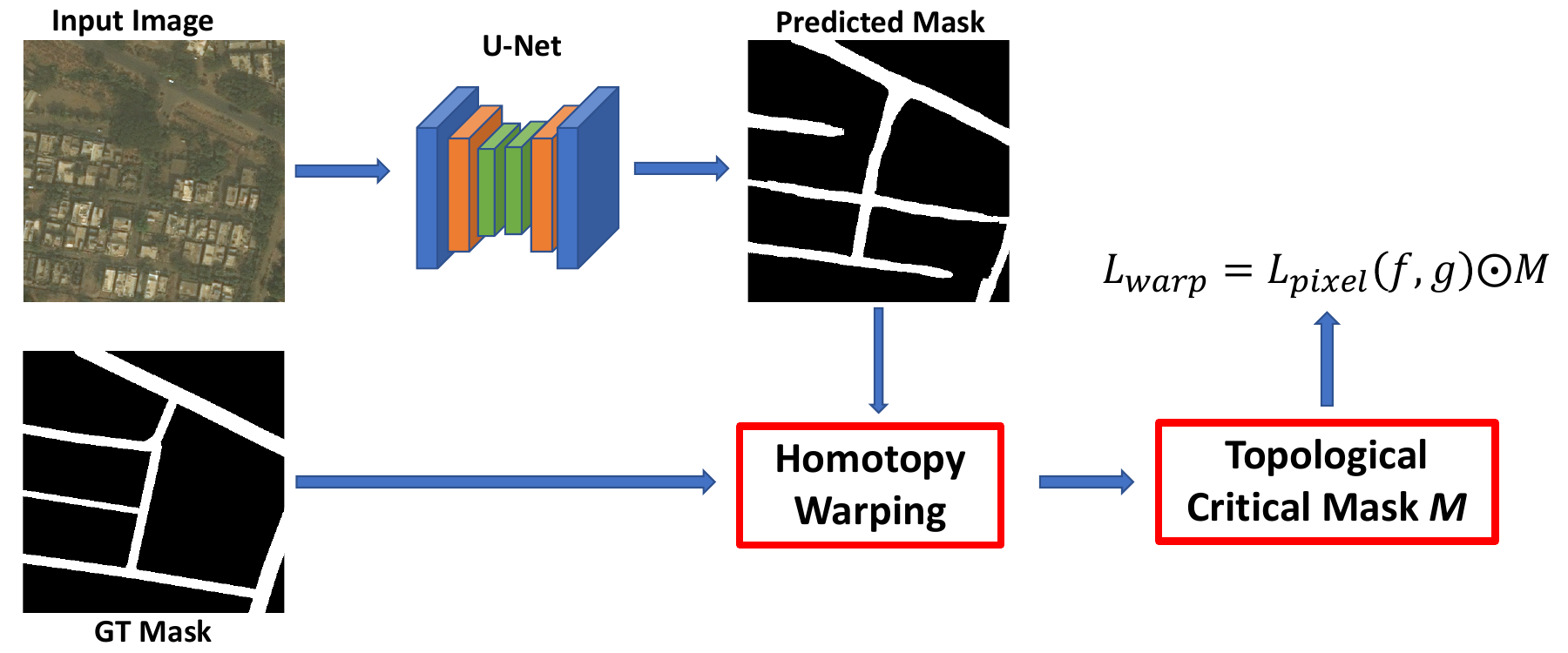}
\caption{The illustration of the proposed \textit{homotopy warping loss} $L_{warp}$. The homotopy warping algorithm tries to identify the topological critical pixels via the binary mask instead of noisy likelihood maps. These identified topological critical pixels/masks are used to define a new loss that is complementary to standard pixel-wise loss functions. The details of \textit{Homotopy Warping} and \textit{Topological Critical Mask M} can be found in Sec.~\ref{sec:warp_error} and Sec.~\ref{warpingloss}, respectively.}
  \label{fig:framework_warping}
\end{figure}

By warping the binary predicted mask to the ground truth or conversely, we can accurately and efficiently identify the critical pixels. And then we propose a novel homotopy warping loss that targets them to fix topological errors of the model. The overall framework is illustrated in Fig.~\ref{fig:framework_warping}.

This section is organized as follows. We will start with the necessary definitions and notations. In Sec.~\ref{topology}, we give a concise description of digital topology and simple points. Next, we analyze different types of warping errors in Sec.~\ref{sec:warp_error}. The proposed warping loss is introduced in Sec.~\ref{warpingloss}. And finally, we explain the proposed new warping algorithm in Sec.~\ref{distance}.

\subsection{Digital Topology and Simple Points}
\label{topology}

In this section, we briefly introduce the simple point definition from the classic digital topology~\cite{kong1989digital}. We focus on the 2D setting, whereas all definitions generalize to 3D. Details on 3D images are provided in Sec.~\ref{simple_3d}.

\begin{figure}[ht]
\centering 
\subfigure[4-adjacent]{
\includegraphics[width=0.23\textwidth]{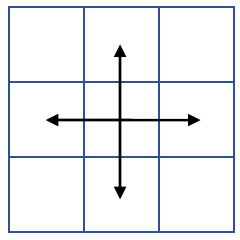}}
\subfigure[8-adjacent]{
\includegraphics[width=0.23\textwidth]{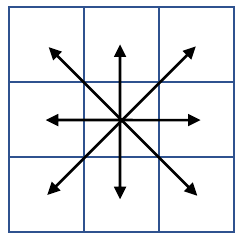}}
\subfigure[Simple Point]{
\includegraphics[width=0.23\textwidth]{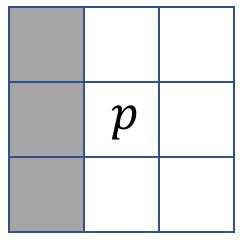}}
\subfigure[Non-simple]{
\includegraphics[width=0.23\textwidth]{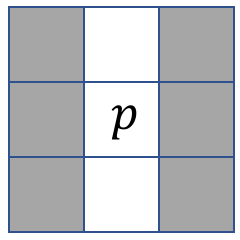}}
\caption{Illustration for 4, 8-adjacency, simple and non-simple points. \textbf{(a)}: 4-adjacency. \textbf{(b)}: 8-adjacency. \textbf{(c)}: a simple point $p$. White and grey pixels are FG and BG, respectively. Flipping the label of $p$ will not change the topology. \textbf{(d)}: a non-simple point $p$. Flipping $p$ will change the topology.}
\label{fig:simple}
\end{figure}

\myparagraph{Connectivities of Pixels.} To discuss the topology of a 2D binary image, we first define the connectivity between pixels. See Fig.~\ref{fig:simple} for an illustration. A pixel $p$ has 8 pixels surrounding it. We can either consider the 4 pixels that share an edge with $p$ as $p$'s neighbors (called \emph{4-adjacency}), or consider all 8 pixels as $p$'s neighbors (called \emph{8-adjacency}). To ensure the Jordan closed curve theorem holds, one has to use one adjacency for foreground (FG) pixels, and the other adjacency for the background (BG) pixels. In this chapter, we use 4-adjacency for FG and 8-adjacency for BG. For 3D binary images, we use 6-adjacency for FG and 26-adjacency for BG.
Denote by $N_4(p)$ the set of 4-adjacency neighbors of $p$, and $N_8(p)$ the set of 8-adjacency neighbors of $p$.

\myparagraph{Simple Points.}
For a binary image (2D/3D), a pixel/voxel is called a \textit{simple point} if it could be flipped from foreground (FG) to background (BG), or from BG to FG, without changing the topology of the image~\cite{kong1989digital}.  The following definition can be used to determine whether a point is simple:

\textbf{Definition 1} (Simple Point Condition~\cite{kong1989digital})
\textit{Let $p$ be a point in a 2D binary image. Denote by $F$ the set of FG pixels. Assume 4-adjacency for FG and 8-adjacent for BG. $p$ is a simple point if and only if both of the following conditions hold: 1) $p$ is 4-adjacent to just one FG connected component in $N_8(p)$, and 2) $p$ is 8-adjacent to just one BG connected component in $N_8(p)$.}

See Fig.~\ref{fig:simple} for an illustration of simple and non-simple points in 2D cases. 
It is easy to check if a pixel $p$ is simple or not by inspecting its $3 \times 3$ neighboring patch.
The \textbf{Definition 1} can also generalize to the 3D setting with 6- and 26-adjacencies for FG and BG, respectively. 

\subsection{Simple Points in 3D}
\label{simple_3d}

For a 3D binary image, a voxel is called a \textit{simple point} if it could be flipped from foreground (FG) to background (BG), or from BG to FG, without changing the topology of the image~\cite{kong1989digital, bertrand1994new}.  The following definition is a natural extension of the \textbf{Definition 1} in 2D case. It can be used to determine whether a voxel is simple:

\textbf{Definition 2} (Simple Point Condition~\cite{kong1989digital})
\textit{Let $p$ be a point in a 3D binary image. Denote by $F$ the set of FG pixels. Assume 6-adjacency for FG and 26-adjacent for BG.}



\subsection{Homotopic Warping Error}
\label{sec:warp_error}
In this section, we introduce the homotopic warping of one mask towards another. 
We warp a mask through a sequence of flipping of simple points. Since we only flip simple points, by definition the warped mask will have the same topology.~\footnote{Note that it is essential to flip these simple points sequentially. The simple/non-simple status of a pixel may change if other adjacent pixels are flipped. Therefore, flipping a set of simple points \textit{simultaneously} is not necessarily topology-preserving.} 
The operation is called a \textit{homotopic warping}. It has been proven that two binary images with the same topology can always be warped into each other by flipping a sequence of simple points~\cite{rosenfeld1998topology}. 

\begin{figure}[ht]
    \centering
     \setlength{\tabcolsep}{0.5pt}
    \begin{tabular}{cccccc}
    \includegraphics[width=.16\textwidth]{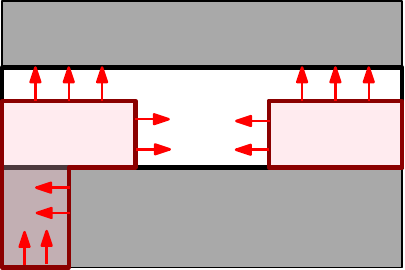} &
    \includegraphics[width=.16\textwidth]{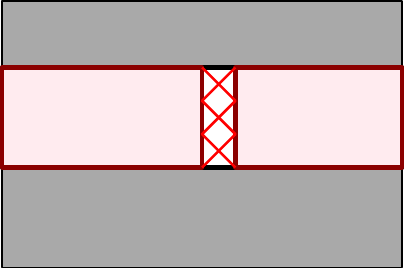} & 
    \includegraphics[width=.16\textwidth]{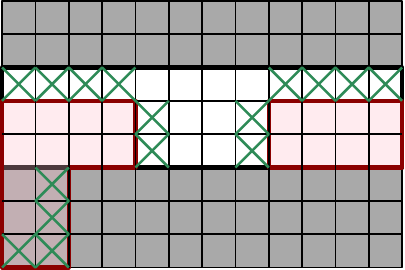} &
    \includegraphics[width=.16\textwidth]{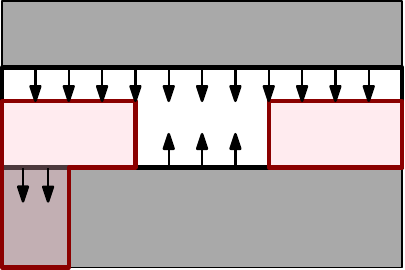} &
    \includegraphics[width=.16\textwidth]{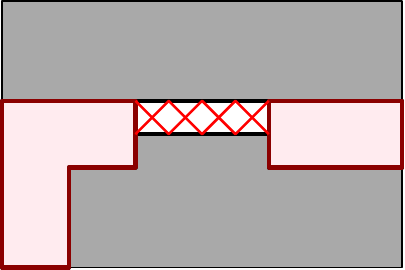} &
    \includegraphics[width=.16\textwidth]{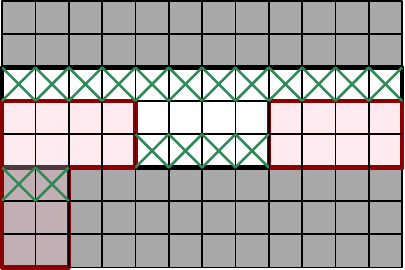} \\
    (a) & (b) & (c) & (d) & (e) & (f)
    \end{tabular}
    \caption{Illustration of homotopic warping between two masks, red and white. If red is the FG of the prediction, this is a false negative topological error; if red is the FG of the ground truth, this is a false positive error. \textbf{(a-c):} warping the red mask towards the white. \textbf{(a):} arrows show the warping direction. \textbf{(b):} the final mask after warping. Only a single-pixel wide gap remains in the middle of the warped red mask. The non-simple/critical pixels are highlighted with red crosses. They correspond to the topological error and will be penalized for the loss. \textbf{(c):} At the beginning of the warping, we highlight (with green crosses) simple points that can be flipped according to our algorithm. \textbf{(d-f):} warping the white mask towards the red mask. Only a single-pixel wide connection remains to ensure the warped white mask is connected. The non-simple/critical pixels are highlighted with red crosses.}
    \label{fig:warping-synthetic}
\end{figure}

Consider two input masks, the prediction mask, and the ground truth mask.
We can warp one of them (source mask) into another (target mask) in the best way possible, i.e., the warped mask has a minimal number of differences from the target mask (formally, the minimal Hamming distance).
Once the warping is finished, the pixels at which the warped mask is different from the target mask, called \emph{critical pixels}, are a sparse set of pixels indicative of the topological errors of the prediction mask.

We will warp in both directions: from the prediction mask to the ground truth mask, and the opposite. They identify different sets of critical pixels for the same topological error. In Fig.~\ref{fig:warping-synthetic}, we show a synthetic example with red and white masks, as well as warping in both directions. Warping the red mask towards the white mask ((a) and (b)) results in a single-pixel wide gap. The pixels in the gap (highlighted with red crosses) are critical pixels; flipping any of them will change the topology of the warped red mask. Warping the white mask towards the red mask ((d) and (e)) results in a single pixel wide link connecting the warped white mask. All pixels along the link (highlighted with red crosses) are critical; flipping any of them will change the topology of the warped white mask. 

Here if the red mask is the prediction mask, then this corresponds to a false negative connection, i.e., a connection that is missed by the prediction. If the red mask is the ground truth mask, then this corresponds to a false positive connection. Note the warping ensures that \emph{only topological errors are represented by the critical pixels}. In the synthetic example (Fig.~\ref{fig:warping-synthetic}), the large area of error in the bottom left corner of the image is completely ignored as it is not topologically relevant. 

\begin{figure*}[ht]
\centering 
\subfigure[GT]{
\includegraphics[width=0.18\textwidth]{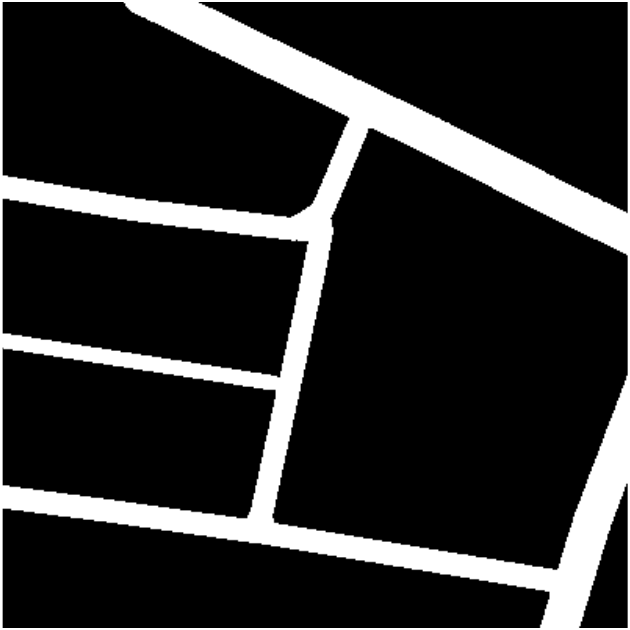}}
\subfigure[Pred.]{
\includegraphics[width=0.18\textwidth]{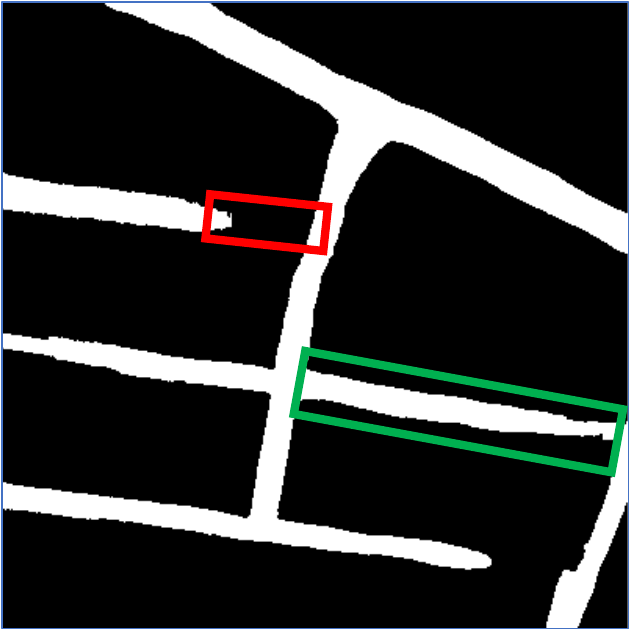}}
\subfigure[Warp GT]{
\includegraphics[width=0.18\textwidth]{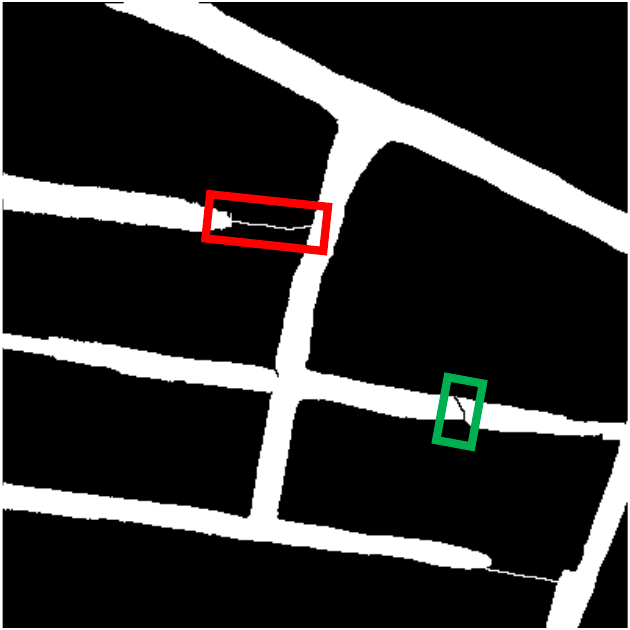}}
\subfigure[Zoom-in]{
\includegraphics[width=0.18\textwidth]{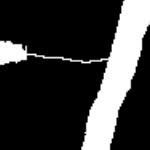}}
\subfigure[Zoom-in]{
\includegraphics[width=0.18\textwidth]{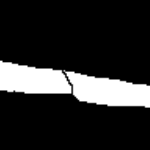}}
\caption{Illustration of warping in a real world example (satellite image). \textbf{(a)} GT mask. \textbf{(b)} The prediction mask. The red box highlights a \textit{false negative connection}, and the green box highlights a \textit{false positive connection}. \textbf{(c)} Warped GT mask (using the prediction mask as the target). \textbf{(d)} Zoomed-in view of the red box in \textbf{(c)}. \textbf{(e)} Zoomed-in view of the green box in \textbf{(c)}. }
\label{fig:error}
\end{figure*}

\begin{figure}[ht]
  \centering
\subfigure[GT]{
   \includegraphics[width=.3\textwidth]{warping/figures/gt_illu.pdf}}
   \subfigure[Prediction]{
     \includegraphics[width=.3\textwidth]{warping/figures/pre_illu.pdf}}
   \subfigure[Warped pred. $f_B^{*}$]{
     \includegraphics[width=.3\textwidth]{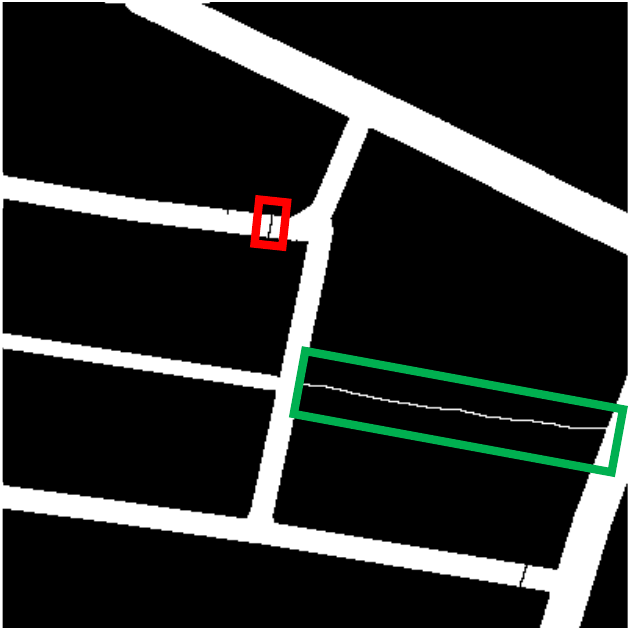}}
      
      \subfigure[Zoom-in]{
     \includegraphics[width=.25\textwidth]{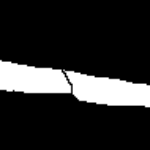}}
      \subfigure[Zoom-in]{
     \includegraphics[width=.67\textwidth]{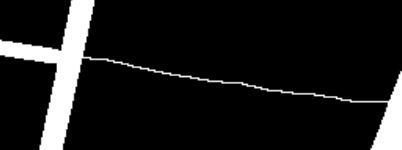}}
  \caption{Illustration of warping in a real world example (satellite image). \textbf{(a)} GT mask. \textbf{(b)} The prediction mask. The red box highlights a \textit{false negative connection}, and the green box highlights a \textit{false positive connection}. \textbf{(c)} Warped prediction mask (using the GT mask as the target). \textbf{(d)} Zoomed-in view of the red box in \textbf{(c)}. \textbf{(e)} Zoomed-in view of the green box in \textbf{(c)}.}
  \label{fig:error_pred}
\end{figure}

In Fig.~\ref{fig:error}, we show a real example from the satellite image dataset, focusing on the errors related to 1D topological structures (connection). In the figure, we illustrate both a false negative connection error (highlighted with a red box) and a false positive connection error (highlighted with a green box). If we warp the ground truth mask towards the prediction mask (c), we observe critical pixels forming a link for the false negative connection (d), and a gap for the false positive connection (e). Similarly, we can warp the prediction mask towards the ground truth, and get different sets of critical pixels for the same topological errors. See Fig.~\ref{fig:error_pred}. 

\begin{figure*}[ht]
  \centering
        \subfigure[GT boundary]{
   \includegraphics[width=.3\textwidth]{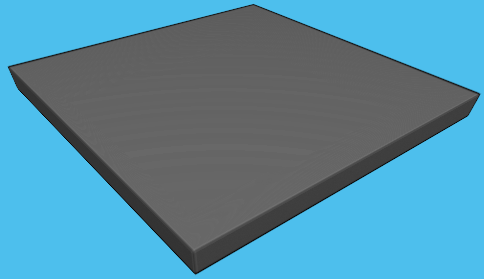}}
  \subfigure[Pred. binary boundary]{
     \includegraphics[width=.324\textwidth]{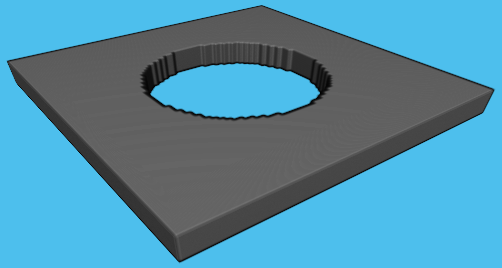}}
     \subfigure[Warped GT boundary]{
     \includegraphics[width=.31\textwidth]{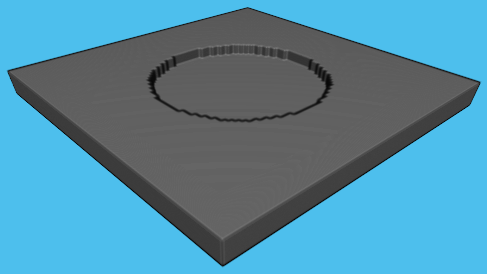}}
  \caption{Illustration of 2D topological structures (holes/voids) for 3D case. \textbf{(a)}: GT boundary. \textbf{(b)}: Prediction binary boundary. \textbf{(c)}: Warped GT boundary. If we warp the GT boundary (Fig.\textbf{(a)}) towards the prediction binary boundary (Fig.\textbf{(b)}), there will be a plane with a thickness of 1 in the middle of the hole/void to keep the original structure, which is illustrated in Fig.~\textbf{(c)}.}
  \label{fig:3d_error}
\end{figure*}

Note that for 2D images with fine structures, errors regarding 1D topological structures are the most crucial. They affect the connectivity of the prediction results. For 3D images, errors on 1D or 2D topological structures are both important, corresponding to broken connections for tubular structures and holes in membranes. We also provide a more comprehensive characterization of different types of topological structures and errors in Fig.~\ref{fig:3d_error}.

Note that for 3D vessel data, it's also the 1D topological structures (connections) that matter. The process of identifying the topological critical pixels is the same as Fig.~\ref{fig:warping-synthetic} and Fig.~\ref{fig:error}. And the only difference is to use 6-adjacency for FG and 26-adjacency
for BG.

\subsection{Homotopy Warping Loss}
\label{warpingloss}

Next, we formalize the proposed \emph{homotopy warping loss}, which is evaluated on the critical pixels due to homotopic warping. As illustrated in the previous section, the warping can be in both directions, from the prediction mask to the ground truth mask, and the opposite. 

Formally, we denote by $f$ the predicted likelihood map of a segmentation network, and $f_B$ the corresponding binarized prediction mask (i.e., $f$ thresholded at 0.5). We denote by $g$ the ground truth mask. First, we warp $g$ towards $f_B$, so that the warped mask, $g^\ast$ has the minimal Hamming distance from the target $f_B$.
\begin{equation}
    \label{eq:warping-g}
    g^{*} = \argmin\nolimits_{g^{w} \lhd g} ||f_B-g^{w}||_H
\end{equation}
where $\lhd$ is the homotopic warping operation. The pixels at which $g^\ast$ and $f_B$ disagree are the critical pixels and will be penalized in the loss. We record these critical pixels due to the warping of $g$ with a mask $M_g$, $M_g = f_B \oplus g^{*}$, 
in which $\oplus$ is the \textit{Exclusive Or} operation. 

We also warp the prediction mask $f_B$ towards $g$.
\begin{equation}
    f_B^{*} = \argmin\nolimits_{f_B^w \lhd f_B} ||f_B^{w}-g||_H
    \label{eq:warping-fb}
\end{equation}
We use the mask $M_f$ to record the remaining critical pixels after warping $f_B$, $M_f = g \oplus f_B^{*}$.
The union of the two critical pixel masks is the complete set of critical pixels corresponding to topological errors, $ M =  M_g \cup M_f$.
$M$ contains all the locations directly related to topological structures. Note this is different from persistent-homology-based method~\cite{hu2019topology}, DMT based method~\cite{hu2021topology} or skeleton based method~\cite{shit2021cldice}, which extract topological locations/structures on the predicted continuous likelihood maps. Our warping loss directly locates the topological critical pixels/structures/locations based on the binary mask. The detected critical pixel set is sparse and less noisy.

$L_{pixel}$ denotes the pixel-wise loss function (e.g., cross-entropy), then $L_{warp}$ can be defined as:
\begin{equation}
\label{loss}
    L_{warp} = L_{pixel}(f, g) \odot M
\end{equation}
where $\odot$ denotes Hadamard product. $L_{warp}$ penalizes the topological critical pixels, forcing the neural network to predict better at these locations, and thus are less prone to topological errors.

The final loss of our method, $L_{total}$, is given by:
\begin{equation}
\label{final_loss}
    L_{total} = L_{dice} + \lambda_{warp} L_{warp}
\end{equation}
where $L_{dice}$ denotes the dice loss. And the loss weight $\lambda_{warp}$ is used to balance the two loss terms.

\subsection{Distance-Ordered Homotopy Warping}
\label{distance}
Even though checking whether a pixel is simple or not is easy, finding the optimal warping as in Eq.~\ref{eq:warping-g} and Eq.~\ref{eq:warping-fb} is challenging. The reason is that there are too many degrees of freedom. At each iteration during the warping, we have to choose a simple point to flip. It is not obvious which simple point will finally lead to a global optimum.

In this section, we provide an efficient heuristic algorithm to find a warping local optimum. 
We explain the algorithm for warping $g$ towards $f_B$. The algorithm generalizes to the opposite warping direction naturally. Recall the warping algorithm iteratively flips simple points. But there are too many choices at each iteration. It is hard to know which flipping choice will lead to the optimal solution. We need good heuristics for choosing a flippable pixel. Below we explain our main intuitions for designing our algorithm. 

First, we restrict the warping so it only sweeps through the area where the two masks disagree. In other words, at each iteration, we restrict the candidate pixels for flipping to not only simple but also pixels on which $g$ and $f_B$ disagree. In Fig.~\ref{fig:warping-synthetic} (c) and (f), we highlight the candidate pixels for flipping at the beginning. Notice that not all simple points are selected as candidates. We only choose simple points within the difference set $\text{Diff}(f_B,g)=f_B \oplus g$. 

Second, since we want to minimize the difference between the warped and target masks, we propose to flip pixels within the difference region $\text{Diff}(f_B,g)$. To implement this strategy efficiently, we order all pixels within $\text{Diff}(f_B,g)$ according to their distance from the FG/BG, and flip them according to this order. A pixel is skipped if it is not simple.

Our algorithm is based on the intuition that a far-away pixel will not become simple until nearby pixels are flipped first. To see this, we first formalize the definition of \emph{distance transform} from the masks, $f_B$ and $g$, denoted by $D^{f_B}$ and $D^{g}$. For a BG pixel of $g$, $p$, its distance value $D^g(p)$ is the shortest distance from $p$ to any FG pixel of $g$, $D^g(p)=\min_{s\in FG_g} \text{dist}(p,s)$. Similarly, for a FG pixel of $g$, $q$, $D^g(q)=\min_{s\in BG_g} \text{dist}(q,s)$. The definition generalizes to $D^{f_B}$.

We observe that a pixel cannot be simple unless it has distance 1 from the FG/BG of a warping mask. The proof is straightforward. Formally,
\begin{lemma}
Given a 2D binary mask $m$, a pixel $p$ cannot be simple for $m$ if its distance function $D^m(p) >1$.
\label{lemma:distance}
\end{lemma}
\myparagraph{Proof.} Assume the foreground has a pixel value of 1 and $p$ is a background pixel with a index of $(i,j)$. Consider the $m$-adjacent ($m$=4) for $p$. Since $D^m(p) > 1$, then we have $m(i-1,j) = m(i+1, j) = m(i, j-1) = m(i, j+1) =0$. In this case, $p$ is not 4-adjacent to any FG connected component, violating the \textbf{1)} of \textbf{Definition 1}. Consequently, pixel $(i,j)$ is not a simple point. This also holds for foreground pixels. This lemma naturally generalizes to 3D case. 
\qed

Lemma~\ref{lemma:distance} implies that only after flipping pixels with distance 1, the other misclassified locations should be considered. This observation gives us the intuition of our algorithm. To warp $g$ towards $f_B$, our algorithm is as follows: (1) compute the difference set $\text{Diff}(f_B,g)$ as the candidate set of pixels; (2) sort candidate pixels in a non-decreasing order of the distance transform $D^g$; (3) enumerate through all candidate pixels according to the order. For each iteration, check if it is simple. If yes, flip the pixel's label. It is possible that this algorithm can miss some pixels. They are not simple when the algorithm checks, but they might become simple as the algorithm continues (since their neighboring pixels get flipped in previous iterations). 

One remedy is to recalculate the distance transform after one round of warping and go through the remaining pixels once more.  But in practice, we found this is not necessary as this scenario is very rare. 

By using Distance-Ordered Homotopy Warping, we are able to only consider all the inconsistent pixels once and flip them if they are simple. Otherwise, we need to iteratively warp all the inconsistent pixels (one non-simple pixel might become simple if its neighbors are flipped in the previous iteration). It takes 1.452s to warp a 512$\times$512 image without the warping strategy, while only 0.317s with the strategy thereby allowing the network to converge much faster.

We conduct extensive experiments to demonstrate the effectiveness of the proposed method. Sec.~\ref{dataset_warping} introduces the datasets used in this chapter, including both 2D and 3D datasets. The benchmark methods are described in Sec.~\ref{baseline}. We mainly focus on topology-aware segmentation methods. Sec.~\ref{sec:metric_warping} describes the evaluation metrics used to assess the quality of the segmentation. To demonstrate the ability to achieve better structural/topological performances, besides standard segmentation metrics, such as the DICE score, we also use several topology-aware metrics to evaluate all the methods. Several ablation studies are then conducted to further demonstrate the efficiency and effectiveness of the technical contributions (Sec.~\ref{sec:ablation}).

\subsection{Datasets}
\label{dataset_warping}
We conduct extensive experiments to validate the efficacy of our method. Specifically, we use four natural and biomedical 2D datasets (\textbf{RoadTracer}~\cite{bastani2018roadtracer}, \textbf{DeepGlobe}~\cite{demir2018deepglobe}, \textbf{Mass.}~\cite{mnih2013machine}, \textbf{DRIVE}~\cite{staal2004ridge}) and one more 3D biomedical dataset (\textbf{CREMI}~\footnote{https://cremi.org/}) to validate the efficacy of the propose method. \textbf{Mass.}~\cite{mnih2013machine}, \textbf{DRIVE} and \textbf{CREMI} have been introduce in Sec.~\ref{sec:dataset_topoloss}. And the details of the other two datasets are listed as follows:
\begin{itemize}
    \item \textbf{RoadTracer}: Roadtracer contains 300 high resolution satellite images, covering urban areas of forty cities from six different countries~\cite{bastani2018roadtracer}. Similar to setting in~\cite{bastani2018roadtracer}, twenty five cities (180 images) are used as the training set, and the rest fifteen cities (120 images) are used as the validation set.
    \item \textbf{DeepGlobe}: DeepGlobe contains aerial images of rural areas in Thailand, Indonesia, and India~\cite{demir2018deepglobe}. Similar to setting in~\cite{batra2019improved}, we use 4696 images as the training set and the rest 1530 images as the validation set.
\end{itemize}

\subsection{Evaluation Metrics}
\label{sec:metric_warping}

We use both pixel-wise (\textbf{DICE}) and topology-aware metrics (\textbf{ARI}, \textbf{Warping Error} and \textbf{Betti number error}) to evaluate the performance of the proposed method. \textbf{ARI} and \textbf{Betti number error} have been introduced in Sec.~\ref{sec:metric_topoloss}.
The details of the other metrics used in this chapter are listed as follows:

\begin{itemize}
    \item \textit{DICE}: DICE score (also known as DICE coefficient, DICE similarity index) is one of the most popular evaluation metrics for image segmentation, which measures the overlapping between the predicted and ground truth masks.
    \item \textit{Warping Error}~\cite{jain2010boundary}: Warping Error is metric that measures topological disagreements instead of simple pixel disagreements. After warping all the simple points of ground truth to the predicted mask, the disagreements left are topological errors. The warping error is defined as the percentage of these topological errors over the image size.
\end{itemize}

\subsection{Baselines}
\label{baseline}

We compare the results of our method with several state-of-the-art methods. The standard/simple UNet (2D/3D)~\textbf{{UNet}}~\cite{ronneberger2015u,cciccek20163d} is used as a strong baseline and the backbone for other methods. The other baselines used in this chapter include \textbf{RoadTracer}~\cite{bastani2018roadtracer}, \textbf{VecRoad}~\cite{tan2020vecroad}, \textbf{iCurb}~\cite{xu2021icurb}, \textbf{{DIVE}}~\cite{fakhry2016deep},  \textbf{{UNet-VGG}}~\cite{mosinska2018beyond}, \textbf{{TopoLoss}}~\cite{hu2019topology}, \textbf{clDice}~\cite{shit2021cldice} and \textbf{DMT}~\cite{hu2021topology}.~\textbf{{UNet}}, \textbf{{DIVE}}~\cite{fakhry2016deep} and \textbf{{UNet-VGG}}~\cite{mosinska2018beyond} have been introduce in Sec.~\ref{sec:baseline_topoloss}.
The other baseline methods used in this chapter are listed as follows:
\begin{itemize}
    \item \textbf{RoadTracer}~\cite{bastani2018roadtracer}: RoadTracer is an iterative graph construction based method where node locations are selected by a CNN.
    \item \textbf{VecRoad}~\cite{tan2020vecroad}: VecRoad is a point-based iterative graph exploration scheme with segmentation-cues guidance and flexible steps.
    \item \textbf{iCurb}~\cite{xu2021icurb}: iCurb is an imitation learning-based solution for an offline road-curb detection method.
    \item \textbf{TopoNet}~\cite{hu2019topology}: TopoNet is a recent work that tries to learn to segment with correct topology based on a novel persistent homology based loss function.
    \item \textbf{clDice}~\cite{shit2021cldice}: Another topology aware method for tubular structure segmentation. The basic idea is to use thinning techniques to extract the skeletons (centerlines) of the likelihood maps and ground truth mask. A new cldice loss is proposed based on the extracted skeletons besides a traditional pixel-wise loss.
    \item \textbf{DMT}~\cite{hu2021topology}: DMT is a topology-aware deep image segmentation method via discrete Morse theory. Instead of identifying topological critical pixels/locations, the DMT loss tries to identify the whole morse structures, and the new loss is defined on the identified morse structures.
\end{itemize}

Note that \textit{RoadTracer}, \textit{VecRoad}, and \textit{iCurb} are graph-based methods for road tracing. Graph-based approaches learn to explicitly detect keypoints and connect them. Since the graph is built iteratively, some detection errors in the early stage can propagate and lead to even more errors. Segmentation-based methods avoid this issue as they make predictions in a global manner. The challenge with segmentation methods in road network reconstruction is they may fail in thin structures especially when the signal is weak. This is exactly what we are addressing in this chapter -- using critical pixels to improve segmentation-based methods.

\textit{UNet-VGG}, \textit{TopoNet}, \textit{clDice} and \textit{DMT} are topology-aware segmentation methods.

\subsection{Implementation Details}
\label{sec:implementation}
For 2D images, we use $(m, n)$ = (4, 8) to check if a pixel is simple or not; while $(m, n)$ = (8, 26) for 3D images. We choose cross-entropy loss as $L_{warp}/L_{pixel}$ for all the experiments, except the ablation studies in Tab.~\ref{losschoose}.

For 2D datasets, the batch size is set as 16, and the initial learning rate is 0.01. We randomly crop patches with the size of $512 \times 512$ and then feed them into the 2D UNet. For 3D case, the batch size is also 16, while the input size is $128 \times 128 \times 16$. We perform the data normalization for the single patch based on its mean and standard deviation.

We use PyTorch framework (Version: 1.7.1) to implement the proposed method. A simple/standard UNet (2D or 3D) is used as the baseline and the backbone. For the proposed method, as well as the other loss function based baselines, to make a fair comparison, we use the same UNet as the backbone. And the training strategy is to train the UNet with dice loss first until converges and then add the proposed losses to fine-tune the models obtained from the initial step. 

\subsection{Results}
\begin{figure}[ht]
\centering 
    \begin{tabular}{cccc}
    \includegraphics[width=.21\textwidth]{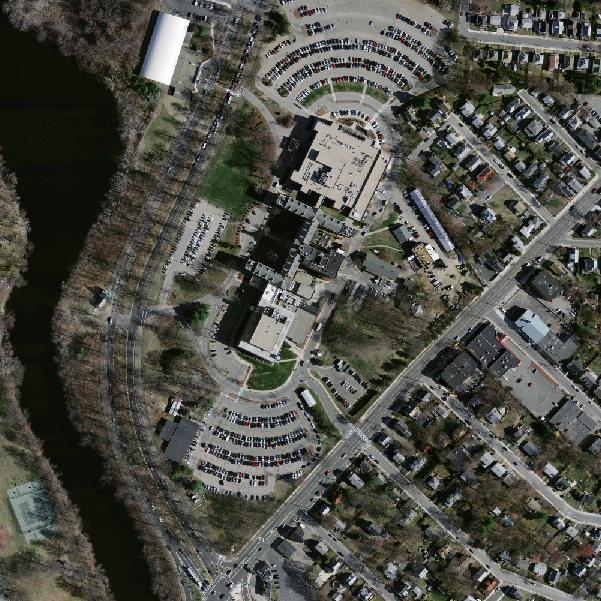} &
    \includegraphics[width=.21\textwidth]{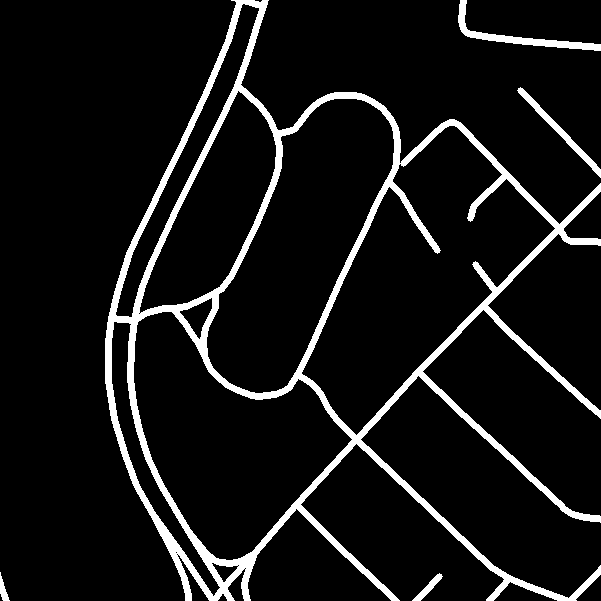} & 
    \includegraphics[width=.21\textwidth]{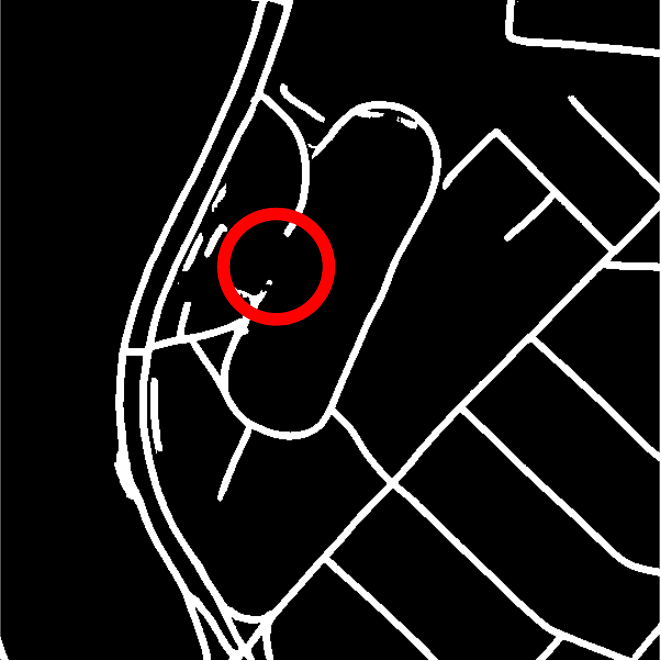} &
    \includegraphics[width=.21\textwidth]{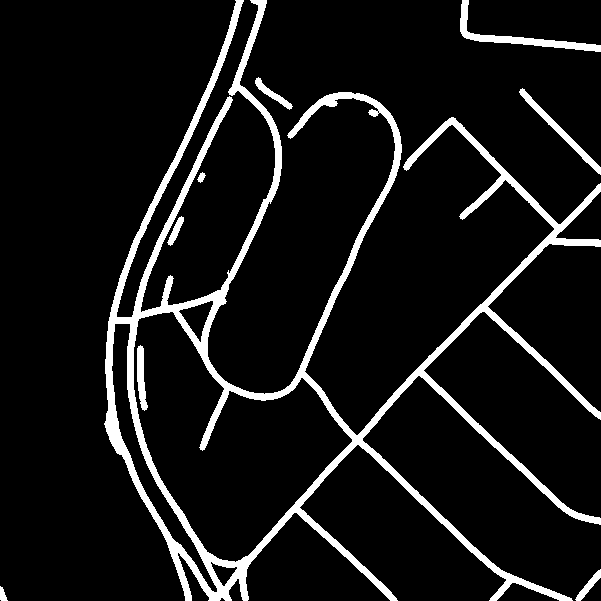} \\
    
  \includegraphics[width=.21\textwidth]{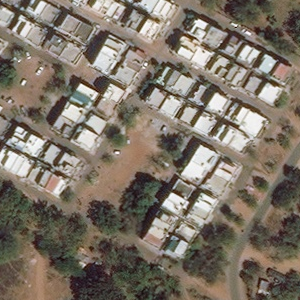} &
     \includegraphics[width=.21\textwidth]{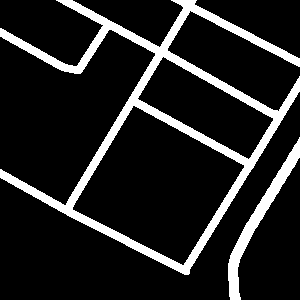} &
     \includegraphics[width=.21\textwidth]{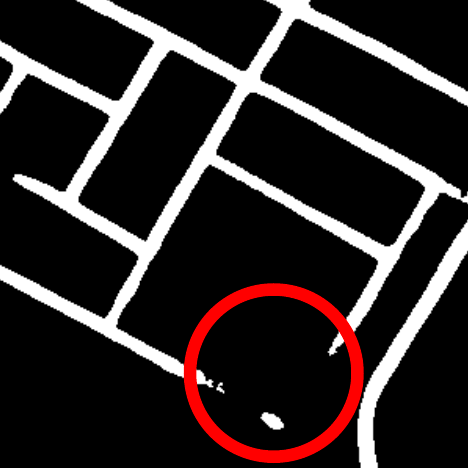} &
     \includegraphics[width=.21\textwidth]{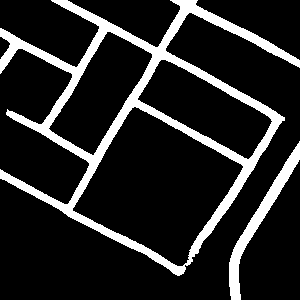} \\
    
        \includegraphics[width=.21\textwidth]{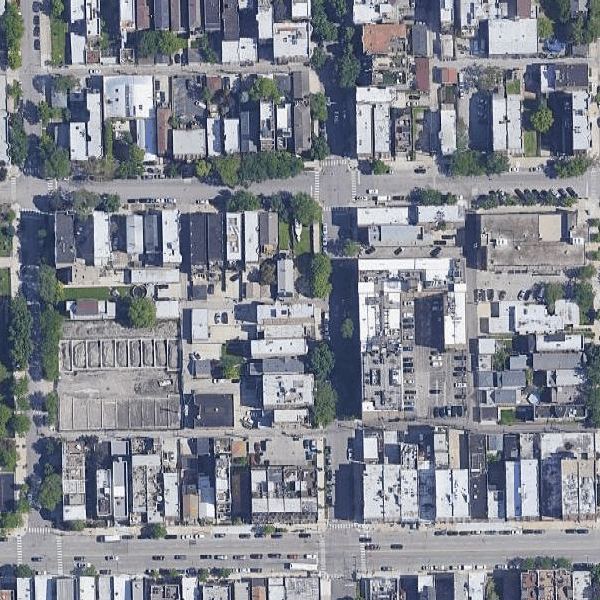} &
    \includegraphics[width=.21\textwidth]{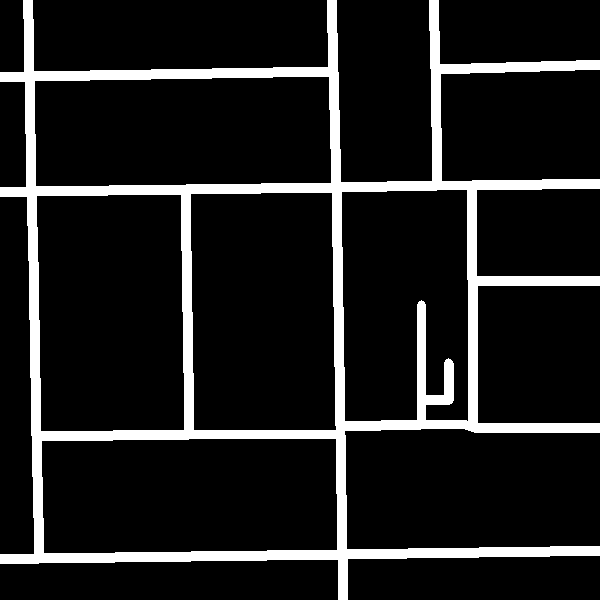} & 
    \includegraphics[width=.21\textwidth]{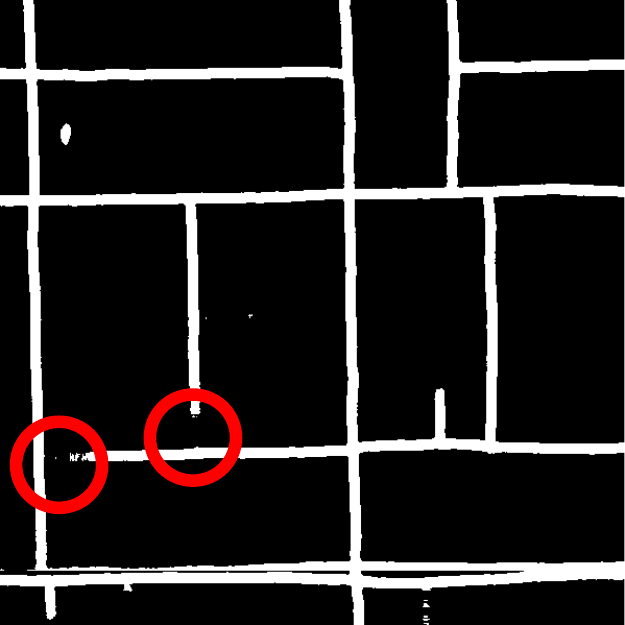} &
    \includegraphics[width=.21\textwidth]{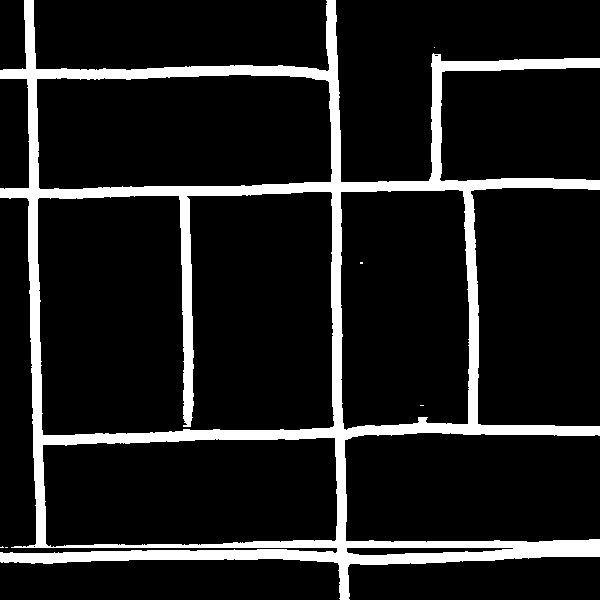} \\
    
        \includegraphics[width=.21\textwidth]{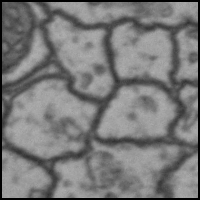} &
    \includegraphics[width=.21\textwidth]{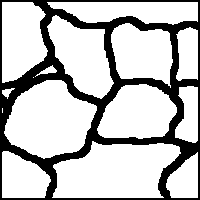} & 
    \includegraphics[width=.21\textwidth]{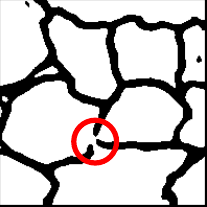} &
    \includegraphics[width=.21\textwidth]{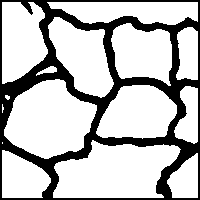} \\
    (a) Original & (b) GT & (c) UNet & (d) Ours
    \end{tabular}
\caption{Qualitative results compared with the standard UNet. The proposed warping loss can help to correct the topological errors (highlighted by red circles). The sampled patches are from four different datasets.}
\label{fig:Qualitative_warping}
\end{figure}

In Fig.~\ref{fig:Qualitative_warping}, we show qualitative results from different datasets. Compared with baseline UNet, our method recovers better structures, such as connections, which are highlighted by red circles. Our final loss is a weighted combination of the dice loss and warping-loss term $L_{warp}$. When $\lambda_{warp}$ = 0, the proposed method degrades to a standard UNet. The recovered better structures (UNet and \textit{Warping} columns in Fig.~\ref{fig:Qualitative_warping}) demonstrate that our warping-loss helps the deep neural networks to achieve better topological segmentations.

\section{Experiments}
\label{sec:experiment_warping}

Tab.~\ref{roadtracer} shows quantitative results for three 2D image datasets, RoadTracer, DeepGlobe, and The Massachusetts dataset, and one 3D image dataset, CREMI. The best performances are highlighted in bold. The proposed warping-loss usually achieves the best performances in both DICE score and topological accuracy (ARI, Warping Error, and Betti Error) over other topology-aware segmentation baselines.

\setlength{\tabcolsep}{5pt}
\begin{table}[ht]
\caption{Quantitative results of different methods.}
\label{roadtracer}
\begin{center}
\footnotesize
\begin{tabular}{ccccc} 
 \hline \hline
 Method & DICE$\uparrow$  & ARI$\uparrow$ & Warping$\downarrow$ & Betti$\downarrow$\\
 \hline \hline
 \multicolumn{5}{c}{RoadTracer} \\
 \hline
UNet~\cite{ronneberger2015u} & 0.587 & 0.544 & 10.412 $\times 10^{-3}$ & 1.591\\
RoadTracer~\cite{bastani2018roadtracer} & 0.547 & 0.521 & 13.224$\times 10^{-3}$ & 2.218 \\
VecRoad~\cite{tan2020vecroad} & 0.552 & 0.533 & 12.819$\times 10^{-3}$ & 2.095\\
iCurb~\cite{xu2021icurb} & 0.571 & 0.535 & 11.683$\times 10^{-3}$ & 1.873 \\
 \hline
UNet-VGG~\cite{mosinska2018beyond} & 0.576 & 0.536 & 11.231 $\times 10^{-3}$& 1.607 \\
TopoNet~\cite{hu2019topology} & 0.584 & 0.556 & 10.008 $\times 10^{-3}$ & 1.378\\
clDice~\cite{shit2021cldice} & 0.591 & 0.550 &9.192 $\times 10^{-3}$&  1.309\\
DMT~\cite{hu2021topology} & 0.593 & 0.561 & 9.452 $\times 10^{-3}$ & 1.419\\
  \textit{Warping} & \textbf{0.603} & \textbf{0.572} & \textbf{8.853} $\times 10^{-3}$& \textbf{1.251}\\
  \hline \hline
 \multicolumn{5}{c}{DeepGlobe} \\
 \hline
UNet~\cite{ronneberger2015u} & 0.764 & 0.758 & 3.212 $\times 10^{-3}$ & 0.827\\
 \hline
UNet-VGG~\cite{mosinska2018beyond} & 0.742 & 0.748 & 3.371 $\times 10^{-3}$ & 0.867\\
TopoNet~\cite{hu2019topology} & 0.765 & 0.763 & 2.908 $\times 10^{-3}$ & 0.695\\
clDice~\cite{shit2021cldice} & 0.771 & 0.767& 2.874 $\times 10^{-3}$ & 0.711\\
DMT~\cite{hu2021topology} & 0.769 & 0.772 &2.751 $\times 10^{-3}$ & 0.609\\
  \textit{Warping} & \textbf{0.780} & \textbf{0.784} & \textbf{2.683}$\times 10^{-3}$ & \textbf{0.569}\\
 \hline \hline
 \multicolumn{5}{c}{Mass.} \\
 \hline
UNet~\cite{ronneberger2015u} & 0.661 & 0.819 & 3.093$\times 10^{-3}$ & 3.439 \\
 \hline
UNet-VGG~\cite{mosinska2018beyond} & 0.667 & 0.846 & 3.185$\times 10^{-3}$ & 2.781\\

TopoNet~\cite{hu2019topology} & 0.690 & 0.867 & 2.871$\times 10^{-3}$ & 1.275\\
clDice~\cite{shit2021cldice} & 0.682 & 0.862 & 2.552$\times 10^{-3}$ & 1.431\\
DMT~\cite{hu2021topology} & 0.706 & \textbf{0.881} & 2.631 $\times 10^{-3}$  & 0.995 \\
 \textit{Warping} &\textbf{0.715} & 0.864 & \textbf{2.440}$\times 10^{-3}$ & \textbf{0.974}\\
 \hline \hline
  \multicolumn{5}{c}{DRIVE} \\
 \hline
UNet~\cite{ronneberger2015u} & 0.749 & 0.834 & 4.781$\times 10^{-3}$ & 3.643 \\
DIVE~\cite{fakhry2016deep} & 0.754 & 0.841 & 4.913$\times 10^{-3}$ & 3.276 \\
 \hline
UNet-VGG~\cite{mosinska2018beyond} & 0.721 & 0.887 & 4.362$\times 10^{-3}$ & 2.784\\
TopoNet~\cite{hu2019topology} & 0.761 & 0.902 & 3.895$\times 10^{-3}$ & 1.076\\
clDice~\cite{shit2021cldice} & 0.753 & 0.896 & 4.012 $\times 10^{-3}$ & 1.218\\
DMT~\cite{hu2021topology} & 0.773 & 0.908 & 3.561 $\times 10^{-3}$  & 0.873 \\
 \textit{Warping} &\textbf{0.781} & \textbf{0.911} & \textbf{3.419}$\times 10^{-3}$ & \textbf{0.812}\\
 \hline \hline
 \multicolumn{5}{c}{CREMI} \\
 \hline
3D UNet~\cite{cciccek20163d} & 0.961 & 0.832 & 11.173 $\times 10^{-3}$ &  2.313\\
DIVE~\cite{fakhry2016deep} & 0.964 & 0.851 & 11.219 $\times 10^{-3}$& 2.674 \\
\hline
TopoNet~\cite{hu2019topology} & 0.967 & 0.872 & 10.454 $\times 10^{-3}$ &  1.076\\
clDice~\cite{shit2021cldice} & 0.965 & 0.845 & 10.576 $\times 10^{-3}$ & 0.756\\
DMT~\cite{hu2021topology} & \textbf{0.973} & 0.901 & 10.318 $\times 10^{-3}$ & 0.726\\
  \textit{Warping} & 0.967 & \textbf{0.907} & \textbf{9.854} $\times 10^{-3}$ & \textbf{0.711}\\
 \hline
\end{tabular}
\end{center}
\end{table}

\subsection{Ablation Studies}
\label{sec:ablation}
To further explore the technical contributions of the proposed method and provide a rough guideline of how to choose the hyperparameters, we conduct several ablation studies. Note that all the ablation studies are conducted on the RoadTracer dataset.  
\setlength{\tabcolsep}{5pt}
\begin{table}[ht]
\caption{Ablation study for loss weight $\lambda_{warp}$.}
\label{weight}
\begin{center}
\small
\begin{tabular}{ccccc} 
 \hline
 $\lambda_{warp}$ & DICE$\uparrow$  & ARI$\uparrow$ & Warping$\downarrow$ & Betti$\downarrow$\\
 \hline\hline
0 & 0.587 & 0.544 & 10.412 $\times 10^{-3}$ & 1.591\\
$2 \times 10^{-5}$ & \textbf{0.603} & 0.561 & 9.012 $\times 10^{-3}$ & 1.307\\
$5 \times 10^{-5}$ & 0.601 & 0.548 & 9.356 $\times 10^{-3}$& 1.412 \\
$1 \times 10^{-4}$  & \textbf{0.603} & \textbf{0.572} & \textbf{8.853} $\times 10^{-3}$ & \textbf{1.251}\\
$2 \times 10^{-4}$ & 0.602 & 0.565 & 9.131 $\times 10^{-3}$& 1.354\\
 \hline
\end{tabular}
\end{center}
\end{table}

\myparagraph{The Impact of the Loss Weights.}
As seen in Eq.~\ref{final_loss}, our final loss function is a combination of dice loss and the proposed warping loss $L_{warp}$. The balanced term $\lambda_{warp}$ controls the influence of the warping loss term, and it's a dataset dependent hyper-parameter. The quantitative results for different choices of $\lambda_{warp}$ are illustrated in Tab.~\ref{weight}. For the RoadTracer dataset, the optimal value is $1 \times 10^{-4}$. From Tab.~\ref{weight}, we can find that different choices of $\lambda_{warp}$ do affect the performances. The reason is that, if $\lambda_{warp}$ is too small, the effect of the warping loss term is negligible. However, if $\lambda_{warp}$ is too large,
the warping loss term will compete with the $L_{dice}$ and decrease the performance of the other easy-classified pixels. Note that within a reasonable range of $\lambda_{warp}$, all the choices contribute to better performances compared to baseline (row `0', standard UNet), demonstrating the effectiveness of the proposed loss term.

\begin{table}[ht]
\caption{Ablation study for the choices of loss.}
\label{losschoose}
\begin{center}
\small
\begin{tabular}{ccccc} 
 \hline
 $L_{pixel}$ & DICE$\uparrow$  & ARI$\uparrow$ & Warping$\downarrow$ & Betti$\downarrow$\\
 \hline\hline
w/o & 0.587 & 0.554 & 10.412 $\times 10^{-3}$ & 1.591\\
MSE & 0.598 & 0.556 & 9.853 $\times 10^{-3}$ & 1.429\\
Dice loss & \textbf{0.606} & 0.563 & 9.471 $\times 10^{-3}$ &  1.368\\
CE   & 0.603 & \textbf{0.572} & \textbf{8.853} $\times 10^{-3}$ & \textbf{1.251}\\
 \hline
\end{tabular}
\end{center}
\end{table}

\myparagraph{The Choice of Loss Functions.}
\label{loss_choice}
The proposed warping loss is defined on the identified topological critical pixels. Consequently, any pixel-wise loss functions can be used to define the warping loss $L_{warp}/L_{pixel}$. In this section, we investigate how the choices of loss functions affect the performances. The quantitative results are shown in Tab.~\ref{losschoose}. Compared with mean-square-error loss (MSE) or Dice loss, the cross-entropy loss (CE) achieves the best performances in terms of topological metrics. On the other hand, all these three choices perform better than the baseline method (row `w/o', standard UNet), which further demonstrates the contribution of the proposed loss term.

 \begin{table}[ht]
\caption{Comparison of different critical pixel selection strategies.}
\label{strategy}
\begin{center}
\small
\begin{tabular}{ccccc} 
 \hline
 Kernel Size & DICE$\uparrow$  & ARI$\uparrow$ & Warping $\downarrow$ & Betti$\downarrow$\\
 \hline \hline
UNet & 0.587 & 0.544 & 10.412 $\times 10^{-3}$& 1.591\\
\hline
w/o DT & 0.586 & 0.547 & 10.256$\times 10^{-3}$ & 1.473 \\
Warping (GT $\rightarrow$ Pred) & 0.594 & 0.567 & 9.171$\times 10^{-3}$ & 1.290\\
Warping (Pred $\rightarrow$ GT) & 0.598 & 0.562 & 9.124 $\times 10^{-3}$& 1.315 \\
 \hline
  \textit{Warping} & \textbf{0.603} & \textbf{0.572} & \textbf{8.853$\times 10^{-3}$} & \textbf{1.251}\\
 \hline
\end{tabular}
\end{center}
\end{table}
\myparagraph{Comparison of Different Critical Pixel Selection Strategies.} We also conduct additional experiments to demonstrate the effectiveness of the proposed critical pixel selection strategy. The first variation is flipping the simple pixels but removing the heuristic that uses the distance transform, and then using the remaining non-simple pixels as our critical pixel set, which is denoted as `w/o' in Tab.~\ref{strategy}. The other two variations (warping only in one direction: ground truth to prediction or prediction to ground truth) achieve reasonably better while still slightly inferior results to the proposed version. The reason might be that the proposed critical point selection strategy contains more complete topologically challenging pixels. Both ablation studies demonstrate the effectiveness of the proposed critical point selection strategy.

\myparagraph{Comparison with Morphology Post-processing.} To verify the necessity of the proposed method, we also compare the proposed method with traditional morphology post-processing, i.e., dilation, and erosion. Dilation and erosion are global operations. 
Though closing operation (dilates image and then erodes dilated image) could bridge some specific gaps/broken connections, it will damage the global structures.
In practice, the gaps/broken will usually be more than a few pixels. If the kernel size is too small, the closing operation (dilate then erode) will hardly affect the final performance; while too big kernel sizes will join the separated regions. Tab.~\ref{post} lists post-processing results on baseline UNet.

\setlength{\tabcolsep}{5pt}
 \begin{table}[ht]
\caption{Comparison against post-processing.}
\label{post}
\begin{center}
\small
\begin{tabular}{ccccc} 
 \hline
 Kernel Size & DICE$\uparrow$  & ARI$\uparrow$ & Warping $\downarrow$ & Betti$\downarrow$\\
 \hline \hline
UNet & 0.587 & 0.544 & 10.412 $\times 10^{-3}$& 1.591\\
\hline
Closing (5) & 0.588 & 0.542 & 10.414$\times 10^{-3}$ & 1.590 \\
Closing (10) & 0.587 & 0.546 & 10.399$\times 10^{-3}$ & 1.583\\
Closing (15) & 0.586 & 0.541 & 10.428 $\times 10^{-3}$& 1.598 \\
 \hline
  \textit{Warping} & \textbf{0.603} & \textbf{0.572} & \textbf{8.853$\times 10^{-3}$} & \textbf{1.251}\\
 \hline
\end{tabular}
\end{center}
\end{table}

\myparagraph{The Efficiency of the Proposed Loss.}
In this section, we'd like to investigate the efficiency of the proposed method. Our warping algorithm contains two parts, the distance transform and the sorting of the distance matrix. The complexity for distance transform is $O(n)$ for a 2D image where $n$ is the size of the 2D image, and $n=H \times W$. $H,W$ are the height and width of the 2D image, respectively. And the complexity for sorting is $mlog(m)$, where $m$ is the number of misclassified pixels. Usually, $m \ll n$, so the overall complexity for the warping algorithm is $O(n)$. As a comparison,~\cite{hu2019topology} needs $O(n^3)$ complexity to compute the persistence diagram. And the computational complexity for~\cite{hu2021topology},~\cite{shit2021cldice} are $O(nlog(n))$ and $O(n)$, respectively. 

\begin{table}[ht]
\caption{Comparison of efficiency.}
\label{efficiency}
\begin{center}
\small
\begin{tabular}{ccccc} 
 \hline
 Method & Complexity & Training time\\
 \hline\hline
TopoNet~\cite{hu2019topology} & $O(n^3)$ & $\approx 12h$\\
clDice~\cite{shit2021cldice} & $O(n)$ & $\approx$ \textbf{3h}\\
DMT~\cite{hu2021topology} & $O(nlogn)$ & $\approx 7h$\\
  \textit{Warping} & $O(n)$ & $\approx 4h$\\
 \hline
\end{tabular}
\end{center}
\end{table}

The comparison in terms of complexity and training time are illustrated in Tab.~\ref{efficiency}. Note that for the proposed method and all the other baselines, we first train a simple/standard UNet, and then add the additional loss terms to fine-tune the models obtained from the initial step. Here, the training time is only for the fine-tune step. The proposed method takes slightly longer training time than clDice, while achieving the best performance over the others. As all these methods use the same backbone, the inference times are the same.

\subsection{Failure Cases}
\label{failure}
In this section, we add a few failure cases from different datasets. Note that inferring topology given an image is a very difficult task, especially near challenging spots, e.g., blurred membrane locations or weak vessel connections. Current methods can help to improve topology-wise accuracy, but they are far from perfect.

\begin{figure}[ht]
\centering 
    \begin{tabular}{cccc}
   \includegraphics[width=.2\textwidth]{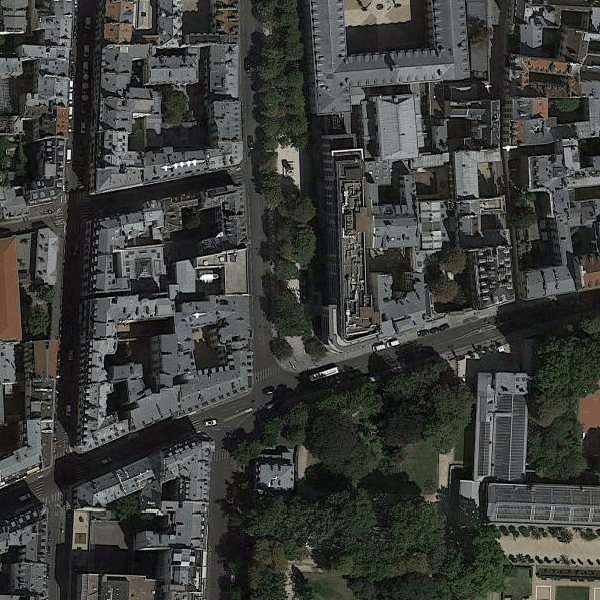} &
     \includegraphics[width=.2\textwidth]{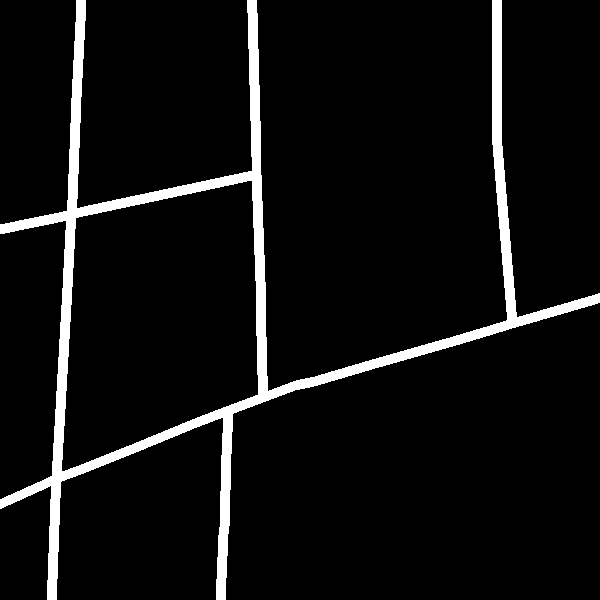}&
     \includegraphics[width=.2\textwidth]{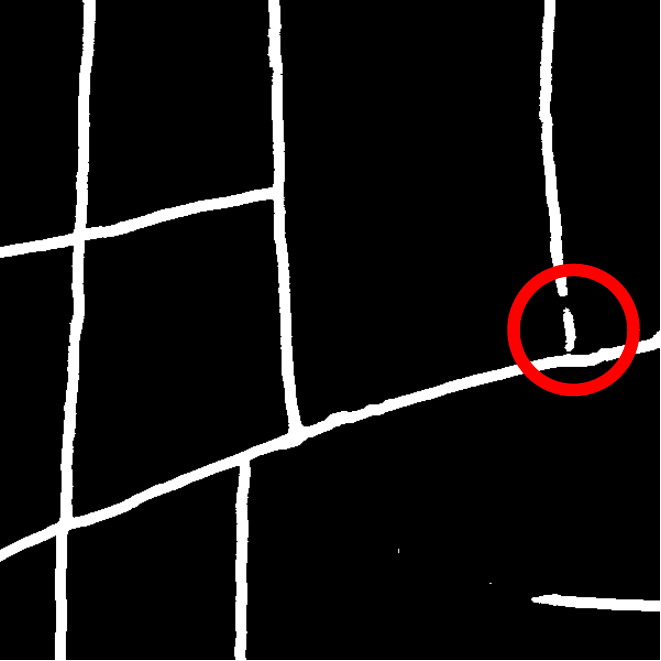}\\

   \includegraphics[width=.2\textwidth]{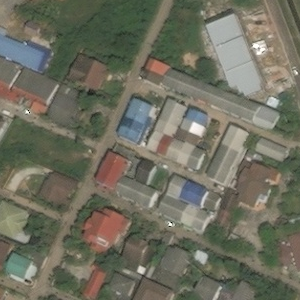} &
     \includegraphics[width=.2\textwidth]{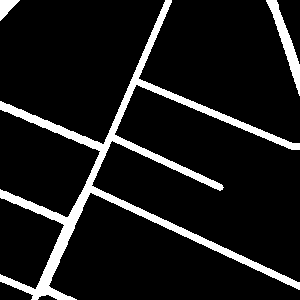}&
     \includegraphics[width=.2\textwidth]{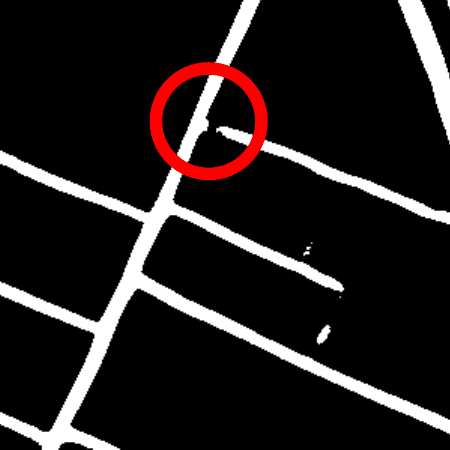}\\
  
   \includegraphics[width=.2\textwidth]{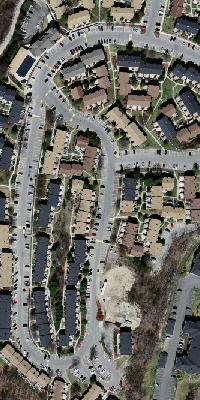} &
     \includegraphics[width=.2\textwidth]{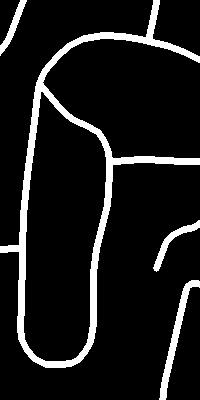}&
     \includegraphics[width=.2\textwidth]{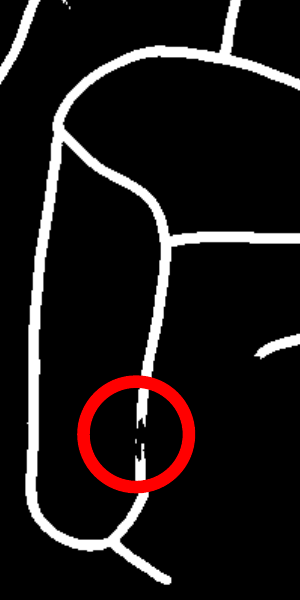}\\
     
        \includegraphics[width=.2\textwidth]{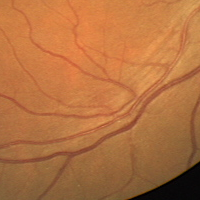} &
     \includegraphics[width=.2\textwidth]{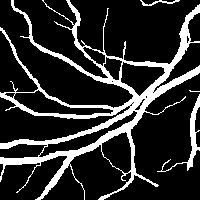}&
     \includegraphics[width=.2\textwidth]{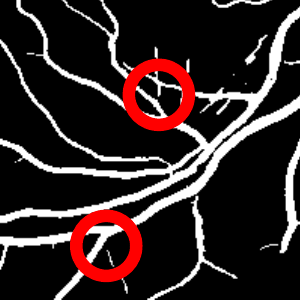}\\
     
        \includegraphics[width=.2\textwidth]{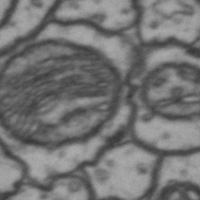} &
     \includegraphics[width=.2\textwidth]{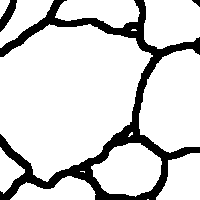}&
     \includegraphics[width=.2\textwidth]{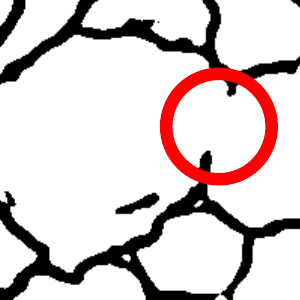}\\

         (a) Original & (b) GT & (c) Ours
         \end{tabular}
\caption{A few failure cases of the proposed method. From top to bottom, the sampled patches are from RoadTracer, DeepGlobe, Mass, DRIVE, and CREMI datasets respectively.}
\label{fig:failure}
\end{figure}

\section{Conclusion}
In this chapter, we propose a novel homotopy warping loss to learn to segment with better structural/topological accuracy. Under the homotopy warping strategy, we can identify the topological critical pixels/locations, and the new loss is defined on these identified pixels/locations. Furthermore, we propose a novel strategy called Distance-Ordered Homotopy Warping to efficiently identify the topological error locations based on distance transform. Extensive experiments on multiple datasets and ablation studies have been conducted to demonstrate the efficacy of the proposed method.

%% file: dmt_loss.tex
\chapter{Topology-Aware Segmentation Using Discrete Morse Theory}
\label{chapter:dmt}

In both Chapter~\ref{chapter:topoloss} and Chapter~\ref{chapter:warping}, we introduce methods identifying a set of critical points of the likelihood function, e.g., saddles and extrema, as
topologically critical locations for the neural network to memorize. However,
only identifying a sparse set of critical points at every epoch is inefficient
in terms of training. In this chapter,
we introduce a novel method to identify critical structures instead of critical points.

\section{Introduction}

\begin{figure*}[ht]
\centering 
\subfigure[]{
\includegraphics[width=0.15\textwidth]{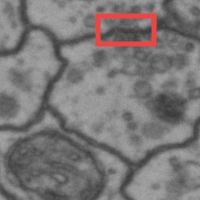}}
\hspace{-6pt}
\subfigure[]{
\includegraphics[width=0.15\textwidth]{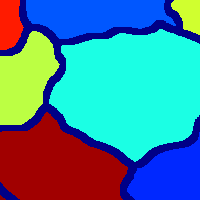}}
\hspace{-6pt}
\subfigure[]{\label{fig:teaser_likelihood}
\includegraphics[width=0.15\textwidth]{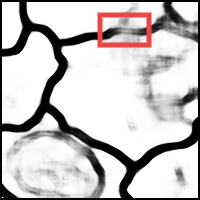}}
\hspace{-6pt}
\subfigure[]{
\includegraphics[width=0.15\textwidth]{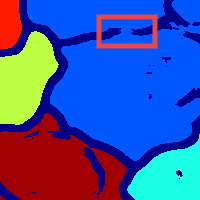}}
\hspace{-6pt}
\subfigure[]{\label{fig:structure}
\includegraphics[width=0.15\textwidth]{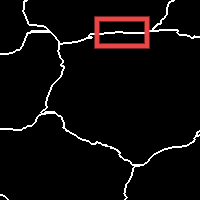}}
\hspace{-6pt}
\subfigure[]{
\includegraphics[width=0.15\textwidth]{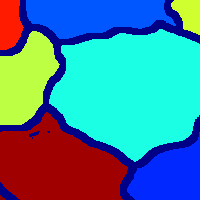}}
\vspace{-8pt}
\caption{Illustration of the importance of topological correctness in a neuron image segmentation task and the effectiveness of the proposed DMT-loss. The goal of this task is to segment membranes that partition the image into regions corresponding to neurons. \textbf{(a)} an input neuron image with challenging locations (blur regions) highlighted. \textbf{(b)} ground truth segmentation of the membranes (dark blue) and the result neuron regions. \textbf{(c)} likelihood map of a baseline method without topological guarantee~\cite{ronneberger2015u}. \textbf{(d)} segmentation results of the baseline method. Small pixel-wise errors lead to broken membranes, resulting in the merging of many neurons into one. \textbf{(e)} The topologically critical structure captured by the proposed DMT-loss (based on the likelihood in (c)). \textbf{(f)} Our method produces the correct topology and the correct partitioning of neurons.}
\label{fig:teaser_dmt}

\end{figure*}

Segmenting objects while preserving their global structure is a challenging yet important problem. Various methods have been proposed to encourage neural networks to preserve fine details of objects~\cite{long2015fully,he2017mask,chen2014semantic,chen2018deeplab,chen2017rethinking}. Despite their high per-pixel accuracy, most of them are still prone to structural errors, such as missing small object instances, breaking thin connections, and leaving holes in membranes. These structural errors can significantly damage downstream analysis. 
For example, in the segmentation of biomedical structures such as membranes and vessels, small pixel errors at a junction will induce significant structure errors, leading to catastrophic functional mistakes.
See Fig.~\ref{fig:teaser_dmt} for an illustration.

Topology is a very global characterization that needs a lot of observations to learn. 
Any training set is insufficient in teaching the network to correctly reason about topology, especially near challenging spots, e.g., blurred membrane locations or weak vessel connections. A neural network tends to learn from clean-cut cases and converge quickly. Meanwhile, topologically-challenging locations remain misclassified, causing structural/topological errors.
We note that this issue cannot be alleviated even with more annotated (yet still unbalanced) images.

We propose a novel approach that identifies critical topological structures during training and teaches a neural network to learn from these structures. Our method can produce segmentations with correct topology, i.e., having the same \emph{Betti number} (i.e., number of connected components and handles/tunnels) as the ground truth. 
Underlying our method is the classic Morse theory~\cite{milnor1963morse}, which captures singularities of the gradient vector field of the likelihood function. Intuitively speaking, we treat the likelihood as a terrain function and Morse theory helps us capture terrain structures such as ridges and valleys.  
These structures, composed of 1D and 2D manifold pieces,
reveal the topological information captured by the (potentially noisy) likelihood function. 

We consider these Morse structures as \emph{topologically critical}; they encompass all potential skeletons of the object. We propose a new loss that identifies these structures and enforce higher penalty along them. This way, we effectively address the sampling bias issue and ensure that the networks predict correctly near these topologically difficult locations. Since the Morse structures are identified based on the (potentially noisy) likelihood function, they can be both false negatives (a structure can be a true structure but was missed in the segmentation) and false positives (a hallucination of the model and should be removed).
Our loss ensures that both kinds of structural mistakes are corrected.

Several technical challenges need to be addressed. First,  classical Morse theory was defined for smooth functions on continuous domains. Computing the Morse structures can be expensive and numerically unstable. Furthermore, the entire set of Morse structures may include an excessive amount of structures, a large portion of which can be noisy, irrelevant ones. 
To address these challenges, we use the discrete version of Morse theory by~\cite{forman,forman2002user}. For efficiency purposes, we also use an approximation algorithm to compute 2D Morse structures with almost linear time. The idea is to compute zero dimensional Morse structures of the dual image, which boils down to a minimum spanning tree computation. Finally, we use the theory of persistent homology~\cite{edelsbrunner2000topological,edelsbrunner2010computational} to prune spurious Morse structures that are not relevant.

Our discrete-Morse-theory based loss called the \emph{DMT-loss}, can be evaluated efficiently and can effectively train the neural network to achieve high performance in both topological accuracy and per-pixel accuracy. Our method outperforms state-of-the-art methods in multiple topology-relevant metrics (e.g., ARI and VOI) on various 2D and 3D benchmarks. It has superior performance in the Betti number error, which is an exact measurement of the topological fidelity of the segmentation. 

\section{Method}
\label{sec:method_dmt}
\begin{figure}[ht]
  \centering
  \noindent\makebox[\textwidth][c] {
    \includegraphics[width=0.6\paperwidth]{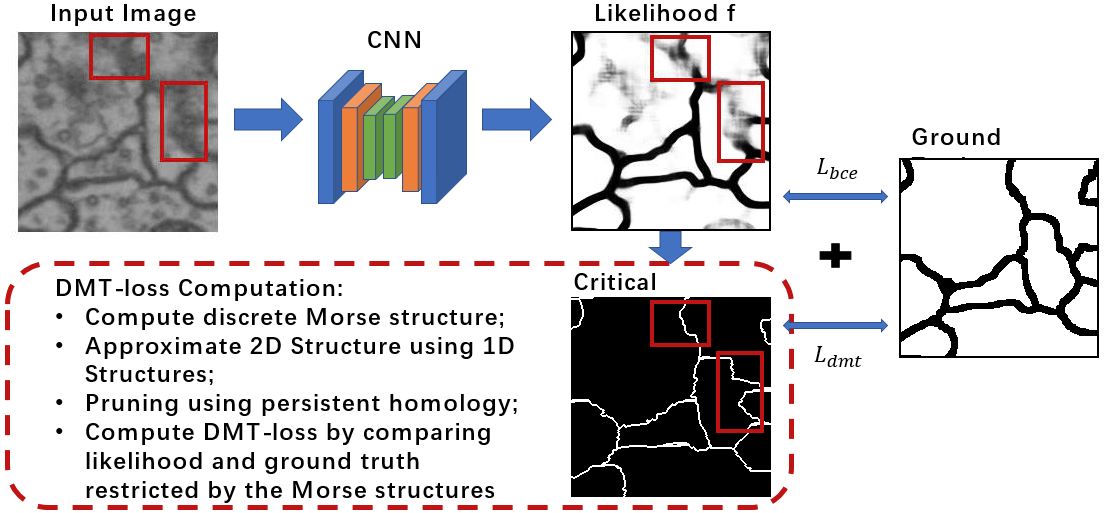}}
    \caption{Overview of our method. Topologically critical and error-prune structures are highlighted.}
      \label{fig:architecture_dmt}
\end{figure}

We propose a novel loss to train a topology-aware network end-to-end. It uses global structures captured by discrete Morse theory (DMT) to discover critical topological structures. In particular, through the language of 1- and 2-stable manifolds, DMT helps identify 1D skeletons or 2D sheets (separating 3D regions) that may be critical for structural accuracy. These Morse structures are used to define a DMT-loss that is essentially the cross-entropy loss constrained to these topologically critical structures. As the training continues, the neural network learns to better predict around these critical structures and eventually achieves better topological accuracy. Please refer to Fig.~\ref{fig:architecture_dmt} for an overview of our method.

\subsection{Morse Theory}
\label{sec:dmt}

\begin{figure*}[ht]
\centering 
\subfigure[]{\label{fig:likelihood}
\includegraphics[width=0.22\textwidth]{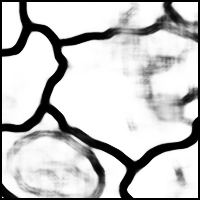}}
\subfigure[]{\label{fig:density}
\includegraphics[width=0.33\textwidth]{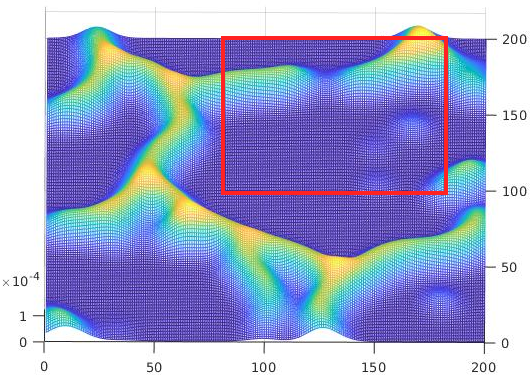}}
\subfigure[]{\label{fig:vpath}
\includegraphics[width=0.33\textwidth]{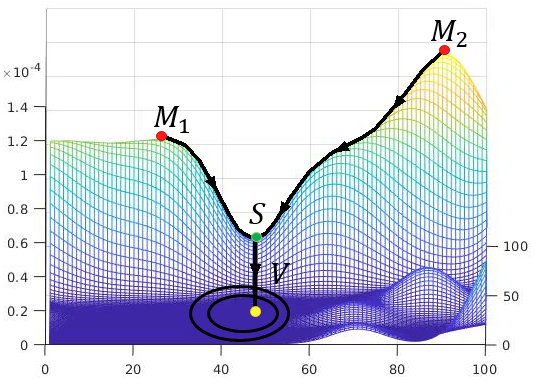}}
\caption{From left to right: \textbf{(a)} Likelihood map. \textbf{(b)} Density map: the $z$-axis value is the probability of the likelihood map in the left figure. \textbf{(c)} Density map for the highlighted region in the middle figure. $M_1$ and $M_2$ are maxima (red dots), $V$ is a minimum (yellow), $S$ is a saddle (green) with its stable manifolds flowing to it from $M_1$ and $M_2$.}
\label{fig:discrete}
\end{figure*}

Morse theory~\cite{milnor1963morse} identifies topologically critical structures from a likelihood map (Fig.~\ref{fig:likelihood}).
In particular, it views the likelihood as a terrain function (Fig.~\ref{fig:density}) and extracts its landscape features such as mountain ridges and their high-dimensional counterparts. 
The broken connection in the likelihood map corresponds to a local dip in the mountain ridge of the terrain in Fig.~\ref{fig:density} and Fig.~\ref{fig:vpath}. The bottom of this dip is captured by a so-called saddle point ($S$ in Fig.~\ref{fig:vpath}) of the likelihood map. The mountain ridge connected to this bottom point captures the main part of the missing pixels. Such ``mountain ridges" can be captured by the so-called stable manifold w.r.t. the saddle point using the language of Morse theory. By finding the saddle points and the stable manifold of the saddle points on the likelihood map, we can ensure the model learns to ``correctly'' handle pixels near these structures. We note that an analogous scenario can also happen with such 1D signals (such as blood vessels) as well as 2D signals (such as membranes of cells) in 3D images -- they can also be captured by saddles (of different indices) and their stable manifolds.

In this chapter, we focus on the application of segmenting 2D and 3D images. Specifically, suppose we have a smooth function $f: \R^{d} \rightarrow \R$ to model the likelihood (density) map. Given any point $x\in \R^d$, the negative gradient $-\nabla f(x) = -[\frac{\partial f}{ \partial x_{1}},\frac{\partial f}{\partial x_{2}},\dots, \frac{\partial f}{\partial x_{d}}]^T$ indicates the steepest descending direction of $f$. 
A point  $x = (x_{1},x_{2},\dots,x_{k})$ is \emph{critical} if the function gradient at this point vanishes (i.e., $\nabla f(x) = 0$).
For a well-behaved function (more formally, called Morse function) defined on $\R^d$, a critical point could be a minimum, a maximum, or $d-1$ type of saddle point. See Fig.~\ref{fig:vpath} for an example.
For $d=2$, there is only one saddle point type. For $d=3$, there are two saddle point types, referred to as index-1 and index-2 saddles. Formally, when taking the eigenvalues of the Hessian matrix at a critical point, its index is equal to the number of negative eigenvalues.

Intuitively, imagine we put a drop of water on the graph of $f$ (i.e, the terrain in Fig.~\ref{fig:density}) at the lift of $x$ onto this terrain, then $-\nabla f(x)$ indicates the direction along which the water will flow down. If we track the trajectory of this water drop as it flows down, this gives rise to a so-called \emph{integral line} (a flow line). Such flow lines can only start and end at critical points~\footnote{More precisely, flow lines only tend to critical points in the limit and never reach them.}, where the gradient vanishes. 

The stable manifold $\Stm(p)$ of a critical point $p$ is defined as the collection of points whose flow line ends at $p$. For a 2D function $f:\R^2\to \R$, for a saddle $q$, its stable manifold $\Stm(q)$ starts from local maxima (mountain peaks in the terrain) and ends at $q$, tracing out the mountain ridges separating different valleys  (Fig.~\ref{fig:vpath}). The stable manifold $\Stm(p)$ of a minimum $p$, on the other hand, corresponds to the entire valley around this minimum $p$. See the valley point $V$ and its corresponding stable field in Fig.~\ref{fig:vpath}. 
For a 3D function $f: \R^3 \to \R$, the stable manifold w.r.t.~an index-2 saddle connects mountain peaks to saddles, tracing 1D mountain ridges as in the case of a 2D function.
The stable manifold w.r.t.~an index-1 saddle $q$ consists of flow lines starting at index-2 saddles and ending at $q$. Their union, called the 2-stable manifold of $f$, consists of a collection of 2-manifold pieces. 

These stable manifolds indicate important topological structures (graph-like or sheet-like) based on the likelihood of the current neural network. Using these structures, we will propose a novel loss (Sec.~\ref{sec:loss}) to improve the topological awareness of the model.
In practice, for images, we will leverage the discrete version of Morse theory for both numerical stability and easier simplification.

\myparagraph{Discrete Morse Theory.}
We view a $d$D image, $d = 2$ or $3$, as a $d$-dimensional cubical complex, meaning it consists of $0$-, $1$-, $2$- and $3$-dimensional cells corresponding to vertices, edges, squares, and voxels (cubes) as its building blocks. 

 Discrete Morse theory (DMT), originally introduced in~\cite{forman,forman2002user}, is a combinatorial version of Morse theory for general cell complexes. There are many beautiful results established for DMT, analogous to classical Morse theory. We will however only briefly introduce some relevant concepts for the present chapter, and we will describe them in the setting of cubical complexes (instead of simplicial complexes) as they are more suitable for images. 

Let $K$ be a cubical complex. Given a $p$-cell $\tau$, we denote by $\sigma < \tau$ if $\sigma$ is a ($p-1$)-dimensional face for $\tau$. 
A \emph{discrete gradient vector} (also called \emph{a V-pair} for simplicity) is a pair $(\tau, \sigma)$ where $\sigma < \tau$. 
Now suppose we are given a collection of V-pairs $\myVF(K)$ over the cubical complex $K$. 
A sequence of cells  $\pi: \tau_{0}^{p+1},\sigma_{1}^{p},\tau_{1}^{p+1},\sigma_{2}^{p},\cdots,\sigma_{k}^{p},\tau_{k}^{p+1}, \sigma_{k+1}^p$, where the superscript $p$ in $\alpha^p$ stands for the dimension of this cell $\alpha$, 
form a \emph{V-path} if $(\tau_i, \sigma_i) \in \myVF(K)$ for for any $i\in [1, k]$ and $\sigma_i < \tau_{i-1}$ for any $i\in [1, k+1]$. A V-path $\pi$ is \emph{acyclic} if $(\tau_0, \sigma_{k+1}) \notin \myVF(K)$. 
This collection of V-pairs $\myVF(K)$ form a \emph{discrete gradient vector field}~\footnote{We will not introduce the concept of discrete Morse function, as the discrete gradient vector field is sufficient to define all relevant notations.} if (cond-i) each cell in $\myVF(K)$ can only appear in at most one pair in $\myVF(K)$; and (cond-ii) all V-paths in $\myVF(K)$ are acyclic. 
Given a discrete gradient vector field $\myVF(K)$, a simplex $\sigma \in K$ is \emph{critical} if it is not in any V-pair in $\myVF(K)$. 

Even though a discrete gradient vector (a V-pair), say $(\tau, \sigma)$ is a combinatorial pair instead of a real vector, it still indicates a ``flow" from $\tau$ to its face $\sigma$. A V-path thus corresponds to a flow path (integral line) in the smooth setting. However, to make a collection of V-pairs a valid analog of gradient field, (cond-i) says that at each simplex there should only be one ``flow" direction; while (cond-ii) is necessary as flow lines traced by gradient can only go down in function values and thus never come back (thus acyclic). 

A critical simplex has a ``vanishing gradient" as it is not involved in any V-pair in $\myVF(K)$ (i.e, there is no flow at this simplex). 
Given a 2D cubical complex $K$ a discrete gradient vector field $\myVF(K)$, we can view critical $0$-, $1$-, $2$- and $3$-cells as minima, saddle points, and maxima, respectively. 
If $K$ is 3D, then we can view critical $0$-, $1$-, $2$- and $3$-cells as minima, index-1 saddle, index-2 saddle and maxima, respectively. 

Hence, a 1-stable manifold in 2D will correspond to a V-path connecting a critical square (a maximum) and a critical edge (a saddle), while in 3D, it will be a V-path connecting a critical cube and a critical square. 

\myparagraph{Morse Cancellation.} 
A given discrete gradient field $\myVF(K)$ could be noisy, e.g, there are shallow valleys where the mountain ridge around it should be ignored. Fortunately, the discrete Morse theory provides an elegant and purely combinatorial way to cancel pairs of critical simplices (and thus reduce their stable manifolds). In particular, given $\myVF(K)$, a pair of critical simplices $\langle \delta^{(p+1), \gamma^p} \rangle$ is \emph{cancellable} if there is a unique V-path $\pi = \delta=\delta_0, \gamma_1, \delta_1, \ldots, \delta_s, \gamma_{s+1} = \gamma$ from $\delta$ to $\gamma$. The \emph{Morse cancellation operation} simply reverses all V-pairs along this path by removing all V-pairs along this path and adding $(\delta_{i-1}, \gamma_i)$ to $\myVF(K)$ for any $i \in [1, s+1]$. It is easy to check that after the cancellation neither $\delta$ nor $\gamma$ is critical. 

\subsection{Simplification and Computation}
\label{sec:computation}
In this section, we describe how we extract discrete Morse structures corresponding to the 1-stable and  2-stable manifolds in the continuous analog. First, we prune unnecessary Morse structures, based on the theory of persistent homology~\cite{edelsbrunner2000topological,edelsbrunner2010computational}. Second, we approximate the 2-stable manifold structures using 0-stable manifolds of the dual to achieve high efficiency in practice, because it is rather involved to compute based on the original definition.

\begin{figure*}[ht]
\centering
\subfigure[]{\label{fig:sample}
\includegraphics[width=0.22\textwidth]{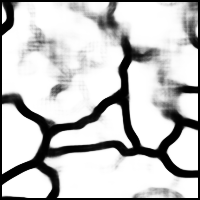}}
\subfigure[]{\label{fig:sample_gt}
\includegraphics[width=0.22\textwidth]{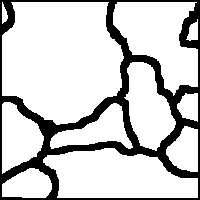}}
\subfigure[]{\label{fig:under_pruning}
\includegraphics[width=0.22\textwidth]{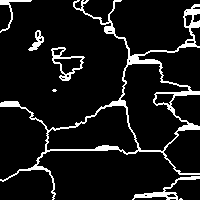}}
\subfigure[]{\label{fig:proper_pruning}
\includegraphics[width=0.22\textwidth]{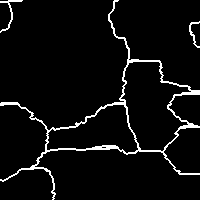}}
\caption{From left to right: \textbf{(a)} Sample likelihood map, \textbf{(b)} Ground truth, \textbf{(c)} improperly pruned structures and \textbf{(d)} properly pruned structures.}
\label{fig:pruning}
\end{figure*}

\myparagraph{Persistence-based Structure Pruning.} 
While Morse structures reveal important structural information, they can be sensitive to noise. Without proper pruning, there can be an excessive amount of Morse structures, many of which are spurious and not relevant to the true signal. See Fig.~\ref{fig:under_pruning} for an example.
Similar to previous approaches e.g., \cite{2011MNRAS,DRS15,WWL15}, we will prune these structures using persistent homology.

Persistent homology is one of the most important developments in the field of topological data analysis in the past two decades~\cite{edelsbrunner2010computational,edelsbrunner2000topological}. Intuitively speaking, we grow the complex by starting from the empty set and gradually including more and more cells using a decreasing threshold. Through this course, new topological features can be created upon adding a critical cell, and sometimes a feature can be destroyed upon adding another critical cell. The persistence algorithm~\cite{edelsbrunner2000topological} will pair up these critical cells; that is, its output is a set of critical cell pairs, where each pair captures the birth and death of topological features during this evolution. The persistence of a pair is defined as the difference of function values of the two critical cells, intuitively measuring how long the topological feature lives in terms of $f$. 

Using persistence, we can prune critical cells that are less topologically salient, and thus their corresponding Morse structures. Recall each 1- and 2-stable Morse structure is constituted by V-paths flowing into a critical cell (corresponding to a saddle in the continuous setting). We then use the persistence associated with this critical cell to determine the saliency of the corresponding Morse structure. If the persistence is below a certain threshold $\epsilon$, we prune the corresponding Morse structure via an operation called \emph{Morse cancellation}.~\footnote{Technically, not all spurious structures can be pruned/canceled. But in practice, most of them can.} See Fig.~\ref{fig:proper_pruning} for example Morse structures after pruning. We denote by $\myS_1(\epsilon)$ and $\myS_2(\epsilon)$ the remaining sets of 1- and 2-stable manifolds after pruning. We'll use these Morse structures to define the loss (Sec.~\ref{sec:loss}).

\myparagraph{Persistence Pruning.}
\label{sec:prune}
We can extend this vertex-valued function $\rho$ to a function $\rho: K \to \reals$, by setting $\rho(\sigma)$ for each cell to be the maximum $\rho$-value of each vertex in $\sigma$. 
How to obtain a discrete gradient vector field from such function $\rho: K \to \reals$? Following the approach developed in~\cite{WWL15,DWW18}, we initialize a trivial discrete gradient vector field where all cells are initially critical. Let $\epsilon > 0$ be a threshold for simplification. We then perform persistence algorithm~\cite{edelsbrunner2000topological} induced by the super-level set filtration of $\rho$ and pair up all cells in $K$, denoted by $\myP_\rho(K)$.

Persistent homology is one of the most important developments in the field of topological data analysis in the past two decades~\cite{edelsbrunner2010computational,edelsbrunner2000topological,zomorodian2005computing}. We will not introduce it formally here. Imagine we grow the complex $K$ by starting from the empty set and gradually including more and more cells in decreasing $\rho$ values. (More formally, this is the so-called super-level set filtration of $K$ induced by $\rho$.) Through this course, the new topological features can be created upon adding a simplex $\sigma$, and sometimes a feature can be destroyed upon adding a simplex $\tau$. Persistence algorithm~\cite{edelsbrunner2000topological} will pair up simplices; that is, its output is a set of pairs of simplices $\myP_\rho(K) = \{ (\sigma, \tau) \}$, where each pair captures the birth and death of topological features during this evolution. The persistence of a pair, say $\pp = (\sigma, \tau)$, is defined as $\pers(\pp) = \rho(\sigma) - \rho(\tau)$, measuring how long the topological feature captured by $\pp$ lives in term of $\rho$. In this case, we also write $\pers(\sigma) = \pers(\tau) = \pers(\pp)$ -- the persistence of a simplex (say $\sigma$ or $\tau$) can be viewed as the importance of this simplex. 

With this intuition of the persistence pairings, we next perform Morse cancellation operation to all pairs of these cells $(\sigma, \tau) \in \myP_\rho(K)$ in increasing order their persistence if (i) its persistence $\pers(\delta, \gamma) < \epsilon$ (i.e, this pair has low persistence and thus not important); and (ii) this pair $(\delta, \gamma)$ is cancellable. 

Let $\myVF_\epsilon(K)$ be the resulting discrete gradient field after simplifying all low-persistence critical simplices. We then construct the 1-stable and 2-stable manifolds for the remaining (high persistence, and thus important) saddles (critical 1-cell and 2-cells) from $\myVF_\epsilon(K)$. 
Let $\myS_1(\epsilon)$ and $\myS_2(\epsilon)$ be the
resulting collection of 1- and 2-stable manifolds respectively. 
In particular, see an illustration of a V-path (highlighted in black) corresponding to a 1-stable manifold of the green saddle in Fig. 3(c).

\myparagraph{Computation.} 
We need an efficient algorithm to compute $\myS_1(\epsilon)$ and $\myS_2(\epsilon)$ from a given likelihood $f$, because this computation needs to be carried out at each epoch. 
It is significantly more involved to define and compute $\myS_2(\epsilon)$ in the discrete Morse setting~\cite{DRS15}. Furthermore, this also requires the computation of persistent homology up to 2-dimensions, which takes time $T = O(n^\omega)$ (where $\omega \approx 2.37$ is the exponent in the matrix multiplication time, i.e., the time to multiply two $n\times n$ matrices). 
To this end, we propose to approximate $\myS_2(\epsilon)$ by $\hatS_2(\epsilon)$ which intuitively comes from the ``boundary'' of the stable manifold for minima. Note that in the smooth case for a function $f: \R^3\to \R$, the closure of 2-stable manifolds exactly corresponds to the $2$D sheets on the boundary of stable manifolds of minima.~\footnote{This however is not always true for the discrete Morse setting.} This is both conceptually clear and also avoids the costly persistence computation. 
In particular, given a minimum $q$ with persistence  greater than the pruning threshold $\epsilon$, 
the collections of V-paths ending at the minimum $q$ form a spanning tree $T_q$. 
In fact, consider all minima $\{q_1, \ldots, q_\ell\}$ with persistence at least $\epsilon$. $\{ T_{q_i}\}$ form a maximum spanning forest among all edges with persistence value smaller than $\epsilon$~\cite{Bauer2012, DWW18}. Hence it can be computed easily in $O(n\log n)$ time, where $n$ is the image size.

\begin{wrapfigure}{r}{0.3\textwidth}
\centering 
    \includegraphics[width=0.3\textwidth]{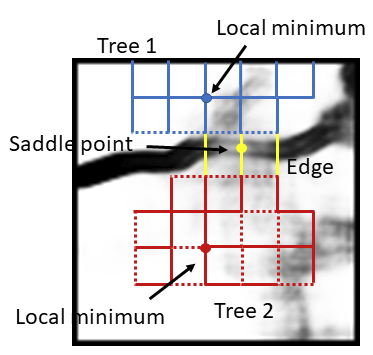}
  \caption{Spanning tree illustration.}
\label{fig:approximation}
\end{wrapfigure}

\myparagraph{More Details on the Approximation of $\myS_2$ via $\hat{S_2}$.}
We approximate $S_2$ by taking the boundary of the stable manifold of the minima (basins/valleys in the terrain). This is like a watershed algorithm: growing the basins from all minima until they meet. The stable manifolds of the minima are approximated using spanning trees. This algorithm is inspired by the continuous analog for Morse functions.

We then take all the edges incident to nodes from different trees. The dual of these edges, denoted as $\hatS_2(\epsilon)$, serves as the ``boundaries'' separating different spanning trees (representing stable manifolds to different minima with persistence $\ge \epsilon$). See Fig.~\ref{fig:approximation}. 
Overall, the computation of $\hatS_2(\epsilon)$ takes only $O(n\log n)$ by a maximum spanning tree algorithm.

As for $\myS_1(\epsilon)$, we use a simplified algorithm of~\cite{DWW18}, which can compute $\myS_1(\epsilon)$ in $O(n\log n)$ time for a 2D image, in which $n$ is the image size. For a 3D image, the time is $O(n\log n + T)$, where $T = O(n^\omega)$ is the time to compute persistent homology, where $\omega \approx 2.37$ is the exponent in matrix multiplication time.

\subsection{The DMT-based Loss Function and Training Details}

\label{subsec:lossfunc}
\label{sec:loss}

Our loss has two terms, the cross-entropy term, $L_{bce}$ and the DMT-loss, $L_{dmt}$: $L(f,g)= L_{bce}(f,g)+\beta L_{dmt}(f,g)$, in which $f$ is the likelihood, $g$ is the ground truth, and $\beta$ is the weight of $L_{dmt}$. Here we focus on one single image, while the actual loss is aggregated over the whole training set. 

The DMT-loss enforces the correctness of the topologically challenging locations discovered by our algorithm. These locations are pixels of the (approximation of) 1- and 2-stable manifolds $\myS_1(\epsilon)$ and $\hatS_2(\epsilon)$ of the likelihood, $f$. 
Denote by $\mathcal{M}_f$ a binary mask of the union of pixels of all Morse structures in $\myS_1(\epsilon)\cup \hatS_2(\epsilon)$.
We want to enforce these locations to be correctly segmented. We use the cross-entropy between the likelihood map $f$ and ground truth $g$ restricted to the Morse structures, formally, 
$L_{dmt}(f, g) = L_{bce}(f\circ \mathcal{M}_f, g\circ \mathcal{M}_f)$,
in which $\circ$ is the Hadamard product.

\myparagraph{Different Topological Error Types.} Recall that the Morse structures are computed over the potentially noisy likelihood function of a neural network, which can help identify two types of structural errors: (1) \textbf{false negative}: a true structure that is incomplete in the segmentation, but can be visible in the Morse structures. These types of false negatives (broken connections, holes in membrane) can be restored as the additional cross-entropy loss near the Morse structures will force the network to increase its likelihood value on these structures. (2) \textbf{false positive}: phantom structures hallucinated by the network when they do not exist (spurious branches, membrane pieces). These errors can be eliminated as the extra cross entropy loss on these structures will force the network to decrease the likelihood values along these structures. 
\myparagraph{Illustration of 3D Topological Errors.}
We have already introduced discrete Morse theory with a 2D example. Here, we would like to illustrate 3D topological errors with 3D examples.

\begin{figure*}[ht]
\centering 
\subfigure{
\includegraphics[width=.8\textwidth]{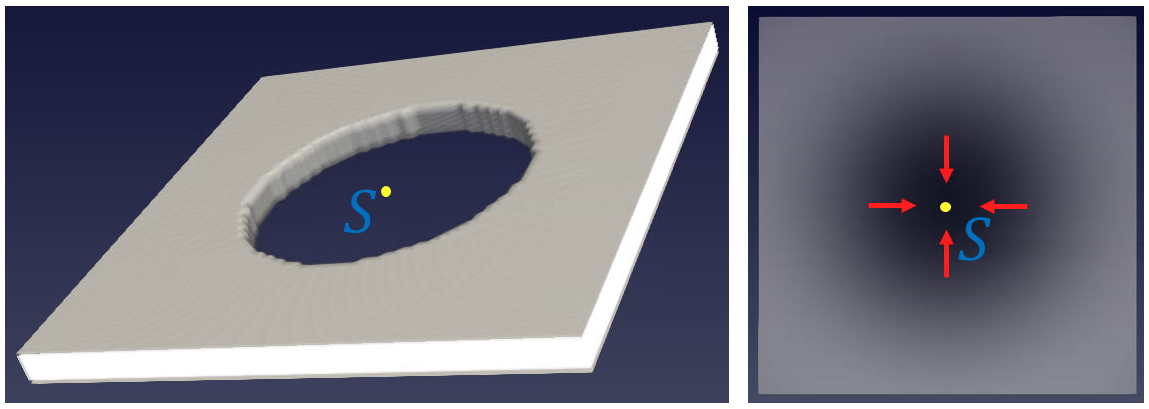}}
\caption{Illustration of an index-1 topological error (3D hole in the middle) for a 3D synthetic data. The ground truth is a complete sheet without a hole. We intentionally weaken the likelihood function in the middle. So the segmentation has a hole in the middle. \textbf{Left}: 3D segmentation result. $S$ is a saddle point of the likelihood function. Its Hessian has 1 negative and 2 positive eigenvalues. The stable manifold of the saddle point $S$ is a 2D plane going through the saddle point and cutting the segmentation into two thin slices. \textbf{Right}: the likelihood function visualized on the 2D stable manifold of $S$. Red arrows illustrate how different $V$-paths (streamlines of negative gradient) flow to the saddle $S$.}
\label{fig:index1}
\end{figure*}

\begin{figure*}[ht]
\centering 
\subfigure{
\includegraphics[width=0.8\textwidth]{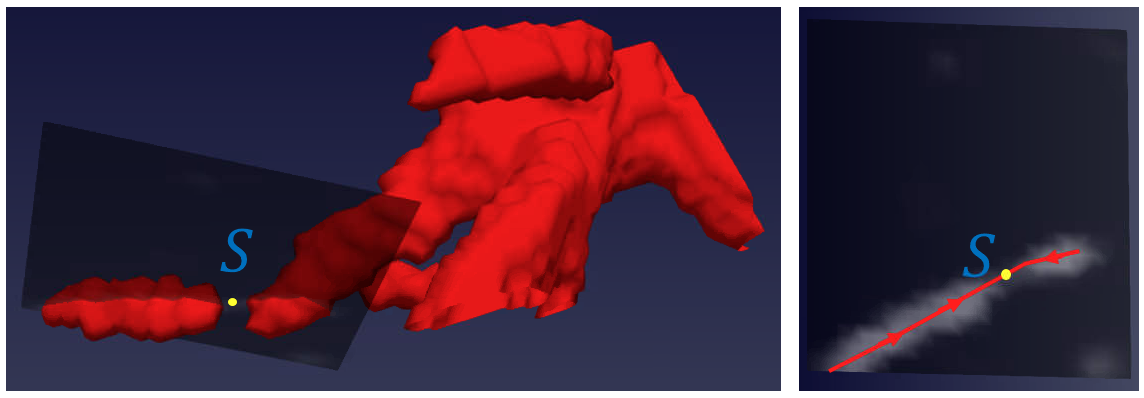}}
\caption{Illustration of an index-2 topological error from a 3D vessel image. The vessel segmentation has a broken connection near the bottom-left of the Left image.
\textbf{Left}: part of the 3D segmentation result. The saddle point $S$ corresponds to a broken connection. The Hessian of the likelihood at the saddle point has 2 negative and 1 positive eigenvalues. \textbf{Right}: One slice of the 3D likelihood map passing the saddle point. The saddle point (yellow) and its 1-stable manifold (red) are also drawn.}
\label{fig:index2}
\end{figure*}

Fig.~\ref{fig:index1} and Fig.~\ref{fig:index2} illustrate two different types of topological errors for 3D data. 
Fig.\ref{fig:index1} illustrates an index-1 topological error for 3D synthetic data. 3D EM/neuron has the same type of topological error as the synthetic data.
Fig.\ref{fig:index2} illustrates index-2 topological error for 3D vessel data. 

\begin{figure*}[ht]
\centering 
\subfigure[]{
\includegraphics[width=0.22\textwidth]{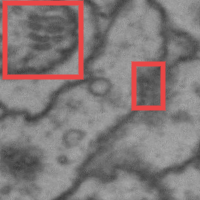}}
\subfigure[]{\label{fig:supple_likelihood}
\includegraphics[width=0.22\textwidth]{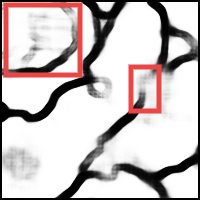}}
\subfigure[]{\label{fig:supple_gt}
\includegraphics[width=0.22\textwidth]{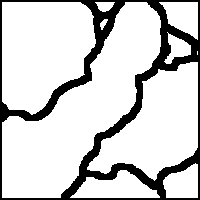}}
\subfigure[]{\label{fig:supple_dmt}
\includegraphics[width=0.22\textwidth]{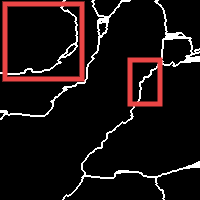}}
\caption{Illustration of two different types of topological errors captured by DMT-loss. \textbf{(a)} an input neuron image with challenging locations highlighted. \textbf{(b)} likelihood map of a baseline method without topological guarantee~\cite{ronneberger2015u}. \textbf{(c)} ground truth. \textbf{(d)} The topologically critical structure from the likelihood, captured by the proposed discrete Morse algorithm. These structures will be used in the DMT-Loss. }
\label{fig:illu}
\end{figure*}

\myparagraph{False Negative and False Positive Errors.}
\label{sec:false}
We have mentioned that the proposed DMT-loss can capture and fix two different types of topological errors: false negative and false positive.
We illustrate these two types in Fig.~\ref{fig:illu}. The two highlighted red rectangles represent the two types of topological errors: 1) The red rectangle on the right represents a sample of false negative error; part of the membrane structure is missing, due to a blurred region near the membrane. 2) The red rectangle on the left represents a sample of a false positive error. In this specific case, it is caused by mitochondria which are not the boundary of neurons. 

In summary, with the help of the proposed DMT-loss, we can identify both these two types of
topological errors, and then force the network to increase/decrease its likelihood value on these structures to correctly segment the images with the correct topology.

\myparagraph{Differentiability.} We note that the Morse structures are recomputed at every epoch. The structures, as well as their mask $\mathcal{M}_f$, may change with $f$. However, the change is not continuous; the output of the discrete Morse algorithm is a combinatorial solution that does not change continuously with $f$. Instead, it only changes at singularities, i.e., when the function values of $f$ at different pixels/voxels are the same. In other words, for a general $f$, the likelihood function is real-valued, so it is unlikely two pixels share the exact same value. In case they are, the persistence homology algorithm by default will break the tie and choose one as critical. 
The mask $\mathcal{M}_f$ remains a constant within a small neighborhood of current $f$. Therefore, the gradient of $L_{dmt}$ exists and can be computed naturally.

\myparagraph{Training Details.} 
Although our method is architecture-agnostic, for 2D datasets, we select an architecture driven by a 2D UNet~\cite{ronneberger2015u}; for 3D datasets, we select an architecture inspired by a 3D UNet~\cite{cciccek20163d}. Both UNet and 3D UNet were originally designed for neuron segmentation tasks, capturing the fine-structures of images. In practice, we first pretrain the network with only the cross-entropy loss, and then train the network with the combined loss.

\section{Experiments on 2D Datasets}
\subsection{Datasets}
Six natural and biomedical 2D datasets are used: \textbf{ISBI12}~\cite{arganda2015crowdsourcing}, \textbf{ISBI13}~\cite{arganda20133d},  \textbf{CREMI},
\textbf{CrackTree}~\cite{zou2012cracktree}, \textbf{Mass.}~\cite{mnih2013machine} and  \textbf{DRIVE}~\cite{staal2004ridge}. The details of the datasets have been introduced in Sec.~\ref{sec:dataset_topoloss}.
For all the experiments, we use a 3-fold cross-validation to tune hyperparameters for both the proposed method and other baselines, and report the mean performance over the validation set. This also holds for 3D experiments.

\subsection{Evaluation Metrics} 
We use five different evaluation metrics: \textbf{Pixel-wise accuracy}, \textbf{DICE score}, \textbf{ARI}, \textbf{VOI}, and the most important one is \textbf{Betti number error}, which directly compares the topology (number of handles/voids) between the segmentation and the ground truth.
More details about the evaluation metrics have been provided in Sec.~\ref{sec:metric_topoloss} and Sec.~\ref{sec:metric_warping}. The last three metrics are topology-aware.

\subsection{Baselines} We use \textbf{{DIVE}}~\cite{fakhry2016deep}, \textbf{{UNet}}~\cite{ronneberger2015u}, \textbf{{UNet-VGG}}~\cite{mosinska2018beyond} and \textbf{{TopoLoss}}~\cite{hu2019topology} as baselines. All the baselines have been described in Sec.~\ref{sec:baseline_topoloss}. For all methods, we generate segmentations by thresholding the predicted likelihood maps at 0.5, and this also holds for 3D experiments. 

\subsection{Results}
Tab.~\ref{table:2d_dmt} shows quantitative results for 2D image datasets. The results are highlighted when they are significantly better, and statistical significance is determined by t-tests. The DMT-loss outperforms others in both DICE score and topological accuracy (ARI, VOI, and Betti Error).
Please note that here the backbone of TopoLoss is the same as in~\cite{hu2019topology}, a heavily engineered network. The performance of TopoLoss will be worse if we implement it using the same UNet backbone as DMT-Loss. 

\begin{figure*}[ht]
\centering 
\subfigure{
\includegraphics[width=0.12\textwidth]{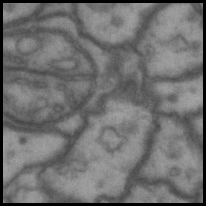}}
\subfigure{
\includegraphics[width=0.12\textwidth]{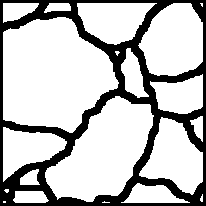}}
\subfigure{
\includegraphics[width=0.12\textwidth]{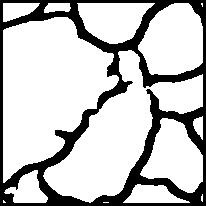}}
\subfigure{
\includegraphics[width=0.12\textwidth]{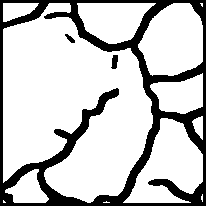}}
\subfigure{ 
\includegraphics[width=0.12\textwidth]{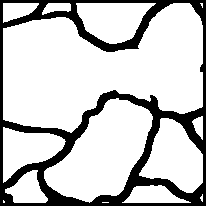}}
\subfigure{
\includegraphics[width=0.12\textwidth]{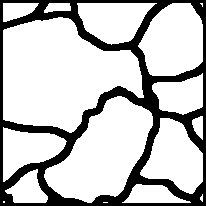}}
\subfigure{
\includegraphics[width=0.12\textwidth]{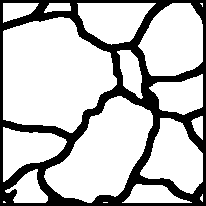}}

\subfigure{
\includegraphics[width=0.12\textwidth]{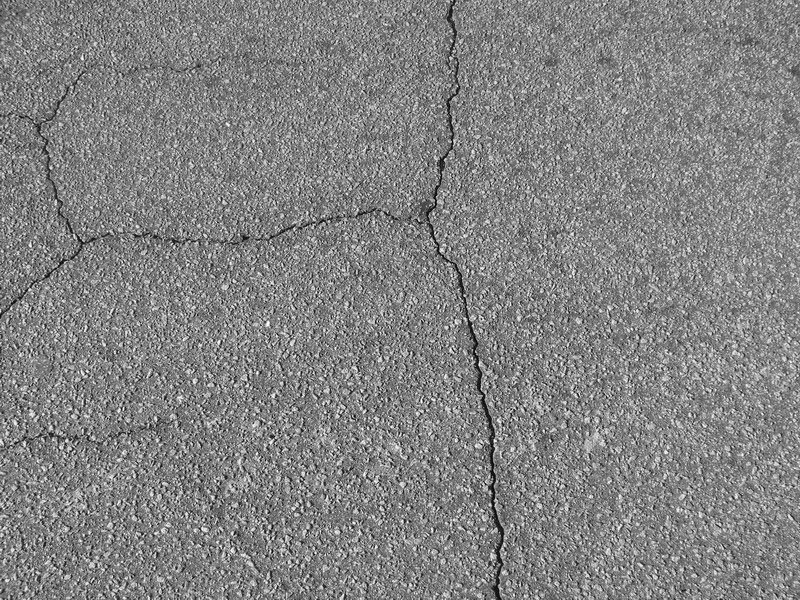}}
\subfigure{
\includegraphics[width=0.12\textwidth]{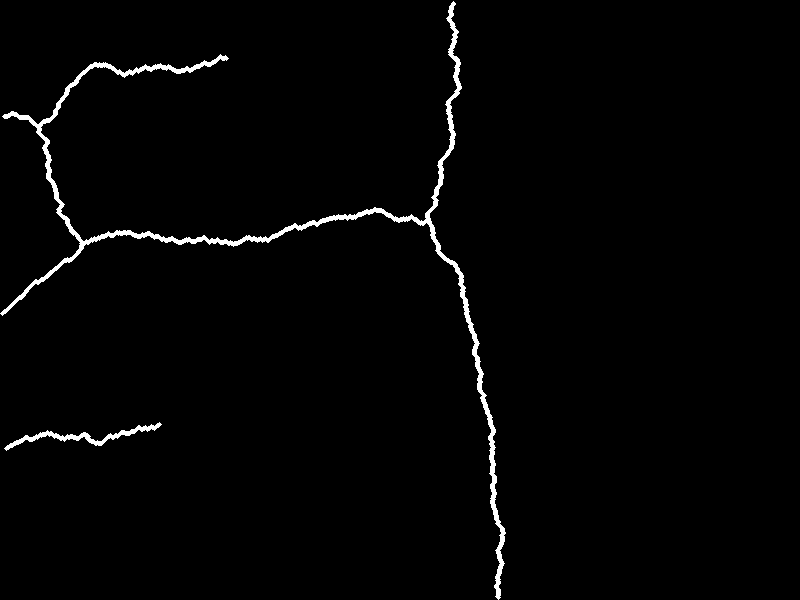}}
\subfigure{
\includegraphics[width=0.12\textwidth]{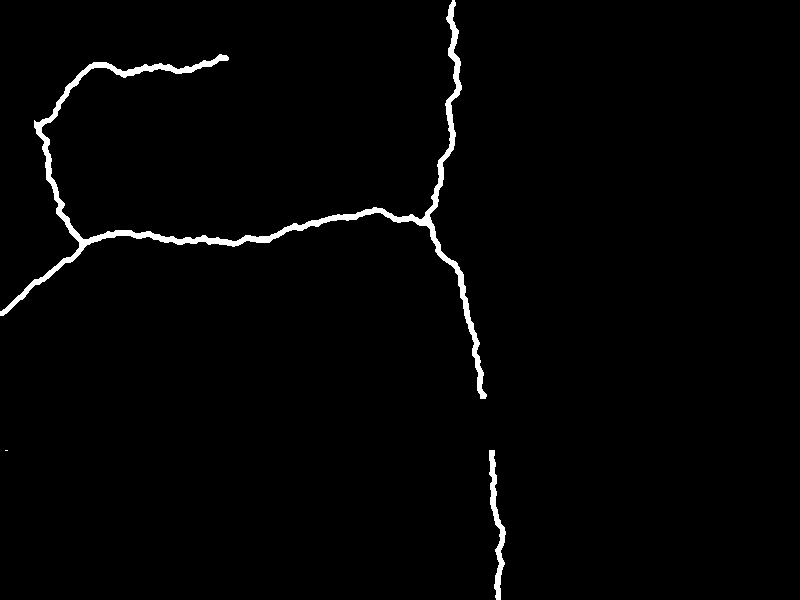}}
\subfigure{
\includegraphics[width=0.12\textwidth]{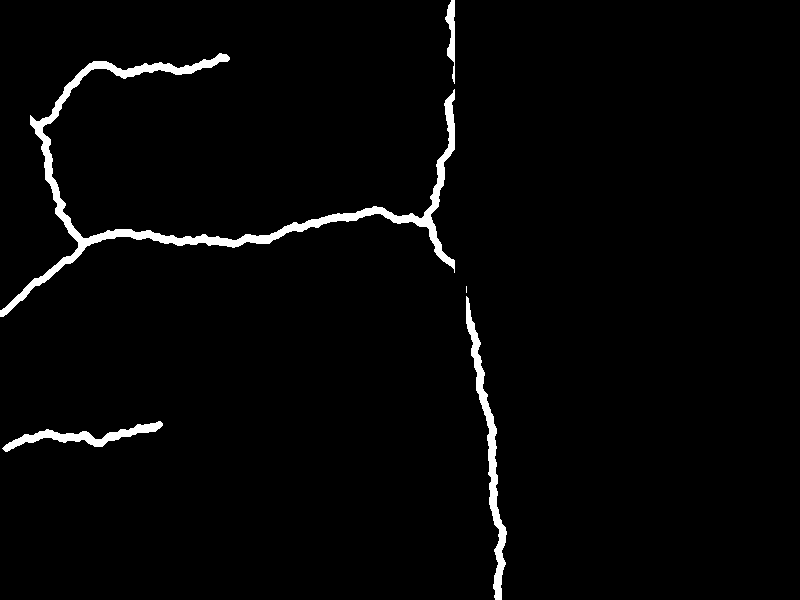}}
\subfigure{ 
\includegraphics[width=0.12\textwidth]{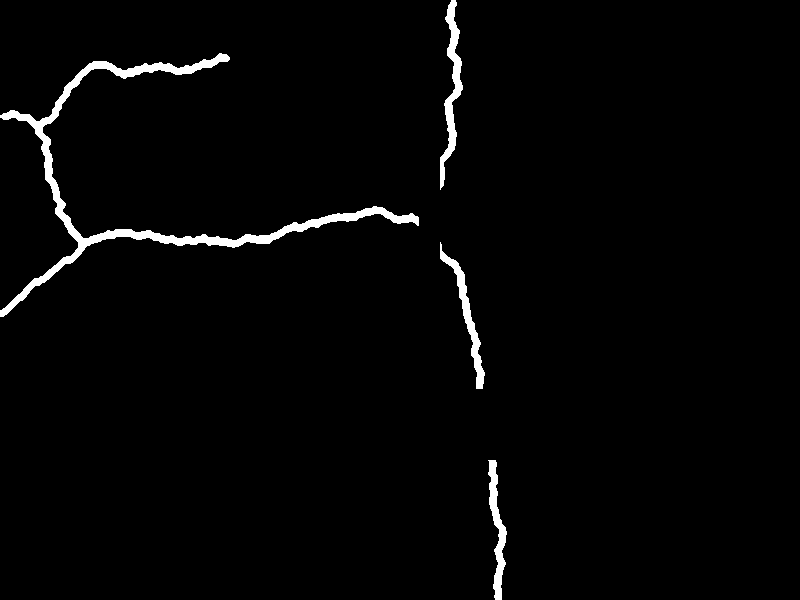}}
\subfigure{
\includegraphics[width=0.12\textwidth]{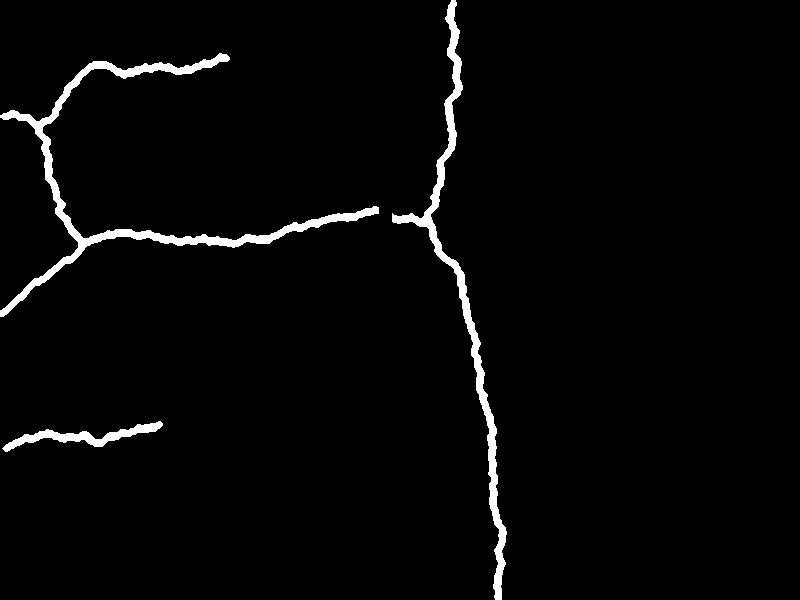}}
\subfigure{
\includegraphics[width=0.12\textwidth]{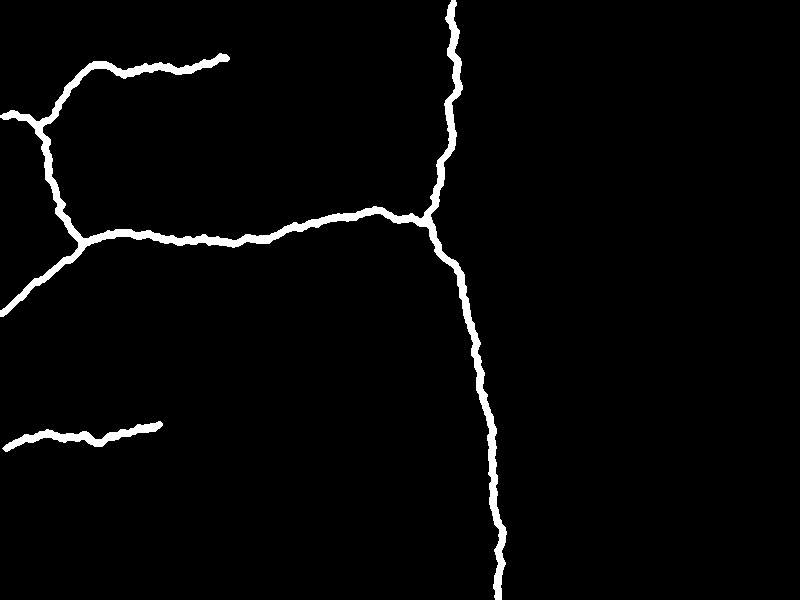}}

\subfigure{
\stackunder{\includegraphics[width=0.12\textwidth]{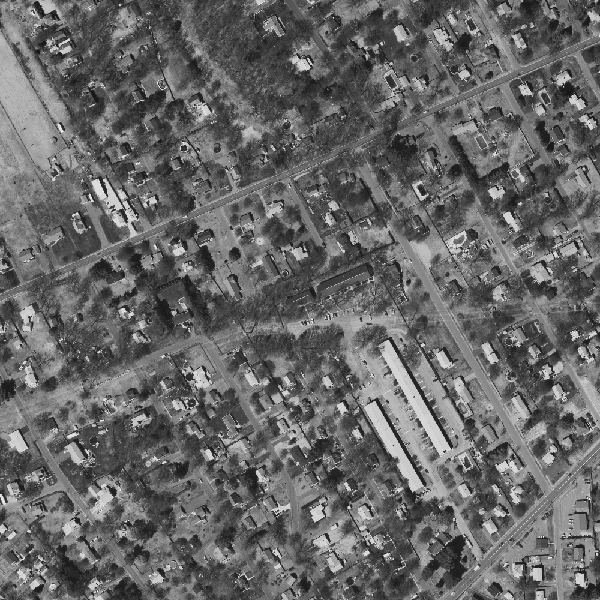}}{(a)}}
\subfigure{
\stackunder{\includegraphics[width=0.12\textwidth]{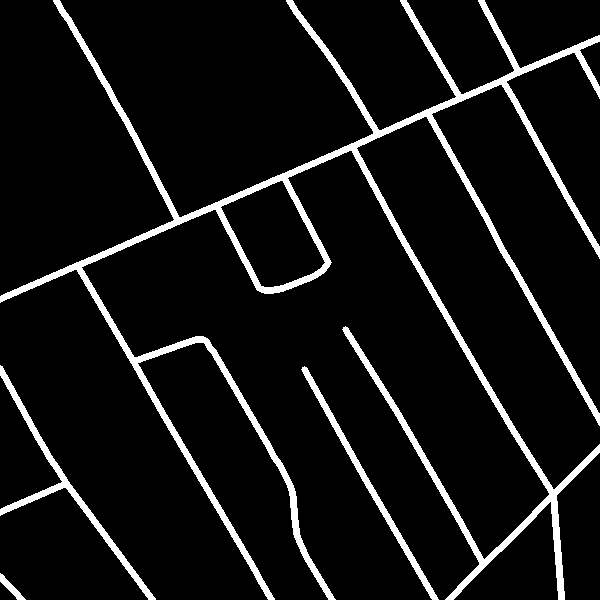}}{(b)}}
\subfigure{
\stackunder{\includegraphics[width=0.12\textwidth]{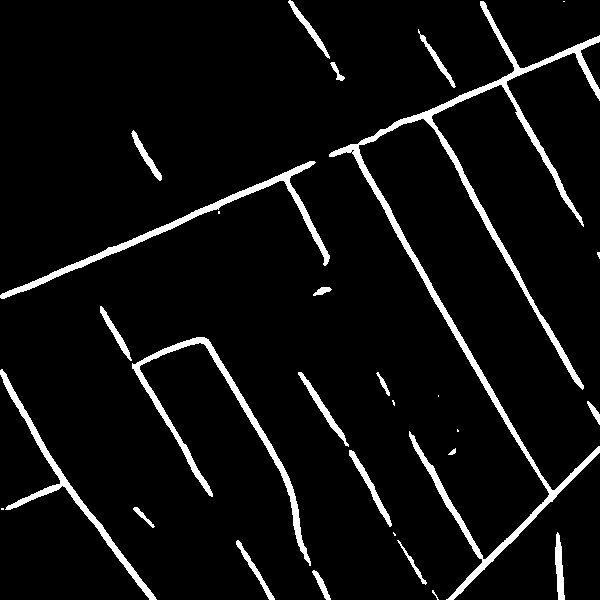}}{(c)}}
\subfigure{
\stackunder{\includegraphics[width=0.12\textwidth]{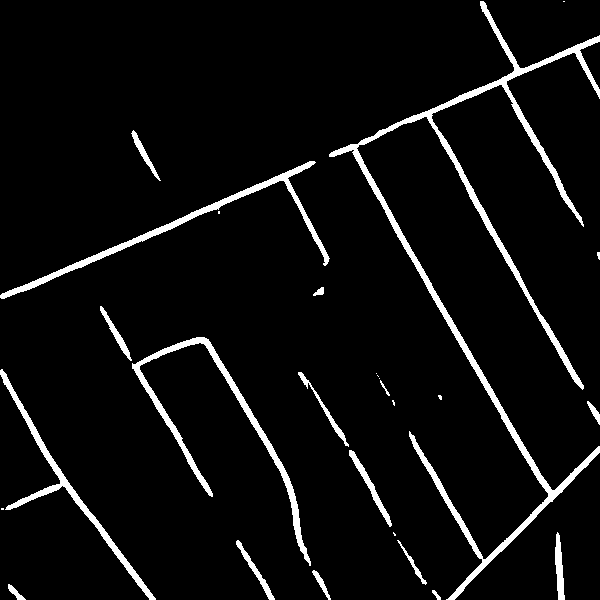}}{(d)}}
\subfigure{ 
\stackunder{\includegraphics[width=0.12\textwidth]{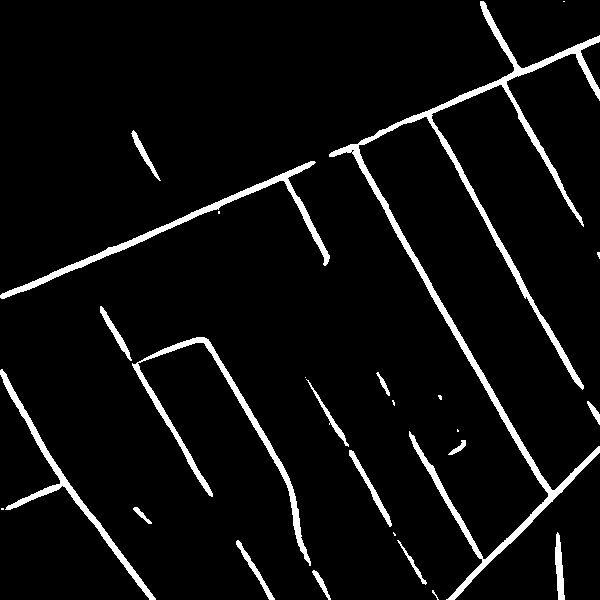}}{(e)}}
\subfigure{
\stackunder{\includegraphics[width=0.12\textwidth]{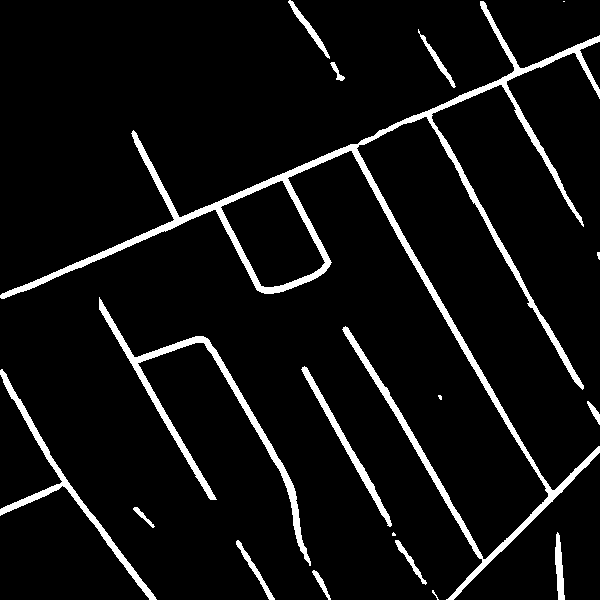}}{(f)}}
\subfigure{
\stackunder{\includegraphics[width=0.12\textwidth]{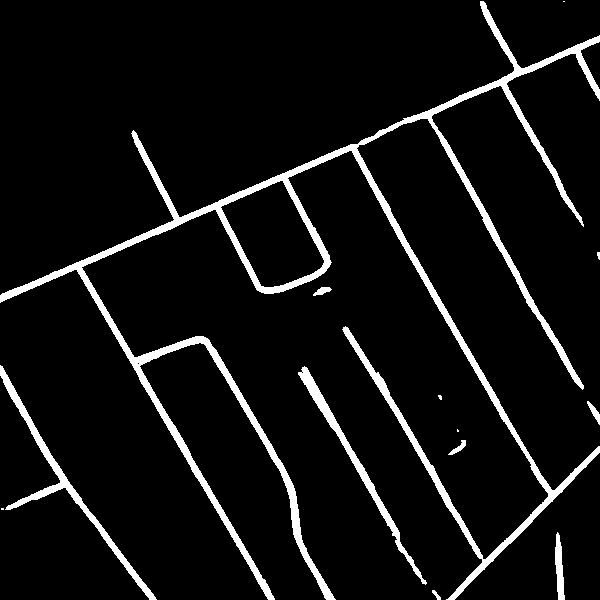}}{(g)}}
\caption{Qualitative results of the proposed method compared to other models. From left to right: sample images, ground truth, results for \textbf{DIVE}, \textbf{UNet}, \textbf{Mosin.}, \textbf{TopoLoss} and our proposed \textbf{DMT}.}
\label{fig:quantitative}
\end{figure*}

Fig.~\ref{fig:quantitative} shows qualitative results. Our method correctly segments fine structures such as membranes, roads, and vessels. Our loss is a weighted combination of the cross entropy and DMT-losses. When $\beta = 0$, the proposed method degrades to a standard UNet. The performance improvement over all datasets (UNet and DMT line in Tab.~\ref{table:2d_dmt}) demonstrates that our DMT-loss is helping the deep neural nets to learn a better structural segmentation.

\setlength{\tabcolsep}{3pt}
\begin{table*}[ht]
\begin{center}
\scriptsize
\caption{Quantitative results for different models on several 2D datasets.}
\label{table:2d_dmt}
\begin{tabular}{ccccccc}

Method & Accuracy & DICE & ARI & VOI  & Betti Error\\
\hline
\hline
\multicolumn{6}{c}{ISBI13} \\
\hline
DIVE & 0.9642 $\pm$ 0.0018 & 0.9658 $\pm$ 0.0020 & 0.6923 $\pm$ 0.0134  & 2.790 $\pm$ 0.025 & 3.875 $\pm$ 0.326\\
UNet & 0.9631 $\pm$ 0.0024 &0.9649 $\pm$ 0.0057 &0.7031 $\pm$ 0.0256 &  2.583 $\pm$ 0.078 &3.463 $\pm$ 0.435\\
Mosin. & 0.9578 $\pm$ 0.0029 & 0.9623 $\pm$ 0.0047 & 0.7483 $\pm$ 0.0367 & 1.534 $\pm$ 0.063 & 2.952 $\pm$ 0.379\\
TopoLoss & 0.9569 $\pm$ 0.0031 & \textbf{0.9689 $\pm$ 0.0026} & 0.8064 $\pm$ 0.0112 & 1.436 $\pm$ 0.008& \textbf{1.253 $\pm$ 0.172}\\
DMT & 0.9625 $\pm$ 0.0027 & \textbf{0.9712 $\pm$ 0.0047} &\textbf{0.8289 $\pm$ 0.0189} & \textbf{1.176 $\pm$ 0.052}& \textbf{1.102 $\pm$ 0.203}\\
\hline
\multicolumn{6}{c}{CREMI} \\
\hline
DIVE & 0.9498 $\pm$ 0.0029 & 0.9542 $\pm$ 0.0037 & 0.6532 $\pm$ 0.0247 & 2.513 $\pm$ 0.047  & 4.378 $\pm$ 0.152\\
UNet & 0.9468 $\pm$ 0.0048 & 0.9523 $\pm$ 0.0049 & 0.6723 $\pm$ 0.0312 & 2.346 $\pm$ 0.105 & 3.016 $\pm$ 0.253\\
Mosin. & 0.9467 $\pm$ 0.0058 & 0.9489 $\pm$ 0.0053 & 0.7853 $\pm$ 0.0281 & 1.623 $\pm$ 0.083 & 1.973 $\pm$ 0.310\\
TopoLoss & 0.9456 $\pm$ 0.0053 & 0.9596 $\pm$ 0.0029 &0.8083 $\pm$ 0.0104 & 1.462 $\pm$ 0.028  & \textbf{1.113 $\pm$ 0.224}\\
DMT & 0.9475 $\pm$ 0.0031 & \textbf{0.9653 $\pm$ 0.0019} & \textbf{0.8203 $\pm$ 0.0147} & \textbf{1.089 $\pm$ 0.061}  & \textbf{0.982 $\pm$ 0.179}\\
\hline
\multicolumn{6}{c}{ISBI12} \\
\hline

DIVE & 0.9640 $\pm$ 0.0042 & 0.9709 $\pm$ 0.0029& 0.9434 $\pm$ 0.0087 & 1.235 $\pm$ 0.025 & 3.187 $\pm$ 0.307\\

UNet & 0.9678 $\pm$ 0.0021 &0.9699 $\pm$ 0.0048 & 0.9338 $\pm$ 0.0072 & 1.367 $\pm$ 0.031 & 2.785 $\pm$ 0.269\\
Mosin. & 0.9532 $\pm$ 0.0063 & 0.9716 $\pm$ 0.0022 & 0.9312 $\pm$ 0.0052 & 0.983 $\pm$ 0.035 &1.238 $\pm$ 0.251\\
TopoLoss & 0.9626 $\pm$ 0.0038 & 0.9755 $\pm$ 0.0041 & 0.9444 $\pm$ 0.0076 & 0.782 $\pm$ 0.019  & \textbf{0.429 $\pm$ 0.104} \\
DMT & 0.9593 $\pm$ 0.0035 & \textbf{0.9796 $\pm$ 0.0033} & \textbf{0.9527 $\pm$ 0.0052} & \textbf{0.671 $\pm$ 0.027}  & \textbf{0.391 $\pm$ 0.114} \\
\hline
\multicolumn{6}{c}{DRIVE} \\
\hline
DIVE & 0.9549 $\pm$ 0.0023 & 0.7543 $\pm$ 0.0008 & 0.8407 $\pm$ 0.0257 & 1.936 $\pm$ 0.127 & 3.276 $\pm$ 0.642 \\

UNet & 0.9452 $\pm$ 0.0058 & 0.7491 $\pm$ 0.0027 & 0.8343 $\pm$ 0.0413 &  1.975 $\pm$ 0.046 & 3.643 $\pm$ 0.536\\
Mosin. & 0.9543 $\pm$ 0.0047 & 0.7218 $\pm$ 0.0013 &0.8870 $\pm$ 0.0386  & 1.167 $\pm$ 0.026 & 2.784 $\pm$ 0.293\\
TopoLoss & 0.9521 $\pm$ 0.0042 & 0.7621 $\pm$ 0.0036 & 0.9024 $\pm$ 0.0113 & 1.083 $\pm$ 0.006 & \textbf{1.076 $\pm$ 0.265}\\
DMT & 0.9495 $\pm$ 0.0036 & \textbf{0.7733 $\pm$ 0.0039} &\textbf{0.9077 $\pm$ 0.0021} & \textbf{0.876 $\pm$ 0.038} & \textbf{0.873 $\pm$ 0.402}\\
\hline
\multicolumn{6}{c}{CrackTree} \\
\hline
DIVE & 0.9854 $\pm$ 0.0052  & 0.6530 $\pm$ 0.0017 & 0.8634 $\pm$ 0.0376 & 1.570 $\pm$ 0.078  & 1.576 $\pm$ 0.287 \\

UNet & 0.9821 $\pm$ 0.0097 & 0.6491 $\pm$ 0.0029 & 0.8749 $\pm$ 0.0421 & 1.625 $\pm$ 0.104  & 1.785 $\pm$ 0.303\\
Mosin. & 0.9833 $\pm$ 0.0067 &0.6527 $\pm$ 0.0010 &0.8897 $\pm$ 0.0201 & 1.113 $\pm$ 0.057 & 1.045 $\pm$ 0.214\\
TopoLoss & 0.9826 $\pm$ 0.0084 & 0.6732 $\pm$ 0.0041 & \textbf{0.9291 $\pm$ 0.0123} & 0.997 $\pm$ 0.011 & \textbf{0.672 $\pm$ 0.176} \\
DMT & 0.9842 $\pm$ 0.0041 & \textbf{0.6811 $\pm$ 0.0047} &\textbf{0.9307 $\pm$ 0.0172} & \textbf{0.901 $\pm$ 0.081} & \textbf{0.518 $\pm$ 0.189} \\

\hline
\multicolumn{6}{c}{Road} \\
\hline
DIVE & 0.9734 $\pm$ 0.0077 & 0.6743 $\pm$ 0.0051 & 0.8201 $\pm$ 0.0128 & 2.368 $\pm$ 0.203  & 3.598 $\pm$ 0.783\\

UNet & 0.9786 $\pm$ 0.0052 & 0.6612 $\pm$ 0.0016 & 0.8189 $\pm$ 0.0097 & 2.249 $\pm$ 0.175  & 3.439 $\pm$ 0.621\\
Mosin. & 0.9754 $\pm$ 0.0043 & 0.6673 $\pm$ 0.0044 & 0.8456 $\pm$ 0.0174 & 1.457 $\pm$ 0.096 & 2.781 $\pm$ 0.237\\
TopoLoss & 0.9728 $\pm$ 0.0063  & 0.6903 $\pm$ 0.0038 & 0.8671 $\pm$ 0.0068 & 1.234 $\pm$ 0.037  & \textbf{1.275 $\pm$ 0.192} \\
DMT & 0.9744 $\pm$ 0.0049  & \textbf{0.7056 $\pm$ 0.0022} & \textbf{0.8819 $\pm$ 0.0104} & \textbf{1.092 $\pm$ 0.129}  & \textbf{0.995 $\pm$ 0.301}\\
\hline
\end{tabular}
\end{center}
\end{table*}

\myparagraph{Fairness Comparisons with Same Backbone Networks.}
\label{sec:fairness}
We copy the numbers of TopoLoss from~\cite{hu2019topology}, which is TopoLoss+DIVE. And in this chapter, we use UNet as the backbone. Indeed, with UNet, TopoLoss will be worse and the gap will be even bigger. The DIVE used in~\cite{hu2019topology} is more expensive and better designed specifically for EM images. We choose UNet in this manuscript as it is lightweight and easy to generalize to many datasets.
We also apply our backbone-agnostic DMT-loss to the DIVE network~\cite{fakhry2016deep}. All the experiments are conducted on CREMI 2D dataset. The quantitative results (Betti Error) are shown in Tab.~\ref{table:fairness}.

\setlength{\tabcolsep}{5pt}
\begin{table*}[ht]
\begin{center}
\small
\caption{Comparison with same backbones.}
\label{table:fairness}
\begin{tabular}{ccc}
\hline
 & UNet & DIVE\\
\hline
TopoLoss & 1.451$\pm$ 0.216 & 1.113 $\pm$ 0.224\\
\hline
DMT-Loss & \textbf{0.982 $\pm$ 0.179} & \textbf{0.956 $\pm$ 0.142}\\
\hline
\end{tabular}
\end{center}
\end{table*}

\section{Experiments on 3D Datasets} 

\subsection{Datasets}
We use three different biomedical 3D datasets: \textbf{ISBI13}, \textbf{CREMI} and \textbf{3Dircadb}~\cite{soler20103d}.
\textbf{ISBI13} and \textbf{CREMI}, which have been discussed above, are originally 3D datasets, can be used in both 2D and 3D evaluations. We also evaluate our method on the open dataset \textbf{3Dircadb}, which contains 20 enhanced CT volumes with artery annotations.

\subsection{Evaluation Metrics} We use similar evaluation metrics as the 2D part. Note that, in terms of 2D images, for ARI and VOI, we compute the numbers for each slice and then average the numbers for different slices as the final performance; for 3D images, we compute the performance for the whole volume. For 2D images, we compute the 1D Betti number (number of holes) to obtain the Betti Error; while for 3D images, we compute the 2D Betti number (number of voids) to obtain the Betti Error.

\subsection{Baselines} \textbf{{3D DIVE}}~\cite{zeng2017deepem3d}, \textbf{{3D UNet}}~\cite{cciccek20163d}, \textbf{{3D TopoLoss}}~\cite{hu2019topology} are the 3D versions for \textbf{DIVE}, \textbf{UNet} and \textbf{TopoLoss}. \textbf{MALA}~\cite{funke2018large} trains the UNet using a new structured loss function.

\setlength{\tabcolsep}{1.5pt}
\begin{table*}[ht]
\begin{center}
\scriptsize
\caption{Quantitative results for different models on several 3D datasets.}
\label{table:medical_dmt_3d}
\begin{tabular}{ccccccc}
Method & Accuracy & DICE  & ARI & VOI  & Betti Error\\
\hline
\multicolumn{6}{c}{ISBI13} \\
\hline
3D DIVE &  0.9723 $\pm$ 0.0021 & 0.9681 $\pm$ 0.0043 & 0.8719 $\pm$ 0.0189 & 1.208 $\pm$ 0.149 & 2.375 $\pm$ 0.419\\

3D UNet & 0.9746 $\pm$ 0.0025 & 0.9701 $\pm$ 0.0012 & \textbf{0.8956 $\pm$ 0.0391} & 1.123 $\pm$ 0.091 & 1.954 $\pm$ 0.585\\
MALA & 0.9701 $\pm$ 0.0018 & 0.9699 $\pm$ 0.0013 & \textbf{0.8945 $\pm$ 0.0481} & 0.901 $\pm$ 0.106 & 1.103 $\pm$ 0.207\\
3D TopoLoss & 0.9689 $\pm$ 0.0031 & 0.9752 $\pm$ 0.0045 & \textbf{0.9043 $\pm$ 0.0283} & 0.792 $\pm$ 0.086 & 0.972 $\pm$ 0.245\\
DMT & 0.9701 $\pm$ 0.0026 & \textbf{0.9803 $\pm$ 0.0019} & \textbf{0.9149 $\pm$ 0.0217} & \textbf{0.634 $\pm$ 0.086} & \textbf{0.812 $\pm$ 0.134}\\

\hline
\multicolumn{6}{c}{CREMI} \\
\hline
3D DIVE & 0.9503 $\pm$ 0.0061 & 0.9641 $\pm$ 0.0011 & 0.8514 $\pm$ 0.0387 & 1.219 $\pm$ 0.103 & 2.674 $\pm$ 0.473\\
3D UNet & 0.9547 $\pm$ 0.0038 & 0.9618 $\pm$ 0.0026 & 0.8322 $\pm$ 0.0315 & 1.416 $\pm$ 0.097 & 2.313 $\pm$ 0.501\\
MALA & 0.9472 $\pm$ 0.0027 & 0.9583 $\pm$ 0.0023 &0.8713 $\pm$ 0.0286 & 1.109 $\pm$ 0.093 & 1.114 $\pm$ 0.309\\
3D TopoLoss & 0.9523 $\pm$ 0.0043 & 0.9672 $\pm$ 0.0010 & 0.8726 $\pm$ 0.0194 & 1.044 $\pm$ 0.128 & 1.076 $\pm$ 0.206\\
DMT & 0.9529 $\pm$ 0.0031 & \textbf{0.9731 $\pm$ 0.0045} & \textbf{0.9013 $\pm$ 0.0202} & \textbf{0.891 $\pm$ 0.099} & \textbf{0.726 $\pm$ 0.187}\\
  
  \hline
\multicolumn{6}{c}{3Dircadb} \\
\hline
3D DIVE & 0.9618 $\pm$ 0.0054 & 0.6097 $\pm$ 0.0034 & / & / & 4.571 $\pm$ 0.505\\
3D UNet & 0.9632 $\pm$ 0.0009 & 0.5898 $\pm$ 0.0025 & / & / & 4.131 $\pm$ 0.483\\
MALA & 0.9546 $\pm$ 0.0033 & 0.5719 $\pm$ 0.0043 & / & / & 2.982 $\pm$ 0.105\\
3D TopoLoss & 0.9561 $\pm$ 0.0019 & 0.6138 $\pm$ 0.0029 & / & / & 2.245 $\pm$ 0.255 \\
DMT & 0.9587 $\pm$ 0.0023 & \textbf{0.6257 $\pm$ 0.0021} & / & / & \textbf{1.415 $\pm$ 0.305}\\
\hline
\end{tabular}
\end{center}
\end{table*}

\subsection{Results}
\begin{figure*}[ht]
\centering
\subfigure{
\includegraphics[width=0.23\textwidth]{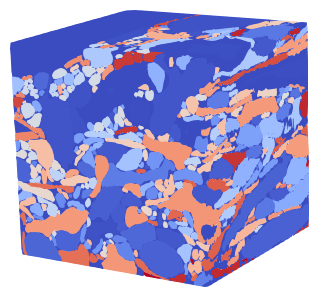}}
\subfigure{
\includegraphics[width=0.23\textwidth]{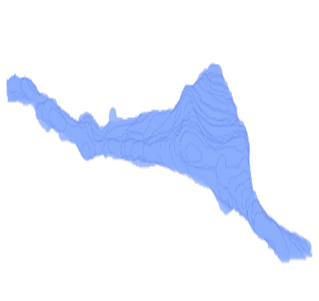}}
\subfigure{
\includegraphics[width=0.23\textwidth]{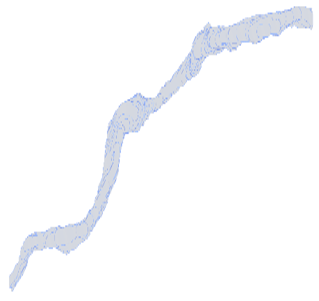}}
\subfigure{
\includegraphics[width=0.23\textwidth]{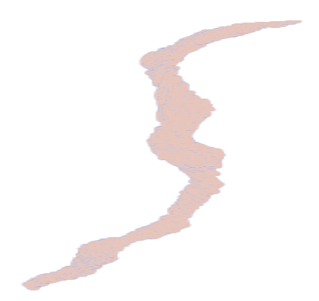}}
\caption{Segmentation results for ISBI13 dataset and 3 randomly selected neurons.}
\label{fig:3d_result}
\end{figure*}

Tab.~\ref{table:medical_dmt_3d} shows the quantitative results for three different 3D image datasets, ISBI13, CREMI, and 3Dircadb. Our method outperforms existing methods in topological accuracy (in all three topology-aware metrics), which demonstrates the effectiveness of the proposed method.
Fig.~\ref{fig:3d_result} shows qualitative results for the ISBI13 dataset.

\myparagraph{The Benefit of the Proposed DMT-loss.} Instead of capturing isolated critical points in TopoLoss~\cite{hu2019topology}, the proposed DMT-loss captures the whole V-path as critical structures. Taking the patch in Fig.~\ref{fig:sample} as an example, TopoLoss identifies $\approx$ 80 isolated critical pixels for further training, whereas the critical structures captured by the DMT-loss contain $\approx$ 1000 critical pixels (Fig.~\ref{fig:proper_pruning}). We compare the efficiency of DMT-loss and TopoLoss using the same backbone network, evaluated on the CREMI 2D dataset. Both methods start from a reasonable pre-trained likelihood map. TopoLoss achieves 1.113 (Betti Error), taking $\approx$3h to converge; while DMT-loss achieves 0.956 (Betti Error), taking $\approx$1.2h to converge (the standard UNet takes $\approx$0.5h instead). Aside from converging faster, the DMT-loss is also less likely to converge to low-quality local minima. We hypothesize that the loss landscape of the topological loss will have more local minima than that of the DMT-loss, even though the global minima of both landscapes may have the same quality.

\myparagraph{Ablation Study for Persistence Threshold $\epsilon$.} As illustrated in Fig.~\ref{fig:pruning}, different persistence thresholds $\epsilon$ will identify different critical structures. The ablation experiment is also conducted on the CREMI 2D dataset. When $\epsilon = 0.2$ (See Fig.~\ref{fig:proper_pruning}), the proposed DMT-loss achieves the best performance of 0.982 (Betti Error). When $\epsilon = 0.1$ and $\epsilon = 0.3$, the performance drops to 1.206 and 2.105 (both in Betti Error), respectively. This makes sense for the following reasons: 1) for $\epsilon = 0.1$, the DMT-loss captures lots of unnecessary structures which mislead the neural networks; 2) for $\epsilon = 0.3$, the DMT-loss misses lots of important critical structures, making the performance drop significantly. The $\epsilon$ is chosen via cross-validation. 

\myparagraph{The Ablation Study for Balanced Term $\beta$.}
We conduct another ablation study for the balanced weight of parameter $\beta$. Note that, the parameter $\beta$ is dataset dependent. We conduct the ablation experiment on CREMI 2D dataset. When $\beta = 3$, the proposed DMT-loss achieves the best performance of 0.982 (Betti Error). When $\beta = 2$ and $\beta = 4$, the performance drops to 1.074 and 1.181 (both in Betti Error), respectively. The parameter $\beta$ is also chosen via cross-validation.

\myparagraph{Comparison with Other Simpler Choices.}  The proposed method essentially highlights geometrically rich locations. To demonstrate the effectiveness of the proposed method, we also compare with two baselines: canny edge detection, and ridge detection, which achieve 2.971 and 2.507 in terms of Betti Error respectively, much worse than our results (Betti Error: 0.982). Although our focus is the Betti error, we also report per-pixel accuracy for reference (See Tab.~\ref{table:simpler} for details). From the results, we observe that the baseline models could not solve topological errors, even though they achieve high per-pixel accuracy. Without a persistent-homology based pruning, these baselines generate too many noisy structures and thus are not as effective as DMT-loss. 

\setlength{\tabcolsep}{5pt}
\begin{table*}[ht]
\begin{center}
\small
\caption{Comparison with other simpler choices.}
\label{table:simpler}
\begin{tabular}{ccc}

Method & Accuracy & Betti Error\\

\hline
DMT & 0.9475 & 0.982\\

Canny edge detection & 0.9386 & 2.971\\
Ridge detection & 0.9443 & 2.507 \\
\hline

\end{tabular}
\end{center}
\end{table*}

\myparagraph{Robustness of the Proposed Method.} We run another ablation study on images corrupted with Gaussian noise. The experiment is also conducted on the CREMI 2D dataset. 
From the results (See Tab.~\ref{table:robust} for details), the DMT-loss is fairly robust and maintains good performance even with high noise levels. The reason is that the Morse structures are computed on the likelihood map, which is already robust to noise. 

In Tab.~\ref{table:robust}, 10$\%$ means the percentage of corrupted pixels, and $\delta/2\delta$ means the sdv of the added Gaussian noise. For reference, we note that the performance of the standard UNet is 3.016 (Betti Error).

\setlength{\tabcolsep}{5pt}
\begin{table*}[ht]
\begin{center}
\small
\caption{Results with different noise levels.}
\label{table:robust}
\begin{tabular}{ccc}

Method & Accuracy & Betti Error\\

\hline
DMT & 0.9475 & 0.982\\

Gaussian ($10\%, \delta$) & 0.9393 & 1.086\\
Gaussian ($20\%, \delta$) & 0.9272 & 1.391 \\
\hline

\end{tabular}
\end{center}
\end{table*}

\begin{figure*}[ht]
\centering 
\subfigure[]{
\includegraphics[width=0.22\textwidth]{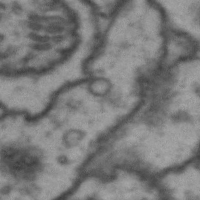}}
\subfigure[]{
\includegraphics[width=0.22\textwidth]{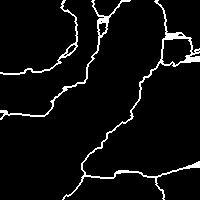}}
\subfigure[]{
\includegraphics[width=0.22\textwidth]{dmt/illu_gt_padding.png}}
\subfigure[]{
\includegraphics[width=0.22\textwidth]{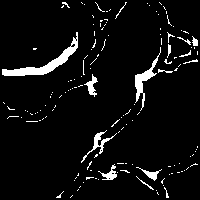}}
\caption{Illustration of comparison between the proposed DMT-loss and a simple reweighted cross entropy loss. Please refer to Fig.~\ref{fig:supple_likelihood} for the likelihood map of a baseline method without topological guarantee. \textbf{(a)} an input neuron image. \textbf{(b)} The topologically critical structures from the likelihood, captured by the proposed discrete Morse algorithm. These structures will be used in the DMT-Loss. \textbf{(c)} ground truth. \textbf{(d)} The FP/FN pixels identified by simple re-weighting cross entropy loss.}
\label{fig:reweight}
\end{figure*}

\myparagraph{Comparison with Reweighted Cross Entropy Loss.} We run an additional ablation study to compare with a baseline of reweighting the FP and FN pixels in the cross-entropy loss (Reweighting CE). The weights of the FP/FN pixels are hyperparameters tuned via cross-validation. The reweighting CE strategy achieves 2.753 in terms of Betti Error (on CREMI 2D data), and the DMT-loss is better than this baseline. The reason is that the DMT-loss specifically penalizes FP and FN pixels which are topology-critical. Meanwhile, reweighting CE adds weights to FP/FN pixels without discrimination. A majority of these misclassified pixels are not topology-critical. They are near the boundary of the foreground. Please see Fig.~\ref{fig:reweight} for an illustration.

%% file: uncertainty.tex
\chapter{Learning Probabilistic Topological Representations Using Discrete Morse Theory}
\label{chapter:uncertainty}
In the previous chapters, we have developed algorithms that can improve topology-aware segmentation accuracy, while all these methods are essentially learning pixel-wise representations. In this chapter,
we propose to learn topological/structural representations directly.

\section{Introduction}
\label{sec:intro_uncertainty}
Accurate segmentation of fine-scale structures, e.g., vessels, neurons, and membranes is crucial for downstream analysis. 
In recent years, topology-inspired losses have been proposed to improve structural accuracy~\cite{hu2019topology,hu2021topology,shit2021cldice,mosinska2018beyond,clough2020topological}. These losses identify topologically critical locations at which a segmentation network is error-prone, and force the network to improve its prediction at these critical locations. 

\begin{figure*}[ht]
\centering 
\subfigure[Image]{
\includegraphics[width=0.14\textwidth]{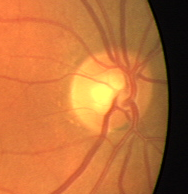}}
\subfigure[GT]{
\includegraphics[width=0.14\textwidth]{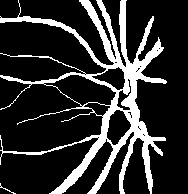}}
\subfigure[Pro.UNet]{
\includegraphics[width=0.14\textwidth]{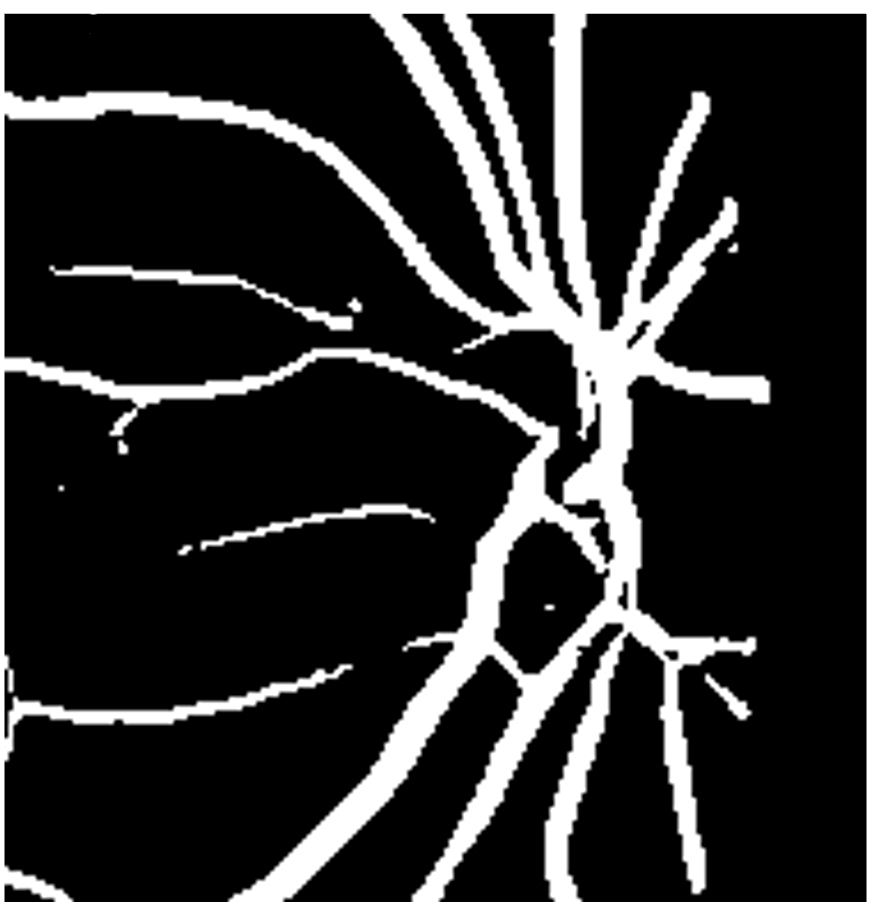}}
\subfigure[Pixel Uncer.]{
\includegraphics[width=0.166\textwidth]{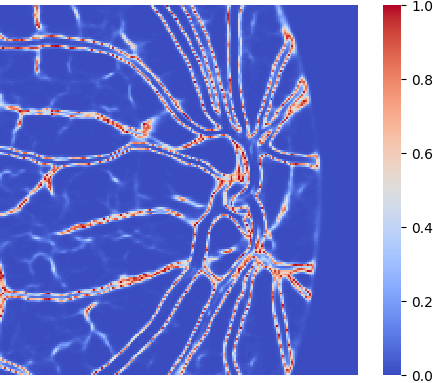}}
\subfigure[Ours]{
\includegraphics[width=0.14\textwidth]{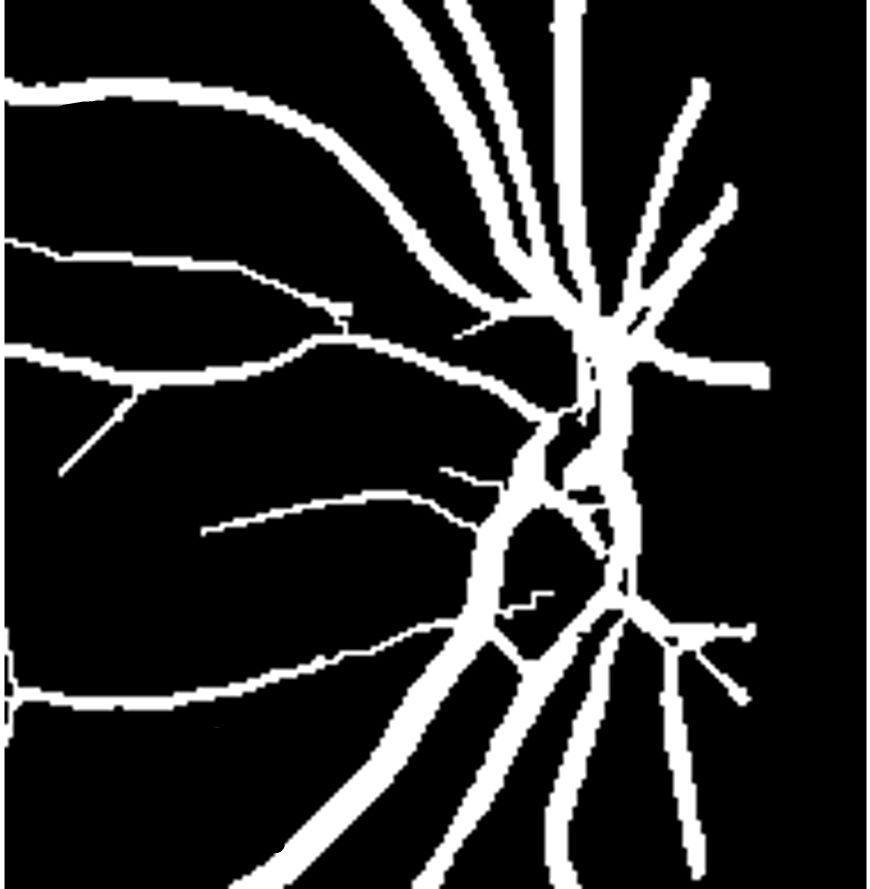}}
\subfigure[Stru. Uncer.]{
\includegraphics[width=0.166\textwidth]{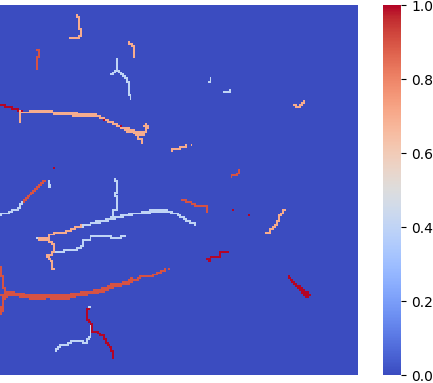}}
\caption{Illustration of structural segmentation and structure-level uncertainty. Compared with Probabilistic-UNet~\cite{kohl2018probabilistic} (Fig.~\ref{fig:more_uncertain}(c)-(d)), the proposed method is able to generate a structure-preserving segmentation map (Fig.~\ref{fig:more_uncertain}(e)), and structure-level uncertainty (Fig.~\ref{fig:more_uncertain}(f)).
}
\label{fig:more_uncertain}
\end{figure*}

However, these loss-based methods are still not ideal. They are based on a standard segmentation network, and thus \emph{only learn pixel-wise feature representations}. 
This causes several issues. First, a standard segmentation network makes pixel-wise predictions. Thus, at the inference stage, topological errors, e.g.~broken connections, can still happen, even though they may be mitigated by the topology-inspired losses.
Another issue is in uncertainty estimation, i.e., estimating how certain a segmentation network is at different locations. Uncertainty maps can direct the focus of human annotators for efficient proofreading~\cite{budd2021survey, wang2022troubleshooting}. However, for fine-scale structures, the existing pixel-wise uncertainty map is not effective. As shown in Fig.~\ref{fig:more_uncertain}(d), every pixel adjacent to a vessel branch is highly uncertain, in spite of whether the branch is salient or not. What is more desirable is a structural uncertainty map that can highlight uncertain branches (e.g., Fig.~\ref{fig:more_uncertain}(f)).

\begin{figure*}[ht]
\centering
\includegraphics[width=1\textwidth]{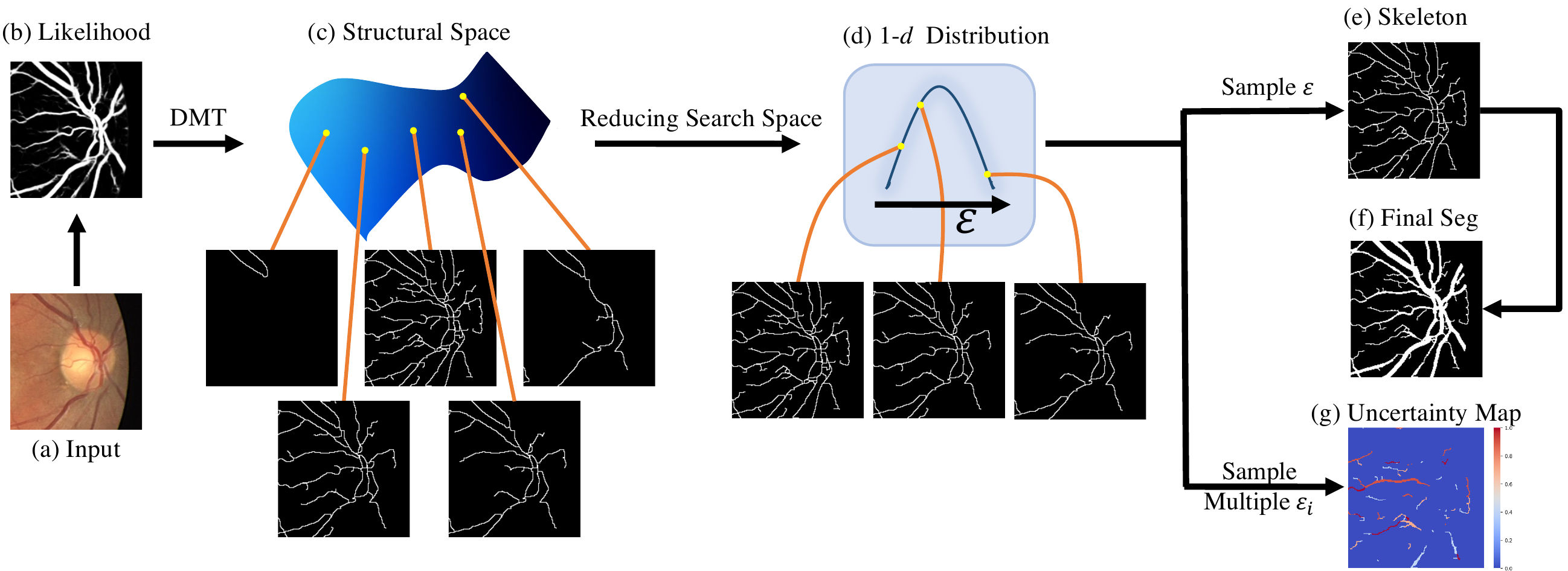}
\caption{The probabilistic topological/structural representation. \textbf{(a)} is a sample input, \textbf{(b)} is the predicted likelihood map from the deep neural network, \textbf{(c)} is the whole structural space obtained by running a discrete Morse theory algorithm on the likelihood map, \textbf{(d)} the 1-$d$ structural family parametrized by the persistence threshold $\epsilon$, as well as a Gaussian distribution over $\epsilon$, \textbf{(e)} a sampled skeleton, \textbf{(f)} the final structural segmentation map generated using the skeleton sample, and \textbf{(g)} the uncertainty map generated by multiple segmentations. 
}
\label{fig:structure_space}
\end{figure*}

To fundamentally address these issues, we propose to directly model and reason about the structures. In this chapter, we propose \emph{the first deep learning based method that directly learns the topological/structural~\footnote{We will be using the terms topology/topological and structure/structural interchangeably in this chapter.}
representation of images}. 
To move from pixel space to structural space~\footnote{\highlight{Structures/branches instead of pixels construct the space and the operations are conducted at the structure/branches level instead of pixel level.}}, we apply the classic discrete Morse theory~\cite{milnor1963morse,forman2002user} to decompose an image into a Morse complex, consisting of structural elements like branches, patches, etc. These structural elements are hypothetical structures one can infer from the input image.
Their combinations constitute a space of structures arising from the input image. See Fig.~\ref{fig:structure_space}(c) for an illustration.

For further reasoning with structures, we propose to learn a probabilistic model over the structural space. The challenge is that the space consists of exponentially many branches and is thus of very high dimension. To reduce the learning burden, we introduce the theory of persistent homology~\cite{2011MNRAS,DRS15,WWL15} for structure pruning. Each branch has its own persistence measuring its relative saliency. By continuously thresholding the complete Morse complex in terms of persistence, we obtain a sequence of Morse complexes parameterized by the persistence threshold, $\epsilon$. See Fig.~\ref{fig:structure_space}(d). By learning a Gaussian over $\epsilon$, we learn a parametric probabilistic model over these structures.

This parametric probabilistic model over structural space allows us to make direct structural predictions via sampling (Fig.~\ref{fig:structure_space}(e)), and to estimate the empirical structure-level uncertainty via sampling (Fig.~\ref{fig:structure_space}(g)). The benefit is two-fold: First, direct prediction of structures will ensure the model outputs always have structural integrity, even at the inference stage. This is illustrated in Fig.~\ref{fig:more_uncertain}(e). Samples from the probabilistic model are all feasible structural hypotheses based on the input image, with certain variations at uncertain locations. This is in contrast to state-of-the-art methods using pixel-wise representations (Fig.~\ref{fig:more_uncertain}(c)-(d)). Note the original output structure (Fig.~\ref{fig:structure_space}(e), also called skeleton) is only 1-pixel wide and may not serve as a good segmentation output. In the inference stage, we use a post-processing step to grow the structures without changing topology as the final segmentation prediction (Fig.~\ref{fig:structure_space}(f)). More details are provided in Sec.~\ref{sec:inference} and Fig.~\ref{fig:inference}. 

Second, the probabilistic structural model can be seamlessly incorporated into semi-automatic interactive annotation/proofreading workflows to facilitate large scale annotation of these complex structures. This is especially important in the biomedical domain where fine-scale structures are notoriously difficult to annotate, due to the complex 2D/3D morphology and low contrast near extremely thin structures. Our probabilistic model makes it possible to identify uncertain structures for efficient interactive annotation/proofreading. Note that the structural space is crucial for uncertainty reasoning. As shown in Fig.~\ref{fig:more_uncertain}(f) and Fig.~\ref{fig:structure_space}(g), our structural uncertainty map highlights uncertain branches for efficient proofreading. On the contrary, the traditional pixel-wise uncertainty map (Fig.~\ref{fig:more_uncertain}(d)) is not helpful at all; it highlights all pixels on the boundary of a branch. 

The main contributions of this chapter are:
\begin{enumerate}[topsep=0pt, partopsep=0pt,itemsep=0pt,parsep=0pt]
    \item We propose the first deep learning method which learns structural representations, based on discrete Morse theory and persistent homology.
    \item We learn a probabilistic model over the structural space, which facilitates different tasks such as topology-aware segmentation, uncertainty estimation and interactive proofreading. 
\item We validate our method on various datasets with \textit{rich and complex structures}. It outperforms state-of-the-art methods in both deterministic and probabilistic categories. 
\end{enumerate}
\section{Method}
Our key innovation is to restructure the output of a neural network so that it is indeed making predictions over a space of structures. This is achieved through insights into the topological structures of an image and the usage of several important tools in topological data analysis.

To move from pixel space to structural space, we apply discrete Morse theory to decompose an image into a Morse complex, consisting of structures like branches, patches, etc. For simplification, we will use the term "branch" to denote a single piece of Morse structure. These Morse branches are the hypothetical structures one can infer from the input image. This decomposition is based on a likelihood function produced by a pixel-wise segmentation network trained in parallel. Thus it is of good quality, i.e., the structures are close enough to the true structures.

Any binary labeling of these Morse branches is a legitimate segmentation; we call it a \emph{structural segmentation}. But for full-scope reasoning of the structural space, instead of classifying these branches one-by-one, we would like to have the full inference, i.e., predicting a probability distribution for each branch.  
To further reduce the degrees of freedom to make the inference easier, we apply persistent homology to filter these branches with regard to their saliency. This gives us a linear size family of structural segmentations, parameterized by a threshold $\epsilon$. Finally, we learn a 1D Gaussian distribution for the $\epsilon$ as our probabilistic model. This gives us the opportunity not only to sample segmentations, but also to provide a probability for each branch, which can be useful in downstream tasks including proofreading. 
In Sec.~\ref{sec:structure_space}, we introduce the discrete Morse theory and how to construct the space of Morse structures. We also explain how to use persistent homology to reduce the search space of reasoning into a 1-parameter family. 
In Sec.~\ref{sec:method_uncertainty}, we will provide details on how our deep neural network is constructed to learn the probabilistic model over the structural space, as illustrated in Fig.~\ref{fig:pipeline}.

\subsection{Constructing the Structural Space}
\label{sec:structure_space}
In this section, we explain how to construct a structural representation space using discrete Morse theory. The resulting structural representation space will be used to build a probabilistic model.
We will then discuss how to reduce the structural space into a 1-parameter family of structural segmentations, using persistent homology. 
We assume a 2D input image, although the method naturally extends to 3D images.

\begin{figure}[ht]
\centering 
\includegraphics[width=0.9\textwidth]{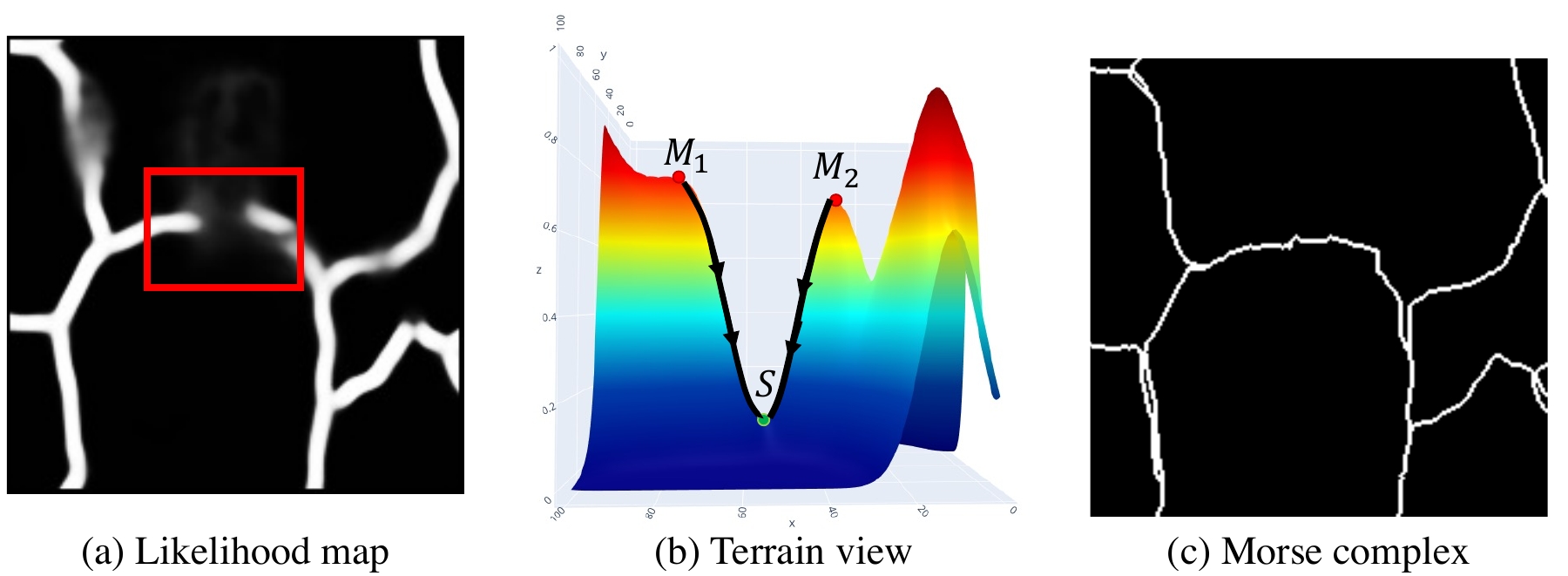}
\caption{\textbf{(a)} shows a sample likelihood map from the deep neural network, and \textbf{(b)} is the terrain view of the red patch in \textbf{(a)} and illustrates the stable manifold of a saddle point in 2D case for a line-like structure. \textbf{(c)} is the 2D Morse complex generated by DMT from \textbf{(a)}.}
\label{fig:discrete_uncertainty}
\end{figure}

Given a reasonably clean input (e.g., the likelihood map of a deep neural network, Fig.~\ref{fig:structure_space}(b)), we treat the 2D likelihood as a terrain function, and  Morse theory~\cite{milnor1963morse} can help to capture the structures regardless of weak/blurred conditions. See Fig.~\ref{fig:discrete_uncertainty} for an illustration. The weak part of a line in the continuous map can be viewed as the local dip in the mountain ridge of the terrain. In the language of Morse theory, the lowest point of this dip is a saddle point ($S$ in Fig.~\ref{fig:discrete_uncertainty}(b)), and the mountain ridges which are connected to the saddle point ($M_1S$ and $M_2S$) are called the stable manifolds of the saddle point.

We mainly focus on 2D images in this chapter, although extending to 3D images is natural. We consider two dimensional continuous functions $f: \R^{2} \rightarrow \R$. For a point $x\in \R^2$, the gradient can be computed as $\nabla f(x) = [\frac{\partial f}{ \partial x_{1}},\frac{\partial f}{\partial x_{2}}]^T$. We call a point $x = (x_{1},x_{2})$ \emph{critical} if $\nabla f(x) = 0$.
For a Morse function defined on $\R^2$, a critical point could be a minimum, a saddle or a maximum.

Consider a continuous line (the red rectangle region in Fig.~\ref{fig:discrete_uncertainty}(a)) in a 2D likelihood map. Imagine if we put a ball on one point of the line, then $-\nabla f(x)$ indicates the direction in which the ball will flow down. By definition, the ball will eventually flow to the critical points where $\nabla f(x) = 0$. The collection of points whose ball eventually flows to $p$ ($\nabla f(p) = 0$) is defined as the stable manifold (denoted as $S(p)$) of point $p$. Intuitively, for a 2D function $f$, the stable manifold $S(p)$ of a minimum $p$ is the entire valley of $p$ (similar to the watershed algorithm); similarly, the stable manifold $S(q)$ of a saddle point $q$ consists of the whole ridge line which connects two local maxima and goes through the saddle point. See Fig.~\ref{fig:discrete_uncertainty}(b) as an illustration. 

For a link-like structure, the stable manifold of a saddle contains the topological structures (usually curvilinear) of the continuous likelihood map predicted by deep neural networks, and they are exactly what we want to recover from noisy images.  In practice, we adopt the discrete version of Morse theory for images.

\myparagraph{Discrete Morse Theory.}
Take a 2D image as a 2-dimensional cubical complex. A 2-dimensional cubical complex then contains $0$-, $1$-, and $2$-dimensional cells, which correspond to vertices (pixels), edges, and squares, respectively. In the setting of discrete Morse theory (DMT)~\cite{forman,forman2002user}, a pair of adjacent cells, termed discrete gradient vectors, compose the gradient vector.
Critical points ($\nabla f(x) = 0$) are those critical cells that are not in any discrete gradient vectors. In the 2D domain, a minimum, a saddle, and a maximum correspond to a critical vertex, a critical edge, and a critical square respectively. 
A 1-stable manifold (the stable manifold of a saddle point) in 2D corresponds to a \emph{V-path}, i.e., connecting two local maxima and a saddle. See Fig.~\ref{fig:discrete_uncertainty}(b). And the Morse complex generated by the DMT algorithm is illustrated in Fig.~\ref{fig:discrete_uncertainty}(c).

\myparagraph{Constructing the Full Structural Space.} In this way, by using discrete Morse theory, for a likelihood map from the deep neural network, we can extract all the stable manifolds of the saddles, whose compositions constitute the full structural space. \textit{Formally, we call any combinations of these stable manifolds a structure}. 
Fig.~\ref{fig:structure_space}(c) illustrates 5 different structures. This structural space, however, is of exponential size. Assume there are $N$ pieces of stable manifolds/branches for a given likelihood map. Any combinations of these stable manifolds/branches will be a potential structure. We will have $2^{N}$ possible structures in total. This can be computationally prohibitive to construct and model. We need a principled way to reduce the structural search space.

\myparagraph{Reducing the Structural Search Space with Persistent Homology.} 
We propose to use persistent homology~\cite{2011MNRAS,DRS15,WWL15} to reduce the structural search space. 
Persistent homology is an important tool for topological data analysis~\cite{edelsbrunner2010computational,edelsbrunner2000topological}. Intuitively, we grow a Morse complex by gradually including more and more discrete elements (called cells) from empty. A branch of the Morse complex is a special type of cell. Other types include vertices, patches, etc. Cells will be continuously added to the complex. New branches will be born and existing branches will die. The persistence algorithm~\cite{edelsbrunner2000topological} pairs up all these critical cells as birth and death pairs. The difference of their function values is essentially the life time of the specific topological structure/branch, which is called the \emph{persistence}.
The importance of a branch is associated with its persistence. Intuitively, the longer the persistence of a specific branch is, the more important the branch is. 

Recall that our original construction of the structural space considers all possible combinations of branches, and thus can have exponentially many combinations. Instead, we propose to only select branches with high persistence as important ones. By doing this, we will be able to prune the less important/noisy branches very efficiently and recover the branches with true signals. Specifically, the structure pruning is done via  \emph{Morse cancellation} operation. The persistence thresholding provides us the opportunity to obtain a \textit{structural space of linear size}. We start with the complete Morse complex, and continuously increase the threshold $\epsilon$. At each threshold, we obtain a structure by filtering with $\epsilon$ and only keeping the branches whose persistence is above $\epsilon$. This gives a sequence of structures parametrized by $\epsilon$. As shown in Fig.~\ref{fig:structure_space}(d), the family of structures represents different structural densities. 

The one-parameter space allows us to easily learn a probabilistic model and carry out various inference tasks such as segmentation, sampling, uncertainty estimation, and interactive proofreading.
Specifically, we will learn a Gaussian distribution over the persistence threshold $\epsilon$, $\epsilon\sim N(\mu, \sigma)$. Denote the persistence of a branch $b$ as $\epsilon_b$. 
Any branch $b$ belongs to the structure map $M$ (we also call the structure map $M$ a structural segmentation) as long as its persistence is smaller or equal to the maximal persistence threshold of $M$, i.e., $b\in M$ if and only if $\epsilon_b\leq \epsilon_M$, where $\epsilon_M$ is used to generate $M$ ($\epsilon_M \geq \max_{b\in M} \epsilon_b$). More details will be provided in Sec.~\ref{sec:method_uncertainty}.

\myparagraph{Morse Cancellation.}
As the predicted likelihood map is noisy, the extracted discrete gradient field $\myVF(K)$ could also be noisy. Fortunately, the discrete Morse theory provides an elegant way to cancel critical simplicial pairs and ignore the unimportant Morse branches. Particularly, if there is a unique V-path $\pi = \delta=\delta_0, \gamma_1, \delta_1, \ldots, \delta_s, \gamma_{s+1} = \gamma$ from $\delta$ to $\gamma$, then the pair of critical simplices $\langle \delta^{(p+1)}, \gamma^p \rangle$ is \emph{cancellable} . By removing all V-pairs along these path, and adding $(\delta_{i-1}, \gamma_i)$ to $\myVF(K)$ for any $i \in [1, s+1]$, the \emph{Morse cancellation operation} reverses all V-pairs along this path. In this way, neither $\delta$ nor $\gamma$ is critical after the cancellation operation and we can prune/remove the corresponding stable manifold/branch. More details can be found in~\cite{hu2021topology}.

\myparagraph{Approximation of Morse Structures for Volume Data.} In the 2D setting, the stable manifold of saddles is composed of curvilinear structures, and the captured Morse structures will essentially contain the \textit{non-boundary edges}, which fits well with vessel data. However, the output structures should always be \textit{boundary edges} for volume data, which can't be dealt with the original discrete Morse theory. Consequently, we approximate the Morse structures of 2D volume data with the boundaries of the stable manifolds of local minima. As mentioned above, the stable manifold of a local minimum $p$ in the 2D setting corresponds to the whole valley, and the boundaries of these valleys construct the approximation of the Morse structures for volume data. Similar to the original discrete Morse theory, we also introduce a persistence threshold parameter $\epsilon$ and use persistent homology to prune the less important branches. The details of the proposed persistent-homology filtered topology watershed algorithm are illustrated in Alg.~\ref{alg:ph_watershed}.

\begin{table}[ht]
\begin{algorithm}[H]
\caption{Persistent-Homology filtered Topology Watershed Algorithm}
\label{alg:ph_watershed}
\KwIn{a grid 2D image, and a threshold $\theta$}
\KwOut{Morse structures for volume data}
\textbf{Definition}: $G =(V, E)$ denote a graph; 
$f(v)$ is the intensity value of node $v$; lower\_star$(v)$ = $\{(u,v) \in E| f(u) < f(v) \}$; 
$cc(v)$ is the connected component id of node $v$.  
\begin{algorithmic}[1]
\STATE PD =$\emptyset$; Build the proximity graph (4-connectivity) for 2D grid image; 

\STATE $U = V$ sorted according to $f(v)$; $T$ a sub-graph, which includes all the nodes and edges whose value $<t$.
\FOR {$v$ in $U$}
\STATE $t= f(v)$, $T = T + \{v\}$

\FOR {$(u,v)$ in lower\_star($v$)}
\item Assert $u \in T$

\IF{$cc(u) = cc(v)$}
\item Edge\_tag$(u,v)$ = loop edge
\item Continue
\ELSE
\item Edge\_tag$(u,v)$ = tree edge
\item younger\_cc = $\argmaxA_{w=cc(u), cc(v)} f(w)$
\item older\_cc = $\argminA_{w=cc(u), cc(v)} f(w)$

\item \textit{pers = $t$-$f$(younger\_cc)}
\IF{pers >= $\theta$}
\item \textit{Edge\_tag$(u,v)$ = watershed edge}
\item Continue
\ENDIF

\FOR {$w$ in younger\_cc} 
\item $cc(w)$ = older\_cc
\ENDFOR
\item PD = PD + ($f$(younger\_cc), $t$)
\ENDIF
\ENDFOR
\ENDFOR

\STATE \textbf{return} Membrane\_vertex\_set = $\cup$ vertices of watershed\_edge\_set
\end{algorithmic}
\end{algorithm}
\end{table}

\subsection{Neural Network Architecture}
\label{sec:method_uncertainty}

\begin{figure*}[ht]
\centering
\includegraphics[width=.9\textwidth]{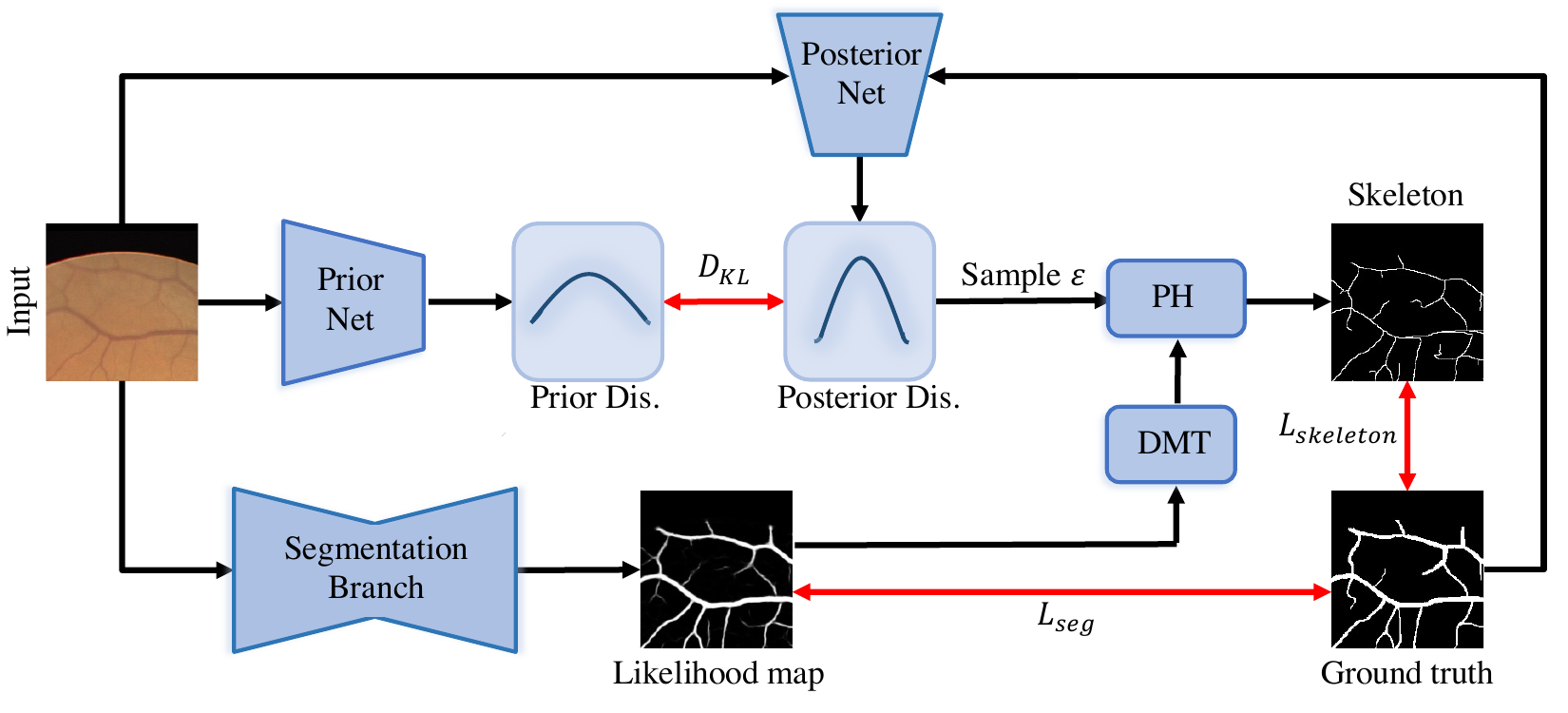}
\caption{The overall workflow of the training stage. The red arrows indicate supervision.}
\label{fig:pipeline}
\end{figure*}

In this section, we introduce our neural network that learns a probabilistic model over the structural representation to obtain structural segmentations. See Fig.~\ref{fig:pipeline} for an illustration of the overall pipeline.

Since the structural reasoning needs a sufficiently clean input to construct discrete Morse complexes, our method first obtains such a likelihood map by training a segmentation branch which is supervised by the standard segmentation loss, cross-entropy loss. Formally, $L_{seg} = L_{bce}(Y, S(X;\omega_{seg}))$, in which $S(X;\omega_{seg})$ is the output likelihood map, $\omega_{seg}$ is the segmentation branch's weight. 

The output likelihood map, $S(X;\omega_{seg})$, is used as the input for the discrete Morse theory algorithm (DMT), which generates a discrete Morse complex consisting of all possible Morse branches from the likelihood map. Thresholding these branches using persistent homology with different $\epsilon$'s will produce different structures. We refer to the DMT computation and the persistent homology thresholding operation as $f_{DMT}$ and $f_{PH}$. So given a likelihood map $S(X;\omega_{seg})$ and a threshold $\epsilon$, we can generate a structure (which we call a skeleton):
$
    S_{skeleton}(\epsilon) = f_{PH}(f_{DMT}(S(X; \omega_{seg})); \epsilon)
$.
Next, we discuss how to learn the probabilistic model. Recall we want to learn a Gaussian distribution over the persistent homology threshold,
$\epsilon\sim N(\mu, \sigma)$.
The parameters $\mu$ and $\sigma$ are learned by a neural network called the \emph{posterior network}. The network uses the input image $X$ and the corresponding ground truth mask $Y$ as input, and outputs the parameters $\mu(X,Y;\omega_{post})$ and $\sigma(X,Y;\omega_{post})$. $\omega_{post}$ is the parameter of the network. 

During training, at each iteration, we draw a sample $\epsilon$ from the distribution ($\epsilon \sim N(\mu, \sigma)$). Using the sample $\epsilon$, together with the likelihood map, we can generate the corresponding sampled structure, $S_{skeleton}(\epsilon)$. This skeleton will be compared with the ground truth for supervision. 
To compare a sampled skeleton, $S_{skeleton}(\epsilon)$, with ground truth $Y$, we use the skeleton to mask both $Y$ and the likelihood map $S(X;\omega_{seg})$, and then compare the skeleton-masked ground truth and the likelihood using cross-entropy loss:
$L_{bce}(Y \circ S_{skeleton}(\epsilon), S(X; \omega_{seg}) \circ S_{skeleton}(\epsilon))$. 

To learn the distribution, we use the expected loss:
\begin{equation}
    L_{skeleton} = \mathbb{E}_{\epsilon \sim N(\mu, \sigma)} L_{bce}(Y \circ S_{skeleton}(\epsilon), S(X; \omega_{seg}) \circ S_{skeleton}(\epsilon))
\end{equation}
\highlight{The skeleton is actually the inference structure from the probabilistic model and the supervision of the skeleton loss ensures the topological correctness of the inference results.} The loss can be backpropagated through the posterior network through the reparameterization technique~\cite{kingma2013auto}. Note that this loss will also provide supervision to the segmentation network through the likelihood map.

\myparagraph{Reparameterization Technique.}
\label{sec:reparameterization}
We adopt the reparameterization technique of VAE to make the network differentiable and be able to backpropagate.

The posterior net randomly draws samples from the posterior distribution $\epsilon \sim N(\mu_{post}, \sigma_{post})$. To implement the posterior net as a neural network, we will need to backpropagate through random sampling. The issue is that backpropagation cannot flow through random nodes; to overcome this obstacle, we adopt the reparameterization technique proposed in~\cite{kingma2013auto}.

Assuming the posterior is normally distributed, we can approximate it with another normal distribution. We approximate $\epsilon$ with normally distribution $Z$ ($Z \sim N(0, \textbf{I})$).

\begin{equation}
    \epsilon \sim N(\mu, \sigma), \quad \quad \epsilon = \mu + \sigma Z.
\end{equation}

Now instead of saying that $\epsilon$ is sampled from $Q(X,Y;\omega_{post})$ , we can say $\epsilon$ is a function that takes parameter ($Z$,($\mu$, $\sigma)$) and these $\mu$, $\sigma$ come from deep neural network. Therefore all we need is partial derivatives w.r.t. $\mu$, $\sigma$, and $Z$ is irrelevant for taking derivatives for backpropagation.

\myparagraph{Learning a Prior Network from the Posterior Network.} Although our posterior network can learn the distribution well, it relies on the ground truth mask $Y$ as input. This is not available at the inference stage. To address this issue, inspired by Probabilistic-UNet~\cite{kohl2018probabilistic}, we use another network to learn the distribution of $\epsilon$ with only the image $X$ as input. We call this network the \textit{prior net}. We denote by $P$ the distribution using parameters predicted by the prior network, and denote by $Q$ the distribution predicted by the posterior network. 

During the training, we want to use the prior net to mimic the posterior net; and then in the inference stage, we can use the prior net to obtain a reliable distribution over $\epsilon$ with only the image $X$. Thus, we incorporate the Kullback-Leibler divergence of these two distributions,
$
    D_{KL}(Q||P) = \mathbb{E}_{\epsilon \sim Q} (\log \frac{Q}{P})
$
, which measures the differences of prior distribution $P$ ($N(\mu_{prior}, \sigma_{prior})$) and the posterior distribution $Q$ ($N(\mu_{post}, \sigma_{post})$). 

\myparagraph{Training the Neural Network.} The final loss is composed by the standard segmentation loss, the skeleton loss $L_{skeleton}$, and the KL divergence loss, with two hyperparameters $\alpha$ and $\beta$ to balance the three terms,
\begin{equation}
    L(X, Y) = L_{seg} + \alpha  L_{skeleton} + \beta D_{KL}(Q||P)
\end{equation}
The network is trained to jointly optimize the segmentation branch and the probabilistic branch (containing both prior and posterior nets) simultaneously. 
During the training stage, the KL divergence loss ($D_{KL}$) pushes the prior distribution towards the posterior distribution. The training scheme is also illustrated in Fig.~\ref{fig:pipeline}. 

\begin{figure*}[ht]
\centering
\includegraphics[width=.9\textwidth]{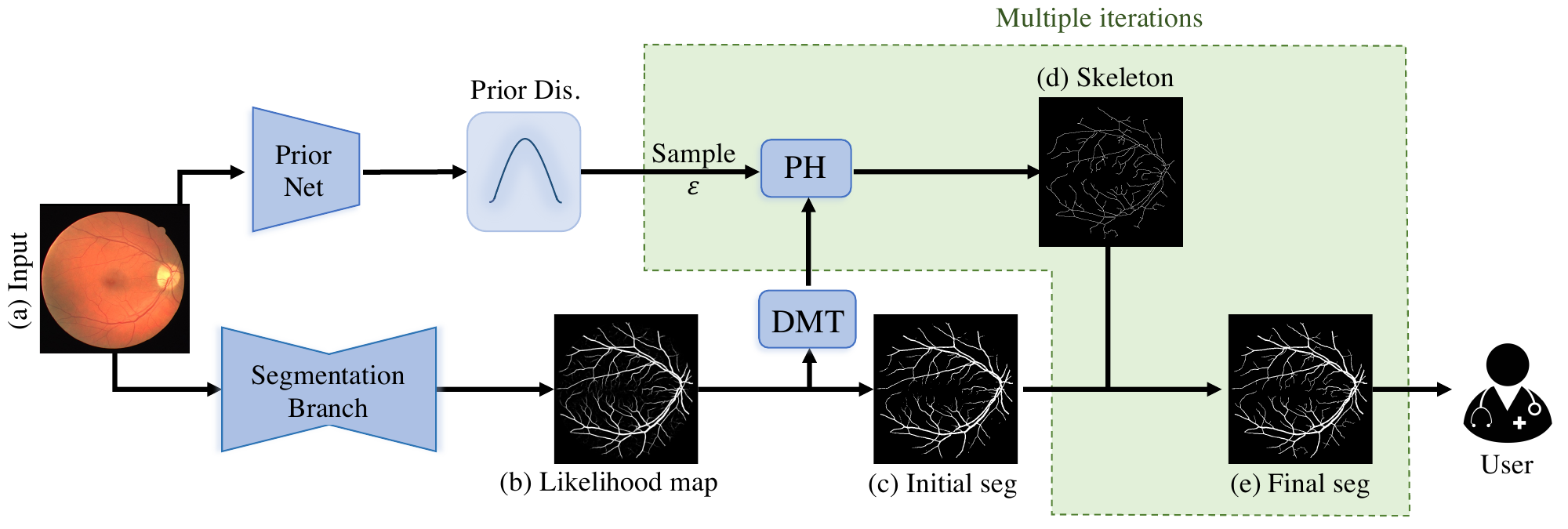}
\caption{The inference and interactive annotation/proofreading pipeline.}
\label{fig:inference}
\end{figure*}

\myparagraph{Inference Stage: Generating Structure-Preserving Segmentation Maps.} 
\label{sec:inference}
In the inference stage,  given an input image, we are able to produce an unlimited number of plausible structure-preserving skeletons via sampling. 
We use a postprocessing step to grow the 1-pixel wide structures/skeletons without changing their topology as the final structural segmentation. Specifically, the skeletons are overlaid on the binarized initial segmentation map (Fig.~\ref{fig:inference}(c)), and only the connected components which exist in the skeletons are kept as the final segmentation maps (Fig.~\ref{fig:inference}(e)). In this way, each plausible skeleton (Fig.~\ref{fig:inference}(d)) generates one final segmentation map (Fig.~\ref{fig:inference}(e)) and it has exactly the same topology as the corresponding skeleton. The pipeline of the procedure is illustrated in Fig.~\ref{fig:inference}.

\myparagraph{Uncertainty of Structures.} \label{sec:uncertain}
Given a learned prior distribution, $P$, over the family of structural segmentations, we can naturally calculate the probability of each Morse structure/branch. Recall a branch $b$ has its persistence $\epsilon_b$. 
And the prior probability of a structural segmentation map $M$ is $P(\epsilon_M)$, in which $\epsilon_M$ is used to generate $M$. Also, any branch $b$ whose persistence is smaller or equal to the maximal persistence threshold of $M$ belongs to $M$, i.e., $b\in M$ if and only if $\epsilon_b\leq \epsilon_M$. Thus we have $\epsilon_M \geq \max_{b\in M} \epsilon_b$. 
Therefore, the probability of a branch $b$ being in a segmentation map $M$ such as $\epsilon_M\sim P$ follows a Bernoulli distribution with the probability $Pr(b)$ being the cumulative distribution function (CDF) of $P$, $CDF_P(\epsilon_b) = P(\epsilon \leq \epsilon_b) $. 
This can be directly calculated at inference, and the absolute difference of the CDF from 0.5 is the confidence $($which equals 1-uncertainty$)$ of the Morse structure/branch $b$.

\section{Experiments}
\label{sec:experiment_uncertainty}

Our method directly makes predictions and inferences on structures rather than on pixels. This can significantly benefit downstream tasks. While probabilities of structures can certainly be used for further analysis of the structural system, in this chapter we focus on both automatic image segmentation and semi-automatic annotation/proofreading tasks. On automatic image segmentation, we show that direct prediction can ensure topological integrity even better than previous topology-aware losses. This is not surprising as our prediction is on structures. On the semi-automatic proofreading task, we show our structure-level uncertainty can assist human annotators to obtain satisfying segmentation annotations in a much more efficient manner than previous methods. 

\subsection{Automatic Topology-Aware Image Segmentation}
We introduce the datasets, evaluation metrics, and baselines used in this chapter to demonstrate the effectiveness of the proposed method.
\subsection{Datasets}
\label{sec:dataset_uncertainty}
We use three datasets to validate the efficacy of the proposed method: \textbf{ISBI13}~\cite{arganda20133d} (volume),  \textbf{CREMI} (volume), and \textbf{DRIVE}~\cite{staal2004ridge} (vessel).
Both volume and vessel datasets are used to validate the efficacy of the proposed method, and the details of the datasets have been introduced in Sec.~\ref{sec:dataset_topoloss}.
We use a 3-fold cross-validation for all the methods to report the numbers over the validation set.

\subsection{Evaluation Metrics}
\label{sec:metric_uncertainty}
We use four different evaluation metrics: \textbf{Dice score}, \textbf{ARI}, \textbf{VOI}, and \textbf{Betti number error}. More details about the evaluation metrics have been provided in Sec.~\ref{sec:metric_topoloss} and Sec.~\ref{sec:metric_warping}.

\subsection{Baselines}
\label{sec:baseline}

We compare the proposed method with two kinds of baselines: 1) Standard segmentation baselines: \textbf{{DIVE}}~\cite{fakhry2016deep}, \textbf{{UNet}}~\cite{ronneberger2015u}, \textbf{{UNet-VGG}}~\cite{mosinska2018beyond},  \textbf{TopoLoss}~\cite{hu2019topology} and \textbf{DMT}~\cite{hu2021topology}. 2) Probabilistic-based segmentation methods: \textbf{Dropout UNet}~\cite{kendall2015bayesian} and \textbf{Probabilistic-UNet}~\cite{kohl2018probabilistic}.
The details of Probabilistic-based segmentation methods are listed as follows: 
\begin{itemize}
\item Dropout UNet~\cite{kendall2015bayesian} dropouts the three inner-most encoder
and decoder blocks with a probability of 0.5 during both the training and inference. 
\item Probabilistic-UNet~\cite{kohl2018probabilistic} introduces a probabilistic segmentation method by combining UNet with a VAE. 
\end{itemize}
For all methods, we generate binary segmentations by thresholding predicted likelihood maps at 0.5. 

\myparagraph{Quantitative and Qualitative Results.} Tab.~\ref{table:quantitative} shows the quantitative results compared to several baselines. Note that for deterministic methods, the numbers are computed directly based on the outputs; while for probabilistic methods, we generate five segmentation masks and report the averaged numbers over the five segmentation masks for each image (for both the baselines and the proposed method). We use the t-test (95\% confidence interval) to determine the statistical significance and highlight the significantly better results. From the table, we can observe that the proposed method achieves significantly better performances in terms of topology-aware metrics (ARI, VOI, and Betti Error).

\setlength{\tabcolsep}{3pt}
\begin{table*}[ht]
\begin{center}
\footnotesize
\caption{Quantitative results for different models on three different biomedical datasets.}
\label{table:quantitative}
\begin{tabular}{ccccc}
\hline
Method  & Dice $\uparrow$ &  ARI $\uparrow$ & VOI $\downarrow$ & Betti Error $\downarrow$\\
\hline\hline
\multicolumn{5}{c}{ISBI13 (Volume)} \\
\hline
DIVE & 0.9658 $\pm$ 0.0020 & 0.6923 $\pm$ 0.0134  & 2.790 $\pm$ 0.025 & 3.875 $\pm$ 0.326\\
UNet & 0.9649 $\pm$ 0.0057 & 0.7031 $\pm$ 0.0256 &  2.583 $\pm$ 0.078 &3.463 $\pm$ 0.435\\
UNet-VGG & 0.9623 $\pm$ 0.0047 & 0.7483 $\pm$ 0.0367 & 1.534 $\pm$ 0.063 & 2.952 $\pm$ 0.379\\
TopoLoss & 0.9689 $\pm$ 0.0026 & 0.8064 $\pm$ 0.0112 & 1.436 $\pm$ 0.008& 1.253 $\pm$ 0.172\\
DMT & \textbf{0.9712 $\pm$ 0.0047} & 0.8289 $\pm$ 0.0189 & 1.176 $\pm$ 0.052 & 1.102 $\pm$ 0.203\\
\hline 
Dropout UNet & 0.9591 $\pm$ 0.0031 & 0.7127 $\pm$ 0.0181 & 2.483 $\pm$ 0.046 & 3.189 $\pm$ 0.371 \\
Prob.-UNet &0.9618 $\pm$ 0.0019 & 0.7091 $\pm$ 0.0201 & 2.319 $\pm$ 0.041 & 3.019 $\pm$ 0.233\\
\hline
\textbf{Ours} & 0.9637 $\pm$ 0.0032 & \textbf{0.8417 $\pm$ 0.0114} & \textbf{1.013 $\pm$ 0.081} & \textbf{0.972 $\pm$ 0.141} \\
\hline\hline
\multicolumn{5}{c}{CREMI (Volume)} \\
\hline
DIVE & 0.9542 $\pm$ 0.0037 & 0.6532 $\pm$ 0.0247 & 2.513 $\pm$ 0.047  & 4.378 $\pm$ 0.152\\
UNet &  0.9523 $\pm$ 0.0049 & 0.6723 $\pm$ 0.0312 & 2.346 $\pm$ 0.105 & 3.016 $\pm$ 0.253\\
UNet-VGG & 0.9489 $\pm$ 0.0053 & 0.7853 $\pm$ 0.0281 & 1.623 $\pm$ 0.083 & 1.973 $\pm$ 0.310\\
TopoLoss &  0.9596 $\pm$ 0.0029 & 0.8083 $\pm$ 0.0104 & 1.462 $\pm$ 0.028  & 1.113 $\pm$ 0.224\\
DMT & 0.9653 $\pm$ 0.0019 & 0.8203 $\pm$ 0.0147 & 1.089 $\pm$ 0.061  & 0.982 $\pm$ 0.179\\
\hline
Dropout UNet & 0.9518 $\pm$ 0.0018 & 0.6814 $\pm$ 0.0202 & 2.195 $\pm$ 0.087 & 3.190 $\pm$ 0.198 \\
Prob.-UNet & 0.9531 $\pm$ 0.0022 & 0.6961 $\pm$ 0.0115 & 1.901 $\pm$ 0.107 & 2.931 $\pm$ 0.177\\
\hline
\textbf{Ours} & \textbf{0.9681 $\pm$ 0.0016} & \textbf{0.8475 $\pm$ 0.0043} & \textbf{0.935 $\pm$ 0.069} & \textbf{0.919 $\pm$ 0.059}\\
\hline\hline
\multicolumn{5}{c}{DRIVE (Vessel)} \\
\hline
DIVE & 0.7543 $\pm$ 0.0008 & 0.8407 $\pm$ 0.0257 & 1.936 $\pm$ 0.127 & 3.276 $\pm$ 0.642 \\
UNet & 0.7491 $\pm$ 0.0027 & 0.8343 $\pm$ 0.0413 &  1.975 $\pm$ 0.046 & 3.643 $\pm$ 0.536\\
UNet-VGG & 0.7218 $\pm$ 0.0013 & 0.8870 $\pm$ 0.0386  & 1.167 $\pm$ 0.026 & 2.784 $\pm$ 0.293\\
TopoLoss & 0.7621 $\pm$ 0.0036 & 0.9024 $\pm$ 0.0113 & 1.083 $\pm$ 0.006 & 1.076 $\pm$ 0.265\\
DMT & 0.7733 $\pm$ 0.0039 & 0.9077 $\pm$ 0.0021 & 0.876 $\pm$ 0.038 & 0.873 $\pm$ 0.402\\
\hline
Dropout UNet & 0.7410 $\pm$ 0.0019 & 0.8331 $\pm$ 0.0152 & 2.013 $\pm$ 0.072 & 3.121 $\pm$ 0.334 \\
Prob.-UNet & 0.7429 $\pm$ 0.0020 & 0.8401 $\pm$ 0.1881 & 1.873 $\pm$ 0.081 & 3.080 $\pm$ 0.206\\
\hline
\textbf{Ours} & \textbf{0.7814 $\pm$ 0.0026} & \textbf{0.9109 $\pm$ 0.0019} & \textbf{0.804 $\pm$ 0.047} & \textbf{0.767 $\pm$ 0.098}\\
\hline
\end{tabular}
\end{center}
\end{table*}

\begin{figure*}[ht]
\centering 
\subfigure{
\includegraphics[width=0.12\textwidth]{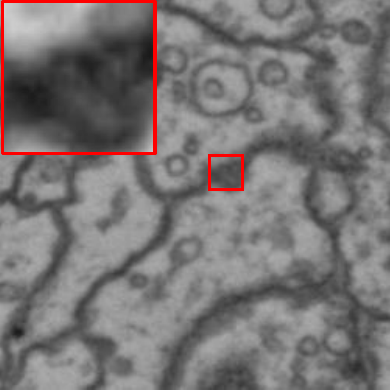}}
\subfigure{
\includegraphics[width=0.12\textwidth]{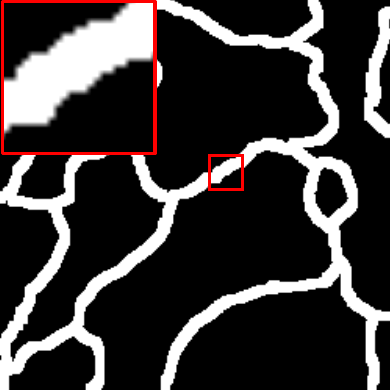}}
\subfigure{
\includegraphics[width=0.12\textwidth]{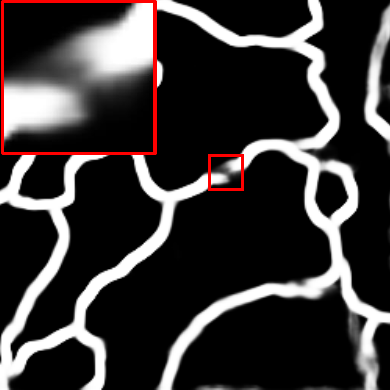}}
\subfigure{
\includegraphics[width=0.12\textwidth]{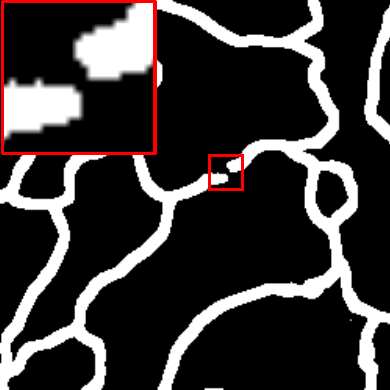}}
\subfigure{
\includegraphics[width=0.12\textwidth]{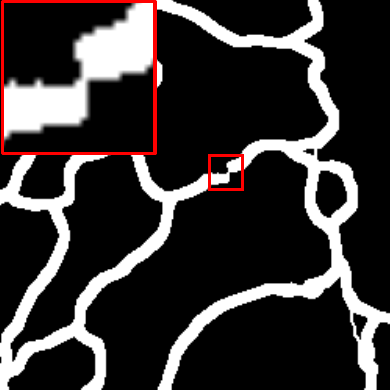}}
\subfigure{
\includegraphics[width=0.12\textwidth]{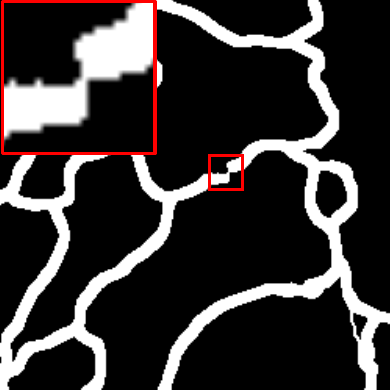}}
\subfigure{
\includegraphics[width=0.12\textwidth]{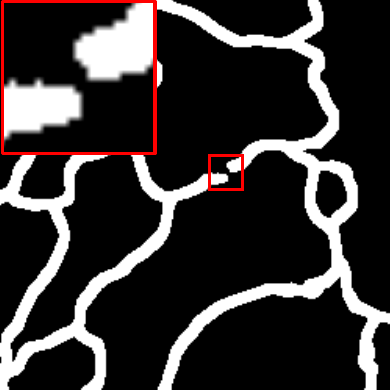}}

\subfigure{
\includegraphics[width=0.12\textwidth]{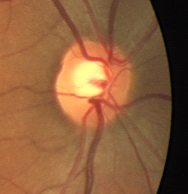}}
\subfigure{
\includegraphics[width=0.12\textwidth]{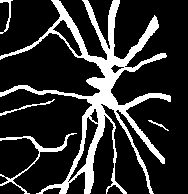}}
\subfigure{
\includegraphics[width=0.12\textwidth]{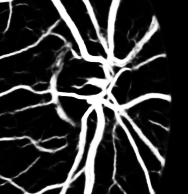}}
\subfigure{
\includegraphics[width=0.12\textwidth]{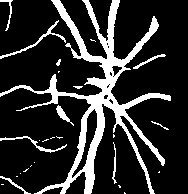}}
\subfigure{
\includegraphics[width=0.12\textwidth]{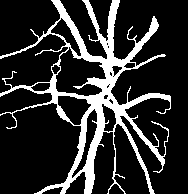}}
\subfigure{
\includegraphics[width=0.12\textwidth]{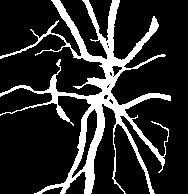}}
\subfigure{
\includegraphics[width=0.12\textwidth]{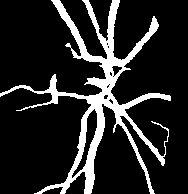}}

\subfigure{
\includegraphics[width=0.12\textwidth]{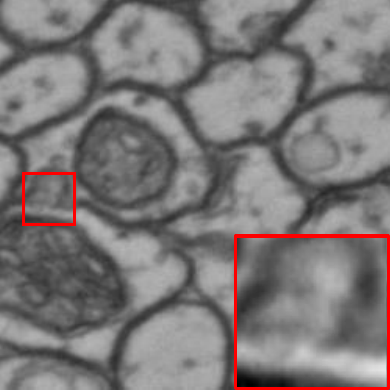}}
\subfigure{
\includegraphics[width=0.12\textwidth]{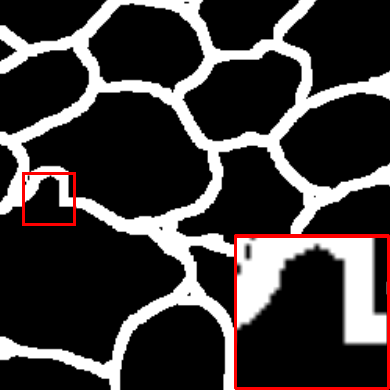}}
\subfigure{
\includegraphics[width=0.12\textwidth]{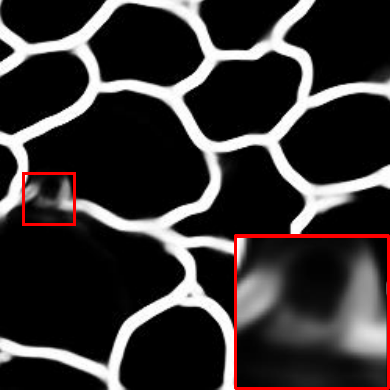}}
\subfigure{
\includegraphics[width=0.12\textwidth]{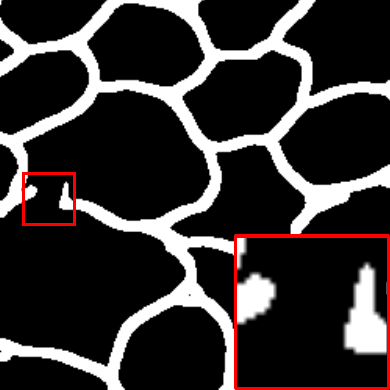}}
\subfigure{
\includegraphics[width=0.12\textwidth]{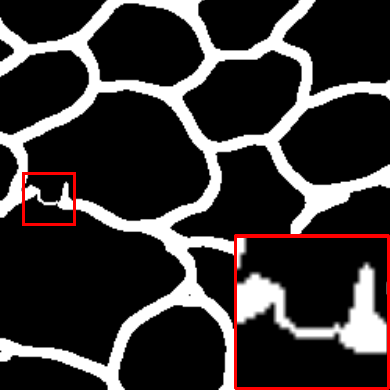}}
\subfigure{
\includegraphics[width=0.12\textwidth]{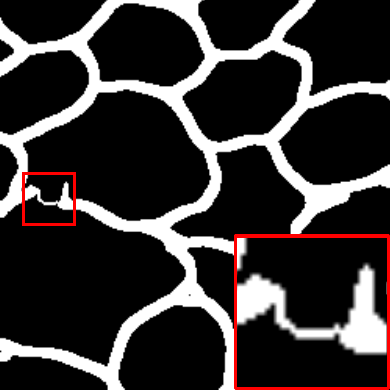}}
\subfigure{
\includegraphics[width=0.12\textwidth]{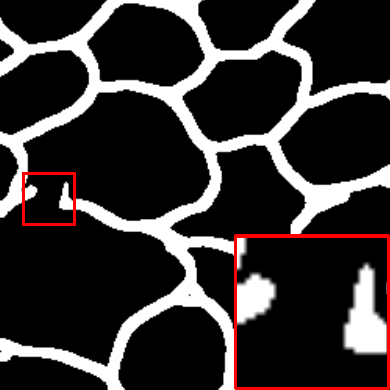}}

\subfigure{
\includegraphics[width=0.12\textwidth]{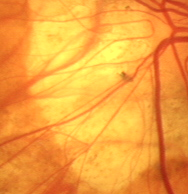}}
\subfigure{
\includegraphics[width=0.12\textwidth]{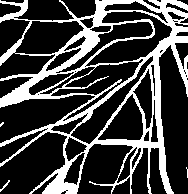}}
\subfigure{
\includegraphics[width=0.12\textwidth]{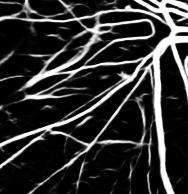}}
\subfigure{
\includegraphics[width=0.12\textwidth]{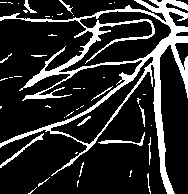}}
\subfigure{
\includegraphics[width=0.12\textwidth]{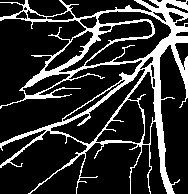}}
\subfigure{
\includegraphics[width=0.12\textwidth]{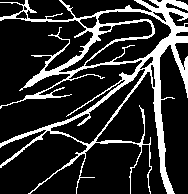}}
\subfigure{
\includegraphics[width=0.12\textwidth]{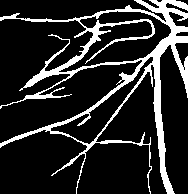}}

\subfigure{
\includegraphics[width=0.12\textwidth]{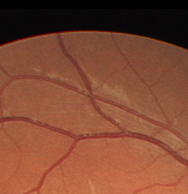}}
\subfigure{
\includegraphics[width=0.12\textwidth]{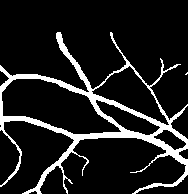}}
\subfigure{
\includegraphics[width=0.12\textwidth]{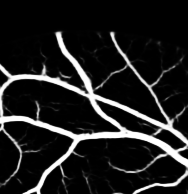}}
\subfigure{
\includegraphics[width=0.12\textwidth]{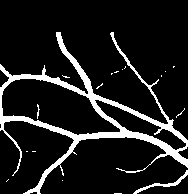}}
\subfigure{
\includegraphics[width=0.12\textwidth]{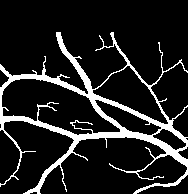}}
\subfigure{
\includegraphics[width=0.12\textwidth]{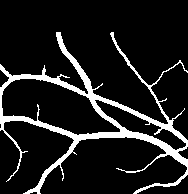}}
\subfigure{
\includegraphics[width=0.12\textwidth]{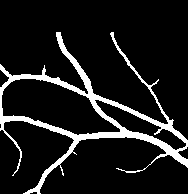}}

\subfigure{
\includegraphics[width=0.12\textwidth]{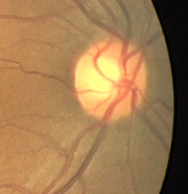}}
\subfigure{
\includegraphics[width=0.12\textwidth]{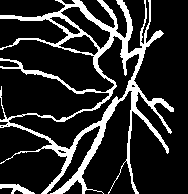}}
\subfigure{
\includegraphics[width=0.12\textwidth]{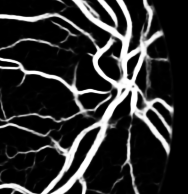}}
\subfigure{
\includegraphics[width=0.12\textwidth]{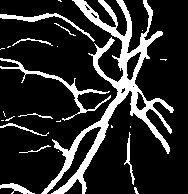}}
\subfigure{
\includegraphics[width=0.12\textwidth]{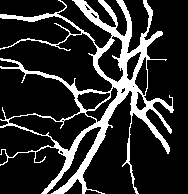}}
\subfigure{
\includegraphics[width=0.12\textwidth]{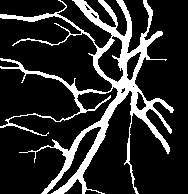}}
\subfigure{
\includegraphics[width=0.12\textwidth]{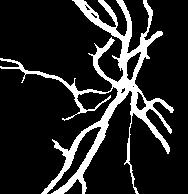}}

\subfigure{
\stackunder{\includegraphics[width=0.12\textwidth]{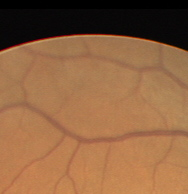}}{(a)}}
\subfigure{
\stackunder{\includegraphics[width=0.12\textwidth]{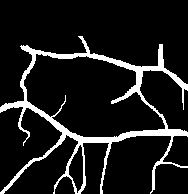}}{(b)}}
\subfigure{
\stackunder{\includegraphics[width=0.12\textwidth]{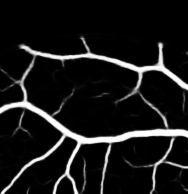}}{(c)}}
\subfigure{
\stackunder{\includegraphics[width=0.12\textwidth]{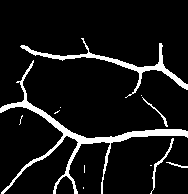}}{(d)}}
\subfigure{
\stackunder{\includegraphics[width=0.12\textwidth]{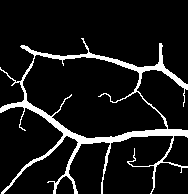}}{(e)}}
\subfigure{
\stackunder{\includegraphics[width=0.12\textwidth]{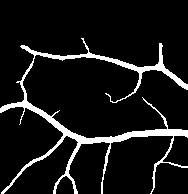}}{(f)}}
\subfigure{
\stackunder{\includegraphics[width=0.12\textwidth]{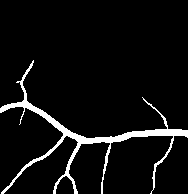}}{(g)}}

\caption{Qualitative results of our method compared to DMT-loss~\cite{hu2021topology}. From left to right: \textbf{(a)} image, \textbf{(b)} ground truth, \textbf{(c)} continuous likelihood map and \textbf{(d)} thresholded binary mask for DMT~\cite{hu2021topology}, and \textbf{(e-g)} three sampled segmentation maps generated by our method.}
\label{fig:qualitative}
\end{figure*}

Fig.~\ref{fig:qualitative} shows qualitative results. Compared with DMT~\cite{hu2021topology}, our method is able to produce a set of true structure-preserving segmentation maps, as illustrated in Fig.~\ref{fig:qualitative}(e-g). Note that compared with the existing topology-aware segmentation methods, our method is more capable of recovering the weak connections by using Morse skeletons as hints.

\begin{wrapfigure}{r}{0.5\textwidth}
\centering 
      \vspace{-.2in}
    \includegraphics[width=0.5\textwidth]{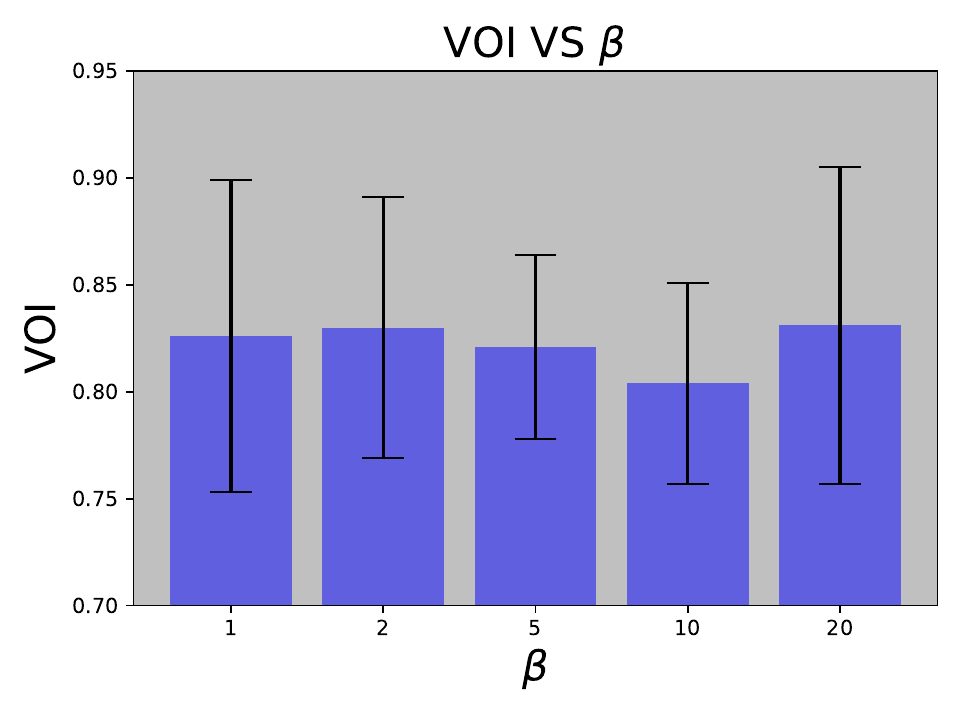}
  \caption{Ablation study for $\beta$.}
\label{fig:ablation_uncertainty}
\end{wrapfigure}

\myparagraph{Ablation Study of Loss Weights.} We observe that the performances of our method are quite robust to the loss weights $\alpha$ and $\beta$. As the learned distribution over the persistence threshold might affect the final performances, we conduct an ablation study in terms of the weight of KL divergence loss ($\beta$) on the DRIVE dataset. The results are reported in Fig.~\ref{fig:ablation_uncertainty}. When $\beta=10$, the model achieves slightly better performance in terms of VOI (0.804 $\pm$ 0.047, the smaller the better) than other choices. Note that, for all the experiments, we set $\alpha=1$.

\begin{figure*}[ht]
\centering 
\subfigure[Image]{
\includegraphics[width=0.14\textwidth]{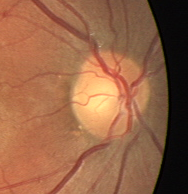}}
\subfigure[GT]{
\includegraphics[width=0.14\textwidth]{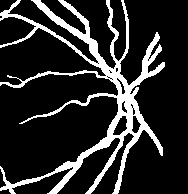}}
\subfigure[Sample1]{
\includegraphics[width=0.14\textwidth]{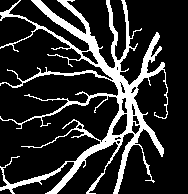}}
\subfigure[Sample2]{
\includegraphics[width=0.14\textwidth]{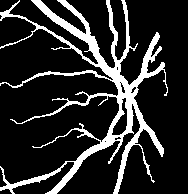}}
\subfigure[Sample3]{
\includegraphics[width=0.14\textwidth]{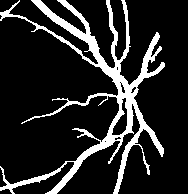}}
\subfigure[Uncertainty]{
\includegraphics[width=0.166\textwidth]{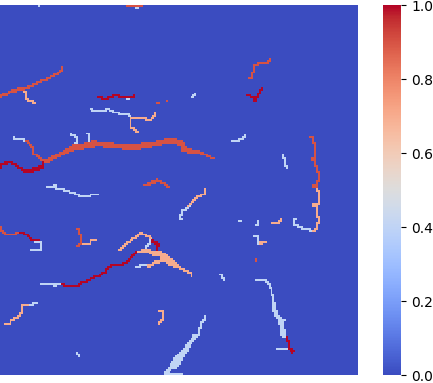}}
\caption{An illustration of structure-level uncertainty.}
\label{fig:uncertain}
\end{figure*}

\myparagraph{Illustration of the Structure-level Uncertainty.} We also explore the structure-level uncertainty with the proposed method here. We show three sampled masks (Fig.~\ref{fig:uncertain}(c-e)) in the inference stage for a given image (Fig.~\ref{fig:uncertain}(a)), and the structure-level uncertainty map (Fig.~\ref{fig:uncertain}(f)). Note that in practice, for simplification, different from Sec.~\ref{sec:uncertain}, we empirically generate an uncertainty map by taking variance across all the samples (the number is 10 in our case). Different from pixel-wise uncertainty, each small branch has the same uncertainty value as our method. By looking at the original image, we find that the uncertainties are usually caused by the weak signals of the original image. The weak signals of the original image make the model difficult to predict these locations correctly and confidently, especially structure wise. Actually, this also makes sense in real cases. Different from natural images, even experts can not always reach a consensus for biomedical image annotation~\cite{armato2011lung,clark2013cancer}. 

Fig.~\ref{fig:more_uncertain} shows the comparison of traditional pixel-wise uncertainty and the proposed structure-level uncertainty. Specifically, Fig.~\ref{fig:more_uncertain}(c) is a sampled segmentation result by Prob.-UNet~\cite{kohl2018probabilistic}, and Fig.~\ref{fig:more_uncertain}(d) is the pixel uncertainty map from Prob.-UNet~\cite{kohl2018probabilistic}. Different from traditional pixel-wise uncertainty, our proposed structure-level uncertainty (Fig.~\ref{fig:more_uncertain}(f)) can focus on the structures.  

\begin{figure*}[ht]
\centering 

\subfigure[Image]{
\includegraphics[width=0.3\textwidth]{uncertainty/img9_1_2_ori.png}}
\subfigure[Overlay]{
\includegraphics[width=0.3\textwidth]{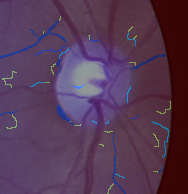}}
\caption{Overlaying the structure-level uncertainty on the original image.}
\label{fig:overlay}
\end{figure*}

We also overlay the structure-level uncertainty (Fig.~\ref{fig:more_uncertain}(f)) on the original image (Fig.~\ref{fig:more_uncertain}(a)), which is shown in Fig.~\ref{fig:overlay}. By comparison with the original image, we can observe that the structure-level uncertainty is mainly caused by the weak signals in the original image. 

\myparagraph{The Advantage of the Joint Training and Optimization.} Another straightforward alternative to the proposed approach is to use the discrete Morse theory to postprocess the continuous likelihood map obtained from the standard segmentation networks. 

In this way, we can still obtain structure-preserving segmentation maps, but there are two main issues: 1) if the segmentation network itself is structure-agnostic, we'll not be able to generate satisfactory results even with the postprocessing, and 2) we have to manually choose the persistence threshold to prune the unnecessary branches for each image, which is cumbersome and unrealistic in practice. The proposed joint training strategy overcomes both of these issues. First, during the training, we incorporate the structure-aware loss ($L_{skeleton}$). Consequently, the trained segmentation branch itself is structure-aware essentially. On the other hand, with the prior and posterior nets, we are able to learn a reliable distribution of the persistence threshold ($\epsilon$) given an image in the inference stage. Sampling over the distribution makes it possible to generate satisfactory structure-preserving segmentation maps within a few trials (the inference won't take long), which is more much efficient. 

We conduct an empirical experiment to demonstrate the advantage of joint training and optimization. For the postprocessing, given the predicted likelihood maps from the standard segmentation networks, we randomly choose five persistence thresholds and generate the segmentation masks separately, and select the most reasonable one as the final segmentation mask to report the performances. The results in Tab.~\ref{table:ablation_joint} demonstrate the advantage of our joint training and optimization strategy.

\setlength{\tabcolsep}{3pt}
\begin{table*}[ht]
\begin{center}
\footnotesize
\caption{Quantitative results for comparison of postprocessing on DRIVE dataset.}
\label{table:ablation_joint}
\begin{tabular}{ccccc}
\hline
Method  & Dice $\uparrow$ &  ARI $\uparrow$ & VOI $\downarrow$ & Betti Error $\downarrow$\\
\hline
Postprocessing & 0.7653 $\pm$ 0.0052 & 0.8841 $\pm$ 0.0046 & 1.165 $\pm$ 0.086 & 1.249 $\pm$ 0.388\\
\hline
\textbf{Ours} & \textbf{0.7814 $\pm$ 0.0026} & \textbf{0.9109 $\pm$ 0.0019} & \textbf{0.804 $\pm$ 0.047} & \textbf{0.767 $\pm$ 0.098}\\
\hline
\end{tabular}
\end{center}
\end{table*}

\subsection{Semi-Automatic Efficient Annotation/Proofreading with User Interaction}
\label{sec:semi-proofreading}
Proofreading is a struggling while essential step for the annotation of images with fine-scale structures. We develop a semi-automatic interactive annotation or proofreading pipeline based on the proposed method. 
As the proposed method is able to generate structural segmentations (Fig.~\ref{fig:uncertain}(c)-(e)), the annotators can efficiently proofread one image with rich structures by removing the unnecessary/redrawing the missing structures with the help of structure-level uncertainty map (Fig.~\ref{fig:uncertain}(f)). The whole inference and structure-aware interactive image annotation pipeline is illustrated in Fig.~\ref{fig:inference}.

\begin{wrapfigure}{r}{0.4\textwidth}
\centering 
  \includegraphics[width=1\linewidth]{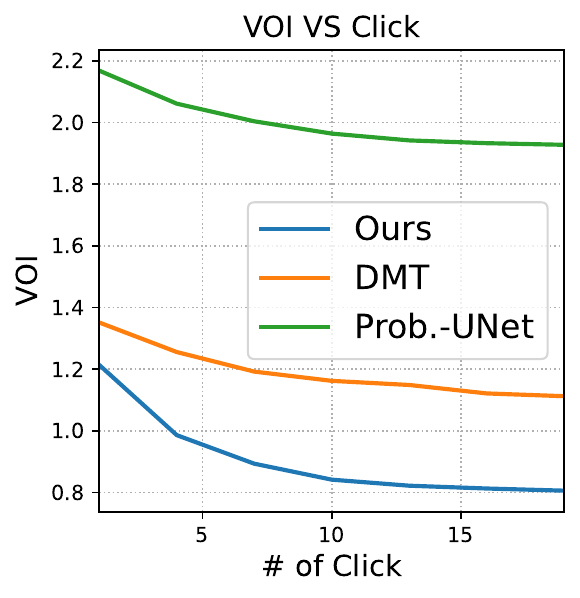}
    \caption{Interaction simulation.}
\label{fig:user_study}
\end{wrapfigure}

We conduct empirical experiments to demonstrate the efficiency of proofreading by using the proposed method. We randomly select a few samples from the ISBI13 dataset and simulate the user interaction process. For both the proposed and baseline methods, we get started from the final segmentations and correct \textit{one misclassified branch} each time. For the deterministic method (DMT), the user draws one false-negative or erases one false-positive branch for each click. For the pixel-wise probabilistic method (Prob.-UNet), the user does the same while taking the uncertainty map as guidance. For the proposed method, the user checks each branch based on the descending order of structure-level uncertainty. VOI is used to evaluate the performances.

Fig.~\ref{fig:user_study} shows the comparative results of semi-automatic proofreading with user interactions. By always checking branches with the highest uncertainty, the proposed method clearly achieves better results and improves the results much faster than baseline methods. Our developed pipeline achieves higher efficiency because of the following two perspectives: 1) the generated structural segmentations are essentially topology/structure-preserving; 2) the proposed method provides the structure-level uncertainty map to guide the human proofreading.

\section{Applications: Mitochondria Annotation}

Mitochondria play a vital role as the main energy suppliers for cellular functions, making them indispensable for metabolism. Accurate measurement of the size and shape of mitochondria is not only critical for fundamental neuroscience investigations~\cite{schubert2019learning} but also valuable for clinical studies examining various diseases~\cite{kasahara2016depression, zeviani2004mitochondrial}.

Electron microscopy (EM) provides detailed snapshots of cellular and subcellular ultrastructure at high resolutions. Recent advancements in volume EM have made it possible to obtain imaging of larger specimens~\cite{peddie2014exploring, titze2016volume}, and deep learning algorithms play an active role in enabling quantitative analyses~\cite{januszewski2018high, funke2018large}. These methods have also been applied to investigate organellar structures in various systems~\cite{heinrich2021whole, mekuvc2020automatic}, leading to unprecedented biological insights on a large scale. The typical deep learning workflow for such applications involves densely annotating within regions of interest (2D or 3D), training a model based on the annotations, conducting inference on the unannotated dataset, and then involving humans to proofread the model's output~\cite{berning2015segem, januszewski2018high}. Although achieving impressive performances, both predicting and proofreading are conducted at the pixel level, which is not efficient enough.  

The original method is developed for curvilinear structures. In this section, we mainly focus on the problem of human-in-the-loop mitochondria annotation. In order to extend from curvilinear structure data to non-curvilinear structure data, we propose to adopt distance transform and transform the instance segmentation task to a boundary segmentation problem. By measuring the existing confidence/uncertainty of the boundaries, we are able to measure the probability of instances getting connected or not. Based on the proposed probabilistic method, we aim to develop a semi-automatic human-in-the-loop mitochondria annotation pipeline. Our intuition is, together with the mitochondria segmentation mask, the generated probability map can be used to guide the proofreading process and develop the semi-automatic human-in-the-loop annotation pipeline.

\begin{figure}[ht]
    \centering
    \includegraphics[width=1\linewidth]{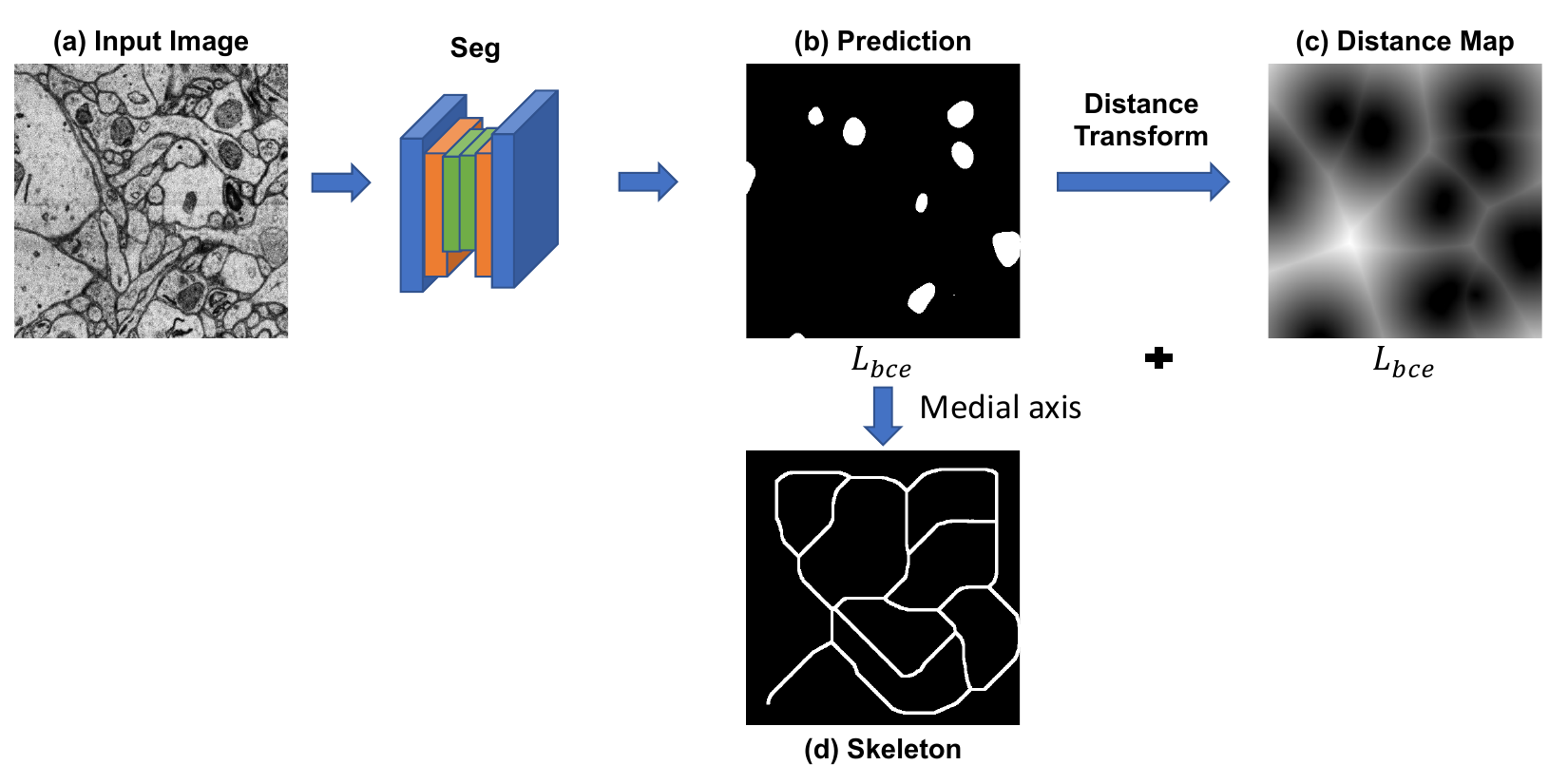}
    \caption{Illustration of the segmentation branch in 2D view.}
    \label{fig:seg_branch}
\end{figure}

\subsection{Distance Transform for Mitochondria}

As aforementioned, we propose to transform the discretely labeled masks to distance maps for mitochondria data. Given the ground-truth label $Y$ of the mitochondria, let $S_V$ be the set of pixels/voxels on the mitochondria surface, which can be defined by

\begin{equation}
     S_{V}=\{v|y_v = 1, \exists u \in \mathcal{N}(v), y_u = 0 \},
\end{equation}
where $\mathcal{N}(v)$ denotes the 4 or 6-neigobor voxels (in 2D or 3D scenarios) of $v$. The distance transform map is formally defined as:
\begin{equation}
     d_{v}=\begin{cases}
\min \limits_{u \in S_V}||v-u|| & \text{ if } y_v= 0, \\
0 & \text{ if } y_v= 1.
\end{cases}
\end{equation}

For each voxel $v$, the distance transform operation assigns it a distance transform value which is the nearest distance from $v$ to the mitochondria surface $S_V$.

To learn more accurate distance transform maps, we additionally impose a distance loss term used to train the network, which indicates a penalty if the predicted distance transform value is different from its counterpart of ground-truth. Details will be introduced later.

\subsection{Segmentation Branch}
\begin{figure*}[ht]
    \centering
    \includegraphics[width=1\linewidth]{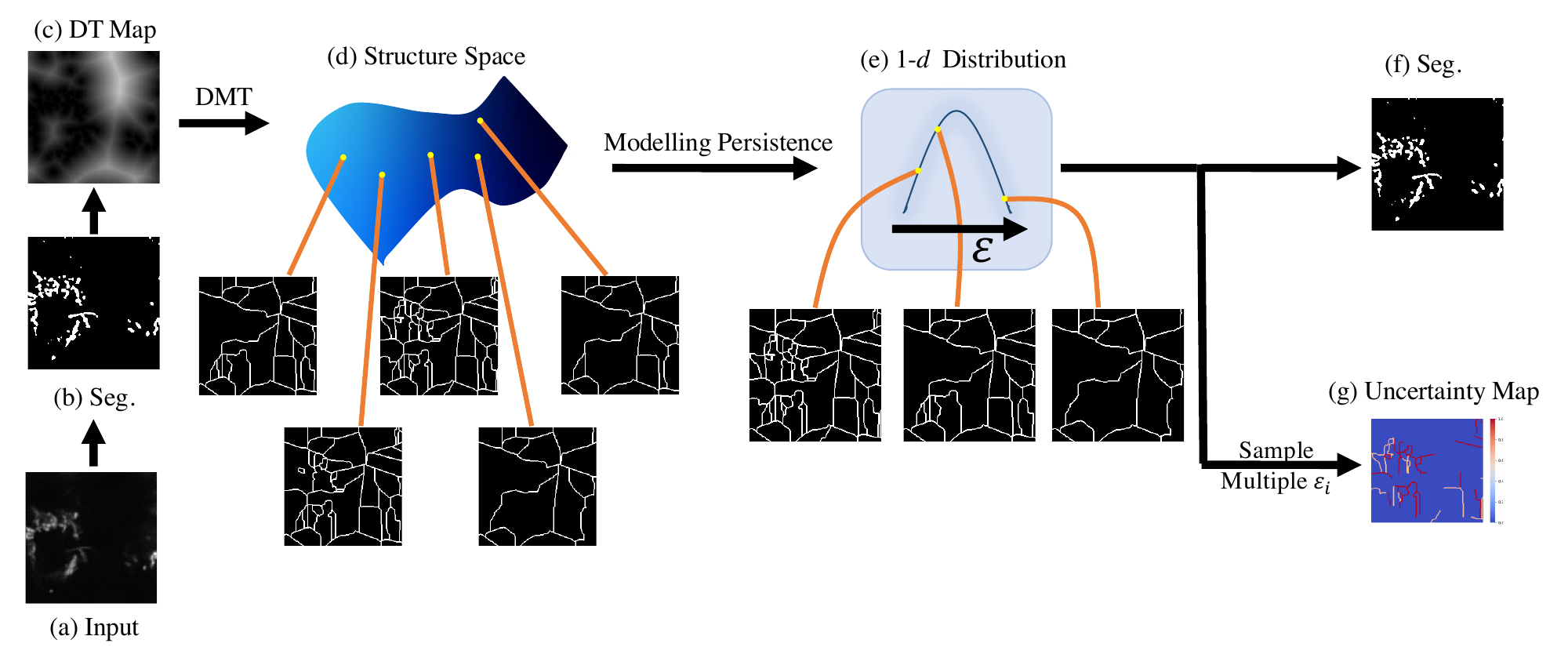}
    \caption{The overall framework of the proposed method. Part of the framework credits to ~\cite{hu2022learning}.}
    \label{fig:mito_framework}
\end{figure*}

Since structural reasoning needs a curvilinear-like structure to construct discrete Morse complexes, the proposed method first obtains such a likelihood map by training a segmentation branch. More specifically, we adopt distance transform to transform the binary mask to continuous distance maps. Besides enforcing the penalty on the predicted likelihood map, we also impose an additional penalty term on the distance maps. Formally, 
\begin{equation}
L_{seg} = L_{bce}(Y, S(X;\omega_{seg})) + \lambda L_{bce}(d^{GT}, d^{pred}),
\end{equation}
in which $Y$ is the ground truth. $X$ is the input, $\omega_{seg}$ denotes the parameters of the segmentation branch, and $S(X;\omega_{seg})$ is the output likelihood map. $d_{GT}$ and $d_{pred}$ denote the distance maps for binary ground truth and binarized predicted result, respectively. See Fig.~\ref{fig:seg_branch} for an illustration.
 
In Fig.~\ref{fig:seg_branch}(d), we also show the medial axis of the background region. The medial axis partitions the whole image/patch into different sub-regions, and each sub-region corresponds to one mitochondria component (illustrated in 2D). The distance map of the binarized predicted result, $d_{pred}$, is a curvilinear structure and thus can be fed to the discrete Morse theory algorithm (DMT). 

The discrete Morse theory (DMT) algorithm can decompose the distance map into many more small Morse branches, and each branch denotes a possible boundary to separate different mitochondria components. Formally, we call any combination of these branches a sample from the structural space, and each branch has a specific persistence value. Similarly, a threshold value $\epsilon$ can be used to filter out the most salient branches, and different $\epsilon$'s result in different branch combinations, each of them corresponding to one specific mitochondria segmentation result. 

\subsection{Skeleton Extraction}

The DMT computation is referred to as $d^{pred}_{DMT}$, and the persistent homology thresholding is referred to as $d^{pred}_{PH}$. So given a distance map $d^{pred}_{PH}$ and a threshold $\epsilon$, a structure/skeleton can be generated as follow:
\begin{equation}
\label{eq:skeleton}
    S_{skeleton}(\epsilon) = d^{pred}_{PH}(d^{pred}_{DMT}; \epsilon).
\end{equation}

Besides the segmentation branch, we also want to make sure the $S_{skeleton}(\epsilon)$ is correctly predicted. More specifically, we impose additional supervision regarding the predicted skeleton. We use the skeleton $S_{skeleton}(\epsilon)$ to mask both the predicted distance map $d^{pred}$ and ground truth distance map $d^{GT}$. And then we compute the binary cross-entropy loss over the sampled skeleton 
\begin{equation}
    L_{bce}(d^{GT} \circ S_{skeleton}(\epsilon), S(d^{pred} \circ S_{skeleton}(\epsilon)),
\end{equation}
in which $\circ$ denotes the Hadamard product. One of our goals is to learn a 1-D Gaussian distribution, $\epsilon\sim N(\mu, \sigma)$, regarding the persistent homology threshold. Involving the 1-D Gaussian distribution, more formally, we have the expected loss:
\begin{equation}
\begin{split}
    L_{skeleton} = \mathbb{E}_{\epsilon \sim N(\mu, \sigma)} L_{bce}(d^{GT} \circ S_{skeleton}(\epsilon), \\  S(d^{pred} \circ S_{skeleton}(\epsilon)).
\end{split}
\end{equation}

This loss term also supervised the training of the segmentation branch via the predicted distance map. The backpropagation can be achieved through the reparameterization technique in~\cite{kingma2013auto}.

We also employ the VAE framework to learn better distributions over $\epsilon$. Thus, the Kullback-Leibler divergence of these two distributions is calculated, 
\begin{equation}
    D_{KL}(Q||P) = \mathbb{E}_{\epsilon \sim Q} (\log \frac{Q}{P}),
\end{equation}
which measures how distance between the prior distribution $P$ ($N(\mu_{prior}, \sigma_{prior})$) and the posterior distribution $Q$ ($N(\mu_{post}, \sigma_{post})$). 

\subsection{Training Neural Network}
Our final loss consists of standard segmentation loss (including the distance map term), the skeleton term $L_{skeleton}$, and the KL divergence term.
\begin{equation}
    L(X, Y) = L_{seg} + \alpha  L_{skeleton} + \beta D_{KL}(Q||P),
\end{equation}
and it can be rewritten as,
\begin{equation}
\begin{split}
        L(X, Y) = L_{bce}(Y, S(X;\omega_{seg})) + \lambda L_{bce}(d^{GT}, d^{pred}) + \\
    \alpha  L_{skeleton} + \beta D_{KL}(Q||P),
\end{split}
\end{equation}
The parameters $\lambda$, $\alpha$, and $\beta$ are weights to balance the four different loss terms. 

\subsection{Experiment Design and Results}
\myparagraph{Datasets:} We conduct our experiments on popular mitochondria datasets, MitoEM dataset~\cite{wei2020mitoem}. \textbf{MitoEM~\cite{wei2020mitoem}}: MitoEM dataset consists of two volumes with resolutions of $8 \times 8 \times 30$ $nm^3$, which come from human (MitoEM-H) and rat (MitoEM-R) cortices, respectively. The authors cropped a (30 $\mu$m)$^3$ sub-volume to contain the mitochondria. The carefully curated dataset contains complex mitochondria without introducing much of the domain adaptation problem due to the diverse image appearance. Each subset has a training set ($400 \times 4096 \times 4096$) and a validation set ($100 \times 4096 \times 4096$).

\myparagraph{Baselines:} We compare the proposed method with two kinds of baselines: 1) Standard segmentation baselines: \textbf{DIVE}~\cite{fakhry2016deep}, \textbf{UNet}~\cite{ronneberger2015u},
and 2) Probabilistic-based segmentation methods: \textbf{Dropout UNet}~\cite{kendall2015bayesian}, \textbf{Probabilistic-UNet}~\cite{kohl2018probabilistic}. The details of these baselines have been introduced in Sec.~\ref{sec:baseline}.

\myparagraph{Evaluation Metrics:} We use four different evaluation metrics: \textbf{Jaccard-index coefficient}, \textbf{Dice score}, \textbf{AP-75}, and \textbf{VOI}. The details of \textbf{Dice score} and \textbf{VOI} can be found in Sec.~\ref{sec:metric_topoloss} and Sec.~\ref{sec:metric_warping}. The other two metrics are as followings:
\begin{itemize}
    \item \textit{Jaccard-index coefficient}: Jaccard-index coefficient score is usually used to gauge the similarity and diversity of sample sets, and it's often used for segmentation tasks.
        \item \textit{AP-75}: Following~\cite{wei2020mitoem}, we also choose AP-75, which requires at
least 75\% intersection over union (IoU) with the ground truth for a detection to be a true positive.
\end{itemize}



\myparagraph{Results:} Table~\ref{table:mito_quantitative} shows the quantitative results regarding the segmentation, comparing to baselines. Note that for deterministic methods (standard segmentation methods), the numbers are computed directly; while for probabilistic-base segmentation methods, five segmentation masks are generated at once, and the averaged numbers over the five segmentation masks for each image (for both the baselines and the proposed method). The proposed method obviously outperforms the baseline methods. 
\begin{figure*}[t]
\centering 
\subfigure{
\includegraphics[width=0.18\textwidth]{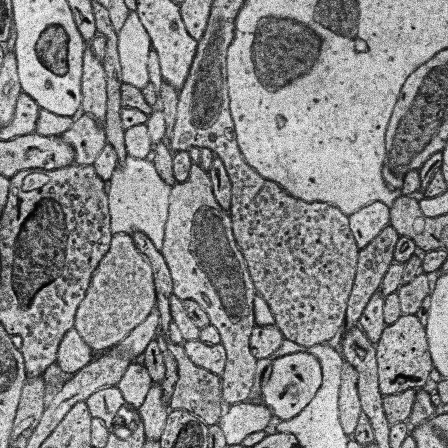}}
\subfigure{
\includegraphics[width=0.18\textwidth]{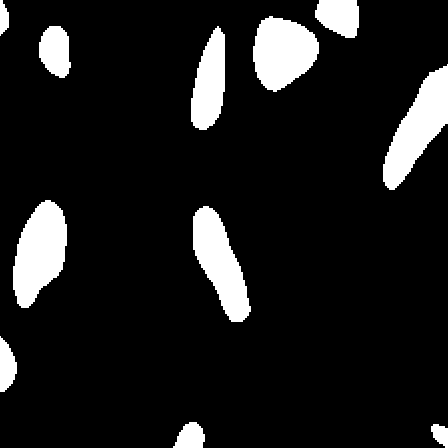}}
\subfigure{
\includegraphics[width=0.18\textwidth]{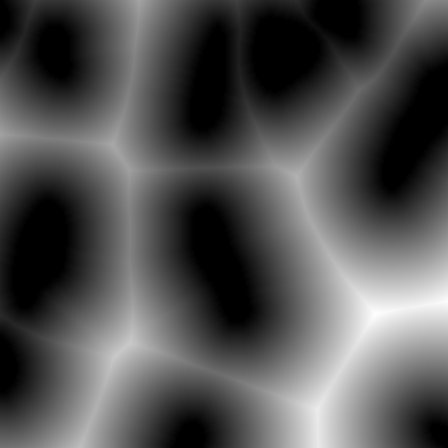}}
\subfigure{
\includegraphics[width=0.18\textwidth]{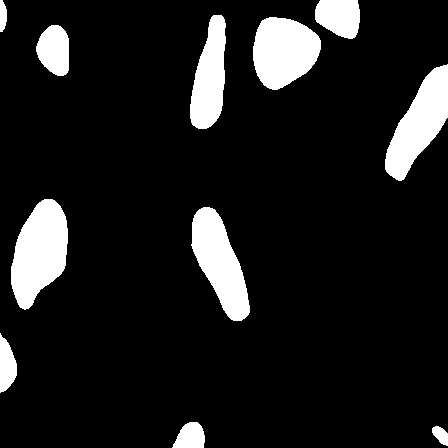}}
\subfigure{
\includegraphics[width=0.18\textwidth]{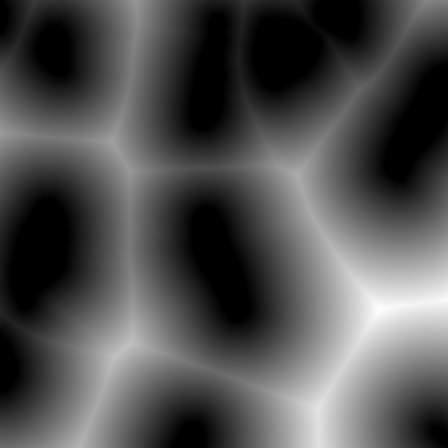}}

\subfigure{
\stackunder{\includegraphics[width=0.18\textwidth]{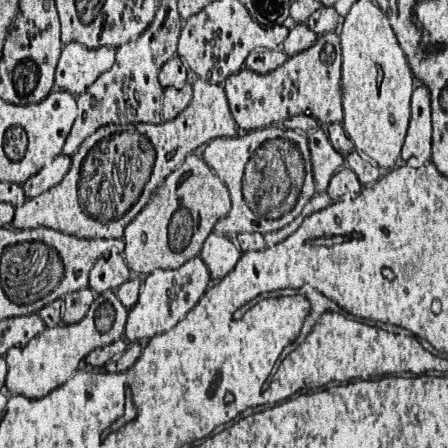}}{(a) Image}}
\subfigure{
\stackunder{\includegraphics[width=0.18\textwidth]{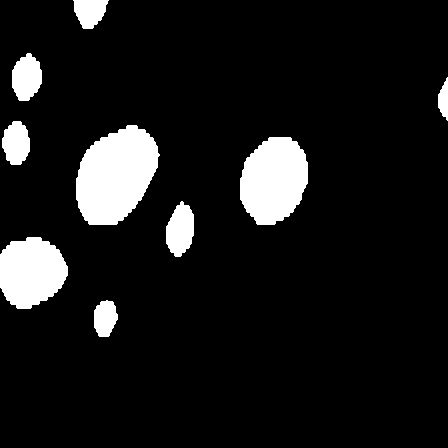}}{(b) GT}}
\subfigure{
\stackunder{\includegraphics[width=0.18\textwidth]{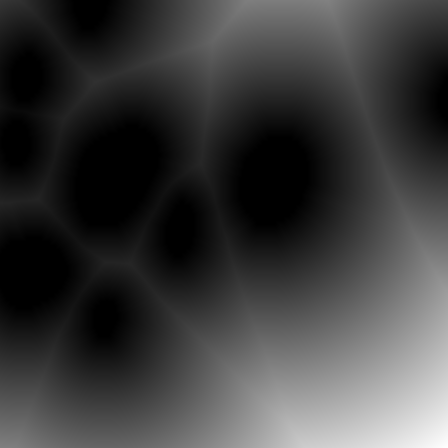}}{(c) $d^{GT}$}}
\subfigure{
\stackunder{\includegraphics[width=0.18\textwidth]{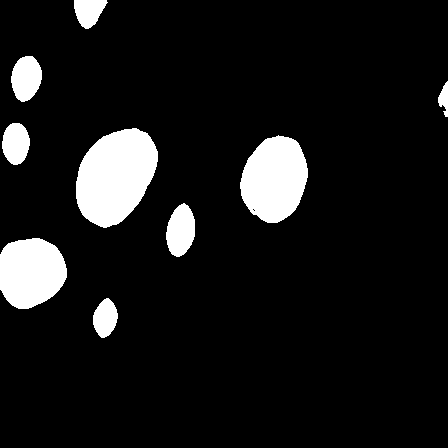}}{(d) Pred}}
\subfigure{
\stackunder{\includegraphics[width=0.18\textwidth]{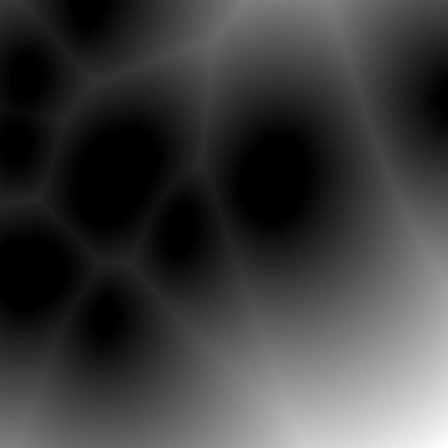}}{(e) $d^{pred}$}}
\caption{Segmentation results of the proposed method. From left to right: \textbf{(a)} image, \textbf{(b)} ground truth, \textbf{(c)} $d^{GT}$, \textbf{(d)} segmentation maps generated by our method, and \textbf{(e)} $d^{pred}$.}
\label{fig:mito_qualitative}
\end{figure*}

Fig.~\ref{fig:mito_qualitative} shows qualitative results. Our method is able to predict both segmentation maps (Fig.~\ref{fig:mito_qualitative}(d)) and distance maps (Fig.~\ref{fig:mito_qualitative}(e)) accurately.

\setlength{\tabcolsep}{6pt}
\begin{table}[t]
\begin{center}
\small
\caption{Quantitative results for different models on mitochondria datasets.}
\label{table:mito_quantitative}
\begin{tabular}{ccccc}
\hline
Method  & Jaccard $\uparrow$ & Dice $\uparrow$ &  AP-75 $\uparrow$ & VOI $\downarrow$\\
\hline\hline
\multicolumn{5}{c}{MitoEM-R} \\
\hline
UNet & 0.821 & 0.897 & 0.481  & 0.262\\
DIVE & 0.792 &  0.874 & 0.407  & 0.291\\
\hline
Dropout UNet & 0.833 & 0.905 & 0.496  & 0.238 \\
Prob.-UNet & 0.841 & 0.911 & 0.541  & 0.284\\
\hline
\textbf{Ours} & \textbf{0.847} & \textbf{0.912} & \textbf{0.604}  & \textbf{0.173}\\
\hline\hline
\multicolumn{5}{c}{MitoEM-H} \\
\hline
UNet & 0.816 & 0.884 & 0.583  & 0.344\\
DIVE & 0.788 & 0.850 & 0.482  & 0.318\\
\hline
Dropout UNet & 0.820 & 0.891 & 0.572  & 0.378 \\
Prob.-UNet & 0.825 & 0.892 & 0.623  & 0.298\\
\hline
\textbf{Ours} & \textbf{0.839} & \textbf{0.905} & \textbf{0.635} & \textbf{0.216}\\
\hline
\end{tabular}
\end{center}
\end{table}
\subsection{Human-in-the-loop Structure-Aware Mitochondria Annotation Pipeline}
\label{sec:mito-proofreading}

\begin{wrapfigure}{r}{0.4\textwidth}
\centering 
  \includegraphics[width=1\linewidth]{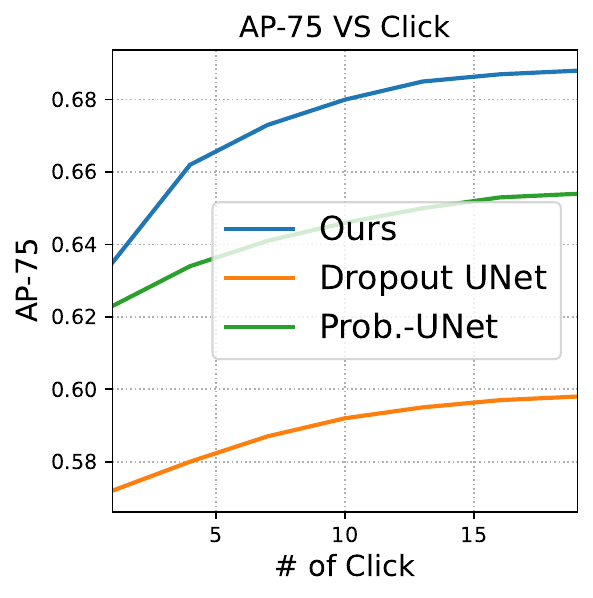}
    \caption{Human-in-the-loop annotation illustration. Experiments are conducted on the MitoEM-H dataset.}
\label{fig:mito_curve}
\vspace{-.2in}
\end{wrapfigure}
Based on segmentation results and the boundary uncertainty estimation, we are able to develop an efficient human-in-the-loop structure-aware mitochondria annotation pipeline. The rationale is that, by always focusing on the most uncertain branches, the human annotations will be able to proofread the segmentation results efficiently.


We conduct empirical experiments on the MitoEm-H dataset to demonstrate the efficiency of the proposed mitochondria annotation pipeline. For both the proposed and baseline methods, we get started from the final segmentations and conduct one click each time. For the pixel-wise probabilistic method (Dropout UNet and Prob.-UNet), the user focuses on one branch each time. For the proposed method, the user checks branch by branch based on the descending order of branch uncertainty. AP-75 is used to evaluate the performances.

Fig.~\ref{fig:mito_curve} shows the results of the human-in-the-loop structure-aware mitochondria annotation pipeline. By always checking the most uncertain branches, the proposed method obviously outperforms baseline methods, which is not surprising, as the developed pipeline is always focusing on the whole branches instead of pixels.

\section{Conclusion}
\label{sec:conslusion}
Instead of making pixel-wise predictions, we propose to learn structural representation with a probabilistic model to obtain structural segmentations. Specifically, we construct the structure space by leveraging classical discrete Morse theory. We then build a probabilistic model to learn a distribution over structures. In the inference stage, we are able to generate a set of structural segmentations and explore the structure-level uncertainty, which is beneficial for interactive proofreading. We further extend the proposed method from curvilinear structure data to non-curvilinear structure data (mitochondria).
Extensive experiments have been conducted to demonstrate the efficacy of the proposed method. 

\clearpage

%% file: conclusion.tex
\chapter{Conclusion and Future Work}
\label{chapter:conclusion}

In the previous chapters, we have introduced our research on \textit{Learning Topological Representations for Deep Image Segmentation}, in five chapters:

\begin{itemize}
    \item \textit{Topology-Preserving Deep Image Segmentation}, Chapter~\ref{chapter:topoloss}~\cite{hu2019topology}.
    \item \textit{Trigger Hunting with a Topological Prior for Trojan Detection}, Chapter~\ref{chapter:trojan}~\cite{hu2022trigger}.
    \item \textit{Structure-Aware Image Segmentation with Homotopy Warping}, Chapter~\ref{chapter:warping}~\cite{hu2022structure}.
    \item \textit{Topology-Aware Segmentation Using Discrete Morse Theory}, Chapter~\ref{chapter:dmt}~\cite{hu2021topology}.
        \item \textit{Learning Probabilistic Topological Representations Using Discrete Morse Theory}, Chapter~\ref{chapter:uncertainty}~\cite{hu2022learning}.
\end{itemize}

My research has been focused on understanding images from the topology/structure view. 
By using persistent homology, we have been proposing differentiable topological loss to segment with correct topology~\cite{hu2019topology, yang2021topological}. By enforcing topological constraints into the trojan detection context~\cite{hu2022trigger}, we further explore the power of the topological loss to recover triggers with as few connected components as possible. To avoid the critical points which are not really relevant to topological errors, we further introduce another warping strategy to efficiently identify critical points~\cite{hu2022structure}. Instead of identifying a sparse set of critical points at every epoch, we use discrete Morse theory to identify critical structures instead of critical points~\cite {hu2021topology}. Furthermore, we propose to learn topological/structural representations directly~\cite{hu2022learning}.

Throughout my thesis research on learning topological representations on deep image understanding, I believe topology plays an important role in analyzing images, especially biomedical images, with rich structures. Exploring the geometry/topology/structure of various data will be helpful to achieve my ultimate goal --- building AI systems that can efficiently assist in disease diagnosis and treatment. Though making solid progress, there is still a long way to go before I reach my long-term goal. Below, we discuss a few directions I plan to pursue in the next stage.

\begin{itemize}
    \item \textbf{\textit{Efficiency of Using Data}}: 
    By using topological tools~\cite{hu2019topology, hu2021topology, hu2022structure, hu2022deep, yang20213d, yang2021topological, gupta2022learning, hu2022trigger}, I have explored the data efficiently by taking their topological properties into consideration. However, as the field of topology-driven image analysis continues to evolve, I am deeply motivated to explore even deeper into the realm of data efficiency. More specifically, I plan to develop novel methods to optimize the utilization of data, capitalizing on its intrinsic topological/geometric/structural information to enhance the efficiency of image analysis methodologies. By using data-driven techniques, I strive to create novel algorithms with both concrete
    theoretical foundations and strong empirical results for imaging data under different contexts, especially biomedical scenarios. 
    \item \textbf{\textit{Uncertainty Estimation and Its Applications}}: Besides achieving good performances, we usually want to measure how the model is confident of its predictions, which is important in different scenarios. \textit{A good model should both know what it knows and what it does not know}. Building on my recent work regarding structure-wise uncertainty estimation for curvilinear structure data~\cite{hu2022learning, gupta2023topology} and confidence estimation using unlabeled data~\cite{li2022confidence, li2023calibrating}, I want to further expand the boundary of uncertainty estimation and its wide-ranging applications. As deep learning methods have been generalized in different contexts, understanding and quantifying the uncertainty of deep models has emerged as a crucial aspect. By demystifying the uncertainty of models, I aim to not only enhance the reliability of predictive models but also extend their applications to diverse domains.
    \item \textbf{\textit{Deep Learning Based Quantification Analysis}}: Based on my work on topology-preserving deep image segmentation and topology-aware uncertainty estimation in Chapter~\ref{chapter:dmt} and Chapter~\ref{chapter:uncertainty}, I am ready for quantifying how the accurate topology/geometry aware segmentation results affect the downstream analysis. By extracting topology/geometry-informed features, we can do a number of interesting analyses, such as diagnosing retinal diseases, predicting aortic aneurysm eruption risk, and so on. Fig.~\ref{fig:downstream} is an illustration pipeline for the downstream analysis. This is an ongoing direction and will be explored further.
\begin{figure*}[ht]
\centering 
\subfigure{
\includegraphics[width=.95\textwidth]{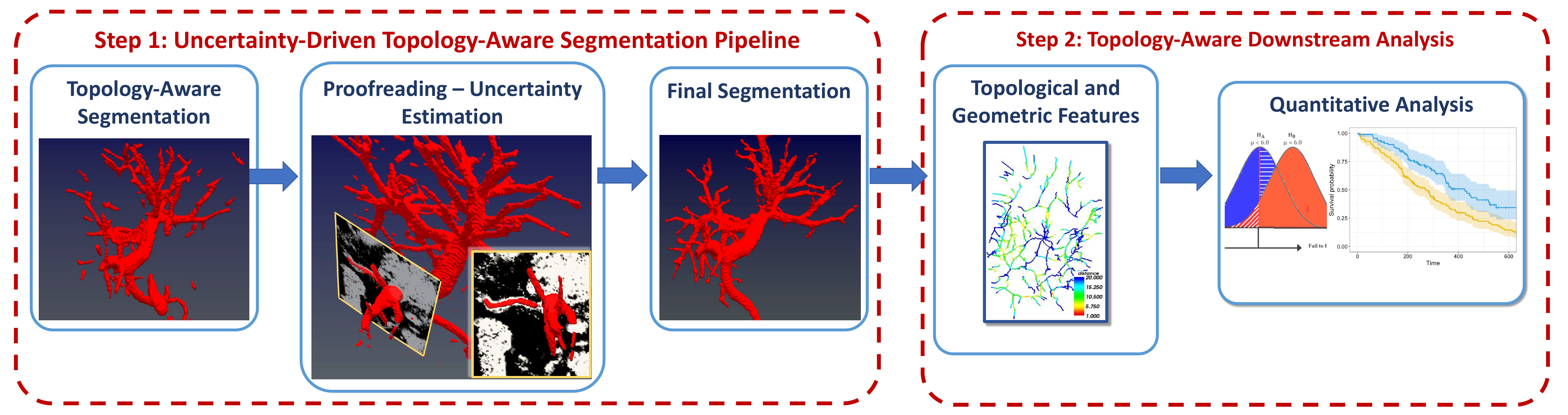}}
\caption{Workflow for uncertainty-driven topology-aware segmentation, and the downstream topology/geometry aware analysis.}
\label{fig:downstream}
\end{figure*}
    \item \textbf{\textit{Learning with Imperfect Data}}: In my thesis, I have been focused on fully supervised scenarios with ground truth. While in practice especially in biomedical contexts, the gathering of labeled data can be cost-prohibitive and time-consuming, which usually requires domain knowledge~\cite{konwer2023enhancing}. The dependency on diverse, high-quality training datasets substantially limits model applicability to complex scenes where data from are imperfect including \textit{missing modality}, and/or \textit{human labeling limited}. Furthermore, the uncertainty estimation~\cite{li2022confidence} I have been working on might provide hints for scenarios with imperfect data. Driven by this, I would like to answer the following question: 1) are we able to achieve satisfactory performances even under scenarios with imperfect data? 2) are we able to focus on the most unconfident samples/regions to facilitate the training with imperfect data?~\cite{konwer2023enhancing}
\end{itemize}

As aforementioned, over the next few years, I will continue to create novel algorithms with both concrete theoretical foundations and strong empirical results for imaging data under different contexts, especially biomedical scenarios, and  build intelligent AI systems that can assist in diagnosis and disease treatment.